\documentclass[CGV,biber]{nowfnt} 

\usepackage[utf8]{inputenc}
\usepackage{bm}
\usepackage{algorithm}
\usepackage{algorithmic}
\usepackage{amsmath,amssymb}
\usepackage{amsthm}
\usepackage{tcolorbox}
\usepackage{xcolor}
\usepackage{lipsum} 
\usepackage{amssymb}
\usepackage{booktabs}
\usepackage{longtable}
\usepackage{stmaryrd}
\usepackage[framemethod=TikZ]{mdframed}
\newcommand{\intset}[1]{\llbracket  #1 \rrbracket}

\newcommand{\Sm}{\mathbb{S}}
\newcommand{\fb}{\bm{\mathrm{f}}}
\newcommand{\gb}{\bm{\mathrm{g}}}
\newcommand{\Rb}{\bm{\mathrm{R}}}
\newcommand{\Cb}{\bm{\mathrm{C}}}
\newcommand{\alphab}{\bm{\alpha}}
\newcommand{\betab}{\bm{\beta}}
\newcommand{\setX}{\mathcal{X}}
\newcommand{\setWR}{\mathcal{WR}}
\newcommand{\setP}{\mathcal{P}}
\newcommand{\setU}{\mathcal{U}}
\newcommand{\setY}{\mathcal{Y}}
\newcommand{\N}{\mathbb{N}}
\newcommand{\setF}{\mathcal{F}}
\newcommand{\setM}{\mathcal{M}}
\newcommand{\setR}{\mathcal{R}}
\newcommand{\setS}{\mathcal{S}}
\newcommand{\setGR}{\mathcal{GR}}
\newcommand{\setPR}{\mathcal{PR}}
\newcommand{\setC}{\mathcal{C}}
\newcommand{\diff}{\mathrm{d}}
\renewcommand{\Re}{\mathbb{R}}
\newcommand{\simiid}{\overset{i.i.d}{\sim}}
\newcommand{\Cm}{\mathbb{C}}

\definecolor{lightgray}{RGB}{245,245,245}

\newmdenv[
  backgroundcolor=lightgray,
  linecolor=lightgray,
  roundcorner=5pt,
  innertopmargin=1pt,
  innerbottommargin=3pt,
  innerleftmargin=3pt,
  innerrightmargin=3pt,
  skipabove=5pt, 
  skipbelow=5pt 
]{thmframe}

\newtheorem{theoremplain}{Theorem}[chapter]
\newtheorem{propositionplain}{Proposition}[chapter]
\newtheorem{definitionplain}{Definition}[chapter]
\newtheorem{remarkplain}{Remark}[chapter]

\renewenvironment{proposition}
  {\begin{thmframe}\begin{propositionplain}}
  {\end{propositionplain}\end{thmframe}}

\renewenvironment{definition}
  {\begin{thmframe}\begin{definitionplain}}
  {\end{definitionplain}\end{thmframe}}

\renewenvironment{remark}
  {\begin{thmframe}\begin{remarkplain}}
  {\end{remarkplain}\end{thmframe}}

\title{An Introduction to Sliced Optimal Transport}

\subtitle{Foundations, Advances, Extensions,  and Applications}

\maintitleauthorlist{
Khai Nguyen \\
Department of Statistics and Data Sciences
\\
University of Texas at Austin \\
khainb@utexas.edu
}

\usepackage{mwe}

\author[1]{Nguyen, Khai}

\affil[1]{Department of Statistics and Data Sciences, University of Texas at Austin; khainb@utexas.edu}

\begin{document}

\makeabstracttitle

\begin{abstract}
Sliced Optimal Transport (SOT) is a rapidly developing branch of optimal transport (OT) that exploits the tractability of one-dimensional OT problems. By combining tools from OT, integral geometry, and computational statistics, SOT enables fast and scalable computation of distances, barycenters, and kernels for probability measures, while retaining rich geometric structure. This work provides a comprehensive review of SOT, covering its mathematical foundations, methodological advances, computational methods, and applications. We discuss key concepts of OT and one-dimensional OT, the role of tools from integral geometry such as Radon transform in projecting measures, and statistical techniques for estimating sliced distances. The work further explores recent methodological advances, including non-linear projections, improved Monte Carlo approximations, statistical estimation techniques for one-dimensional optimal transport,  weighted slicing techniques, and transportation plan estimation methods. Variational problems, such as minimum sliced Wasserstein estimation, barycenters, gradient flows, kernel constructions, and embeddings are examined alongside extensions to unbalanced, partial, multi-marginal, and Gromov-Wasserstein settings. Applications span machine learning, statistics, computer graphics and computer visions, highlighting SOT’s versatility as  a practical computational tool. This work will be of interest to researchers and practitioners in machine learning, data sciences, and computational disciplines seeking efficient alternatives to classical OT.

\end{abstract}

\chapter{Introduction}
\label{chapter:intro} 

Optimal transport (OT) is a mathematical topic with a rich history. From Gaspard Monge to Leonid Kantorovich, Yann Brenier, Cédric Villani, Luis Caffarelli, and Alessio Figalli, many fundamental aspects of OT have been explored. OT has deep connections to various mathematical domains, including optimization, partial differential equations, probability, and measure theory. In recent years, the computational aspects of OT have become a highly active area of research, with significant contributions from researchers such as Marco Cuturi and Gabriel Peyré. As a fundamental tool, OT has been applied in diverse domains including economics, statistics, machine learning, data science, robotics, computer graphics, computer vision, and natural language processing.

Several books have been written on the foundational aspects of OT, for example, the works of Cédric Villani~\citep{villani2021topics,villani2008optimal}, Filippo Santambrogio~\citep{santambrogio2015optimal}, and Alessio Figalli and Federico Glaudo~\citep{figalli2021invitation}. For computational aspects, a recent monograph by Gabriel Peyré and Marco Cuturi~\citep{peyre2019computational} offers a comprehensive overview. There are also domain-specific references on OT, such as a book for OT in economics~\citep{galichon2016optimal}, statistical aspects of OT~\citep{chewi2025statistical}, a survey for OT in signal processing and machine learning~\citep{kolouri2017optimal}, and a survey on OT in computer graphics and computer vision~\citep{bonneel2023survey}, among others.

Sliced Optimal Transport (SOT) is an emerging branch of OT that has gained attention due to its appealing computational benefits and strong theoretical ties to OT. SOT leverages the convenience of one-dimensional OT, such as closed-form solutions for transport maps and costs, and fast statistical estimation rates. While SOT is a relatively new field, originating from seminal work in computer graphics by Julien Rabin, Gabriel Peyré, Julie Delon, and Marc Bernot~\citep{rabin2011wasserstein} in 2011, it has since been widely adopted as a computationally efficient alternative to OT in various applications.

\begin{figure}[!t]
    \centering
    \includegraphics[width=0.8\linewidth]{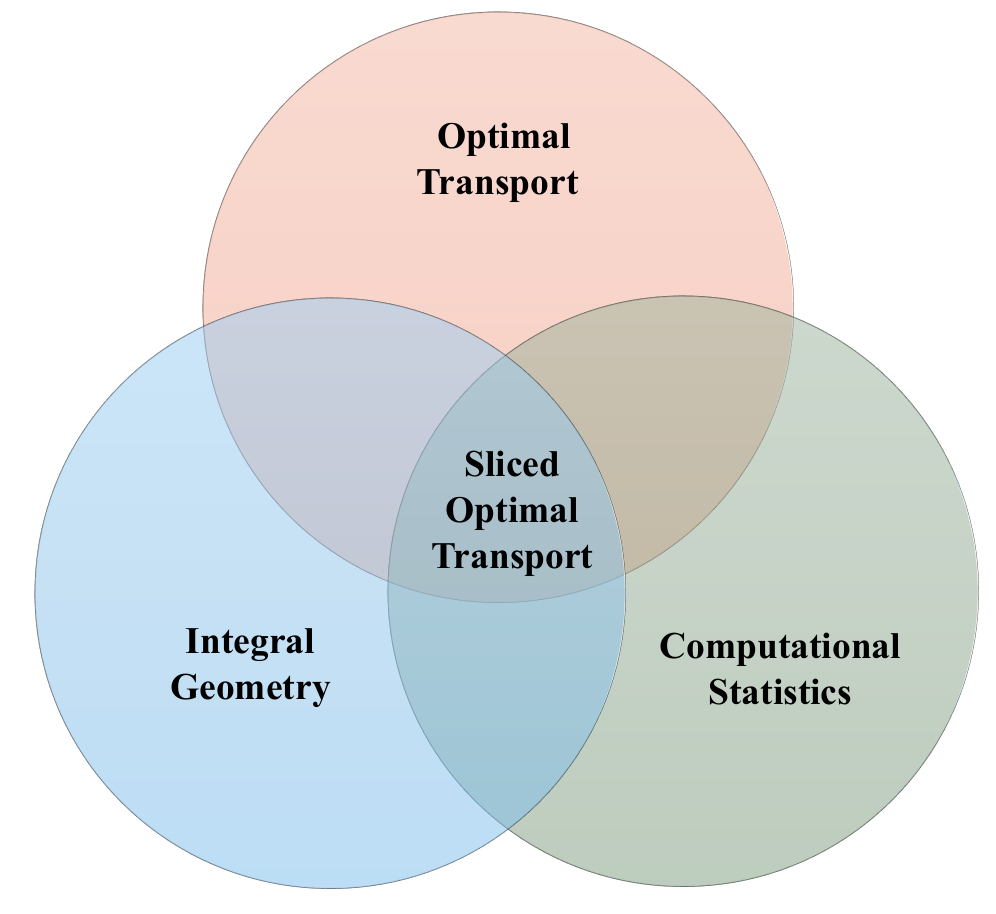} \\
    \caption{Sliced optimal transport is the intersection of optimal transport, integral geometry, and computational statistics.}
    \label{fig:SOT_intersection}
\end{figure}

\begin{figure}[!t]
    \centering
    \includegraphics[width=0.8\linewidth]{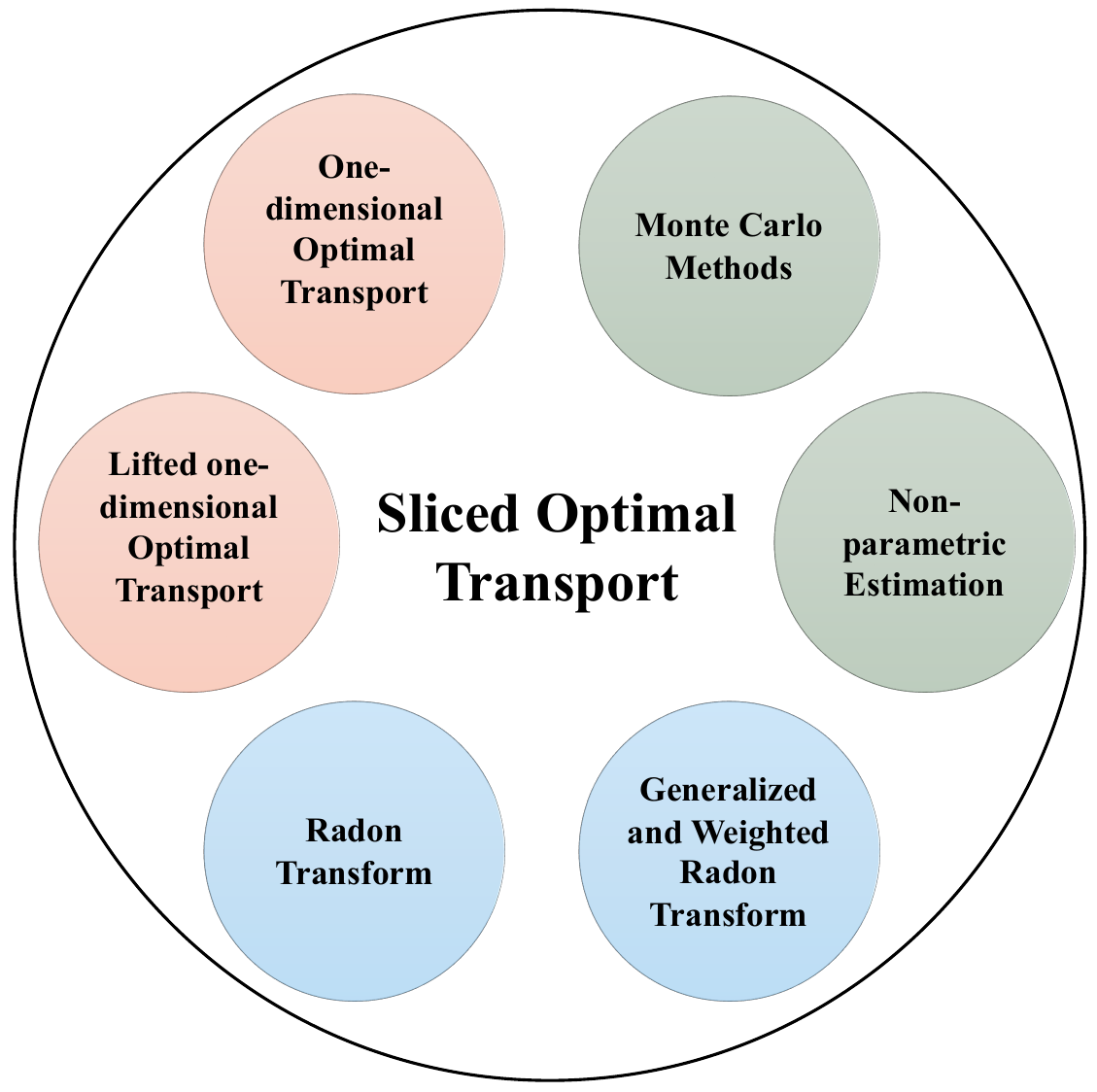}
    \caption{Sub-domains of sliced optimal transport (not exhaustive).}
    \label{fig:SOT_aspects}
\end{figure}

SOT can be viewed as lying at the intersection of OT, integral geometry, and computational statistics (Figure~\ref{fig:SOT_intersection}). In more detail, tools from integral geometry, such as the Radon transform, are used to project high-dimensional probability measures into a new space where OT evaluation can be decomposed into (infinitely many) one-dimensional problems. To make computation tractable, computational statistics methods like Monte Carlo methods and non-parametric estimation techniques (e.g., quantile estimation, CDF estimation) are employed to estimate the transportation in the transformed space. Methodological advances in SOT focus on developing more appropriate projection tools that preserve the geometry of the original space. Many works also explore advanced Monte Carlo methods and non-parametric estimation techniques to improve the quality of SOT approximations~\citep{nguyen2024sliced,nguyen2024quasimonte,leluc2024sliced,sisouk2025user,dai2021sliced}. Additionally, defining meaningful transport on the original space using SOT remains an open and pressing question. We summarize sub-domains of SOT in Figure~\ref{fig:SOT_aspects}.

\begin{figure}[!t]
    \centering
    \includegraphics[width=0.8\linewidth]{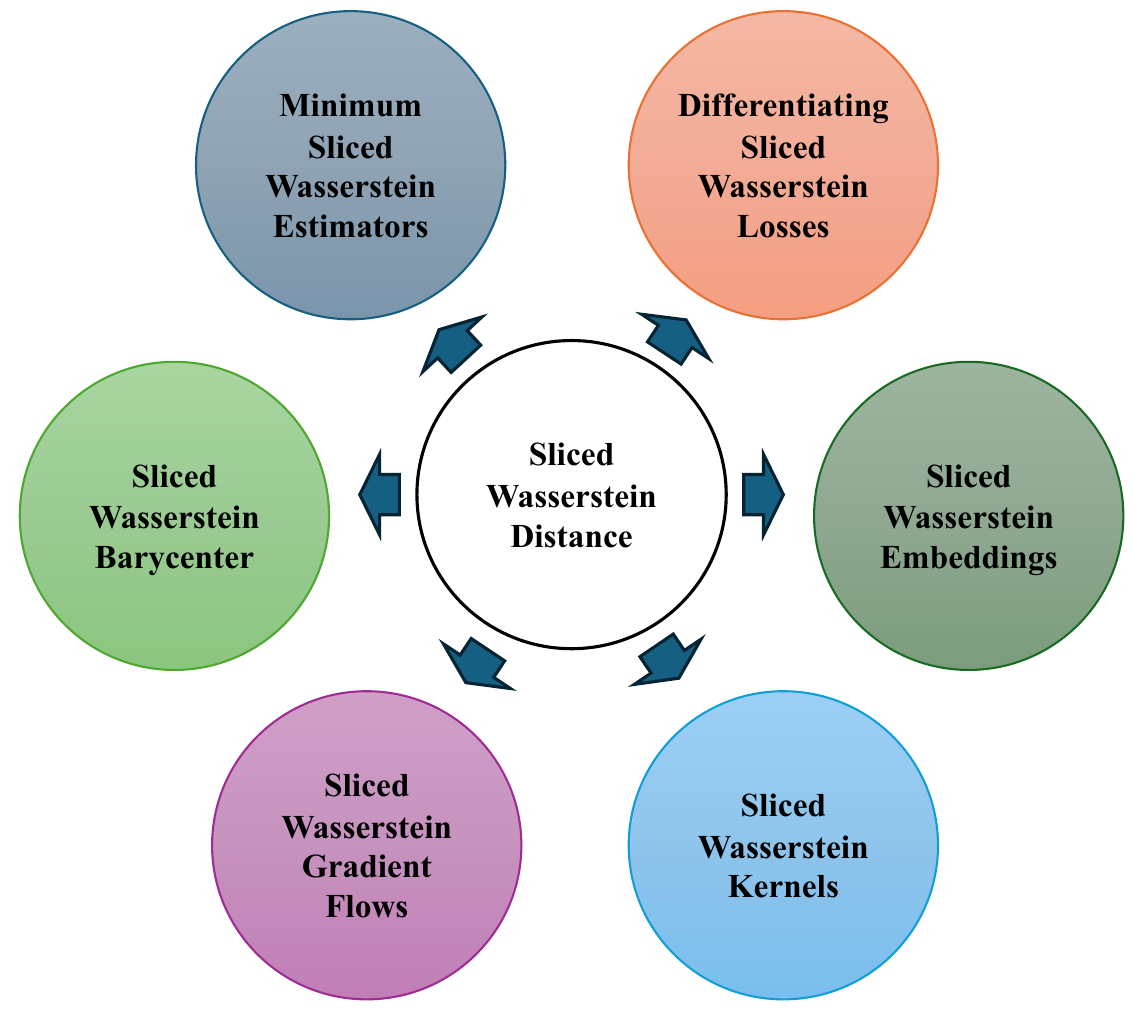} \\
    \caption{Variational problems arising from SOT.}
    \label{fig:SOT_variations}
\end{figure}

Like OT, SOT induces a metric structure on the space of probability measures. This allows for parameter estimation, computation of barycenters, and the definition of gradient flows based on SOT. Notably, SOT provides a framework for defining kernels between probability measures and enables lossless embeddings, capabilities that classical OT does not naturally possess. Beyond developing the core tools and variational formulations of SOT, another exciting direction is the creation of SOT variants for extensions of OT, such as unbalanced OT, partial OT, and multi-marginal OT. These variants not only expand the applicability of OT but also highlight analogies between the OT and SOT frameworks.

While current optimal transport books and surveys typically include only the one-dimensional case or include sliced optimal transport (SOT) only as a subsection, this work aims to fill the gap by providing a comprehensive review of SOT, with an emphasis on methodology, computation, and applications. In addition, some fundamental theoretical properties of SOT are presented to motivate the development of methods and algorithms. To the best of our knowledge, there is no existing work that provides a synthesized view of SOT.

In Chapter~\ref{chapter:foundations}, we cover the foundations of Sliced Optimal Transport (SOT). We begin with a brief review of measures and related concepts such as push-forward measures and disintegration of measures in Section~\ref{sec:measures:chapter:foundations}. Next, Sections~\ref{sec:OT:chapter:foundations} to~\ref{sec:1DOT:chapter:foundations} present key concepts of optimal transport (OT), one-dimensional OT, and relevant computational aspects. Following this, Section~\ref{sec:Radontransform:chapter:foundations} introduces integral geometry tools for functions and measures, including the Radon transform, Fourier transform, the central slice theorem, and their properties. Section~\ref{sec:SOTdistances:chapter:foundations} reviews the sliced Wasserstein distance along with its theoretical and computational properties. Finally, Section~\ref{sec:IDT_Konthe:chapter:foundations} discusses the iterative distributional transfer algorithm and Knothe’s transport.

Chapter~\ref{chapter:advances} discusses recent advances in SOT. Section~\ref{sec:generalized_slicing:chapter:advances} explores nonlinear projection via generalized Radon transforms on manifolds and geometric spaces. Monte Carlo methods for approximating SOT with uniform slicing measure are reviewed in Section~\ref{sec:MC:chapter:advances}. Section~\ref{sec:quantile:chapter:advances} covers various estimation methods for one-dimensional OT based on samples from probability measures. Next, Section~\ref{sec:slicing_measure:chapter:advances} discusses weighted Radon transforms, leading to SOT variants with non-uniform slicing measures. The chapter concludes with Section~\ref{sec:map:chapter:advances}, which reviews recent techniques for obtaining transportation plans from SOT.

\begin{figure}[!t]
    \centering
    \includegraphics[width=1\linewidth]{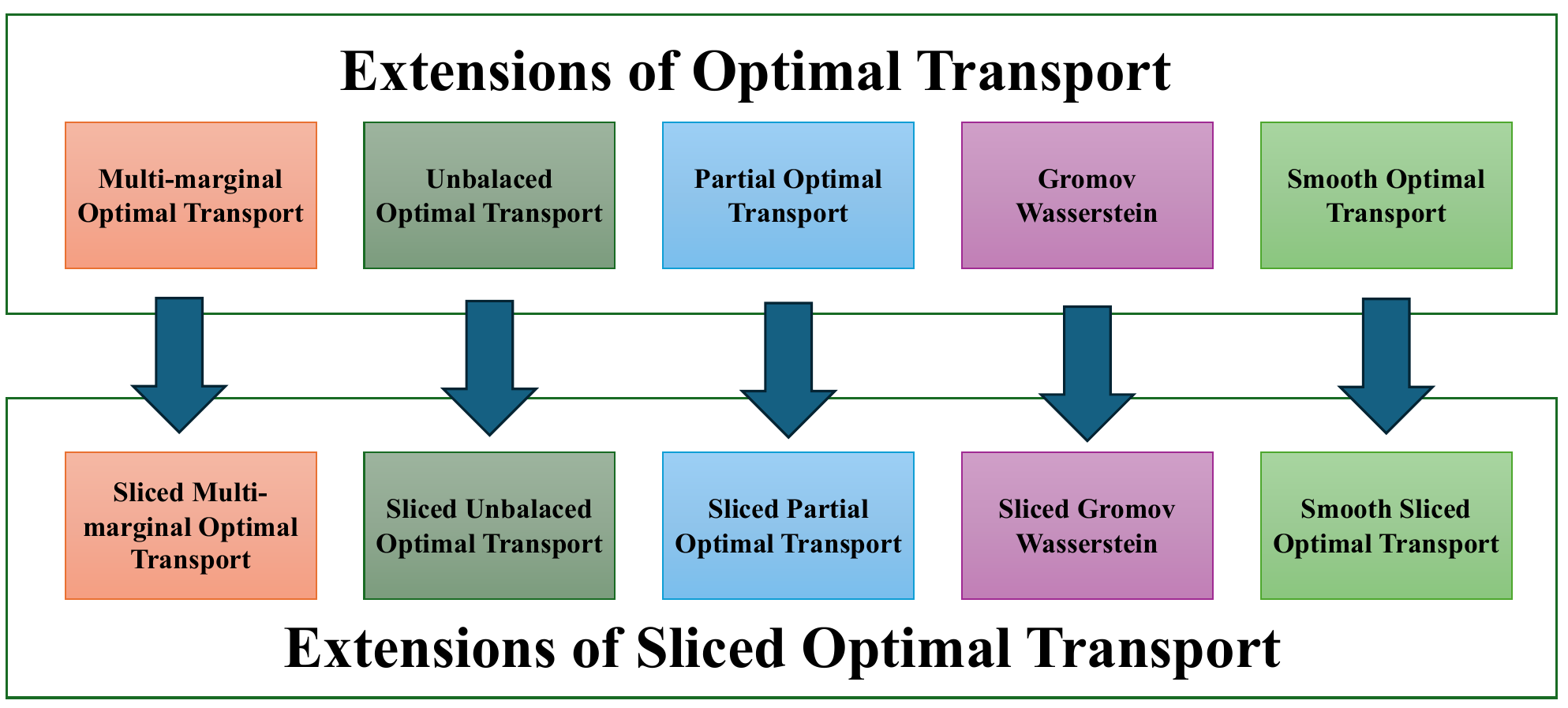}
    \caption{Extensions of sliced optimal transport.}
    \label{fig:SOT_extensions}
\end{figure}

Variational problems arising from SOT (Figure~\ref{fig:SOT_variations}) are discussed in Chapter~\ref{chapter:varitational_SW}. Specifically, Section~\ref{sec:MSWE:chapter:varitational_SW} reviews minimum sliced Wasserstein estimators, followed by a discussion on the differentiability of sliced Wasserstein losses in Section~\ref{sec:differentiatingSW:chapter:varitational_SW}. Section~\ref{sec:SWB:chapter:varitational_SW} covers sliced Wasserstein barycenters, and Section~\ref{sec:SWgradientflow:chapter:varitational_SW} reviews sliced Wasserstein gradient flows. Sliced Wasserstein kernels are presented in Section~\ref{sec:SWKernel:chapter:varitational_SW}, and Section~\ref{sec:SWEmbedding:chapter:varitational_SW} concludes with the use of SOT for embeddings.

Extensions of SOT are the focus of Chapter~\ref{chapter:extension} (Figure~\ref{fig:SOT_extensions}). We first review sliced multi-marginal OT in Section~\ref{sec:SMOT:chapter:extension}, followed by discussions on sliced unbalanced OT and sliced partial OT in Sections~\ref{sec:SUOT:chapter:extension} and~\ref{sec:SPOT:chapter:extension}, respectively. Section~\ref{sec:SGW:chapter:extension} introduces sliced Gromov-Wasserstein methods for cross-domain comparison. Finally, Section~\ref{sec:SmoothSOT:chapter:extension} covers smoothed SOT.

\begin{figure}[!t]
    \centering
    \includegraphics[width=1\linewidth]{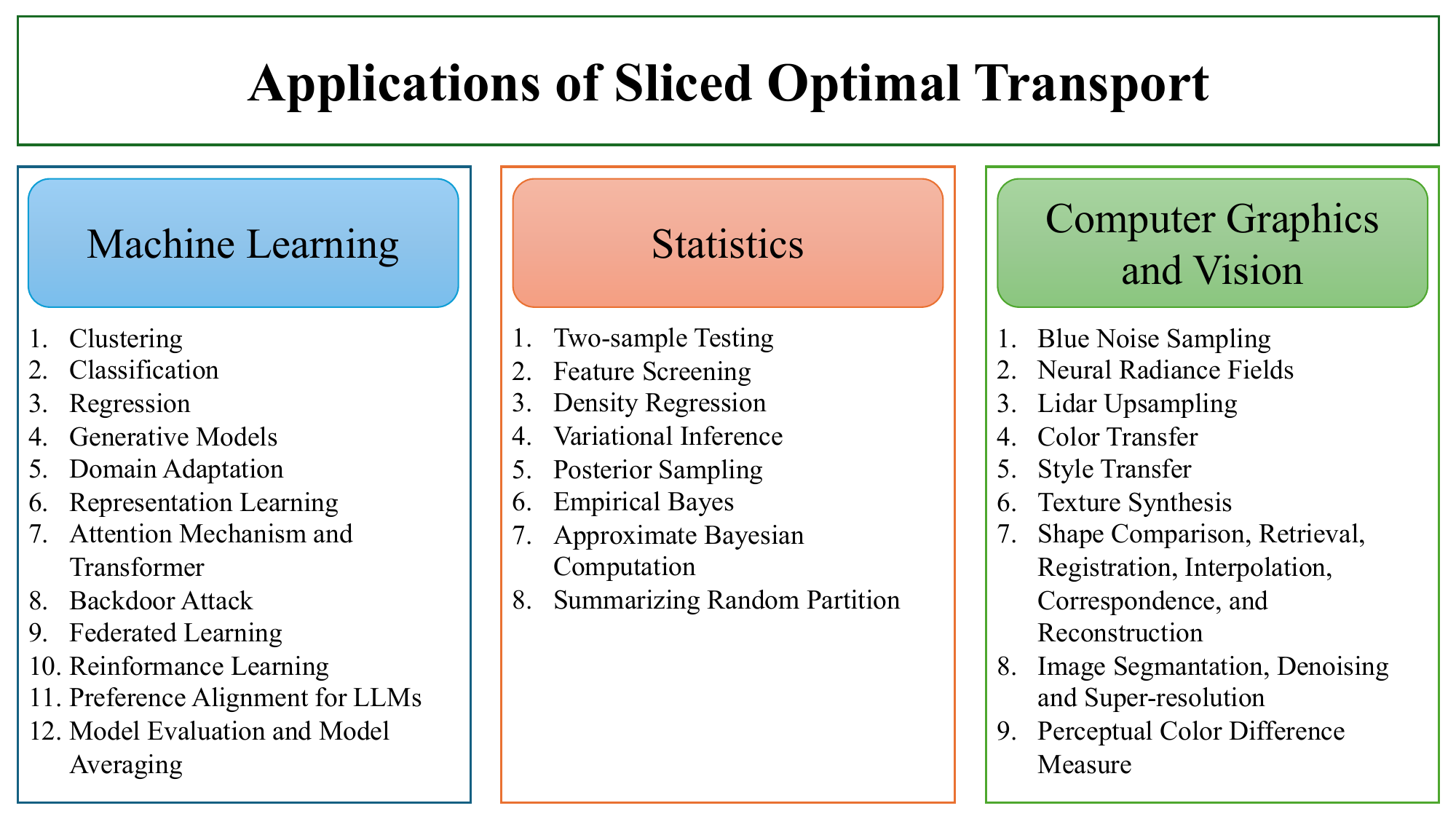}
    \caption{Applications of sliced optimal transport (not exhaustive).}
    \label{fig:SOT_applications}
\end{figure}

In Chapter~\ref{chapter:applications}, we review existing applications of SOT (Figure~\ref{fig:SOT_applications}). For machine learning (Section~\ref{sec:ML:chapter:applications}), topics include clustering, classification, regression, generative models, domain adaptation, representation learning, attention and transformers, backdoor attacks, preference alignment for large language models, federated learning, and reinforcement learning. Applications in statistics are discussed in Section~\ref{sec:stats:chapter:applications}, including two-sample testing, feature screening, density regression, variational inference, empirical Bayes, approximate Bayesian computation, and summarizing random partitions. Lastly, Section~\ref{sec:computer_graphic_and_vision:chapter:applications} reviews applications in computer graphics and vision, such as blue noise sampling, lidar upsampling, neural rendering, color transfer, style transfer, image segmentation, shape comparison and retrieval, shape registration, neural rendering, and more.

We conclude the work by some discussions about some other non-discussed aspects of SOT such as the intersection between optimization and OT, the connection between deep learning and SOT, SOT with discrete one-dimensional structures, and extending from SOT to other sliced probability metrics.

\begin{longtable}{@{} l p{10cm} @{}}
\caption{List of notations used in this work (not exhaustive).} \label{tab:notations} \\

\toprule
\textbf{Notation} & \textbf{Meaning} \\
\midrule
\endfirsthead

\toprule
\textbf{Notation} & \textbf{Meaning} \\
\midrule
\endhead

\bottomrule
\endfoot

$\N$ & Set of natural numbers $\{1,2,\ldots\}$. \\
$\Re$ & Set of real numbers. \\
$\Cm$ & Set of complex numbers. \\
$\Sm^{d-1}$ & Unit hyper sphere in $d$ dimension\\
$S_d^{++}(\Re)$ & Set of positive definite matrices in $\Re^{d\times d}$.\\
$S_d(\Re)$ & Set of symmetric matrices in $\Re^{d\times d}$.\\
$\mathbb{V}_k(\Re^d)$ & Stiefel manifold $\mathbb{V}_k(\Re^d) = \{\Theta \in \Re^{d\times k} \mid \Theta^\top \Theta = I_k\}$. \\
$\Delta^n$ & Simplex of size $n$. \\
$\intset{m}$ & $\{1,2,\ldots,m\}$. \\
$\setM(\setX)$ & Set of Borel measures on $\setX$. \\
$\setM_+(\setX)$ & Set of Borel positive measures on $\setX$. \\
$\setM_a(\setX)$ & Set of Borel positive measures on $\setX$ with total mass $a>0$. \\
$\setC(\setX)$ & Set of continuous functions on $\setX$. \\
$\setP(\setX)$ & Set of Borel probability measures on $\setX$. \\
$\setP_p(\setX)$ & Set of Borel probability measures on $\setX$ with finite $p$-th moment. \\
$\setP_c(\setX)$ & $\{ \mu \in \setP(\setX) \mid \exists\, x_0 \in \setX, \int_{\setX} c(x,x_0) \, \mathrm{d} \mu(x) < \infty \}$ \\
$\Pi(\alphab,\betab)$ & $\{\pi \in \Re_+^{n \times m} \mid \pi\mathbf{1} = \alphab, \pi^\top \mathbf{1} = \betab\}$. \\
$\Pi(\mu,\nu)$ & $\{\pi \in \setM(\setX\times \setY) \mid \pi(A,\setY) = \mu(A), \pi(\setX,B) = \nu(B)\}$. \\
$\delta_{x}$  & Dirac measure at $x$\\
$\setU(\setX)$ & Uniform distribution over set $\setX$. \\
$M_q(\mu)$ & $q$-th moment $M_q(\mu) = \int_{\Re^d} \|x\| \, \mathrm{d} \mu(x)$. \\
$\mathbb{E}_{x\sim p}[f(x)]$ & Expectation of $f(x)$ with $x \sim p(x)$. \\
$\Rb(\Cb)$ & Set of Kantorovich discrete duals $\{(\fb,\gb) \in \Re^n \times \Re^m \mid f_i+g_j \leq C_{ij} \ \forall\, i\in \intset{n}, j\in \intset{m}\}$. \\
$\setR(c)$ & Set of Kantorovich duals $\{(f,g) \in \setC(\setX)\times \setC(\setY) \mid f(x)+g(y)\leq c(x,y) \ \forall\, (x,y) \in \setX\times \setY\}$. \\
$\mathbf{1}$ & Vector with all entries equal to 1. \\
$I(A)$ & Indicator function i.e., equals to 1 when A is true, equals to 0 otherwise\\
$Id$ & Identity function, i.e., $Id(x) = x$. \\
$F_\mu$ & CDF of measure $\mu$, i.e., $F_\mu(x) = \mu((-\infty,x])$. \\
$F_\mu^{-1}$ & Quantile function of $\mu$, i.e., $F_\mu^{-1}(q) = \inf \{y \in \Re \mid q \leq F_\mu(y)\}$. \\
$\setR \mu$ & Radon transform of $\mu$. \\
$\setR f$ & Radon transform of $f$. \\
$\setR_\theta \mu$ & Conditional measure of Radon transform of $\mu$ given $\theta$. \\
$\theta \sharp \mu$ & $\setR_\theta \mu$. \\
$f \sharp \mu$ & Pushforward measure of $\mu$ through function $f$. \\
$p_\mu(x)$ & Density of a measure $\mu$ with respect to Lebesgue measure, i.e., $\frac{\mathrm{d} \mu (x)}{\mathrm{d} x}$. \\
$\mu \otimes \nu$ & Product measure between $\mu$ and $\nu$, i.e., $\int_{\setX\times \setY} f(x,y) \, \mathrm{d}(\mu \otimes \nu)(x,y) = \int_{\setX} \int_{\setY} f(x,y) \, \mathrm{d} \mu(x) \, \mathrm{d} \nu(y)$. \\
$W_c(\mu,\nu)$ & Wasserstein distance between $\mu$ and $\nu$ with ground metric $c(x,y)$. \\
$W_{c,p}(\mu,\nu)$ & Wasserstein-$p$ distance between $\mu$ and $\nu$ with ground metric $c(x,y)$. \\
$W_{p}(\mu,\nu)$ & Wasserstein-$p$ distance between $\mu$ and $\nu$ with ground metric $c(x,y)=\|x-y\|_p$. \\
$SW_c(\mu,\nu)$ & Sliced Wasserstein distance between $\mu$ and $\nu$ with ground metric $c(x,y)$. \\
$SW_{c,p}(\mu,\nu)$ & Sliced Wasserstein-$p$ distance between $\mu$ and $\nu$ with ground metric $c(x,y)$. \\
$SW_{p}(\mu,\nu)$ & Sliced Wasserstein-$p$ distance between $\mu$ and $\nu$ with ground metric $c(x,y)=\|x-y\|_p$. \\
\end{longtable}


\chapter{Foundations of Sliced Optimal Transport}
\label{chapter:foundations}

This chapter describes the foundations of Sliced Optimal Transport (SOT), starting with a brief review of measures and related concepts such as push-forward measures and disintegration of measures. We then introduce the key ideas of optimal transport, including the one-dimensional case and relevant computational methods. After that, we present integral geometry tools for functions and measures, including the Radon transform, Fourier transform, the central slice theorem, and their properties. Next, we discuss the sliced Wasserstein distance, outlining both its theoretical and computational aspects. Finally, we examine the iterative distributional transfer algorithm and Knothe’s transport.

\section{Measures}
\label{sec:measures:chapter:foundations}

Similar to OT, SOT is a mathematical tool to interact with measures, e.g., matching, comparing, and interpolating. In mathematics, a measure is a formal, generalized concept that extends familiar concepts such as length, area, volume, and other similar concepts such as mass, magnitude, and probability. Measures play a fundamental role in areas such as probability theory, statistics, and machine learning. We provide a brief review of measures and some related concepts such as push-foward measure and disintegration of measures in this section.

\begin{definition}[Sigma-algebra]
Let \( \mathcal{X} \) be a nonempty set. A collection \( \mathcal{F} \subseteq 2^\mathcal{X} \) ($2^\mathcal{X}$ is the set of all subsets of 
$\mathcal{X}$) is called a sigma-algebra (or \(\sigma\)-algebra) on \( \setX \) if it satisfies the following properties:
\begin{enumerate}
    \item \( \mathcal{X} \in \mathcal{F} \).
    \item If \( A \in \mathcal{F} \), then \( A^c = \mathcal{X} \setminus A \in \mathcal{F} \).
    \item If \( \{A_n\}_{n=1}^{\infty} \subseteq \mathcal{F} \), then $
    \bigcup_{n=1}^{\infty} A_n \in \mathcal{F}.$
\end{enumerate}
\end{definition}

\begin{definition}[Measurable space and measurable sets]
\label{def:measurable_space}
A measurable space is a pair \((\setX, \mathcal{F})\) consisting of a nonempty set $\setX$ and a \(\sigma\)-algebra $\mathcal{F}$. The elements of \(\mathcal{F}\) are called measurable sets.
\end{definition}

\begin{definition}[Measure]
\label{def:measure}
Let \((\setX, \mathcal{F})\) be a measurable space. 
A measure on \((\setX, \mathcal{F})\) is a function
\[
\mu : \mathcal{F} \to [0, \infty]
\]
satisfying the following properties:
\begin{enumerate}
    \item \textbf{Null empty set:} \(\mu(\emptyset) = 0\).
    \item \textbf{Countable additivity:} For any sequence \(\{A_i\}_{i=1}^{\infty}\) of pairwise disjoint sets in \(\mathcal{F}\),
    \[
    \mu\left( \bigcup_{i=1}^{\infty} A_i \right) 
    = \sum_{i=1}^{\infty} \mu(A_i).
    \]
\end{enumerate}
The triple \((\setX, \mathcal{F}, \mu)\) is called a measure space.
\end{definition}

\begin{definition}[Measurable function]
\label{def:measurable_function}
Let \((\setX, \mathcal{F})\) and \((\mathcal{Y}, \mathcal{G})\) be measurable spaces.  
A function \(f : \setX \to \mathcal{Y}\) is called measurable if, for every set \(B \in \mathcal{G}\),
\begin{align}
f^{-1}(B) = \{ x \in \setX : f(x) \in B \} \in \mathcal{F}.
\end{align}
The preimage of every measurable set in \(\mathcal{Y}\) is measurable in \(\setX\).
\end{definition}

\begin{definition}[Discrete measure]
\label{def:discrete_measure}
Let \((\setX, \mathcal{F})\) be a measurable space.  
A discrete measure \(\mu\) on \((\setX, \mathcal{F})\) with weights \(\alphab = (\alpha_1, \ldots, \alpha_n)\) and support points \(x_1, \ldots, x_n \in \setX\) (where \(n \in \mathbb{N}\)) is defined by
\begin{align}
    \mu = \sum_{i=1}^n \alpha_i \, \delta_{x_i},
\end{align}
where \(\delta_{x_i}\) denotes the Dirac measure at the point \(x_i\).  
For any measurable function \(f : \setX \to \mathbb{R}\) that is continuous at \(x_i\),
\begin{align}
    \int_{\setX} f(x) \, \mathrm{d}\delta_{x_i}(x) = f(x_i).
\end{align}
Equivalently, in terms of the Dirac delta function \(\delta(\cdot)\),
\begin{align}
    \int_{\setX} f(x) \, \mathrm{d}\delta_{x_i}(x) 
    = \int_{\mathbb{R}} f(x)\,\delta(x - x_i)\,\mathrm{d}x,
\end{align}
where
\begin{align}
    \delta(t) =
    \begin{cases}
        \infty, & t = 0,\\[0.5em]
        0, & t \neq 0,
    \end{cases}
    \quad \text{and} \quad \int_{\mathbb{R}} \delta(t) \, \mathrm{d}t = 1.
\end{align}
\end{definition}

The discrete measure $\mu$ is a probability measure if the weights belong to the simplex of size $n$, i.e., $\alphab \in \Delta^n:=\{\alphab=(\alpha_1,\ldots,\alpha_n)\mid \sum_{i=1}^n \alpha_i=1, \alpha_i >0 \,\forall i\}$. The discrete measure $\mu$ is a positive measure if $\sum_{i=1}^n \alpha_i=a$ for $a >0$ and $\alpha_i >0\,\forall i$.

\begin{definition}[Continuous measure]
\label{def:continuous_measure}
Let \((\setX, \mathcal{F})\) be a measurable space.  
A measure \(\mu\) on \((\setX, \mathcal{F})\) is called continuous (or absolutely continuous with respect to the Lebesgue measure) if there exists a non-negative function \(p_\mu : \setX \to [0, \infty)\), called the density of \(\mu\), such that for every measurable function \(f : \setX \to \mathbb{R}\) that is continuous and integrable,
\begin{align}
    \int_\setX f(x) \, \mathrm{d}\mu(x) = \int_\setX f(x) \, p_\mu(x) \, \mathrm{d}x.
\end{align}
The function \(p_\mu(x)\) is the Radon–Nikodym derivative of \(\mu\) with respect to the Lebesgue measure:
$p_\mu(x) = \frac{\mathrm{d}\mu}{\mathrm{d}x}.$
\end{definition}

The nonnegative continuous measure $\mu$ is a probability measure if $\mu(\setX) =1$. For example, we have a Gaussian probability measure $\mu(A)=\frac{1}{\sqrt{(2\pi)^d \det(\Sigma)}} \int_A \exp\Bigg(-\frac{1}{2} (x-\mathbf{m})^\top \Sigma^{-1} (x-\mathbf{m})\Bigg)$ for $A \in \mathcal{F}$, with a mean vector $\mathbf{m} \in \mathbb{R}^d$ and a positive semidefinite covariance matrix $\Sigma \in \mathbb{R}^{d \times d}$. While probability measures and probability distributions are not the same, they are often used interchangeably in machine learning contexts. Therefore, we will use them interchangeably. We will omit explicitly specifying the $\sigma$-algebra. In most contexts, the choice of measurable sets $\mathcal{F}$ is either standard (e.g., the Borel $\sigma$-algebra, which is the smallest $\sigma$-algebra containing all open subsets of $\setX$) or is clear from the underlying space $\setX$. For convenience, we denote by $\setM(\setX)$ the set of Borel measures on $\setX$, $\setM_+(\setX)$ the set of  Borel positive measures on $\setX$, $\setM_a(\setX)$ the set of Borel positive measures on $\setX$ with total mass $a$ ($\mu(\mathcal{X)}=a$), and  $\setP(\setX)$ the set of Borel probability measures on $\setX$.

Next, we discuss some important concepts related to measures, including the push-forward measure, change of variables, and disintegration of measures. The push-forward measure will be used later to define optimal transport, while the disintegration of measures is later used to define sliced optimal transport.

\begin{definition}[Push-forward]
\label{def:push_foward} Given a measure $\mu \in \setM(\setX)$ and a measurable mapping $T:\setX\to\setY$, the push-forward measure $\nu = T\sharp \mu$ satisfies:
\begin{align}
    \int_\setY f(y)\diff \nu(y) = \int_\setX f(T(x)) \diff \mu(x),
\end{align}
for any continuous function $f$. Equivalently, for any measurable set $B\in \setY$, we have
\begin{align}
    \nu(B) = \mu(\{x \in \setX\mid T(x) \in B\}) = \mu(T^{-1}(B)).
\end{align}
$T_\sharp$ preserves positivity and total mass.
\end{definition}

\begin{remark}[Push-forward discrete measure]
    \label{remark:push_forward_discrete} Given a discrete measure $\mu = \sum_{i=1}^n \alpha_i \delta_{x_i}$ ($x_i \in \setX \, \forall \,i$) and a measurable mapping $T:\setX\to\setY$, the push-forward measure is:
    \begin{align}
        T\sharp \mu = \sum_{i=1}^n \alpha_i \delta_{T(x_i)}.
    \end{align}
    The push-forward operation consists simply of moving the positions of all the points in the support of the measure.
\end{remark}

\begin{definition}[Change of variables]
    \label{def:push_forward_continuous} When $\mu$ is a continuous measure on $\Re^d$ (dimension $d\in \N$) and the mapping $T:\Re^d \to \Re^d$ is smooth and bijective, the density function of $\nu=T\sharp \mu$ is defined as follows:
    \begin{align}
        p_\mu(x) = | \det(T'(x))| p_\nu(T(x)),
    \end{align}
    where $T'(x) \in \Re^d\times \Re^d$ is the Jacobian matrix.
\end{definition}

\begin{definition}[Disintegration of measures]
    \label{def:disintegration_measures} As discussed in~\citet{getoor1980claude,ambrosio2008gradient}, for any measure $\mu$ on $\Re^d$, let $\nu = f \sharp \mu$ for $f:\Re^d \to \setX$ ($\setX\subset \Re^d$). There exists a $\nu$-almost everywhere uniquely determined family of probability measures $\{\mu_x\}_{x \in \setX}$ on $\Re^d$ such that:
    \begin{align}
        \mu_x(\Re^d \backslash f^{-1}(x))=0 \quad \text{for} \quad \nu \text{-a.e.} \, x \in \setX,
    \end{align}
    and 
    \begin{align}
        \int_{\Re^d} f(x) \diff \mu(x) =  \int_{\setX} \int_{f^{-1}(x)} f(y) \diff \mu_x(y) \diff \nu (x),
    \end{align}
    for any continuous function $f:\Re^d \to \Re_+$.
\end{definition}
A property $P(x)$ holds \emph{$\nu$-almost everywhere} ($\nu$-a.e.) if it fails only on a set of $\nu$-measure zero, i.e., $\nu(\{x \in \setX : P(x) \text{ does not hold}\}) = 0$.

\section{Optimal Transport}
\label{sec:OT:chapter:foundations}

We now review some fundamental concepts in optimal transport (OT) to establish a solid foundation for understanding sliced optimal transport (SOT). This includes key concepts such as the Monge problem, the Kantorovich problem and Kantorovich duality, and the Wasserstein distance and its properties.

\begin{definition}[Monge problem between discrete measures]
    \label{def:Monge_discrete} Given discrete measures $\mu=\sum_{i=1}^n \alpha_i \delta_{x_i}$ and $\nu = \sum_{j=1}^m \beta_j \delta_{y_j}$, the Monge problem~\citep{monge1781memoire} finds a map that assigns each point $x_i$ to a single
point $y_j$ and preserves the weights. In particular, $T:\{x_1,\ldots,x_n\} \to \{y_1,\ldots,y_m\}$ must satisfy:
\begin{align}
    \forall\, j \in [[m]], \quad \beta_j = \sum_{i\mid T(x_i)= y_j} \alpha_i,
\end{align}
which is denoted as $T \sharp \mu = \nu$. The map also needs to be optimal in the sense that it minimizes the transportation cost with respect to a given ground metric $c(x,y)$:
\begin{align}
    \min_{T\mid T \sharp \mu =\nu} \sum_{i=1}^n c(x_i,T(x_i)).
\end{align}
Assuming all $x$'s and $y$'s are distinct, a mapping $T$ can be encoded using indices $\sigma: \intset{n} \to \intset{m}$ such that:
\begin{align}
    \sum_{i \in \sigma^{-1}(j)} \alpha_i = \beta_j,
\end{align}
where the inverse $\sigma^{-1}(j)$ is the preimage set of $j$. When $n=m$, $\alpha_i=1/n \, \forall \, i$ and $\beta_j = 1/m \, \forall \, j$, the Monge problem is equivalent to the optimal assignment problem~\citep{dantzig1983reminiscences}. The optimal map $T$ might not exist in the Monge problem, e.g., when $n<m$.
\end{definition}

\begin{definition}[Monge problem between general measures]
    \label{def:Monge_continous} Given $a>0$, measures $\mu \in \setM_a(\setX)$ and $\nu \in \setM_a(\setY)$, the Monge problem with the ground metric $c:\setX\times \setY \to \Re_+$ finds a measurable map $T:\setX \to \setY$ that minimizes:
\begin{align}
    \min_{T\mid T \sharp \mu =\nu} \int_\setX c(x,T(x)) \diff \mu(x).
\end{align}
Again, the optimal map $T$ might not exist.
\end{definition}

While the Monge problem seeks a deterministic transport map that pushes one measure onto another, it often fails to have a solution, especially when mass splitting is required. Kantorovich addressed this limitation by introducing transport plans, which are probability measures on the product space that allow mass to be split and distributed. 

\begin{definition}[Kantorovich problem between discrete measures]
    \label{def:Kantorovich_discrete}
    Given discrete measures $\mu=\sum_{i=1}^n \alpha_i \delta_{x_i}$ and $\nu = \sum_{j=1}^m \beta_j \delta_{y_j}$, the Kantorovich problem~\citep{kantorovich1942transfer} finds a coupling $\pi$ (or a transportation plan) that assigns each point $x_i$ to multiple
points $y_j$ and preserves the weights. In particular, $\pi \in \Re_+^{n \times m}$ must satisfy:
\begin{align}
    \sum_{i=1}^n \pi_{ij} = \beta_j \, \forall\, j=1,\ldots, m, \quad \sum_{j=1}^m \pi_{ij} = \alpha_i \, \forall\, i=1,\ldots, n.
\end{align}
The set of all $\pi$ satisfying these conditions is denoted as $\Pi(\alphab,\betab)$. The coupling $\pi$ also needs to be optimal in the sense that it minimizes the transportation cost with respect to a given ground metric $c(x,y)$:
\begin{align}
\label{eq:Kantorovich_discrete}
    \min_{\pi \in \Pi(\alphab,\betab)} \sum_{i=1}^n \sum_{j=1}^m \pi_{ij}c(x_i,y_j).
\end{align}
The Kantorovich transportation problem between $\mu$ and $\nu$ can be seen as a transportation problem between $\alphab$ and $\betab$.
\end{definition}

\begin{definition}[Kantorovich dual problem between discrete measures]
    \label{def:Kantorovich_discrete_dual} The Kantorovich problem in~\eqref{eq:Kantorovich_discrete} is a constrained convex minimization problem, hence, it has a dual problem:
    \begin{align}
        \max_{(\fb,\gb) \in \Rb(\Cb)} \sum_{i=1}^n f_i\alpha_i + \sum_{j=1}^m g_j \beta_j,
    \end{align}
    where $\fb=(f_1,\ldots,f_n)$, $\gb=(g_1,\ldots,g_m)$, $\Cb=[[C_{ij}]]$ with $C_{ij}=c(x_i,y_j)$, and $\Rb(\Cb)=\{(\fb,\gb) \in \Re^n \times \Re^m\mid f_i+g_j \leq C_{ij} \, \forall \, i\in \intset{n},j\in \intset{m}\}$. 

\end{definition}
Dual variables $\fb$ and $\gb$ are  referred to as ``Kantorovich potentials". The proof is based directly on  the strong
duality for linear programs~\citep{bertsimas1997introduction}. We refer the reader to Proposition 2.4 in~\citet{peyre2019computational} for a detailed proof.

\begin{definition}[Kantorovich problem between general measures]
    \label{def:Kantorovich_general}
    Given measures $\mu \in \setM_a(\setX)$ and $\nu \in \setM_a(\setY)$, the Kantorovich problem with the ground metric $c:\setX\times \setY \to \Re_+$ finds a coupling $\pi$ that minimizes:
\begin{align}
\label{eq:Kantorovich_general}
    \min_{\pi \in \Pi(\mu,\nu)} \int_{\setX\times \setY} c(x,y) \diff \pi(x,y),
\end{align}
where $\Pi(\mu,\nu)=\{\pi \in \setM_a(\setX\times \setY) \mid \pi(A\times \setY)=\mu(A) \, \forall \, A \subset \setX, \ \pi(\setX\times B)=\nu(B) \, \forall \, B \subset \setY\}$. 
\end{definition}
If $\setX$ and $\setY$ 
are compact spaces and $c$ is continuous, then the Kantorovich problem always has a solution since the product measure $\mu \otimes \nu$ makes $\Pi(\mu,\nu)$ non-empty. 
\begin{definition}[Kantorovich dual problem between general measures]
    \label{def:Kantorovich_general_dual} The Kantorovich problem in~\eqref{eq:Kantorovich_general} has a dual problem:
    \begin{align}
        \sup_{(f,g) \in \setR(c)} \int_\setX f(x) \diff \mu(x) + \int_\setY g(y) \diff \nu(y),
    \end{align}
    where the set of admissible dual potentials is $\setR(c)=\{(f,g) \in \setC(\setX)\times \setC(\setY) \mid f(x)+g(y)\leq c(x,y) \ \forall\, (x,y) \in \setX\times \setY\}$. 
\end{definition}
One can remove the constraint by using the $c$-transform. In particular, the Kantorovich dual problem can be rewritten as:
    \begin{align}
        \sup_{f \in \setC(\setX)} \int_\setX f(x) \diff \mu(x) + \int_\setY f^c(y) \diff \nu(y),
    \end{align}
    where $f^c(y) = \sup_{x} \big( c(x,y) - f(x) \big)$ is the $c$-transform of $f(x)$.
         We refer the reader to Theorem 6.1.1 in~\citet{ambrosio2008gradient} for a detailed proof. 
    The connection between the optimal transport map and the Kantorovich potential~\citep{caffarelli2017allocation,gangbo1995optimal} (Gangbo-McCann Theorem) is
\begin{align}
    T^\star(x) = (\nabla c(x,\cdot))^{-1} \big( \nabla_x(f^\star(x)) \big),
\end{align}
where $f$ is necessarily $c$-concave.

Given the optimal transport problem, we can define the Wasserstein distance, which is the optimal transport cost between two probability measures.

\begin{definition}[Wasserstein distance between discrete measures]
    \label{def:Wasserstein_discrete} Given two discrete measures  $\mu =  \sum_{i=1}^n \alpha_i \delta_{x_i}$ and $\nu = \sum_{j=1}^m \beta_j \delta_{y_j}$,  the Wasserstein distance with the ground metric $c(x,y)$ is the optimal transportation cost of  the Kantorovich problem:
    \begin{align}
        W_c(\mu,\nu)= \min_{\pi \in \Pi(\alphab,\betab)} \sum_{i=1}^n \sum_{j=1}^m \pi_{ij}c(x_i,y_j),
    \end{align}
    where $\Pi(\alphab,\betab)=\{\pi \in \Re_+^{n \times m}\mid \pi\mathbf{1} = \alphab,\pi^\top \mathbf{1} = \betab\}.$ One can extend  the Wasserstein distance to Wasserstein$-p$ ($p\geq 1$) distance:
    \begin{align}
        W_{c,p}^p(\mu,\nu)= \min_{\pi \in \Pi(\alphab,\betab)} \sum_{i=1}^n \sum_{j=1}^m \pi_{ij}c(x_i,y_j)^p.
    \end{align}
\end{definition}

\begin{definition}[Wasserstein distance between general measures]
    \label{def:Wasserstein_general}
    Given a ground metric $c:\Re^d\times \Re^d \to \Re_+$, two measures $\mu\in \setP_{c}(\Re^d)$ and $\nu\in \setP_{c}(\Re^d)$ with $\setP_c(\Re^d) =\{\mu \in \setP(\Re^d) \mid \,\exists\, x_0 \in \Re^{d}, \int_{\Re^d}c(x,x_0) d\mu(x)<\infty \}$, the Wasserstein$-p$ ($p\geq 1$) distance is defined as:
    \begin{align}
\label{eq:Wasserstein_general}
   W_{c,p}^p(\mu,\nu) =\min_{\pi \in \Pi(\mu,\nu)} \int_{\setX\times \setY} c(x,y)^p \diff \pi(x,y),
\end{align}
where $\Pi(\mu,\nu)=\{\pi \in \setP(\Re^d\times \Re^d) \mid \pi(A\times \Re^d)=\mu(A) \, \forall \, A \subset \Re^d, \pi(\Re^d\times B)=\nu(B)\, \forall \,B \subset \Re^d\}$.
\end{definition}

When $c(x,y)=\|x-y\|_p$, we denote the Wasserstein$-p$ distance as $W_p$. We now discuss some basic properties of the Wasserstein distance, including metricity, weak convergence, and sample complexity.

\begin{proposition}[Wasserstein distance is a valid metric]
For $p \geq 1$, the Wasserstein distance ($W_{c,p}$) is a metric on $\mathcal{P}_{c^p}(\Re^d)$. We follow the proofs in~\citet{villani2008optimal,santambrogio2015optimal,peyre2019computational}.

\begin{proof}
We prove that the Wasserstein distance satisfies the four metric properties:

    1. \textbf{Non-negativity:} For any $\mu, \nu \in \mathcal{P}_{c^p}(\Re^d)$, we have
    \begin{align}
    W_{c,p}^p(\mu, \nu) = \inf_{\pi \in \Pi(\mu,\nu)} \int c(x,y)^p \, \diff \pi(x,y) \geq 0,
   \end{align}
    since $c(x,y)^p \geq 0$ due to the nonnegativity of the ground metric.

    2. \textbf{Identity of indiscernibles:} Suppose $W_{c,p}(\mu,\nu) = 0$, then there exists a sequence $\pi_n \in \Pi(\mu,\nu)$ such that
   \begin{align}
    \int c(x,y)^p \, \diff\pi_n(x,y) \to 0.
    \end{align}
    Since $c(x,y)^p \geq 0$, this implies that $\pi_n$ concentrates near the diagonal $\{x=y\}$, and any weak limit of $(\pi_n)$ must be supported on $\{x = y\}$ and have marginals $\mu$ and $\nu$, which implies $\mu = \nu$. Conversely, if $\mu = \nu$, then the identity coupling $\pi = (Id \times Id)_\# \mu$ gives
    \begin{align}
    W_{c,p}(\mu,\mu)^p \leq \int c(x,x)^p \, \diff\mu(x) = 0.
    \end{align}
    Hence $W_{c,p}(\mu,\nu) = 0$ if and only if $\mu = \nu$.

    3. \textbf{Symmetry:} For any $\pi \in \Pi(\mu,\nu)$, the push-forward of $\pi$ under $(x,y) \mapsto (y,x)$ lies in $\Pi(\nu,\mu)$, and since $c(x,y) = c(y,x)$, we have
    \begin{align}
    W_{c,p}(\mu,\nu) = W_{c,p}(\nu,\mu).
    \end{align}

    4. \textbf{Triangle inequality:} For any $\mu,\nu,\lambda \in \mathcal{P}_{c^p}(\Re^d)$, let $\pi_1 \in \Pi(\mu,\nu)$ and $\pi_2 \in \Pi(\nu,\lambda)$ be optimal (or near-optimal) couplings. By the Gluing Lemma (Theorem 7.3 in~\citet{villani2008optimal}), there exists a measure $\gamma$ on $\Re^d \times \Re^d \times \Re^d $ with marginals $\pi_1$ and $\pi_2$. By the triangle inequality of the ground metric $c$,  we  have:
    \begin{align}
    \int c(x,z)^p \, \diff\gamma(x,y,z) \leq \int \left( c(x,y) + c(y,z) \right)^p \diff\gamma(x,y,z).
    \end{align}
    By Minkowski’s inequality (for integrals),
    \begin{align}
         &\left( \int c(x,z)^p \, \diff\gamma (x,y,z)\right)^{1/p} \nonumber \\&\leq \left( \int c(x,y)^p \, \diff\gamma (x,y,z)\right)^{1/p} + \left( \int c(y,z)^p \, \diff\gamma(x,y,z) \right)^{1/p}.
    \end{align}
   
    Since the marginals of $\gamma$ give $\mu$ and $\lambda$, this shows that:
   \begin{align}
    W_{c,p}(\mu,\lambda) \leq W_{c,p}(\mu,\nu) +W_{c,p}(\nu,\lambda).
    \end{align}
\end{proof}
\end{proposition}

\begin{remark}[Weak convergence under Wasserstein distance]
    \label{remark:weak_convergence_Wasserstein}
A sequence of measures $\mu_k$ converges weakly to $\mu$ in $\mathcal{P}(\mathcal{X})$ for a domain $\setX$ if and only if for any continuous function $f \in \mathcal{C}(\mathcal{X})$,
\[
\int_{\mathcal{X}} f(x)\, \diff \mu_k(x) \to \int_{\mathcal{X}} f(x) \, \diff \mu(x).
\]
This convergence can be shown to be equivalent to convergence in Wasserstein distance, i.e., $W_p(\mu_k, \mu) \to 0$ (see \citet[Theorem 5.11, 5.12]{santambrogio2015optimal}) for compact domain $\setX$ and $\Re^d$.
\end{remark}

\begin{remark}[Computational complexity of Wasserstein distance]
    \label{remark:computational_Wasserstein} Computing the Wasserstein distance is an active area of research. In the discrete case, the traditional way to compute Wasserstein distance is through linear programming and the network simplex algorithm~\citep{bertsimas1997introduction,ford1962flows,korte2011combinatorial} with time complexity of $\mathcal{O}(n^3 \log n)$, where $n$ is the largest number of supports from the two measures. The auction algorithm can also be used~\citep{bertsekas1981new,bertsekas1988dual,lahn2019graph} to improve the time complexity to $\mathcal{O}(n^2/\epsilon + n/\epsilon^2)$, where $\epsilon > 0$ is the precision level. By constraining the set of transportation plans via entropic regularization, an approximate solution can be obtained in $\mathcal{O}(n^2/\epsilon)$ with the Sinkhorn algorithm~\citep{cuturi2013sinkhorn}. We refer the reader to~\citet{peyre2019computational} for a detailed discussion about the computation of discrete optimal transport. For continuous Wasserstein distance, neural optimal transport algorithms are introduced based on neural network~\citep{lecun2015deep} parameterization of Kantorovich potentials. For example, the Wasserstein-1 distance is computed in~\citep{arjovsky2017wasserstein}. Wasserstein-2 distance is computed in~\citep{korotin2019wasserstein,makkuva2020optimal} using Brenier's theorem~\citep{brenier1991polar}. More general costs are recently discussed in~\citep{pooladian2024neural,asadulaevneural}. Since we aim to focus on sliced optimal transport, we defer the detailed discussion of computing continuous Wasserstein distance to future work.
\end{remark}

\begin{remark}[Sample complexity of Wasserstein distance]
    \label{remark:sample_complexity_Wasserstein}
    From a statistical point of view, we often observe i.i.d. samples from the two measures $x_1,\ldots,x_n \sim \mu \in \setP_p(\Re^d)$ and $y_1,\ldots,y_m \sim \nu \in \setP_p(\Re^d)$. Here, $\mu$ and $\nu$ are either continuous or unknown. We can compute the discrete Wasserstein distance $W_p(\mu_n,\nu_m)$ (with $\mu_n =  \frac{1}{n}\sum_{i=1}^n \delta_{x_i}$ and $\nu_m =  \frac{1}{m}\sum_{j=1}^m \delta_{y_j}$ being the corresponding empirical measures) as a proxy for $W_p(\mu,\nu)$. The quality of the approximation can be evaluated by 
    \begin{align}
        \mathbb{E}[|W_p(\mu_n,\nu_n) -W_p(\mu,\nu) |].
    \end{align}
    By the triangle inequality of the Wasserstein distance, the non-negativity of Wasserstein distance, and Jensen's inequality, we have:
    \begin{align}
        &\mathbb{E}[|W_p(\mu_n,\nu_n) -W_p(\mu,\nu) |]  \nonumber \\&\leq \mathbb{E}[W_p(\mu_n,\mu)] +\mathbb{E}[W_p(\nu_n,\nu)]  \nonumber \\
        & \leq (\mathbb{E}[W_p^p(\mu_n,\mu)])^{\frac{1}{p}} + \left(\mathbb{E}[W_p^p(\nu_n,\nu)]\right)^{\frac{1}{p}}.
    \end{align}
    Assume that $M_q(\mu) = \int_{\Re^d}\|x\|^q \diff \mu(x) < \infty$ for some $q > p$. From~\citet[Theorem 1]{fournier2015rate}, there exists a constant $C_{d,p,q}$ depending only on $p$, $d$, and $q$ such that, for all $n \geq 1$,
    \begin{align}
    &\mathbb{E}\left[ W_p^p(\mu_n, \mu) \right] 
    \leq C_{d,p,q} \, M_q(\mu)^{p/q} \nonumber\\&
    \times
    \begin{cases}
    n^{-1/2} + n^{-(q - p)/q},\quad & \text{if } p > d/2 \text{ and } q \neq 2p, \\
    n^{-1/2} \log(1 + n) + n^{-(q - p)/q}, \quad  & \text{if } p = d/2 \text{ and } q \neq 2p, \\
    n^{-p/d} + n^{-(q - p)/q},\quad  & \text{if } p \in (0, d/2) \text{ and } q \neq \frac{d}{d - p}.
    \end{cases}
    \end{align}
    Therefore, we have $\mathbb{E}[W_p(\mu_n,\mu)] = \mathcal{O}(n^{-1/d})$, which means that the error $\mathbb{E}[|W_p(\mu_n,\nu_n) -W_p(\mu,\nu) |] = \mathcal{O}(n^{-1/d})$ (assuming $n > m$). This statistical property is referred to as the sample complexity of the Wasserstein distance. The implication of the above result is that the approximation error is reduced very slowly in high dimensions when increasing the number of samples. This problem is referred to as the curse of dimensionality of the Wasserstein distance. Nevertheless, it is hard to increase $n$ to very large values due to the high computational complexity of the Wasserstein distance.
\end{remark}

In addition to sample complexity, the central limit theorem for estimation of Wasserstein is proved in~\citet{del2019central,del2024central}. We refer the reader to the recent book~\citep{chewi2025statistical} for a complete discussion about related statistical properties of the Wasserstein distance and optimal transport.

\section{One-dimensional Optimal Transport}
\label{sec:1DOT:chapter:foundations}

Since SOT is built on one-dimensional OT, we now review one-dimensional OT to explain why OT is nice in one dimension. We first discuss the Monge problem in one dimension.

\begin{proposition}[Monge problem between uniform one-dimensional discrete measures with the same sizes]
\label{proposition:1DMonge_discrete} Given one-dimensional discrete measures $\mu=\frac{1}{n}\sum_{i=1}^n\delta_{x_i}$ and $\nu = \frac{1}{m}\sum_{j=1}^m  \delta_{y_j}$, the Monge map $T$, with the ground metric $c(x,y)=h(x-y)$ where $h$ is a strictly convex function, can be obtained efficiently via sorting. In particular, let $\sigma_1:\intset{n}\to \intset{n}$ and $\sigma_2:\intset{n}\to \intset{n}$ be sorted permutations:
\begin{align}
\label{eq:sorting}
    x_{\sigma_1(1)} \leq \ldots \leq x_{\sigma_1(n)}, \quad y_{\sigma_2(1)} \leq \ldots \leq y_{\sigma_2(n)},
\end{align}
the optimal Monge map $T$ is defined by mapping:
\begin{align}
   T^\star (x_{\sigma_1(i)}) = y_{\sigma_2(i)} \quad \forall \, i = 1, \dots, n,
\end{align}
with the associated permutation $\sigma^\star(\sigma_1(i)) = \sigma_2(i)$ for all $i = 1, \dots, n.$
\begin{proof}
Assume $\sigma$ is a permutation that is not monotone with respect to the sorted orders of $\{x_1,\ldots,x_n\}$ and $\{y_1,\ldots,y_n\}$, then there exist indices $i < j$ such that:
$$
x_i < x_j \quad \text{but} \quad y_{\sigma(i)} > y_{\sigma(j)}.
$$
Define a new permutation $\sigma'$ by swapping the images of $i$ and $j$:
$$
\sigma'(i) := \sigma(j), \quad \sigma'(j) := \sigma(i), \quad \sigma'(k) := \sigma(k) \text{ for } k \neq i, j.
$$
We compare the cost difference:
\begin{align*}
\Delta &:= \big[h(x_i - y_{\sigma(i)}) + h(x_j - y_{\sigma(j)})\big] - \big[h(x_i - y_{\sigma'(i)}) + h(x_j - y_{\sigma'(j)})\big].
\end{align*}
Set
$$
A := x_i - y_{\sigma(i)}, \quad B := x_j - y_{\sigma(j)}, \quad A' := x_i - y_{\sigma'(i)}, \quad B' := x_j - y_{\sigma'(j)}.
$$
Because $x_i < x_j$ and $y_{\sigma(i)} > y_{\sigma(j)}$, we have:
$$
A < A' < B' < B.
$$
Moreover, the pairs $(A, B)$ and $(A', B')$ satisfy
$$
A + B = A' + B'.
$$
By the strict convexity of $h$, we have
$$
h(A) + h(B) \geq h(A') + h(B').
$$
In fact, the swap reduces the cost whenever $h$ is strictly convex and $A \neq A'$, so
$$
\Delta \geq 0.
$$
Therefore, swapping reduces or maintains the cost. By repeatedly applying such swaps, we obtain a monotone permutation $\sigma^\star$ that pairs sorted $\{x_1,\ldots,x_n\}$ with sorted $\{y_1,\ldots,y_n\}$, achieving minimal total cost.
\end{proof}
\end{proposition}

\begin{proposition}[Monge problem between one-dimensional continuous measures]
\label{proposition:1DMonge_continuous}
Given the ground metric $c(x,y)=h(x-y)$ where $h$ is a strictly convex function, and one-dimensional continuous probability measures $\mu \in \setP_c(\Re)$ and $\nu \in \setP_c(\Re)$ (with $\setP_c(\Re)= \{\mu \in \setP(\Re) \mid \exists x_0 \in \Re, \int_\Re c(x_0,x)\diff \mu(x)<\infty \}$), the Monge map $T$ has the following closed form:
\begin{align}
    T^\star = F_\nu^{-1} \circ F_{\mu},
\end{align}
where $F_\mu$ is the cumulative distribution function (CDF) of $\mu$, and $F_{\nu}^{-1}$ is the generalized inverse CDF or quantile function of the CDF of $\nu$, i.e., 
\begin{align}
F_\nu^{-1}(x) = \inf \{y \in \Re \mid x \leq F_\nu(y)\}.
\end{align}
    The proof is given in Proposition 1.2.4 in~\citet{bonnotte2013unidimensional}. We will defer the proof to later when discussing the Kantorovich problem. In fact, we only need $\mu$ to be continuous to have the closed form as proven in Theorem 2.5 in~\citet{santambrogio2015optimal}. 
\end{proposition}

Similar to the Monge problem, the Kantorovich problem can also be solved efficiently in one dimension.

\begin{remark}[Kantorovich problem between one-dimensional discrete measures]
\label{remark:1DKantorovich_discrete} Given one-dimensional discrete measures $\mu=\sum_{i=1}^n \alpha_i \delta_{x_i}$ and $\nu = \sum_{j=1}^m \beta_j \delta_{y_j}$, the Kantorovich problem, with the ground metric $c(x,y)=h(x-y)$ where $h$ is a strictly convex function, can be efficiently solved in $n+m$ operations by using the north-west corner~\citep{peyre2019computational} solution given the sorting permutations $\sigma_1$ and $\sigma_2$ of $\{x_1,\ldots,x_n\}$ and $\{y_1,\ldots,y_m\}$ as in~\eqref{eq:sorting}. The north-west corner algorithm is formally described as follows:
\begin{algorithmic}[1]
\STATE Initialize $i \gets 1$, $j \gets 1$, $a \gets \alpha_{\sigma_1(1)}$, $b \gets \beta_{\sigma_2(1)}$
\WHILE{$i \leq n$ \textbf{and} $j \leq m$}
    \STATE $t \gets \min(a, b)$
    \STATE $\pi_{ij} \gets t$, $a \gets a - t$, $b \gets b - t$
    \IF{$a = 0$}
        \STATE $i \gets i + 1$
        \IF{$i \leq n$}
            \STATE $a \gets \alpha_{\sigma_1(i)}$
        \ENDIF
    \ENDIF
    \IF{$b = 0$}
        \STATE $j \gets j + 1$
        \IF{$j \leq m$}
            \STATE $b \gets \beta_{\sigma_2(j)}$
        \ENDIF
    \ENDIF
\ENDWHILE
\RETURN $\pi^\star= \pi_{\sigma_1^{-1},\sigma_2^{-1}}$.
\end{algorithmic}

Here, $\pi_{\sigma_1^{-1},\sigma_2^{-1}}$ denotes the rearrangement of $\pi$ given the rows and columns of the submentions $\sigma_1$ and $\sigma_2$. The coupling from the algorithm has at most $n+m-1$ non-zero entries. In fact, we can also obtain the Kantorovich potentials by using a modified algorithm (Algorithm 1 in \citet{sejourne2022faster}):
\begin{algorithmic}[1]
\STATE Set $\pi, f, g \gets 0, 0, 0$
\STATE Set $g_1, a, b, i, j \gets c(x_1, y_1), \alpha_{\sigma_1(1)}, \beta_{\sigma_2(1)}, 1, 1$
\WHILE{$i < n$ \textbf{or} $j < m$}
  \IF{$(a \geq b$ \textbf{and} $i < n)$ \textbf{or} $(j = m)$}
    \STATE $\pi_{ij}, b \gets a, b - a$
    \STATE $i \gets i + 1$
    \STATE $f_i, a \gets c(x_i, y_j) - g_j, \alpha_{\sigma_1(i)}$
  \ELSIF{$(a > b$ \textbf{and} $j < m)$ \textbf{or} $(i = n)$}
    \STATE $\pi_{i,j}, a \gets b, a - b$
    \STATE $j \gets j + 1$
    \STATE $g_j, b \gets c(x_i, y_j) - f_i, \beta_{\sigma_2(j)}$
  \ENDIF
\ENDWHILE
\RETURN $\pi^\star= \pi_{\sigma_1^{-1},\sigma_2^{-1}}, \fb^\star = \fb_{\sigma_1^{-1}}, \gb^\star = \gb_{\sigma_2^{-1}}$.
\end{algorithmic}
\end{remark}

\begin{proposition}[Kantorovich problem between one-dimensional general probability measures]
\label{proposition:1DKantorovich_continous} 
Given one-dimensional  probability measures $\mu \in \setP_c(\Re)$ and $\nu  \in \setP_c(\Re)$, the Monge map $T$, with the ground metric $c(x,y)=h(x-y)$ for $h$ is a strictly convex function, has the following closed-form:
\begin{align}
    \pi^\star &= (F_\mu^{-1},F_\nu^{-1})\sharp \mathcal{U}([0,1]) \\
    &= (Id,F_\nu^{-1} \circ F_{\mu})\sharp \mu,
\end{align}
where the second inequality where the second equality follows from the change of variables $t = F_{\mu}(x)$ and requires $\mu$ to be continuous, and $Id$ is the identity mapping.
\begin{proof}
  We give a similar proof to~\citet[Lemma 2.8]{santambrogio2015optimal}. Let \(U \sim \mathcal{U}([0,1])\) be a uniform random variable on the interval \([0,1]\). We define
\begin{align}
X = F_\mu^{-1}(U), \quad Y = F_\nu^{-1}(U).
\end{align}
Since $F_\mu^{-1}$ and $F_\nu^{-1}$ are the quantile functions (generalized inverses of the CDFs) of $\mu$ and $\nu$, respectively, it follows that
\begin{align}
X \sim \mu, \quad Y \sim \nu,
\end{align}
so the joint distribution $\pi^\star$ of $(X,Y)$ is a coupling in $\Pi(\mu,\nu)$. We then compute the expected cost under $\pi^\star$:
\begin{align}
\int_{\mathbb{R} \times \mathbb{R}} h(x - y) \, \diff \pi^\star(x,y) = \mathbb{E}[h(X - Y)] = \int_0^1 h\big(F_\mu^{-1}(t) - F_\nu^{-1}(t)\big) \, \diff t.
\end{align}
Now, to show $\pi^\star$ minimizes the cost, we consider any coupling $\pi \in \Pi(\mu, \nu)$. We disintegrate $\pi$ with respect to its first marginal $\mu$ (Definition~\ref{def:disintegration_measures}):
\begin{align}
\diff \pi(x, y) = \diff \mu(x) \diff \pi_x(y).
\end{align}
Since $X \sim \mu$, the random variable $T = F_\mu(X)$ is uniform on $[0,1]$. Using the push-forward by $F_\mu$, we rewrite the integral as
\begin{align}
\int_{\mathbb{R} \times \mathbb{R}} h(x - y) \, \diff\pi(x,y) = \int_0^1 \left( \int_{\mathbb{R}} h\big(F_\mu^{-1}(t) - y\big) \, \diff \pi_{F_\mu^{-1}(t)}(y) \right) \diff t.
\end{align}
Applying Jensen's inequality to the strictly convex function $h$, we get for each $t \in [0,1]$:
\begin{align}
\int_{\mathbb{R}} h\big(F_\mu^{-1}(t) - y\big) \, \diff \pi_{F_\mu^{-1}(t)}(y) \geq h\left( F_\mu^{-1}(t) - \int_{\mathbb{R}} y \, \diff \pi_{F_\mu^{-1}(t)}(y) \right).
\end{align}

Note that the marginal of $\pi$ on the second coordinate is $\nu$, so integrating over $t \in [0,1]$. we have:
\begin{align}
\int_0^1 \int_{\mathbb{R}} y \, \diff \pi_{F_\mu^{-1}(t)}(y) \, dt = \int_{\mathbb{R}} y \, \diff \nu(y).
\end{align}

However, to minimize the mean $\int_{\mathbb{R}} h\big(F_\mu^{-1}(t) - y\big) \, \diff \pi_{F_\mu^{-1}(t)}(y)$ and to make the Jensen's inequality tight, it requires $ \pi_{F_\mu^{-1}(t)}$ to be a point mass, which the quantile function $F_\nu^{-1}$ achieves. Therefore, the integral satisfies
\begin{align}
\int_{\mathbb{R} \times \mathbb{R}} h(x - y) \, \diff \pi(x,y) \geq \int_0^1 h\big(F_\mu^{-1}(t) - F_\nu^{-1}(t)\big) \, \diff t,
\end{align}
and the quantile coupling $\pi^\star$ attains this minimal cost.

\end{proof}
\end{proposition}

\begin{figure}[!t]
    \centering

    \begin{tabular}{c}
         \includegraphics[width=1\linewidth]{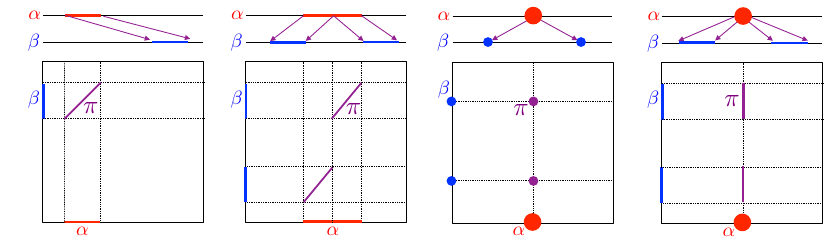} \\  
         \includegraphics[width=1\linewidth]{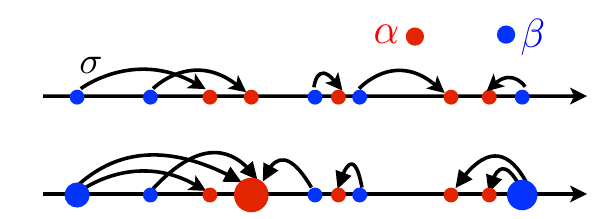}
    \end{tabular}
    \caption{The top figure (Figure 2.7 in~\citet{peyre2019computational}) shows optimal transport plans between 1-D distributions, represented as
maps. The bottom figure (Figure 2.9 in~\citet{peyre2019computational}) shows optimal transport plans for a pair of empirical measures with the same number of points (optimal matching) and a pair of generic discrete measures.} 
    \label{fig:1Dplans}
\end{figure}


The Kantorovich optimal transport plans in one-dimensional are illustrated in Figure~\ref{fig:1Dplans}~\citep{peyre2019computational}. From one-dimensional OT, we can discuss one-dimensional Wasserstein distance.

\begin{remark}[One-dimensional Wasserstein distance between discrete measures]
    \label{remark:1DWasserstein_discrete}
    Given two one-dimensional discrete measures $\mu =  \sum_{i=1}^n \alpha_i \delta_{x_i}$ and $\nu = \sum_{j=1}^m \beta_j \delta_{y_j}$, the Wasserstein$-p$ distance with the ground metric $c(x,y) = h(x-y)$ (where $h$ is a strictly convex function) is:
    \begin{align}
        W_{c,p}^p(\mu,\nu) = \sum_{i=1}^n \sum_{j=1}^m \pi_{ij}^{\star} c(x_i,y_j),
    \end{align}
    where $\pi^{\star}$ is the optimal plan from the north-west corner algorithm (Remark~\ref{remark:1DKantorovich_discrete}). 
\end{remark}

\begin{remark}[One-dimensional Wasserstein distance between general measures]
    \label{remark:1DWasserstein_general}
    Given a ground metric $c(x,y)=h(x-y)$ for a strictly convex function $h$, and two measures $\mu\in \setP_{c}(\Re)$ and $\nu\in \setP_{c}(\Re)$, the one-dimensional Wasserstein$-p$ ($p \geq 1$) distance between $\mu\in \setP_c(\Re)$ and $\nu\in \setP_c(\Re)$ is defined as:
    \begin{align}
\label{eq:1DWasserstein_general}
   W_{c,p}^p(\mu,\nu) &= \int_{0}^1 c(F_{\mu}^{-1}(t),F_{\nu}^{-1}(t))\diff t  \\
   &= \int_{\Re} c(x, F_{\nu}^{-1} \circ F_{\mu} (x)) \diff \mu(x),
\end{align}
where the second equality follows from the change of variables $t = F_{\mu}(x)$ and requires the existence of the OT map. This results directly from Proposition~\ref{proposition:1DKantorovich_continous}.

\end{remark}

\begin{figure}[!t]
    \centering

    \begin{tabular}{c}
         \includegraphics[width=1\linewidth]{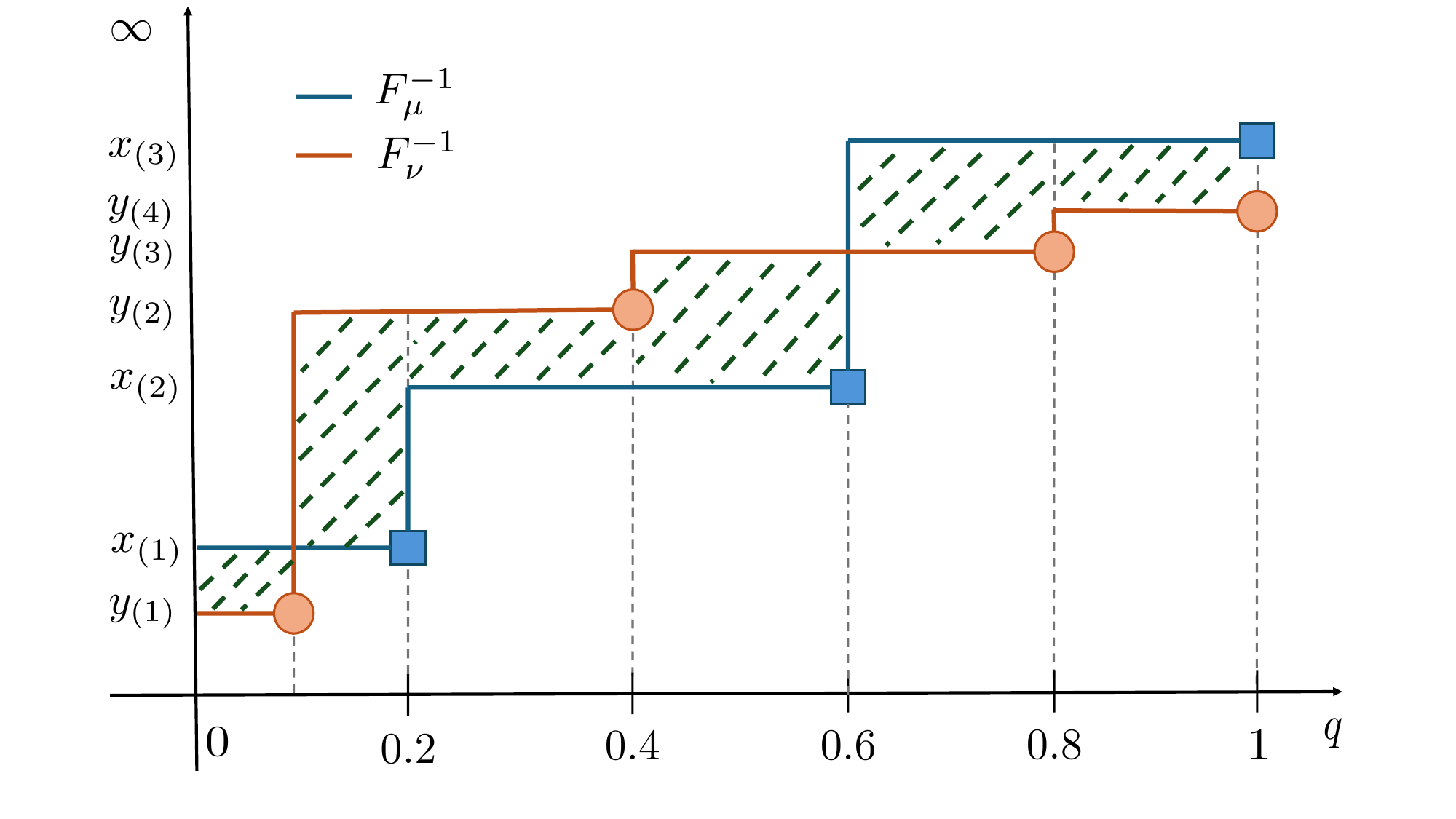}
    \end{tabular}
    \caption{One-dimensional Wasserstein distance between discrete
measures can be seen as one-dimensional Wasserstein distance with quantile approximation.} 
    \label{fig:empirical_quantile}
\end{figure}

\begin{remark}[One-dimensional Wasserstein distance between discrete measures as quantile approximation]
\label{remark:1DWasserstein_as_quantile_approximation}
When $\mu$ and $\nu$ are discrete distributions, i.e., $\mu = \sum_{i=1}^n \alpha_i \delta_{x_i}$ and $\nu = \sum_{j=1}^m \beta_j \delta_{y_j}$, the quantile functions of $\mu$ and $\nu$ can be written as follows:
\begin{align*}
    F_{\mu}^{-1}(t) &=  \sum_{i=1}^n x_{(i)} \, I\left(\sum_{j=1}^{i-1} \alpha_{(j)} < t \leq \sum_{j=1}^i \alpha_{(j)}\right), \\
    F_{\nu}^{-1}(t) &=  \sum_{j=1}^m y_{(j)} \, I\left(\sum_{i=1}^{j-1} \beta_{(i)} < t \leq \sum_{i=1}^j \beta_{(i)}\right),
\end{align*}
where $x_{(1)} \leq \ldots \leq x_{(n)}$ and $y_{(1)} \leq \ldots \leq y_{(m)}$ are the sorted supports (order statistics). Define the cumulative sums of weights:
\[
A_i = \sum_{j=1}^i \alpha_{(j)}, \quad B_j = \sum_{i=1}^j \beta_{(i)},
\]
with $A_0 = B_0 = 0$. The corresponding quantile functions $F_\mu^{-1}$ and $F_\nu^{-1}$ are right-continuous step functions given by:
\[
F_\mu^{-1}(t) = x_i \quad \text{for } t \in (A_{i-1}, A_i], \quad
F_\nu^{-1}(t) = y_j \quad \text{for } t \in (B_{j-1}, B_j].
\]
We construct a coupling $\pi$ by matching the quantile blocks over the intervals where the CDFs overlap. Specifically, define:
\[
\pi(x_i, y_j) = \ell_{ij} := \mathrm{length}\left([\max(A_{i-1}, B_{j-1}), \min(A_i, B_j)] \right),
\]
whenever the interval is non-empty. This $\ell_{ij}$ represents the common mass in the overlapping quantile blocks and corresponds to transporting $\ell_{ij}$ units of mass from $x_i$ to $y_j$. This transport plan is precisely the one produced by the north-west corner rule in the case where $x_i$ and $y_j$ are ordered. Hence, the quantile coupling and the north-west corner rule define the same transport plan. We visualize this in Figure~\ref{fig:empirical_quantile}.
\end{remark}

\begin{remark}[Computational complexities of one-dimensional Wasserstein distance]
    \label{remark:computaitonal_1DWasserstein} 
    For discrete cases, solving one-dimensional Wasserstein distance requires only $\mathcal{O}(n \log n)$ time complexity due to sorting the supports of measures and using the north-west corner algorithm. Moreover, the space complexity is only $\mathcal{O}(n)$ since there is no need to store the cost matrix as in general Wasserstein distance. For continuous cases, one-dimensional Wasserstein distance can be computed as the difference between quantile functions. When quantile functions are not tractable, approximations can be used. We will cover these approaches in detail later in Section~\ref{sec:quantile:chapter:advances}.
\end{remark}

\begin{remark}[Sample complexity of one-dimensional Wasserstein distance]
    \label{remark:sample_complexity_1DWasserstein}
    From Remark~\ref{remark:sample_complexity_Wasserstein}, when $d=1$, we assume that $M_q(\mu) = \int_{\Re^d} |x|^q \diff \mu(x) < \infty$ and $M_q(\nu) = \int_{\Re^d} |y|^q \diff \nu(y) < \infty$ for some $q > p$, there exists a constant $C_{p,q}$ such that:
    \begin{align}
       & \mathbb{E}[|W_p(\mu_n,\nu_n) - W_p(\mu,\nu)|]\leq C_{p,q}^{1/p} \, \bigl(M_q(\mu)^{1/q} + M_q(\nu)^{1/q}\bigr) \nonumber \\
        & 
        \times
        \begin{cases}
            n^{-1/(2p)}, & \text{if } q > 2p, \\
            n^{-1/(2p)} \log(1 + n)^{1/p}, & \text{if } q = 2p, \\
            n^{-(q - p)/(p q)}, & \text{if } q \in (p, 2p).
        \end{cases}
    \end{align} 
\end{remark}

From Remark~\ref{remark:computaitonal_1DWasserstein} and Remark~\ref{remark:sample_complexity_1DWasserstein}, we see that a large number of samples $n$ can be used for one-dimensional Wasserstein distance to reduce the statistical error thanks to its low time complexity and low memory complexity.

\section{Radon Transform and Integral Geometry}
\label{sec:Radontransform:chapter:foundations}

To exploit the benefits of one-dimensional OT in higher dimensions, we need to utilize tools from integral geometry to transform original measures to new spaces where OT can be done efficiently via one-dimensional OT. The conventional transform is the Radon Transform.

\begin{definition}[Radon Transform]
    \label{def:Radon_Transform} Given an integrable function $f:\Re^d \to \Re$, the Radon transform (RT)~\citep{radon27bestimmung,helgason1999radon} is an operator that maps $f$ to a function on the product of the unit hypersphere in $d$ dimensions and the real line ($\Sm^{d-1} \times \Re$), which is defined as:
    \begin{align}
        \setR f(\theta,t) = \int_{\Re^d} f(x)\, \delta(t-\langle x,\theta \rangle) \, dx,
    \end{align}
    where $\langle x,\theta\rangle = \theta^\top x$ is the inner product, $\theta \in \Sm^{d-1}$ is referred to as the direction, and $t \in \Re$. We denote the conditional function of $t$ given $\theta$ as:
    \begin{align}
        \setR_\theta f(t) = \setR f(\theta,t).
    \end{align}
\end{definition}
The RT gives you the integral of the function restricted to a particular hyperplane, parametrized
by  $\theta$ and $t$. To define the inverse of RT, we need to define the Fourier Transform and discuss its injectivity.

\begin{definition}[Fourier Transform]
    \label{def:Fourier_Transform} Given an integrable function $f:\Re^d \to \Cm$, the Fourier transform (FT)~\citep{pinsky2023introduction} is an operator that maps $f$ to a function $\setF f : \Re^d \to \Cm$ defined by:
    \begin{align}
        \setF f (\xi) = \int_{\Re^d} e^{-2\pi i \langle x,\xi \rangle} f(x) \, \diff x,
    \end{align}
    where $\xi \in \Re^d$.
\end{definition}

\begin{remark}[Bijectivity of Fourier Transform]
    \label{remark:bijectivity_Fourier} The Fourier transform is a bijection~\citep{pinsky2023introduction}. The inverse Fourier transform is defined as:
    \begin{align}
        \setF^{-1} g(x) = \int_{\Re^d} e^{2\pi i \langle x,\xi \rangle} g(\xi) \, \diff \xi.
    \end{align} 
\end{remark}

\begin{proposition}[Fourier slice theorem (central slice theorem)]
    \label{proposition:Fourier_slice_theorem} The Fourier transform of the initial function along a line in the direction $\theta$ is the one-dimensional Fourier transform of the Radon transform (acquired at direction $\theta$) of that function:
    \begin{align}
        \mathcal{F} f(t\theta) = \mathcal{F}\mathcal{R}_\theta f(t).
    \end{align}
    The proof is simply based on a change of variables.
    \begin{proof} 
    From the definition of FT, we have:
        \begin{align*}
    \mathcal{F} f(t\theta) 
    &= \int_{\mathbb{R}^d} e^{-2\pi i \langle x, t\theta \rangle} f(x)  \diff x \\
    &= \int_{\mathbb{R}^d} e^{-2\pi i t \langle x, \theta \rangle} f(x) \diff x \quad \text{(linearity of inner product)} \\
    &= \int_{\mathbb{R}^d} \int_{\Re} e^{-2\pi i t s}  f(x) \delta(s-\langle x,\theta \rangle) \diff s \diff x \quad \text{(change of variables)} \\
    &=  \int_{\Re} e^{-2\pi i t s} \int_{\mathbb{R}^d} f(x) \delta(s-\langle x,\theta \rangle) \diff x  \diff s \quad \text{(Fubini's theorem)} \\
    &= \int_\Re e^{-2\pi i t s} \setR_\theta f(s)  \diff s \\
    &= \mathcal{F}\mathcal{R}_\theta f(t),
\end{align*}
which completes the proof.
    \end{proof}
    
\end{proposition}

With the Fourier slice theorem, we can invert the Radon transform.

\begin{remark}[Bijectivity of Radon Transform]
    \label{remark:bijectivity_Radon} The Radon transform is a linear bijection~\citep{natterer2001mathematics,helgason2011radon}. The inverse Radon transform is defined as:
    \begin{align}
        f(x) = \setR^{-1} (\setR f(\theta,t)) = \int_{\Sm^{d-1}} (\setR f (\theta, \langle x,\theta \rangle ) * \eta( \langle x,\theta \rangle)) \, d\theta,
    \end{align} 
    where $\eta(\cdot)$ is a one-dimensional high-pass filter with corresponding Fourier transform $\setF \eta(\omega) = c |\omega|^{d-1}$, where $c=(4\pi)^{(d-1)/2} \frac{\Gamma(d/2)}{\Gamma(1/2)}$, and $*$ denotes the convolution operator.
\end{remark}

 The above formula is also known as the filtered back-projection formula. There are also iterative reconstruction methods for the Radon transform. Above definitions of Fourier transform and Radon transform are for functions. We now extend the Fourier transform and Radon transform to measures.

\begin{definition}[Fourier transform of measures]
    \label{def:Fourier_Transform_measures} Given a measure $\mu \in \setM(\Re^d)$, the Fourier transform of $\mu$~\citep{pinsky2023introduction} is the function $\setF \mu$ defined as:
    \begin{align}
       \setF \mu(\xi) = \int_{\Re^d} e^{-2\pi i \langle x,\xi\rangle} \diff \mu(x),
    \end{align}
    where $\setF \mu \in \setC(\Re^d)$ is a continuous function.
\end{definition}

\begin{definition}[Radon transform of measures]
    \label{def:Radon_Transform_measures} Given a measure $\mu \in \setM(\Re^d)$, the Radon transform (RT) measure~\citep{helgason2011radon} $\setR \mu \in \setM(\Sm^{d-1} \times \Re)$ is defined by:
    \begin{align}
    \label{eq:RT_measure}
        \int_{\Sm^{d-1} \times \Re} f(\theta,t) \diff \setR \mu(\theta,t) = \int_{\Re^d} \int_{\Sm^{d-1}} f(\theta, \langle x,\theta \rangle) \diff \theta \, \diff \mu(x),
    \end{align} 
    for any integrable function $f$ on $\Sm^{d-1} \times \Re$. We denote $\setR_\theta \mu(t)$ as the Radon conditional measure i.e., disintegrating $\diff  \setR \mu(\theta,t) = \diff \theta \diff \setR_\theta\mu(t) $. For brevity, we denote $\setR_\theta \mu$ as $\theta \sharp \mu$.
\end{definition}

\begin{definition}[Inverse Radon transform of measures]
    \label{def:Inverse_Radon_Transform_measures} Given a measure $\nu \in \setM(\Sm^{d-1} \times \Re)$, the inverse RT measure $\mu = \setR^{-1} \nu$ is defined by:
    \begin{align}
        \int_{\Re^d} f(x) \diff \mu(x) = \int_{\Sm^{d-1} \times \Re} \setR (\setR^* \setR)^{-1} f(\theta,t) \diff \nu(\theta,t),
    \end{align}
    where $\setR^* g(x) = \int_{\Sm^{d-1}} g(\theta, \langle \theta, x \rangle) \diff \theta$ is the back-projection operator.
\end{definition}

\begin{figure}[!t]
    \centering
    \includegraphics[width=0.8\linewidth]{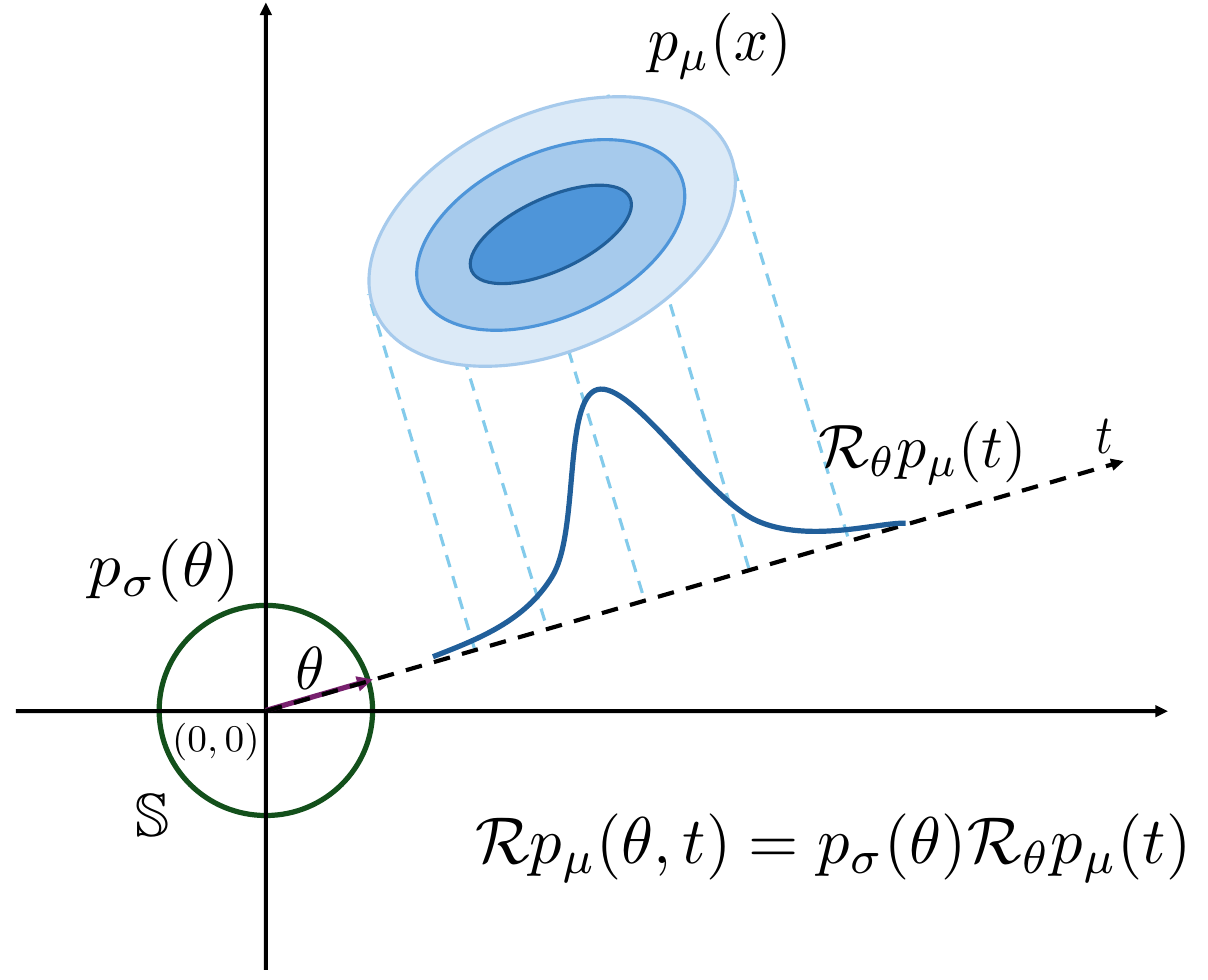}
    \caption{Radon Transform of density function $p_\mu(x)$ in two dimensions with $\sigma(\theta) = \mathcal{U}(\Sm)$ and $p_\sigma(\theta)$ is the density function of $\sigma(\theta)$ .}
    \label{fig:RT}
\end{figure}

\begin{remark}[Connection between Radon Transform of measures and Radon Transform of functions]
    \label{remark:connection_Radon_function_measure} Given a measure $\mu \in \setM(\Re^d)$ with density $p_\mu(x)$, we have:
    \begin{align*}
        &\int_{\Re^d}\int_{\Sm^{d-1}} f(\theta, \langle x,\theta \rangle) \diff \theta \diff \mu(x) 
        = \int_{\Re^d}\int_{\Sm^{d-1}} f(\theta, \langle x,\theta \rangle) \diff \theta \, p_\mu(x) \diff x \\
        &= \int_{\Sm^{d-1}} \int_{\Re^d} f(\theta, \langle x,\theta \rangle) p_\mu(x) \diff x \diff \theta \\
        &= \int_{\Sm^{d-1}} \int_{\Re^d} \int_\Re f(\theta, t) p_\mu(x) \delta(t-\langle x,\theta \rangle) \diff t \diff x \diff \theta \\
        &= \int_{\Sm^{d-1}} \int_\Re f(\theta, t) \int_{\Re^d} p_\mu(x) \delta(t-\langle x,\theta \rangle) \diff x \diff t \diff \theta \\
        &= \int_{\Sm^{d-1}} \int_\Re f(\theta, t) \setR p_\mu(\theta,t) \diff t \diff \theta,
    \end{align*}
    which means that 
    \begin{align}
        \int_{\Sm^{d-1} \times \Re} f(\theta,t) \diff \setR \mu(\theta,t) = \int_{\Sm^{d-1}} \int_\Re f(\theta, t) \setR p_\mu(\theta,t) \diff t \diff \theta,
    \end{align}
    implying that $\setR p_\mu(\theta,t)$ is the density of $\setR \mu(\theta,t)$.
\end{remark}

We illustrate the Radon transform of the density function of $\mu$, i.e., $p_\mu(x)$, in Figure~\ref{fig:RT}. Another key property that will be used frequently later is the connection between the Radon conditional measure and the linear push-forward measure.

\begin{proposition}[Radon conditional measure is a linear push-forward measure]
    \label{proposition:Radon_pushforward_conditional} Given a measure $\mu \in \setM(\Re^d)$ and its RT measure $\setR \mu \in \setM(\Sm^{d-1} \times \Re)$, using the disintegration theorem~\citep{getoor1980claude}, the measure $\setR \mu$ can be decomposed into its conditional measure ($\theta \sharp \mu \in \setM(\Re)$) with respect to the uniform measure on $\Sm^{d-1}$ for almost all $\theta \in \Sm^{d-1}$ outside a set of zero measure such that for any function $f$:
    \begin{align}
    \label{eq:Radon_pushforward_conditional_1}
       \int_{\Sm^{d-1} \times \Re} f(\theta,t) \diff \setR \mu(\theta,t) = \int_{\Sm^{d-1}} \int_{\Re} f(\theta,t) \diff \theta \sharp \mu(t) \diff \theta.
    \end{align}
    \begin{proof}
    As discussed in~\citet{bonneel2015sliced}, from~\eqref{eq:RT_measure}, we have:
    \begin{align}
    \label{eq:Radon_pushforward_conditional_2}
         \int_{\Sm^{d-1} \times \Re} f(\theta,t) \diff \setR \mu(\theta,t) &= \int_{\Re^d} \int_{\Sm^{d-1}} f(\theta, \langle x,\theta \rangle) \diff \theta \diff \mu(x) \nonumber \\
        &= \int_{\Sm^{d-1}} \int_{\Re^d} f(\theta, \langle x,\theta \rangle) \diff \mu(x) \diff \theta \nonumber \\
        &= \int_{\Sm^{d-1}} \int_{\Re} f(\theta, t) \diff g_\theta \sharp \mu(t) \diff \theta,
    \end{align}
    where $g_\theta(x) = \langle \theta,x \rangle$.
    Combining~\eqref{eq:Radon_pushforward_conditional_1} and~\eqref{eq:Radon_pushforward_conditional_2}, we have $\theta \sharp \mu = g_\theta \sharp \mu$. When $\mu = \sum_{i=1}^n \alpha_i \delta_{x_i}$ is a discrete measure, we have $\theta \sharp \mu = \sum_{i=1}^n \alpha_i \delta_{\langle \theta,x_i \rangle}$.
    \end{proof}
\end{proposition}

\section{Sliced Wasserstein Distance}
\label{sec:SOTdistances:chapter:foundations}

With all the materials discussed in the previous section, we are able to define the sliced Wasserstein distance, which is the probability metric form of SOT.

\begin{figure}[!t]
    \centering
    \includegraphics[width=1\linewidth]{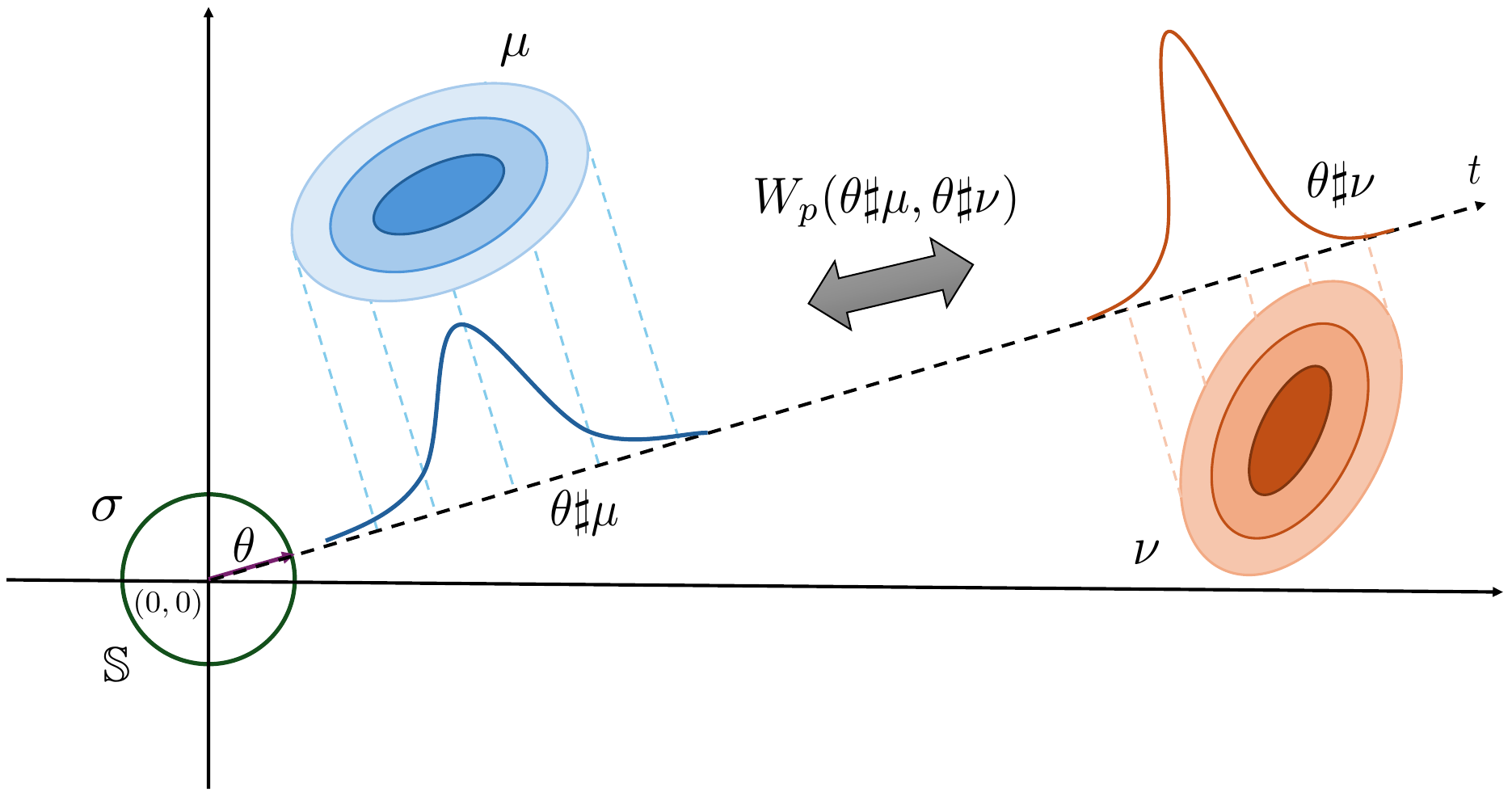}
    \caption{One-dimensional Wasserstein distance between Radon conditional measures.}
    \label{fig:SWdef}
\end{figure}

\begin{definition}[Sliced Wasserstein distance between general measures]
    \label{def:SW} Given $p \geq 1$, the sliced Wasserstein-$p$~\citep{rabin2011wasserstein,bonneel2015sliced} (SW) distance between two measures $\mu \in \setP_p(\Re^d)$ and $\nu \in \setP_p(\Re^d)$ is defined as:
    \begin{align}
        \label{eq:SW} 
        SW_p^p(\mu,\nu) = \mathbb{E}_{\theta \sim \setU(\Sm^{d-1})}[W_p^p(\theta \sharp \mu, \theta \sharp \nu)],
    \end{align}
    where $\setU(\Sm^{d-1})$ denotes the uniform distribution over the unit hypersphere $\Sm^{d-1}$, and $\theta \sharp \mu$ and $\theta \sharp \nu$ denote the Radon conditional measures of $\mu$ and $\nu$, respectively. 
\end{definition}

\begin{figure}[!t]
    \centering
    \includegraphics[width=1\linewidth]{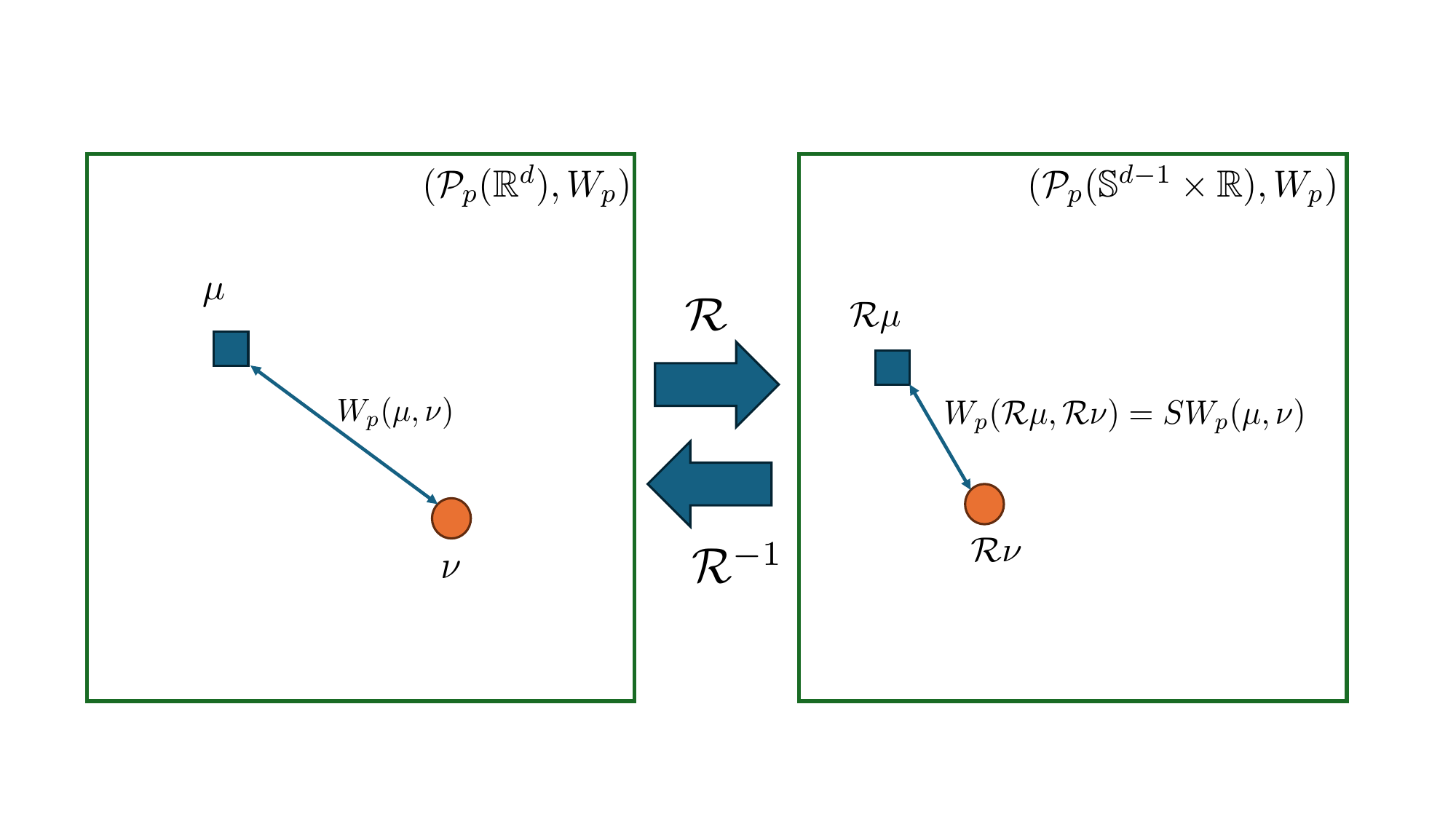}
    \caption{Sliced Wasserstein distance is the Wasserstein distance on Radon spaces.}
    \label{fig:SW}
\end{figure}

We can also use $W_{c,p}^p$ with a ground metric $c: \Re \times \Re \to \Re_+$ to replace $W_p^p$. In that case, we obtain $SW_{c,p}^p$. As discussed in Proposition~\ref{proposition:Radon_pushforward_conditional}, $\theta \sharp \mu$ and $\theta \sharp \nu$ are also the push-forward measures of $\mu$ and $\nu$ through the function $g_\theta(x) = \langle \theta,x \rangle$. We give a visualization of the one-dimensional Wasserstein distance between Radon conditional measures in Figure~\ref{fig:SWdef}. We now show that the SW distance is the Wasserstein distance between transformed measures using RT in Proposition~\ref{proposition:SW_is_W_Radon} (an illustration is given in Figure~\ref{fig:SW}).

\begin{proposition}[Sliced Wasserstein distance is the Wasserstein distance between Radon transform measures]
    \label{proposition:SW_is_W_Radon} Given $\mu \in \setP_p(\Re^d)$ and $\nu \in \setP_p(\Re^d)$, we have:
    \begin{align}
        W_p^p(\setR \mu,\setR \nu) = SW_p^p(\mu,\nu),
    \end{align}
    where $\setR \mu$ is the Radon transform of $\mu$ (Definition~\ref{def:Radon_Transform_measures}).

    \begin{proof}
        We first prove the following lemma.

        \begin{lemma}
            Let $\mu$ and $\nu$ be probability measures on the product space $\mathcal{X} \times \mathcal{Y}$, and suppose they have the same marginal on $\mathcal{X}$, denoted by $\sigma$, i.e.,
            \begin{align}
            \mu(A \times \mathcal{Y}) = \nu(A \times \mathcal{Y}) = \sigma(A), \quad \forall A \subset \mathcal{X}.
            \end{align}
            $\mu$ and $\nu$ admit the following disintegrations (Definitions~\ref{def:disintegration_measures}):
            \begin{align}
            \diff \mu(x, y) = \diff \mu_x(y) \diff \sigma(x), \quad \diff \nu(x, y) = \diff \nu_x(y) \diff \sigma(x),
            \end{align}
            where $\mu_x$ and $\nu_x$ are the conditional measures on $\mathcal{Y}$ given $x \in \mathcal{X}$. Then for any $p \geq 1$, the Wasserstein distance between $\mu$ and $\nu$ with ground metric $c$ satisfies
            \begin{align}
                W_c^p(\mu, \nu) = \int_{\mathcal{X}} W_{c_{\mathcal{Y}}}^p(\mu_x, \nu_x) \diff \sigma(x) = \mathbb{E}_{x \sim \sigma}[W_{c_{\mathcal{Y}}}^p(\mu_x, \nu_x)],
            \end{align}
            where $c_{\mathcal{Y}}^p(y,y') = c^p\big((x,y), (x,y')\big)$ for $x \in \mathcal{X}$ and $y,y' \in \mathcal{Y}$.
        \end{lemma}

        \begin{proof}
            For each $x \in \mathcal{X}$, let $\pi_x \in \Pi(\mu_x, \nu_x)$ be an optimal coupling between the conditional measures $\mu_x$ and $\nu_x$. Define the joint coupling $\pi$ on $\mathcal{X} \times \mathcal{X} \times \mathcal{Y} \times \mathcal{Y}$ by
            \begin{align}
            \diff \pi(x, x', y, y') := \diff \sigma(x) \diff \delta_x(x') \diff \pi_x(y, y').
            \end{align}
            We see that $\pi$ is a coupling between $\mu$ and $\nu$ (with shared $x$), since its marginals are $\mu$ and $\nu$ respectively:
            \begin{align*}
            \int_{\mathcal{X} \times \mathcal{Y}} f(x, y) \diff \mu(x, y) = \int_{\mathcal{X} \times \mathcal{X} \times \mathcal{Y} \times \mathcal{Y}} f(x, y) \diff \sigma(x) \diff \delta_x(x') \diff \pi_x(y, y'),
            \end{align*}
            and
            \begin{align*}
            \int_{\mathcal{X} \times \mathcal{Y}} f(x', y') \diff \nu(x', y') = \int_{\mathcal{X} \times \mathcal{X} \times \mathcal{Y} \times \mathcal{Y}} f(x', y') \diff \sigma(x) \diff \delta_x(x') \diff \pi_x(y, y').
            \end{align*}
            The cost of the coupling $\pi$ is
            \begin{align*}
            \int_{\mathcal{X} \times \mathcal{X} \times \mathcal{Y} \times \mathcal{Y}} c_{\mathcal{Y}}^p(y, y') \diff \pi_x(y, y') \diff \sigma(x) = \int_{\mathcal{X}} W_{c_{\mathcal{Y}}}^p(\mu_x, \nu_x) \diff \sigma(x).
            \end{align*}
            Therefore,
            \begin{align*}
            W_c^p(\mu, \nu) \leq \int_{\mathcal{X}} W_{c_{\mathcal{Y}}}^p(\mu_x, \nu_x) \diff \sigma(x).
            \end{align*}

            To show equality, suppose $\gamma$ is any coupling between $\mu$ and $\nu$. Since both have marginal $\sigma$ on $\mathcal{X}$, disintegrate
            \begin{align*}            
            \diff \gamma(x, x', y, y') = \diff \gamma_x(y, y') \diff \sigma(x) \diff \delta_x(x'),
            \end{align*}
            where $\gamma_x \in \Pi(\mu_x, \nu_x)$. The expected cost under $\gamma$ is
            \begin{align*}
            \int_{\mathcal{X}} \left( \int c_{\mathcal{Y}}^p(y, y') \diff \gamma_x(y, y') \right) \diff \sigma(x) \geq \int_{\mathcal{X}} W_p^p(\mu_x, \nu_x) \diff \sigma(x),
            \end{align*}
            with equality if and only if $\gamma_x$ is optimal for each $x$. Thus, the coupling $\pi$ constructed from optimal conditionals achieves the minimum cost, and we conclude:
            \begin{align}
            W_c^p(\mu, \nu) = \int_{\mathcal{X}} W_{c_{\mathcal{Y}}}^p(\mu_x, \nu_x) \diff \sigma(x).
            \end{align}
        \end{proof}

        Returning to the proof of the proposition, we have $d \setR \mu(t, \theta) = \diff \theta \sharp \mu(t) \diff \sigma(\theta)$ and $d \setR \nu(t, \theta) = \diff \theta \sharp \nu(t) \diff \sigma(\theta)$ (Proposition~\ref{proposition:Radon_pushforward_conditional}), where $\sigma(\theta)$ is the uniform measure over $\Sm^{d-1}$. Applying the lemma, we get:
        \begin{align}
            W_p^p(\setR \mu, \setR \nu) &= \int_{\Sm^{d-1}} W_p^p(\theta \sharp \mu, \theta \sharp \nu) \diff \sigma(\theta) \nonumber \\
            &= \mathbb{E}_{\theta \sim \mathcal{U}(\Sm^{d-1})}[W_p^p(\theta \sharp \mu, \theta \sharp \nu)] = SW_p^p(\mu, \nu),
        \end{align}
        which concludes the proof.
    \end{proof}
\end{proposition}
We now discuss some properties of the SW distance, including metricity, weak convergence, sample complexity, and its equivalence to the Wasserstein distance.

\begin{proposition}[SW distance is a valid metric]
    \label{proposition:SW_metricity} 
    For $p \geq 1$, the sliced Wasserstein distance ($SW_{p}$) is a metric on $\mathcal{P}_{p}(\Re^d)$. We follow the proofs in~\citet{bonnotte2013unidimensional}.
    \begin{proof}
    We prove that the SW distance satisfies the four metric properties:

    1. \textbf{Non-negativity:} For any $\mu, \nu \in \mathcal{P}_{p}(\Re^d)$,
    \[
    SW_{p}^p(\mu, \nu) = \mathbb{E}_{\theta \sim \setU(\Sm^{d-1})}[W_p^p(\theta \sharp \mu, \theta \sharp \nu)] \geq 0,
    \]
    since $W_p^p(\theta \sharp \mu, \theta \sharp \nu) \geq 0$ due to the nonnegativity of the Wasserstein distance.

    2. \textbf{Identity of indiscernibles:} Suppose $SW_{p}(\mu,\nu) = 0$. This means
    \[
    \mathbb{E}_{\theta \sim \setU(\Sm^{d-1})}[W_p^p(\theta \sharp \mu, \theta \sharp \nu)] = 0.
    \]
    Since $W_p^p(\theta \sharp \mu, \theta \sharp \nu) \geq 0$, this implies
    $$
    W_p^p(\theta \sharp \mu, \theta \sharp \nu) = 0,
    $$
    for $\setU(\Sm^{d-1})$-almost every $\theta \in \Sm^{d-1}$. From the identity of indiscernibles property of the Wasserstein distance, it follows that
    $$
    \theta \sharp \mu = \theta \sharp \nu,
    $$
    for $\setU(\Sm^{d-1})$-almost every $\theta \in \Sm^{d-1}$. Applying the central slice theorem (Proposition~\ref{proposition:Fourier_slice_theorem}), we have:
    \begin{align*}
       \mathcal{F}[\mu](t\theta) &= \int_{\mathbb{R}^d} e^{-it\langle \theta,x \rangle} d\mu(x) =  \int_{\mathbb{R}} e^{-itz} d (\theta \sharp \mu)(z) = \mathcal{F}[\theta \sharp \mu](t) \\
       &= \mathcal{F}[\theta \sharp \nu](t) = \int_{\mathbb{R}} e^{-itz} d (\theta \sharp \nu)(z) = \int_{\mathbb{R}^d} e^{-it\langle \theta,x \rangle} d\nu(x) = \mathcal{F}[\nu](t\theta), 
    \end{align*}
    for $\setU(\Sm^{d-1})$-almost every $\theta \in \Sm^{d-1}$. Since the Fourier transform is injective, we get $\mu = \nu$.

    Conversely, when $\mu = \nu$, we have
    $$
    SW_{p}^p(\mu, \nu) = \mathbb{E}_{\theta \sim \setU(\Sm^{d-1})}[W_p^p(\theta \sharp \mu, \theta \sharp \nu)] = \mathbb{E}_{\theta \sim \setU(\Sm^{d-1})}[0] = 0,
    $$
    due to the identity of indiscernibles property of the Wasserstein distance.

    3. \textbf{Symmetry:} From the symmetry of the Wasserstein distance, $W_p(\theta \sharp \mu,\theta \sharp \nu) = W_p(\theta \sharp \nu,\theta \sharp \mu)$, we have
    \[
    \mathbb{E}_{\theta \sim \setU(\Sm^{d-1})}[W_p^p(\theta \sharp \mu, \theta \sharp \nu)] = \mathbb{E}_{\theta \sim \setU(\Sm^{d-1})}[W_p^p(\theta \sharp \nu, \theta \sharp \mu)],
    \]
    which implies $SW_p(\mu,\nu) = SW_p(\nu,\mu)$.

    4. \textbf{Triangle inequality:} For any $\mu,\nu,\lambda \in \mathcal{P}_{p}(\Re^d)$, we have:
    \begin{align*}
        SW_p(\mu,\nu) &= \left( \mathbb{E}_{\theta \sim \setU(\Sm^{d-1})}[W_p^p(\theta \sharp \mu, \theta \sharp \nu)] \right)^{\frac{1}{p}} \\
        &\leq \left( \mathbb{E}_{\theta \sim \setU(\Sm^{d-1})} \big( W_p(\theta \sharp \mu, \theta \sharp \lambda) + W_p(\theta \sharp \lambda, \theta \sharp \nu) \big)^p \right)^{\frac{1}{p}} \\
        &\leq \left( \mathbb{E}_{\theta \sim \setU(\Sm^{d-1})}[W_p^p(\theta \sharp \mu, \theta \sharp \lambda)] \right)^{\frac{1}{p}} + \left( \mathbb{E}_{\theta \sim \setU(\Sm^{d-1})}[W_p^p(\theta \sharp \lambda, \theta \sharp \nu)] \right)^{\frac{1}{p}} \\
        &= SW_p(\mu,\lambda) + SW_p(\lambda,\nu),
    \end{align*}
    where the first inequality is due to the triangle inequality of the Wasserstein distance and the second inequality follows from Minkowski's inequality.
    \end{proof}
\end{proposition}

\begin{remark}[Weak convergence under Sliced Wasserstein distance]
    \label{remark:weak_convergence_sw}
    A sequence of measures $\mu_k$ converges weakly to $\mu$ in $\mathcal{P}(\Re^d)$ if and only if the sliced Wasserstein distance, i.e., $SW_p(\mu_k, \mu) \to 0$~\citep[Theorem 1]{nadjahi2020statistical}.
\end{remark}
\begin{figure}[!t]
    \centering
    \includegraphics[width=1\linewidth]{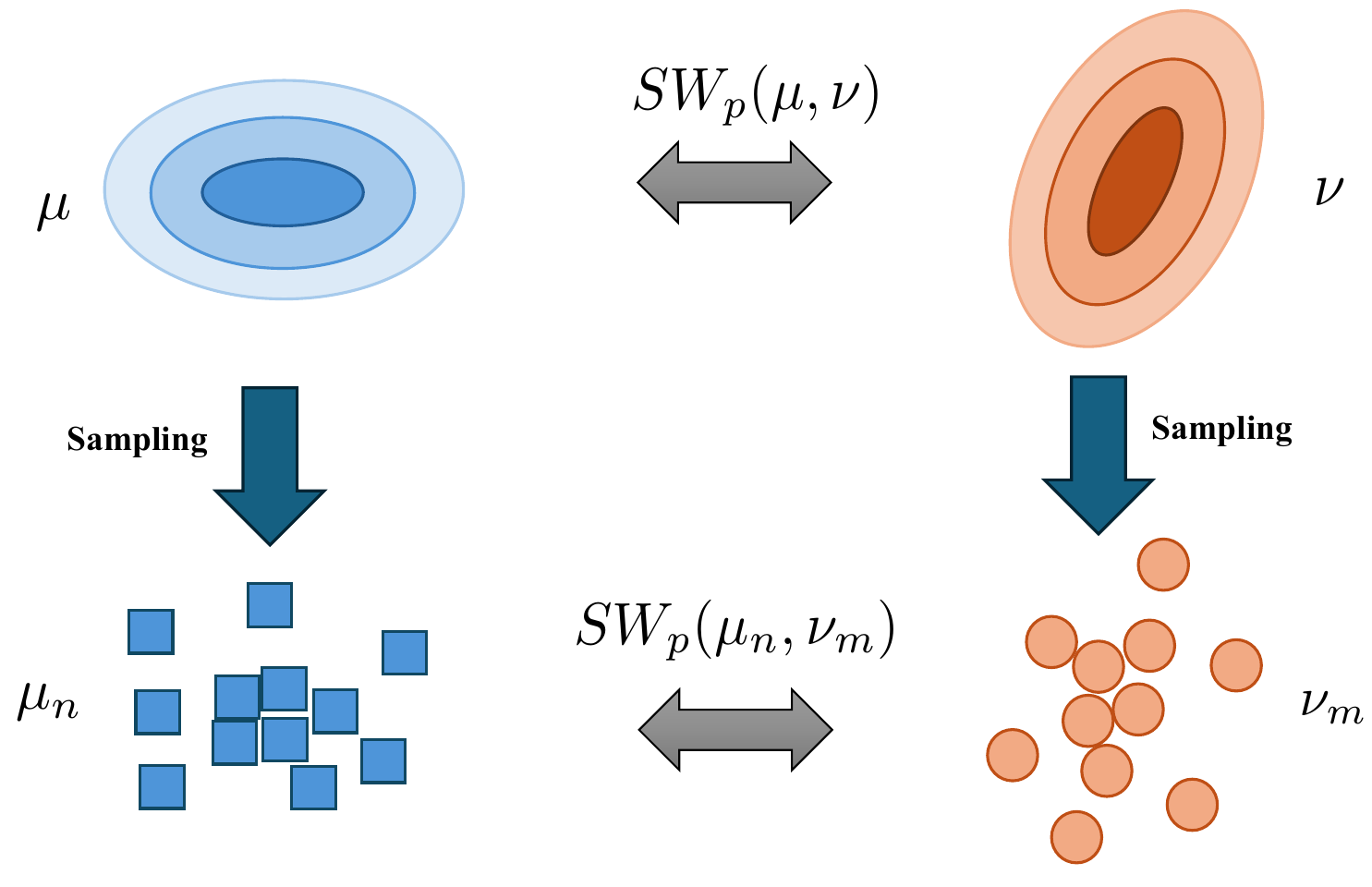}
    \caption{Sliced Wasserstein distance between empirical probability measures.}
    \label{fig:SW_empirical}
\end{figure}

\begin{remark}[Sample complexity of sliced Wasserstein distance]
    \label{remark:sample_complexity_SW} 
    Given $x_1,\ldots,x_n \sim \mu \in \setP_p(\Re^d)$, $y_1,\ldots,y_m \sim \nu \in \setP_p(\Re^d)$, and empirical measures $\mu_n$ and $\nu_m$ with $n > m$, assume that $M_q(\mu) = \int_{\Re^d} \|x\|^q \, d\mu(x) < \infty$ and $M_q(\nu) = \int_{\Re^d} \|y\|^q \, d\nu(y) < \infty$ for some $q > p$. Then, there exists a constant $C_{p,q}$ such that:
    \begin{align}
        &\mathbb{E}[|SW_p(\mu_n,\nu_m) - SW_p(\mu,\nu)| ]  \nonumber \\
        & \leq C_{p,q}^{1/p} \, \bigl(M_q(\mu)^{1/q} + M_q(\nu)^{1/q}\bigr)
        \times
        \begin{cases}
            n^{-1/(2p)}, & \text{if } q > 2p, \\
            n^{-1/(2p)} \log(1 + n)^{1/p}, & \text{if } q = 2p, \\
            n^{-(q - p)/(p q)}, & \text{if } q \in (p, 2p).
        \end{cases}
    \end{align} 
    This follows from Remark~\ref{remark:sample_complexity_1DWasserstein} together with the fact that the moment of the projected measure $\theta \sharp \mu$ is smaller than that of the original measure $\mu$ due to the Cauchy–Schwarz inequality. We refer the reader to~\citet[Corollary 2]{nadjahi2020statistical} for a detailed proof. By restricting measures to $\setP_p(\setX)$ for a compact set $\setX$, the sample complexity can be improved~\citep{nadjahi2020statistical,nguyen2021distributional}. We visualize the idea of empirical approximation of SW in Figure~\ref{fig:SW_empirical}.
\end{remark}

In addition to sample complexity, distributional limits for estimations of SW (e.g., central limit theorem) have been investigated in~\citet{xi2022distributional,goldfeld2024statistical,hundrieser2024empirical,rodriguez2025improved}.

\begin{remark}[Connection between SW and Wasserstein distance]
\label{remark:SW_connect_Wasserstein}
For $\mu, \nu \in \setP_p(\Re^d)$, we have:
\begin{align}
    SW_p(\mu,\nu) \leq \left(\frac{1}{d} \mathbb{E}_{\theta \sim \setU(\Sm^{d-1})}[\|\theta\|_p^p] \right)^{\frac{1}{p}} W_p(\mu,\nu).
\end{align}
The proof can be found in~\citet[Proposition 5.1.3]{bonnotte2013unidimensional}. For $\mu,\nu \in \setP_p(B(0, R))$ with $B(0,R) = \{x \in \Re^d \mid \|x\|_p \leq R \}$, there exists a constant $C_{d,p} > 0$ such that:
\begin{align}
    W_p(\mu,\nu) \leq \left( C_{d,p} R^{p - \frac{1}{d+1}} SW_p(\mu,\nu)^{\frac{1}{d+1}} \right)^{\frac{1}{p}}.
\end{align}
We refer the reader to Theorem 5.1.5 in~\citet{bonnotte2013unidimensional}. From this connection, we see that $SW_p$ and $W_p$ are strongly equivalent when the measures are supported on $B(0,R)$. For measures supported on all of $\Re^d$, only the inequality $SW_p \leq C W_p$ holds; they are not strongly equivalent~\citep{seale2024sliced}.
\end{remark}

We now discuss how SW is computed in practice. Since SW in often intractable for general measures due to the intractable expectation, Monte Carlo estimation is often used.

\begin{definition}[Monte Carlo estimation of sliced Wasserstein distance]
    \label{def:MC_SW}
    The SW distance is usually intractable due to the expectation in Definition~\ref{def:SW}. Therefore, Monte Carlo estimation is often used to approximate:
    \begin{align}
        \widehat{SW}_p^p(\mu,\nu;\theta_1,\ldots,\theta_L) = \frac{1}{L} \sum_{l=1}^L W_p^p(\theta_l \sharp \mu, \theta_l \sharp \nu),
    \end{align}
    where $\theta_1,\ldots,\theta_L \stackrel{i.i.d.}{\sim} \setU(\Sm^{d-1})$ with $L \geq 1$ is the number of Monte Carlo samples, often referred to as the ``number of projections".
\end{definition}
Sampling $\theta \sim \setU(\Sm^{d-1})$ can be done  in practice by setting $\theta = \frac{z}{\|z\|_2}$ with $z \sim \mathcal{N}(0,I)$.  We then discuss the estimation error from Monte Carlo estimation.
\begin{remark}[Monte Carlo estimation error]
    \label{remark:MC_SW}
The Monte Carlo estimation error is bounded as follows:
            \begin{align}
        &\mathbb{E} \left| \widehat{SW}_p^p(\mu,\nu;\theta_1,\ldots,\theta_L) - SW_p^p(\mu,\nu) \right| \nonumber \\
        &\leq \left(\mathbb{E} \left| \widehat{SW}_p^p(\mu,\nu;\theta_1,\ldots,\theta_L) - SW_p^p(\mu,\nu) \right|^2 \right)^{\frac{1}{2}} \nonumber\\
        &= \left(\mathbb{E} \left| \frac{1}{L} \sum_{l=1}^L W_p^p(\theta_l \sharp \mu, \theta_l \sharp \nu) - \mathbb{E}_{\theta \sim \setU(\Sm^{d-1})} \left[ W_p^p(\theta \sharp \mu, \theta \sharp \nu) \right] \right|^2 \right)^{\frac{1}{2}} \nonumber\\
        &= \left( \operatorname{Var}_{\theta \sim \setU(\Sm^{d-1})} \left[ \frac{1}{L} \sum_{l=1}^L W_p^p(\theta_l \sharp \mu, \theta_l \sharp \nu) \right] \right)^{\frac{1}{2}} \nonumber\\
        &= \frac{1}{\sqrt{L}} \operatorname{Var}_{\theta \sim \setU(\Sm^{d-1})} \left[ W_p^p(\theta \sharp \mu, \theta \sharp \nu) \right]^{\frac{1}{2}},
    \end{align}
    where the first inequality is due to Hölder’s inequality, and the last two equalities are due to the i.i.d. property of $\theta_1,\ldots,\theta_L$. One can further bound the variance term by constraining the measures $\mu$ and $\nu$ to have log-concave densities~\citep{nietert2022statistical}. Overall, the Monte Carlo approximation error satisfies
    \begin{align}
    \mathbb{E} \left| \widehat{SW}_p^p(\mu,\nu;\theta_1,\ldots,\theta_L) - SW_p^p(\mu,\nu) \right| = \mathcal{O}(L^{-1/2}).
    \end{align}
\end{remark}

\begin{remark}[Overall complexity]
    \label{remark:overall_complexity_sw}
    Using the triangle inequality, we have:
    \begin{align}
        &\mathbb{E} \big[ |\widehat{SW}_p^p(\mu_n,\nu_m) - SW_p^p(\mu,\nu)| \big] \nonumber \\
        & \leq \mathbb{E} \big[ |\widehat{SW}_p^p(\mu_n,\nu_m) - SW_p^p(\mu_n,\nu_m)| \big]  \nonumber \\
        & \quad + \mathbb{E} \big[ |SW_p^p(\mu_n,\nu_m) - SW_p^p(\mu,\nu)| \big],
    \end{align}
    where the first term corresponds to the Monte Carlo estimation error (Remark~\ref{remark:MC_SW}) and the second term corresponds to the sample complexity (Remark~\ref{remark:sample_complexity_SW}, with the $p$-th power). Combining these results yields the overall estimation complexity.
\end{remark}

\begin{remark}[Computational complexity of SW distance]
    \label{remark:computational_complexity_sw}
    Suppose $\mu$ and $\nu$ are discrete measures each supported on at most $n$ points. The computation of SW distance involves three main steps:
    \begin{enumerate}
        \item Sampling directions $\theta_1, \ldots, \theta_L$,
        \item Computing the projections $\theta_l \sharp \mu$ and $\theta_l \sharp \nu$ for $l=1,\ldots,L$,
        \item Solving the one-dimensional Wasserstein distance $W_p(\theta_l \sharp \mu, \theta_l \sharp \nu)$ for each $l$.
    \end{enumerate}
    The complexities are:
    \begin{itemize}
        \item Sampling directions: time $\mathcal{O}(L d)$, space $\mathcal{O}(L d)$,
        \item Computing projections: time $\mathcal{O}(L d n)$, space $\mathcal{O}(L n)$,
        \item Computing 1D Wasserstein distances: time $\mathcal{O}(L n \log n)$, space $\mathcal{O}(L n)$.
    \end{itemize}
    Overall, the total time complexity is
    $
        \mathcal{O}\bigl(L (d n + n \log n)\bigr),
    $
    and the space complexity is
    $
        \mathcal{O}(L (d + n)).
    $
\end{remark}

From Remarks~\ref{remark:sample_complexity_SW} and~\ref{remark:computational_complexity_sw}, we observe that sliced Wasserstein distance not only avoids the curse of dimensionality but is also scalable in the number of samples $n$.

\section{Iterative Distribution Transfer and Knothe's transport}
\label{sec:IDT_Konthe:chapter:foundations}

In this section, we discuss the transportation aspect of SOT. In particular, we discuss how to transport from one measure to another measure with SOT. In contrast to OT, where the optimal transport plan (map) between two measures is found, SOT does not provide such a plan (map). However, SOT can still provide some notions of transportation "map" to transport measures. We start with the iterative distribution transfer algorithm.

\begin{definition}[Iterative distribution transfer]
    \label{def:IDT}  The iterative distribution transfer (IDT) algorithm~\citep{pitie2007automated} starts from a measure $\mu$ and builds a sequence $\mu^{(t)}$ that seems to tend to a target measure $\nu$ when $t \to \infty$.  The algorithm can be stated as follows. For each time step $t$, we first choose an orthogonal basis $\Theta_t = (\theta_{t1}, \ldots, \theta_{td})$ (e.g., sampling uniformly from the Stiefel manifold $\mathbb{V}_d(\Re^d) = \{\Theta \in \Re^{d\times d} \mid \Theta^\top \Theta = I\}$)~\citep{bendokat2024grassmann}. For each $\theta_{ti}$ for $i=1, \ldots, d$, we denote $T_{\theta_{ti}}:\Re \to \Re$ as the unidimensional optimal mapping between $\theta_{ti} \sharp \mu^{(t)}$ and $\theta_{ti} \sharp \nu$ (Proposition~\ref{proposition:1DMonge_continuous}). We then set
    \begin{align}
        \mu^{(t+1)} =  T_t \sharp \mu^{(t)}, \quad T_t(x) = \sum_{i=1}^d T_{\theta_{ti}}(\langle \theta_{ti}, x \rangle) \theta_{ti},
    \end{align}
    where we have $\theta_{ti} \sharp \mu^{(t+1)} = \theta_{ti} \sharp T_t \sharp \mu^{(t)} = (\langle \theta_{ti}, \cdot \rangle \circ T_t) \sharp \mu^{(t)} = \theta_{ti} \sharp \nu$ due to the orthogonality constraint. Thus, the measure $\mu^{(t+1)}$ seems to move closer to the measure $\nu$.
\end{definition}

\begin{remark}[Convergence of IDT for Gaussian measures]
    \label{remark:convergence_IDT_Gaussian} Let $\mu$ be a continuous measure and $\nu$ be a Gaussian measure, then:
    \begin{enumerate}
        \item If $\Theta_1, \ldots, \Theta_t \simiid \setU(\mathbb{V}_d(\Re^d))$, then $\mu_t \to \nu$ almost surely.
        \item If the sequence $\Theta_1, \ldots, \Theta_t$ is dense, then $\mu_t \to \nu$ weakly.
    \end{enumerate}
    We refer the reader to~\citet{pitie2007automated} for the original proof and to~\citet[Theorem 5.2.2]{bonnotte2013unidimensional} for a revised proof.
\end{remark}

For empirical measures, we can also create an implicit map using gradient flow with SW as functional energy.

\begin{figure}[!t]
    \centering
    \includegraphics[width=1\linewidth]{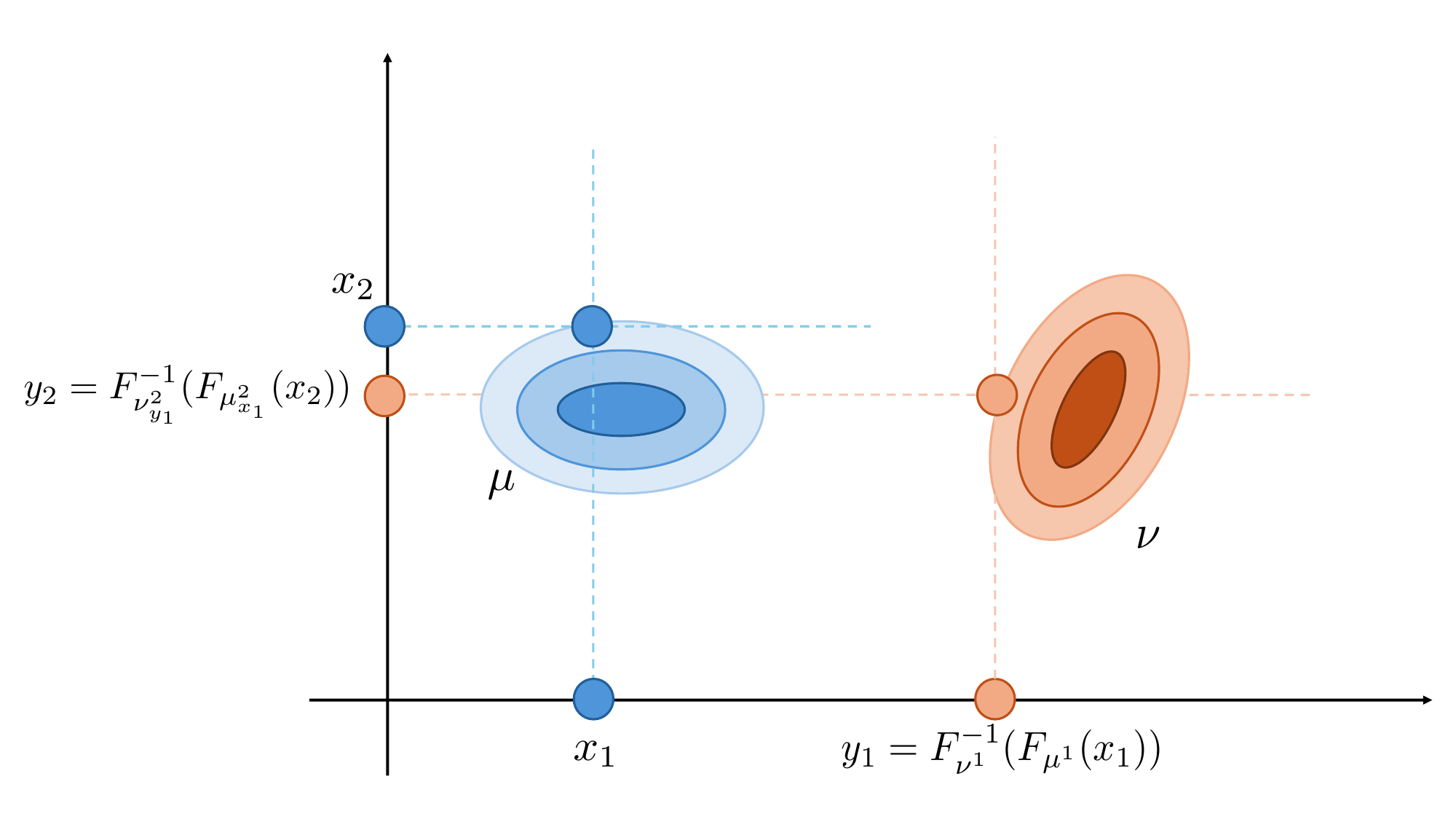}
    \caption{Knothe's transport map in two dimensions.}
    \label{fig:Knothe}
\end{figure}

\begin{remark}[Gradient flows with sliced Wasserstein as functional energy]
    \label{remark:gf_SW_energy} Given two empirical measures $\mu^{(0)} = \frac{1}{n} \sum_{i=1}^n \delta_{x_i(0)}$ and $\nu = \frac{1}{m} \sum_{j=1}^m \delta_{y_j}$, we can implicitly define the transformation map by the following flow:
    \begin{align}
        \frac{\diff (x_1(t), \ldots, x_n(t))}{\diff t} = SW_p^p(\mu^{(t)}, \nu),
    \end{align}
    where $\mu^{(t)} = \frac{1}{n} \sum_{i=1}^n \delta_{x_i(t)}$. Using discretization schemes such as the Euler scheme, we can define the following mapping:
    \begin{align}
        &(x_1(t+1), \ldots, x_n(t+1)) \nonumber \\
        &\quad = (x_1(t), \ldots, x_n(t)) - \nabla_{(x_1(t), \ldots, x_n(t))} SW_p^p(\mu^{(t)}, \nu).
    \end{align}
    This kind of update is used in~\citet{bonneel2015sliced, rabin2011wasserstein} for graphic tasks such as color transfer (grading). We refer the reader to Section~\ref{sec:differentiatingSW:chapter:varitational_SW} for a detailed derivation of the gradient.
\end{remark}

However, the flow does not give an optimal map and might not converge to the target measure even in the case of $n = m$~\citep{cozzi2025long}. We refer the reader to~\citet{vauthiertowards, cozzi2025long} for a more detailed discussion. Finally, we discuss a way to create a valid transportation map between two measures using disintegration of measures.

\begin{definition}[The Knothe–Rosenblatt rearrangement]
    \label{def:Knothe}
From~\citet{rosenblatt1952remarks,knothe1957contributions} Let $\mu$ and $\nu$ be continuous measures on $\Re^d$, by disintegration of measures, $\mu$ and $\nu$ can be disintegrated according to the axes. In particular, there exist families $\{\mu^1,\ldots,\mu^d\}$ and $\{\nu^1,\ldots,\nu^d\}$ such that:
    \begin{align}
        &\int_{\Re^d} f(x) \diff \mu(x) = \int_{\Re} \int_{\Re} \ldots \int_{\Re} f(x) \diff \mu^d_{x_1,\ldots,x_{d-1}}(x_d) \ldots \diff \mu^2_{x_1}(x_2) \diff \mu^1(x_1), \nonumber \\
        &\int_{\Re^d} f(y) \diff \nu(y) = \int_{\Re} \int_{\Re} \ldots \int_{\Re} f(y) \diff \nu^d_{y_1,\ldots,y_{d-1}}(y_d) \ldots \diff \nu^2_{y_1}(y_2) \diff \nu^1(y_1),
    \end{align}
    for any continuous function $f$. We now define:
    \begin{align}
    &T^1(x_1) = F_{\nu^1}^{-1} \circ F_{\mu^1}(x_1), \quad T^2(x_1,x_2) = F_{\nu^2_{T^1(x_1)}}^{-1} \circ F_{\mu^2_{x_1}}(x_2), \nonumber \\
    &T^i(x_1,\ldots,x_i) = F_{\nu^i_{T^{i-1}(x_1,\ldots,x_{i-1})}}^{-1} \circ F_{\mu^i_{x_1,\ldots,x_{i-1}}}(x_i), \quad \forall i=3,\ldots,d.
    \end{align}
    The Knothe's transport map is defined as:
    \begin{align}
        T(x) = (T^1, \ldots, T^d),
    \end{align}
    for $x = (x_1,\ldots,x_d)$. 
\end{definition}

We can show that $T \sharp \mu = \nu$. We have:
    \begin{align*}
        &\int_{\Re^d} f(T(x)) \diff \mu(x) \\ 
        &= \int_{\Re} \int_{\Re} \ldots \int_{\Re} f(T^1(x_1), \ldots, T^d(x_1, \ldots, x_d))  \diff \mu^d_{x_1, \ldots, x_{d-1}}(x_d)   \\ &\quad \quad \ldots \diff \mu^2_{x_1}(x_2) \diff \mu^1(x_1) \\
        &= \int_{\Re} \int_{\Re} \ldots \int_{\Re} f(T^1(x_1), \ldots, y_d) \diff \nu^d_{T^{d-1}(x_1, \ldots, x_{d-1})}(y_d) \ldots \diff \mu^2_{x_1}(x_2) \diff \mu^1(x_1) \\
        &\ldots \\
        &= \int_{\Re} \int_{\Re} \ldots \int_{\Re} f(y_1, \ldots, y_d) \diff \nu^d_{y_1, \ldots, y_{d-1}}(y_d) \ldots \diff \nu^2_{y_1}(y_2) \diff \nu^1(y_1) \\
        &= \int_{\Re^d} f(y) \diff \nu(y),
    \end{align*}
    which completes the proof. We show a visualization of Knothe transport in two dimensions in Figure~\ref{fig:Knothe}. The Knothe map offers a useful alternative by decomposing the transport into a sequence of one-dimensional conditional maps. This triangular structure not only simplifies calculations but also provides insight into how mass is redistributed along each coordinate.  We end this section here. We will further discuss other recent ways to construct transport maps from one-dimensional OT in Section~\ref{sec:map:chapter:advances}.

\chapter{Advances in Sliced Optimal Transport}
\label{chapter:advances}

From Chapter~\ref{chapter:foundations}, we know that Sliced Optimal Transport (SOT) is defined based on the Radon transform and one-dimensional optimal transport. Computationally, it relies on simple Monte Carlo estimation to evaluate a finite number of one-dimensional OT problems. In this section, we discuss recent advances in SOT, including the use of domain-specific integral transforms (i.e., generalized Radon transforms), more advanced Monte Carlo methods, generalized approaches to evaluating one-dimensional OT, the use of weighted Radon transforms to control the contributions of individual one-dimensional OT problems, and new techniques for extracting transport plans.

\section{Generalized Radon Transform on Non-Euclidean spaces}
\label{sec:generalized_slicing:chapter:advances}

Since the Sliced Wasserstein (SW) distance is essentially the Wasserstein distance computed over a transformed version of the original space, the choice of transformation operator plays a crucial role in preserving the intrinsic structure of the data. Although finding an ``optimal'' transformation is non-trivial, improvements over the classical Radon transform are possible, as it is limited to Euclidean space. In this section, we discuss generalized Radon transforms for non-Euclidean spaces such as manifolds of non-positive curvature, spheres, compact manifolds, images, functions, probability measures, and products of spaces.

\subsection{Non-linear Projections}
\label{subsec:non_linear_projections}

In contrast to the Wasserstein distance, where the ground metric \(c\) can be chosen flexibly, the one-dimensional Wasserstein distance admits a closed-form solution only when the cost takes the form \(c(x, y) = h(x - y)\), with \(h\) being a strictly convex function. Given two measures \(\mu\) and \(\nu\), the one-dimensional projected Wasserstein distance between Radon conditional measures (or linearly projected measures) can be expressed as~\citep[Lemma 6]{paty2019subspace}:
\begin{align}
    W_c^p(\theta\sharp \mu,\theta \sharp \nu) = \inf_{\pi \in \Pi(\mu,\nu)} \int_{\mathbb{R}^d \times \mathbb{R}^d} c(\langle \theta,x\rangle,\langle \theta,y\rangle) \, \mathrm{d} \pi(x,y),
\end{align}
which corresponds to the Wasserstein distance between \(\mu\) and \(\nu\) with a directional ground cost \(c_\theta(x,y) = c(\langle \theta,x\rangle,\langle \theta,y\rangle)\). Consequently, we can approximate the Wasserstein distance between \(\mu\) and \(\nu\) induced by a target cost \(c'\)  
\begin{align}
     W_{c'}^p( \mu,\nu) = \inf_{\pi \in \Pi(\mu,\nu)} \int_{\mathbb{R}^d \times \mathbb{R}^d} c'(x,y) \, \mathrm{d} \pi(x,y),
\end{align}
by representing \(c'\) in the form of \(c_\theta\). Since \(c\) is restricted to functions that admit closed-form solutions in one dimension, the expressive power of \(c_\theta\) can be enhanced using nonlinear projections; that is, by defining 
\begin{align}
    c_\theta(x, y) := c(g(\theta, x), g(\theta, y))
\end{align}
for some nonlinear mapping \(g\). Considering all possible \(\theta\), this use of \(g(\theta,x)\) leads to a generalized version of the Radon transform, known as the generalized Radon transform.

The generalized Radon transform (GRT) extends the Radon transform, i.e., from integration over hyperplanes to integration over $(d-1)$-dimensional manifolds~\citep{beylkin1984inversion,beylkin1983inversion,ehrenpreis2003universality,gel1969differential,kuchment2006generalized,homan2017injectivity}. To define the GRT, we first need to define the defining function.

\begin{definition}[Defining function]
    \label{def:defining_functions} A function $g:\setX \times (\Re^{d'} \backslash \{0\})$ (with $\setX \subset \Re^d$) is a
defining function when it satisfies the four conditions below:
\begin{enumerate}
    \item $g$ is real-valued and infinitely differentiable.
    \item $g(x,\theta)$ is homogeneous of degree one in $\theta$, i.e., $$g(x,\lambda\theta) = \lambda g(x,\theta)\, \forall\, \lambda \in \Re.$$
    \item  $\forall (x,\theta) \in \setX \times (\Re^{d'} \backslash \{0\}), \quad \frac{\partial}{\partial x} g(x,\theta) \neq 0.$
    \item The mixed Hessian of $g$ is strictly positive, i.e., $$\det \left( \left(\frac{\partial^2 g }{\partial x_i \partial \theta_j }\right)_{ij}\right) > 0.$$
\end{enumerate}
\end{definition}

\begin{figure}[!t]
    \centering
    \includegraphics[width=1\linewidth]{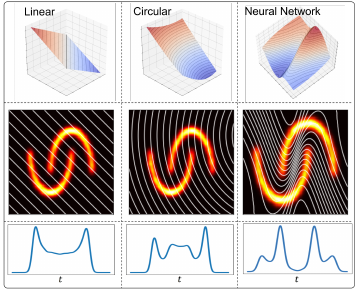}
    \caption{Generalized Radon Transform with some defining functions (Figure 1 in~\citet{kolouri2019generalized}).}
    \label{fig:GRT}
\end{figure}

Now, we can define the GRT and discuss related properties.

\begin{definition}[Generalized Radon Transform] 
\label{def:GRT}
    Given a defining function $g$, the generalized Radon transform (GRT) of an integrable $f:\Re^d \to \Re$ is defined as:
    \begin{align}
        \setGR f (t,\theta) =  \int_{\Re^d}  f(x) \delta(t-g(x,\theta)) \diff x,
    \end{align}
    where $t\in \Re$ and $\theta \in \Omega \subset (\Re^{d'} \backslash \{0\})$ for $d'>0$. The GRT admits the RT as a special case with $g(x,\theta)= \langle x,\theta\rangle$ and $\Omega =\Sm^{d-1}$.
\end{definition}

\begin{remark}[Injectivity of generalized Radon Transform]
    \label{remark:injectivity_GRT}
    The injectivity of GRTs is a long-standing
topic~\citep{beylkin1984inversion,homan2017injectivity,uhlmann2003inside,homan2017injectivity}.  In particular, a GRT is injective if $\setGR f(t,\theta) =  \setGR g(t,\theta)\, \forall (t,\theta) \in \Re \times \Omega $  implies $f=g$. There are some known cases where the GRT is injective, e.g., the circular function~\citep{kuchment2006generalized}, i.e., 
\begin{align}
    g(x,\theta)=\|x-r\theta\|_2,
\end{align}
for $r\in \mathbb{R}^+$ and $\theta \in \Omega=\mathbb{S}^{d-1}$, homogeneous polynomials  with an odd degree ($m$)~\citep{rouviere2015nonlinear}, i.e., 
\begin{align}
    g(x,\theta)=\sum_{|\alpha|=m}\theta_\alpha x^\alpha,
\end{align} 
with $\alpha=(\alpha_1,\ldots,\alpha_{d_\alpha})\in \mathbb{N}^{d_\alpha}$, $|\alpha|=\sum_{i=1}^{d_\alpha }\alpha_i$, $x^\alpha=\prod_{i=1}^{d_\alpha} x_i^{\alpha_i}$, and $\Omega = \mathbb{S}^{d_\alpha}$. A practical way to construct an injective GRT is to use the following defining function~\citep{chen2022augmented}:
\begin{align}
    g(x,\theta) = \langle h(x),\theta \rangle,
\end{align}
where $h:\Re^d \to \Re^{d'}$ is an injective function, e.g., a normalizing flow~\citep{papamakarios2021normalizing}, augmented functions $h(x)=(x,h'(x))$ for any function $h'$.
\end{remark}

\begin{remark}[Neural network defining functions]
    \label{def:nn_defining_function}
    We can use a neural network as $g(x,\theta)$~\citep{kolouri2019generalized}, where $\theta$ is the weight of the neural network. Under some specific network architectures, one can show that the corresponding defining function satisfies all requirements. However, it is non-trivial to show the injectivity of the associated transform. 
\end{remark}

An illustration is given in Figure~\ref{fig:GRT}, which is taken from Figure 1 in~\citet{kolouri2019generalized}. Similar to the RT, we can also extend the GRT to measures and discuss related properties.

\begin{definition}[Generalized Radon Transform of measures]
    \label{def:Generalized_Radon_Transform_measures} Given a measure $\mu \in \setM(\Re^d)$, the GRT measure $\setGR \mu \in \setM( \Re \times \Omega )$ is defined as follows:
    \begin{align}
    \label{eq:GRT_measure}
        \int_{ \Re \times \Omega } f(t,\theta) \diff\setGR \mu(\theta,t) = \int_{\Re^d} \int_{\Omega} f(g(x,\theta),\theta) \diff \theta \diff \mu(x),
    \end{align} 
    for any function $f\in\setC( \Re \times \Omega )$. We denote $\setGR_\theta \mu(t) = \setGR\mu(\theta,t)$ as the generalized Radon conditional measure.
\end{definition}

\begin{remark}[Connection between generalized Radon Transform of measures and generalized Radon Transform of functions]
    \label{remark:connection_generalized_Radon_function_measure} Given a measure $\mu \in \setM(\Re^d)$ with density $p_\mu(x)$, we have:
    \begin{align*}
        &\int_{\Re^d}\int_{\Omega} f(g(x,\theta),\theta) \diff \theta \diff \mu(x) = \int_{\Re^d}\int_{\Omega} f(g(x,\theta),\theta) \diff \theta \, p_\mu(x) \diff x  \\
        &= \int_{\Omega} \int_{\Re^d} f(g(x,\theta),\theta) p_\mu(x) \diff x \diff \theta \\
        &= \int_{\Omega} \int_{\Re^d} \int_\Re f(t,\theta) p_\mu(x) \delta(t-g(x,\theta)) \diff t \diff x \diff \theta \\
        &= \int_{\Omega} \int_\Re f(t,\theta) \int_{\Re^d} p_\mu(x) \delta(t-g(x,\theta)) \diff x \diff t \diff \theta \\
        &= \int_{\Omega} \int_\Re f(t,\theta) \setGR p_\mu(t,\theta) \diff t \diff \theta,
    \end{align*}
    which means that 
    \begin{align}
        \int_{ \Re \times \Omega } f(t,\theta) \diff\setGR \mu(t,\theta) = \int_{\Omega} \int_\Re f(t,\theta) \setGR p_\mu(t,\theta) \diff t \diff \theta,
    \end{align}
    implying that $\setGR p_\mu(t,\theta)$ is the density of $\setGR \mu(t,\theta)$.
\end{remark}

\begin{remark}[Generalized Radon push-forward measure is the generalized Radon conditional measure]
    \label{remark:generalized Radon_pushfoward_conditional}  
    Given a measure $\mu \in \setM(\Re^d)$ and its GRT measure $\setGR \mu \in \setM(\Re\times \Omega)$, using the disintegration theorem~\citep{getoor1980claude}, a measure $\setGR \mu$ can be decomposed into its conditional measure ($\setGR\theta\sharp\mu \in \setM(\Re)$) with respect to the uniform measure on $\Omega$ for almost all $\theta \in \Omega$ outside a set of zero measure such that for any  function$f$:
    \begin{align}
    \label{eq:generalized_Radon_pushfoward_conditional_1}
       \int_{\Re \times \Omega} f(t,\theta) \diff \setGR\mu(t,\theta) = \int_{\Omega} \int_{\Re} f(t,\theta) \diff \setGR\theta \sharp\mu(t) \diff \theta.
    \end{align}
    We have:
    \begin{align}
    \label{eq:generalized_Radon_pushfoward_conditional_2}
         \int_{\Re \times \Omega} f(t,\theta) \diff \setGR \mu(t,\theta) &= \int_{\Re^d} \int_{\Omega} f(g(x,\theta),\theta) \diff \theta \diff \mu(x) \nonumber \\
        &= \int_{\Omega} \int_{\Re^d} f(g(x,\theta),\theta) \diff \mu(x) \diff \theta \nonumber\\
        &= \int_{\Omega} \int_{\Re} f(t,\theta) \diff g_\theta \sharp \mu(t) \diff \theta,
    \end{align}
    where $g_\theta(x)=g(x,\theta)$. From~\eqref{eq:generalized_Radon_pushfoward_conditional_1} and~\eqref{eq:generalized_Radon_pushfoward_conditional_2}, we have $\setGR\theta\sharp\mu=g_\theta \sharp \mu$. We later call $g_\theta$ the projection function, which maps the original measure to one dimension. When $\mu=\sum_{i=1}^n\alpha_i \delta_{x_i}$ is a discrete measure, we have $g_\theta\sharp \mu =  \sum_{i=1}^n\alpha_i \delta_{g(x_i,\theta)}.$ 
\end{remark}

With the GRT, the generalized SW distance can be defined as follows:

\begin{definition}[Generalized Sliced Wasserstein distances]
    \label{def:GSW}  
    For $p\geq 1$ and a defining function $g:\Re^d \times \Omega \to \Re$, a generalized sliced Wasserstein-$p$~\citep{kolouri2019generalized} (GSW) distance between two measures $\mu \in \setP_p(\Re^d)$ and $\nu \in \setP_p(\Re^d)$ is:
    \begin{align}
        \label{eq:GSW} 
        GSW_p^p(\mu,\nu) &= \mathbb{E}_{\theta \sim \setU(\Omega)}[W_p^p(g_\theta \sharp \mu, g_\theta \sharp \nu)], 
    \end{align}
    where $\setU(\Omega)$ denotes the uniform distribution on the space of projection parameters $\Omega$, and $g_\theta \sharp \mu$ and $g_\theta \sharp \nu$ denote the generalized Radon conditional
    measures of $\mu$ and $\nu$, respectively. 
\end{definition}

GSW is a valid metric given the injectivity of the GRT.
\begin{proposition}[GSW distance is a valid metric]
    \label{proposition:GSW_metricity} 
    For $p \geq 1$ and an injective GRT, the generalized sliced Wasserstein distance ($GSW_{p}$) is a metric on $\mathcal{P}_{p}(\Re^d)$.
    \begin{proof}
        The proof is similar to the proof of metricity for SW. We refer the reader to~\citet{kolouri2019generalized} for more detail.
    \end{proof}
\end{proposition}

\subsection{Projections on Manifolds of Non-Positive Curvature}
\label{subsec:projections_manifolds}
GRT is a general family of non-linear transforms. We now discuss specific instances of GRT designed for manifolds with non-positive curvature (Hadamard manifolds). These manifolds are complete connected Riemannian manifolds, meaning any geodesic curve can be extended along $\Re$ as a geodesic line. Moreover, Hadamard manifolds are diffeomorphic to Euclidean space, and their injectivity radius is infinite. Consequently, their geodesics are aperiodic and can be mapped onto the real line, enabling the identification of coordinates along this line. This, in turn, allows for the efficient computation of the OT between the projected measures. Since GRT is defined based on a projection function, we will focus on directly defining such functions. We first review the basic concepts for Riemannian manifolds, including geodesics, the exponential map, and geodesic distance.

\begin{definition}[Riemannian manifold]
    \label{def:Riemannian_manifold}
    A Riemannian manifold $(\mathcal{M},G)$ of dimension $d$ is a smooth manifold where each tangent space $T_x \mathcal{M}$ at $x \in \mathcal{M}$ is equipped with an inner product 
    \begin{align}
        \langle u,v \rangle_x = u^\top G(x) v, \quad u,v \in T_x\mathcal{M},
    \end{align}
    called the Riemannian metric. The associated norm is $\|u\|_x = \sqrt{\langle u,u \rangle_x}$. The tangent bundle of the manifold is the union of all tangent spaces:
    \begin{align}
        T\mathcal{M} = \{(x,v) \mid x \in \mathcal{M}, v \in T_x\mathcal{M}\}.
    \end{align}
\end{definition}

\begin{definition}[Geodesic and geodesic distance]
    \label{def:geodesic}
    Given two points $x,y \in \mathcal{M}$, a \emph{geodesic} is a smooth curve $\gamma:[0,1] \to \mathcal{M}$ satisfying $\gamma(0)=x$ and $\gamma(1)=y$, which locally minimizes the curve length:
    \begin{align}
        \mathcal{L}(\gamma) = \int_0^1 \sqrt{\langle \gamma'(t),\gamma'(t) \rangle_{\gamma(t)}} \,\mathrm{d}t,
    \end{align}
    where $\gamma'(t) \in T_{\gamma(t)} \mathcal{M}$. The geodesic distance between $x$ and $y$ is defined as
    \begin{align}
        c(x,y) = \inf_{\gamma \mid \gamma(0)=x, \gamma(1)=y} \mathcal{L}(\gamma).
    \end{align}
\end{definition}

\begin{definition}[Exponential map]
    \label{def:exponential_map}
    Let $x \in \mathcal{M}$ and $\theta \in T_x \mathcal{M}$. There exists a unique geodesic $\gamma_{x,\theta}$ with $\gamma_{x,\theta}(0)=x$ and $\gamma_{x,\theta}'(0)=\theta$. The exponential map at $x$ is
    \begin{align}
        \exp_x(\theta) = \gamma_{x,\theta}(1),
    \end{align}
    which maps a tangent vector $\theta$ back to the manifold along the geodesic.
\end{definition}

With the definition of geodesic distance, we now define geometry-aware projection functions on manifolds, including geodesic projection and Busemann projection.

\begin{figure}[!t]
    \centering
    \includegraphics[width=1\linewidth]{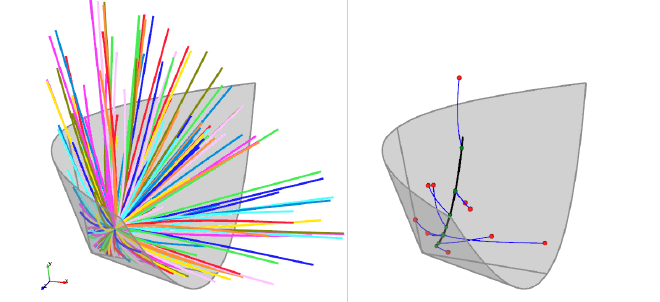}
    \caption{Random geodesics (left) and geodesic projections (right) (Figure~6 in~\citet{bonet2025sliced}).}
    \label{fig:geodesic}
\end{figure}

\begin{definition}[Geodesic projection]
    \label{def:geodesic_projection} Let $\gamma$ be a curve on the manifold $\mathcal{M}$, and denote $\mathcal{A}$ as the set of all points belonging to that curve. The projection of a point $x \in \mathcal{M}$ onto the curve $\gamma$ is defined as:
    \begin{align}
        \tilde{P}_\gamma (x) = \arg\min_{y \in \mathcal{A}} c(x, y),
    \end{align}
    where $c$ is the geodesic distance. If we constrain $\gamma$ to be a geodesic that passes through the origin (denoted $o$) with unit velocity $\theta$ (i.e., $\langle \theta, \theta \rangle_o = 1$), then $\mathcal{A} = \{ \exp_o(t\theta) \mid t \in \mathbb{R} \}$, where $\exp_o(\cdot)$ is the exponential map at the origin. The coordinates of the projection can be determined by solving:
    \begin{align}
        P_{\gamma_{(o,\theta)}}(x) := P_\theta(x) = \arg\min_{t \in \mathbb{R}} c(x, \exp_o(t\theta)).
    \end{align}
\end{definition}

A visualization of random geodesics and geodesic projection is given in Figure~\ref{fig:geodesic} (Figure~6 in~\citet{bonet2025sliced}). We then define  the Busemann projection.

\begin{definition}[Busemann projection]
    \label{def:busemann_projection} Let \( \gamma \) be a curve that passes through the origin with unit velocity \( \theta \). The Busemann function~\citep{bridson2013metric} of $x \in \setM$ associated with $\gamma$ is:
    \begin{align}
        B_\theta(x)  =  \lim_{t\to \infty} c\big(x,\exp_o(t\theta)\big)-t.
    \end{align}
    The projection onto the curve $\gamma$ is:
    \begin{align}
        \tilde{B}_\theta(x) =  \exp_o\big(-B_\theta(x)\theta\big).
    \end{align}
\end{definition}

For the Euclidean space,  the geodesic projection and the Busemann projection appear in the forms of the inner product (up to a sign) as in the conventional Radon Transform.

\begin{remark}[Geodesic projection and Busemann projection in Euclidean space]
    \label{remark:Euclidean_projection}
    In Euclidean space, the exponential map is the identity mapping and $c(x,y)=\|x-y\|_2$. By direct calculation, we obtain
    \begin{align}
        P_\theta(x) =   \langle \theta,x\rangle, \quad B_\theta(x) =   -\langle \theta,x\rangle, 
    \end{align}
    which are the defining functions (up to a sign) of the classical Radon Transform. Here, $\theta \in \Sm^{d-1}$.
\end{remark}

Next, we review specific forms of the geodesic projection and the Busemann projection on some known manifolds, e.g., the pullback Euclidean manifold, hyperbolic spaces, and manifolds of positive definite matrices. These forms are vital to define  SOT  variants on corresponding manifolds.

\begin{remark}[Geodesic projection and Busemann projection in pullback Euclidean manifold]
    \label{remark:pullback_Euclidean_projection} For a pullback Riemannian manifold $\setM$, there exists a diffeomorphism $\phi:\setM \to \mathbb{R}^d$. The geodesic distance is $c(x,y) = \|\phi(x)-\phi(y)\|_2$, and the exponential map is $\exp_x(\theta)= \phi^{-1}\big(\phi(x)+\phi_{*,x}(\theta)\big)$, where $\phi_{*,x}:T_x\setM \to T_{\phi(x)}\mathbb{R}^d$ is the differential of $\phi$ at $x$.  Therefore, we obtain the following forms of the eodesic projection and the Busemann projection:
    \begin{align}
        P_\theta (x) = -B_\theta(x) = \langle \phi(x) -\phi(o),\phi_{*,o}(\theta)\rangle,
    \end{align}
    ~\citep[Proposition 7]{bonet2025sliced}, where $\theta \in S_0 =\{\theta  \in T_o \setM \mid \|\phi_{*,o}(\theta)\|_2^2 =1\}$.
\end{remark}

\begin{remark}[Geodesic projection and Busemann projection in hyperbolic spaces]
\label{remark:Hyperbolic_projection} 
    ~\citet{bonet2023hyperbolic,bonet2025sliced} consider two hyperbolic space models: the Lorentz model and the Poincaré ball.

    \textbf{Lorentz Model.} The Lorentz model of curvature \(K < 0\) is defined as follows:
\[
\mathbb{L}_K^d = \left\{ (x_0, \ldots, x_d) \in \mathbb{R}^{d+1} \ \middle|\ \langle x, x \rangle_{\mathbb{L}} = \frac{1}{K},\ x_0 > 0 \right\},
\]
where
\[
\langle x, y \rangle_{\mathbb{L}} = -x_0 y_0 + \sum_{i=1}^d x_i y_i.
\]
The geodesic distance in this manifold is
\[
c(x, y) = \frac{1}{\sqrt{-K}} \operatorname{arccosh} \big( K \langle x, y \rangle_\mathbb{L} \big).
\]
The exponential map is:
\[
\exp_{x^0}(t\theta) = \cosh\big(\sqrt{-K}t\big) x^0 + \sinh\big(\sqrt{-K}t\big) \frac{\theta}{\sqrt{-K}},
\]
where $\|v\|_{\mathbb{L}} = \sqrt{\langle v, v \rangle_{\mathbb{L}}}$. We obtain:
\begin{align}
    &P_\theta(x) = \frac{1}{\sqrt{-K}} \operatorname{arctanh} \left( 
    - \frac{1}{\sqrt{-K}}  \frac{ \langle x, v \rangle_{\mathbb{L}} }{ \langle x, x^0 \rangle_{\mathbb{L}} }
    \right), \\
    &B_v(x) = \frac{1}{\sqrt{-K}} \log \left( 
    - \sqrt{-K}  \left\langle x,\ \sqrt{-K}x^0 + v \right\rangle_{\mathbb{L}}
    \right),
\end{align}
where $x^0 = \left( \frac{1}{\sqrt{-K}}, 0, \ldots, 0 \right)$ is the origin and $\theta \in S_{x^0} = T_{x^0} \mathbb{L}_K^d \cap \Sm^d$.

\textbf{Poincaré Ball.} The Poincaré ball of curvature \( K < 0 \) is defined as
\[
\mathbb{B}_K^d = \left\{ x \in \mathbb{R}^d \ \middle|\ \|x\|_2^2 < -\frac{1}{K} \right\}.
\]
The origin is \( \boldsymbol{0} \in \mathbb{R}^d \), and the geodesic distance is
\[
c(x, y) = \frac{1}{\sqrt{-K}} \operatorname{arccosh} \left(
1 - \frac{2K \|x - y\|_2^2}{(1 + K \|x\|_2^2)(1 + K \|y\|_2^2)}
\right).
\]
The exponential map is:
\[
    \exp_o(t\theta) = \frac{1}{\sqrt{-K}}\tanh \left(\frac{\sqrt{-K}t}{2}\right)\theta.
\]
We obtain:
\begin{align}
    &P_\theta(x) = \frac{2}{\sqrt{-K}} \operatorname{arctanh} \left( \sqrt{-K} \cdot s(x) \right),
  \\
  &B_v(x) = \frac{1}{\sqrt{-K}} \log \left( 
    \frac{ \| \theta - \sqrt{-K}x \|_2^2 }{ 1 + K \|x\|_2^2 }
  \right),
\end{align}
where
\[
s(x) =
\begin{cases}
\displaystyle \frac{1 - K \|x\|_2^2 - \sqrt{(1 - K \|x\|_2^2)^2 + 4K \langle x, \theta  \rangle^2}}{-2K \langle x, \theta \rangle}, & \text{if } \langle x, \theta \rangle \neq 0, \\
0, & \text{if } \langle x, \theta \rangle = 0,
\end{cases}
\]
and $\theta \in \Sm^{d-1}$.
\end{remark}

\begin{remark}[Geodesic Projection and Busemann Projection in the Manifold of Positive Definite Matrices]
    \label{remark:psdm_projection} Let $S_d(\Re)$ be the set of symmetric matrices in $\Re^{d\times d}$. The manifold of positive definite matrices is denoted as $S_d^{++}(\Re) = \{M \in S_d(\Re) \mid x^\top M x > 0,\, \forall\, x \in \Re^d \setminus \{0\}\}$~\citep{bhatia2009positive}, has the origin $I_d$, and has $T_MS_d^{++}(\Re) = S_d(\Re)$ as its tangent space. The manifold $S_d^{++}(\Re)$ can be endowed with different metrics. 

\textbf{Affine-Invariant Metric.} A classical and widely used metric is the Affine-Invariant metric~\citet{pennec2006riemannian}, which supports the inner product:
\begin{align}
    \langle A,B\rangle_M = \text{Trace}(M^{-1}AM^{-1}B),
\end{align}
and the geodesic distance:
\begin{align*}
    c(X,Y) = \sqrt{\text{Trace}(\log(X^{-1}Y)^2)},
\end{align*}
where $\log(X)$ is the matrix logarithm. The exponential map is:
\begin{align*}
    \exp_{I_d}(t\theta) = \exp(t\theta),
\end{align*}
where $\theta \in S_d(\Re)$ and $\exp(\cdot)$ is the matrix exponential. There is no closed form for
the geodesic projection; however, the Busemann projection is:
\begin{align}
    B_\theta(M) = -\langle \theta, \log (\pi_\theta(M))\rangle_F,
\end{align}
where $\pi_A$ is a projection on the space of commuting matrices, which can be obtained in
practice through a UDU or LDL decomposition~\citep{bonet2023sliced,bonet2025sliced}, and $\langle \cdot,\cdot\rangle_F$ is the Frobenius inner product.

\textbf{Log-Euclidean Metric.} For this type of geometry~\citep{pennec2020manifold}, $S_d^{++}(\Re)$ is a pullback Euclidean manifold with $\phi(X) = \log(X)$ (Remark~\ref{remark:pullback_Euclidean_projection}). From~\citet{bonet2025sliced}, we have:
\begin{align}
    P_\theta(X) = -B_\theta(X) = \langle \log(X),\theta\rangle_F.
\end{align}

\textbf{$\mathcal{O}(n)$-Invariant Log-Euclidean Metric.} For this type of geometry~\citep{thanwerdas2023n}, $S_d^{++}(\Re)$ is a pullback Euclidean manifold with $\phi(X) = q\log(X) + \frac{p-q}{d}\text{Trace}(\log(X))I_d$ for $p,q\geq 0$. From~\citet{bonet2025sliced}, we have:
\begin{align}
    &P_\theta(X) \nonumber =\\ &\langle  q\log(X) + \frac{p-q}{d}\text{Trace}(\log(X))I_d, q\theta + \frac{p-q}{d}\text{Trace}(\theta)I_d \rangle _F.
\end{align}

\textbf{Log-Cholesky Metric.} For this type of geometry~\citep{lin2019riemannian}, $S_d^{++}(\Re)$ is a pullback Euclidean manifold with $\phi(X) = \psi(\mathcal{L}(X))$, where $\mathcal{L}$ denotes the Cholesky decomposition and $\psi(L) = [L] + \log(\text{diag}(L))$ (
$[\cdot]$ denotes the strictly lower triangular part of the matrix). From~\citet{bonet2025sliced}, we have:
\begin{align}
    P_\theta(X) = \langle [\mathcal{L}(X)],\theta\rangle_F.
\end{align}
\end{remark}

As mentioned, we can derive GRTs based on the geodesic projection and the Busemann projection, i.e., using them as the defining function.

\begin{remark}[Integral transforms of geodesic projection and Busemann projection]
    \label{remark:integral_transform_geodesic_projection} By using the defining function $g(x,\theta) = P_\theta(x)$ for the geodesic projection and $g(x,\theta) = B_\theta(x)$ for the Busemann projection in the generalized Radon Transform in Definition~\ref{def:GRT} and Definition~\ref{def:Generalized_Radon_Transform_measures}, we obtain the corresponding integral transforms of geodesic projection and Busemann projection. Such transforms are injective for pullback Euclidean manifolds, however, their injectivity is unknown in other cases~\citep{bonet2025sliced}.
\end{remark}

In this special case, GSW variants, referred to as Cartan–Hadamard Sliced Wasserstein distances, can be defined as follows:

\begin{definition}[Cartan–Hadamard sliced Wasserstein distances]
\label{def:CHSW}
    Let $(\setM,G)$ be a Hadamard manifold (a  Riemannian manifold that is complete and simply connected and has everywhere non-positive sectional curvature) with $o$ as its origin, and let $S_o = \{\theta  \in  T_o\setM \mid \langle \theta,\theta \rangle_o = 1\}$. For $p \geq 1$, the Geodesic Cartan–Hadamard Sliced Wasserstein-$p$ distance~\citep{bonet2025sliced}
    between $\mu,\nu \in \setP_p(\setM)$ is defined as:
    \begin{align}
        GCHSW_p^p(\mu,\nu) = \mathbb{E}_{\theta \sim \setU(S_o)} \left[ W_p^p(P_\theta\sharp \mu, P_\theta \sharp \nu) \right],
    \end{align}
    where $P_\theta$ is the geodesic projection in Definition~\ref{def:geodesic_projection}. Similarly, the Horospherical Cartan–Hadamard Sliced Wasserstein-$p$ distance between $\mu$ and $\nu$ is defined as:
    \begin{align}
        HCHSW_p^p(\mu,\nu) = \mathbb{E}_{\theta \sim \setU(S_o)} \left[ W_p^p(B_\theta\sharp \mu, B_\theta \sharp \nu) \right],
    \end{align}
    where $B_\theta$ is the Busemann projection in Definition~\ref{def:busemann_projection}.
\end{definition}

\begin{proposition}[Metricity of Cartan–Hadamard sliced Wasserstein distances]
    \label{proposition:metricity_CHSW}
    For $p \geq 1$, Cartan–Hadamard Sliced Wasserstein distances are pseudometrics for general manifolds, while they are metrics for pullback Euclidean manifolds.
    \begin{proof}
        We refer the reader to~\citet{bonet2025sliced} for detailed proofs. The pseudometricity is due to the unknown injectivity of the corresponding integral transform of the geodesic projection or the Busemann projection.
    \end{proof}
\end{proposition}

\begin{remark}[Statistical and computational properties of Cartan–Hadamard Sliced Wasserstein distances]
    \label{remark:statistical_computational_properties_CHSW}
    From~\citet{bonet2025sliced}, CHSW also enjoys a low sample complexity, the same as SW, i.e., $\mathcal{O}(n^{-1/2})$. For pullback Euclidean manifolds, CHSW metricizes weak convergence. In practice, CHSW also requires Monte Carlo estimation. The changes in computational complexity compared to SW arise from sampling from $\setU(S_o)$ and from the projection function. 
\end{remark}

\subsection{Spherical Projections}
\label{subsec:spherical_projections}

We now review methods for projection onto the spherical domain. In the previous section, the projection function was used to map a measure onto the real line. In the spherical setting, an additional option for the projected domain is the circle, since the Wasserstein distance on the circle also admits a closed-form expression. 

\begin{definition}[Wasserstein distance on the circle]
    \label{def:1DWasserstein_circle} For continuous measures $\mu,\nu \in \setP_p(\Sm)$, the Wasserstein-$p$ distance with the ground metric $c(x,y) = h(\min(|x-y|,1-|x-y|))$, for an increasing convex function $h$, admits the following form~\citep{delon2010fast,rabin2011transportation}:
    \begin{align}
        W_{c,p}^p(\mu,\nu) = \inf_{\alpha \in \Re} \int_0^1 h\big(|F_{\mu}^{-1}(t) - (F_\nu-\alpha)^{-1}(t)|\big)\,\diff t ,
    \end{align}
    where we parameterize $\Sm = [0,1[$, and $F_\nu-\alpha$ is the shifted CDF. For discrete measures $\mu$ and $\nu$, we can simply use the empirical quantile functions~\citep{delon2010fast}.
\end{definition}

\begin{remark}[Wasserstein-1 distance and Wasserstein-2 distance on the circle]
    \label{remark:W1_W2_circle}
    When $h = \mathrm{Id}$ and $p = 1$, from~\citet{werman1985distance,cabrelli1995kantorovich}, we have:
    \begin{align}
        W_1(\mu,\nu) =  \int_0^1 \big|F_\mu(t) - F_\nu(t) - \mathrm{LevMed}(F_\mu-F_\nu)\big| \,\diff t, 
    \end{align}
    where the level median~\citep{hundrieser2022statistics} is defined as
    \begin{align}
        \mathrm{LevMed}(f) &= \min\left(\arg\min_{\alpha \in \Re} \int_0^1 |f(t)-\alpha| \,\diff t \right) \nonumber \\
        &= \inf \left\{t \in \Re \mid \lambda_0(\{x \in \Sm \mid f(x) \leq t\}) \geq 0.5\right\},
    \end{align}
    where $\lambda_0$ is the Lebesgue measure. 
    When $h = \mathrm{Id}$ and $p = 2$, from~\citet{bonet2023spherical}, we have:
    \begin{align}
        W_2^2(\mu,\nu) =  \int_0^1 \big|F_\mu^{-1}(t) - t - \hat{\alpha}\big|^2 \,\diff t, 
        \quad \text{with} \quad \hat{\alpha} =  \int_{\Sm} x \,\diff \mu(x) - 0.5.
    \end{align}
\end{remark}

We now review the geodesic projection on the hypersphere, which can project a measure onto the circle. After that, we discuss the GSW variant based on such a projection.

\begin{figure}[!t]
    \centering
    \includegraphics[width=0.8\linewidth]{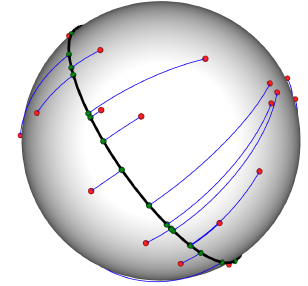}
    \caption{Geodesic projection on the hypersphere (Figure 1 in~\citet{bonet2023spherical}).}
    \label{fig:spherical_projection}
\end{figure}

\begin{definition}[Geodesic projection on hypersphere]
    \label{def:geodesic_projection_sphere} Geodesic projection (Definition~\ref{def:geodesic_projection}) can be applied to the hypersphere $\Sm^{d-1}$. Here, a geodesic is a great circle, and the geodesic distance is $c(x,y) =  \text{argcos}(\langle x,y \rangle)$. The geodesic projection~\citep{bonet2023spherical} is defined as:
    \begin{align}
        P_\theta(x) =  \frac{\theta^\top x}{\|\theta^\top x\|_2},
    \end{align}
    where $\theta \in \mathbb{V}_2(\Re^d)$ (Stiefel manifold). This geodesic projection can yield the semi-circle transform~\citep{quellmalz2023sliced}. We visualize the projection in Figure~\ref{fig:spherical_projection} (Figure 1 in~\citet{bonet2023spherical}).
\end{definition}

\begin{definition}[Spherical Sliced Wasserstein distance]
    \label{def:SSW}
    For $p\geq 1$, the spherical sliced Wasserstein (SSW) distance between two measures $\mu\in \setP_p(\Sm^{d-1})$ and $\nu \in \setP_p(\Sm^{d-1})$ is defined as follows:
    \begin{align}
        SSW_p^p(\mu,\nu) = \mathbb{E}_{\theta \sim \setU(\mathbb{V}_2(\Re^d))}[ W_p^p(P_\theta \sharp \mu,P_\theta \sharp \nu)],
    \end{align}
    where we use the Wasserstein distance on the circle for $W_p^p(P_\theta \sharp \mu,P_\theta \sharp \nu)$ in Definition~\ref{def:1DWasserstein_circle}.
\end{definition}

\begin{proposition}[Metricity of Spherical Sliced Wasserstein distances]
    \label{proposition:metricity_SSW}
    For $p\geq 1$, spherical sliced Wasserstein distances are pseudometrics.
    \begin{proof}
        We refer the reader to~\citet{bonet2023spherical} for a detailed proof. The pseudometricity is due to the unknown injectivity of the corresponding integral transform of the geodesic projection.
    \end{proof}
\end{proposition}

SSW also needs to be approximated in practice with Monte Carlo estimation, i.e., sampling from $\setU(\mathbb{V}_2(\Re^d))$. The sampling can be done by generating a matrix whose entries are i.i.d. from a standard Gaussian distribution and then applying the Gram–Schmidt process to that matrix. We refer the reader to~\citet{bonet2023spherical} for a detailed discussion of computational complexities. We now discuss the second approach, which is projecting a measure onto the real line.

\begin{definition}[Stereographic projection]
    \label{def:stereographic_projection}
    The stereographic projection~\citep{whittaker1984stereographic} $\phi :\Sm^d\backslash \{s^0\} \to \Re^d$ maps a point $s =(s_1,\ldots,s_{d+1})\in \Sm^d$ (excluding the ``North Pole,'' $s^0=(0,\ldots,0,1)$) to a point $x=(x_1,\ldots,x_d) \in \Re^d$ as follows: 
    \begin{align}
        x_i = \frac{2s_i}{1-s_{i+1}}, \, \forall i=1,\ldots, d.
    \end{align}
    This projection is a bijective and smooth mapping. 
\end{definition}

\begin{definition}[Distance distortion]
    \label{def:distance_distortion}  The stereographic projection severely distorts
distances since it is an angle-preserving projection. Therefore, it is necessary to have an injective function $h$ such that:
\begin{align}
    \|h(\phi(s)) -h(\phi(s'))\|_2 \approx \text{argcos}(\langle s,s'\rangle ),\, \forall s,s' \in \Sm^d.
\end{align}
The authors in~\citet{tran2024stereographic} propose to use:
\begin{align}
    h(x) = \text{argcos}\left(\frac{\|x\|_2^2-1}{\|x\|_2^2+1}\right)\frac{x}{\|x\|_2},\, \forall x \in \Re^d.
\end{align}
Another approach is to learn $h(x)$ using neural network parameterization~\citep{tran2024stereographic} with the following objective:
\begin{align}
    \mathbb{E}_{(s,s')\sim \setU(\Sm^d)\otimes \setU(\Sm^d)}[\|h(\phi(s)) -h(\phi(s'))\|_2 -\text{argcos}(\langle s,s'\rangle )];
\end{align}

\end{definition} 
With the stereographic projection and distance distortion, a specific variant of GSW for spheres, named stereographic spherical sliced Wasserstein, can be defined as follows:

\begin{definition}[Stereographic Spherical Sliced Wasserstein distance]
    \label{def:S3W}
    For $p\geq 1$, the stereographic spherical sliced Wasserstein (S3W) distance between two measures $\mu\in \setP_p(\Sm^{d})$ and $\nu \in \setP_p(\Sm^{d})$ is defined as follows:
    \begin{align}
        S3W_p^p(\mu,\nu) = \mathbb{E}_{\theta \sim \setU(\Sm^{d-1})}[ W_p^p(S_\theta \sharp \mu,S_\theta \sharp \nu)],
    \end{align}
    where $S_\theta(s) = \langle h(\phi(s)),\theta \rangle$ with $\phi$ being the stereographic projection (Definition~\ref{def:stereographic_projection}), $h$ the distance distortion function (Definition~\ref{def:distance_distortion}), and we use the conventional Wasserstein distance on the line $W_p^p(S_\theta \sharp \mu,S_\theta \sharp \nu)$.
\end{definition}

Like SSW, S3W is also a pseudometric, owing to the unknown injectivity of the integral transform associated with the geodesic projection. Its main advantage is computational: S3W is faster than SSW because both the projection and the Wasserstein distance evaluation on the line are quicker than on the circle, as required by SSW. 

\begin{proposition}[Metricity of Stereographic Spherical Sliced Wasserstein distances]
    \label{proposition:metricity_S3W}
    For $p\geq 1$, the stereographic spherical sliced Wasserstein distance is a pseudometric~\citep{tran2024stereographic}.
\end{proposition}

We refer the reader to~\citet{tran2024stereographic} for further discussion on computational complexity with Monte Carlo estimation and a rotationally invariant extension of S3W. For the sphere in 3D, i.e., $\mathbb{S}^2$, there is another option for projection, which is vertical slice projection.

\begin{definition}[Vertical Slice Projection]
    \label{def:vertical_slice_projection}
    For $\Sm^2$, the equator can be defined as follows:
    \begin{align}
        \mathcal{E} := \left\{ \Phi(\alpha, 0) = (\cos \alpha,\sin\alpha,0): \alpha \in [0, 2\pi) \right\}.
    \end{align}
    The vertical slice projection~\citep{quinto1991tomography,quellmalz2023sliced} is defined as follows:
    \begin{align}
        V_\theta(x) =  \langle x,\theta\rangle, \quad \text{for } x\in \Sm^2 \text{ and } \theta \in \mathcal{E}.
    \end{align}
\end{definition}

With the vertical slice projection, we can obtain the corresponding GSW distance. It is interesting that this distance is a valid metric, unlike the pseudo-metricity in SSW and S3W.

\begin{definition}[Vertical Sliced Wasserstein Distance]
    \label{def:VSW}
    For $p\geq 1$, the vertical sliced Wasserstein (VSW) distance between two measures $\mu\in \setP_p(\Sm^{2})$ and $\nu \in \setP_p(\Sm^{2})$ is defined as follows:
    \begin{align}
        VSW_p^p(\mu,\nu) = \mathbb{E}_{\theta \sim \setU(\mathcal{E})}\left[ W_p^p\big(V_\theta \sharp \mu,V_\theta \sharp \nu\big)\right],
    \end{align}
    where $V_\theta$ denotes  the vertical slice projection (Definition~\ref{def:vertical_slice_projection}), and  $W_p^p(V_\theta \sharp \mu,V_\theta \sharp \nu)$ denotes the conventional Wasserstein distance on the real line.
\end{definition}

We refer the reader to~\citet[Theorem 3.7]{quellmalz2023sliced} for a detailed proof of metricity and related properties.

\subsection{Projections on Compact Manifolds}
\label{subsec:compact_manifolds}

The first attempt to generalize the SW distance on Riemannian manifolds was made by~\citet{rustamov2023intrinsic} using Laplace–Beltrami operator projections. The eigenvalues $\lambda_l$ and eigenfunctions $\varphi_l$ of the Laplace-Beltrami operator $\Delta$ are defined by the spectral problem
\begin{align}
    -\Delta \varphi_l = \lambda_l \varphi_l, \qquad l = 0,1,2,\ldots,
\end{align}
with $\{\varphi_l\}_{l\ge 0}$ forming an orthonormal basis of $L^2(\mathcal{M})$ and $0=\lambda_0 < \lambda_1 \le \lambda_2 \le \cdots \uparrow \infty$.

\begin{definition}[Laplace–Beltrami operator projection]
    \label{def:laplace-beltrami}
    Let $\mathcal{M}$ be a compact Riemannian manifold. For $l \in \mathbb{N}$, denote by $\lambda_l$ the eigenvalues and by $\varphi_l$ the eigenfunctions of the Laplace–Beltrami operator (see~\citet[Definition~4.7]{do1992riemannian}), sorted in increasing order of eigenvalues. The Laplace–Beltrami operator projections of $\mu \in \setP_2(\setM)$ are:
    \begin{align}
        \varphi_l \sharp \mu, \quad \forall l=1,\ldots,N.
    \end{align}
\end{definition}

\begin{definition}[Intrinsic Sliced Wasserstein Distance]
\label{def:ISW}
    The intrinsic sliced Wasserstein (ISW) distance between $\mu, \nu \in \mathcal{P}_2(\mathcal{M})$ is defined as~\citep{rustamov2023intrinsic}:
    \begin{align}
        ISW_2^2(\mu, \nu) = \sum_{\ell \ge 0} \alpha(\lambda_\ell) W_2^2\big( \varphi_\ell\sharp \mu, \varphi_\ell\sharp \nu \big),
    \end{align}
    where $\alpha : \mathbb{R}^+ \to \mathbb{R}^+$ is a monotonically decreasing function. 
\end{definition}

\begin{proposition}[Metricity of Intrinsic Sliced Wasserstein]
    From~\citet[Theorem 4.6]{rustamov2023intrinsic}, the intrinsic sliced Wasserstein distance is a valid metric on the space of probability measures on the compact manifold $\mathcal{M}$. 
\end{proposition}

In practice, we approximate ISW using the first $L$ eigenvalues and eigenfunctions. Despite being a natural notion of distance, ISW faces computational challenges, since computing eigenvalues and eigenfunctions of the Laplace–Beltrami operator for general manifolds can be difficult.

\subsection{Projections for Images, Functions, and Probability Measures}
\label{subsec:projections_images_functions_measures}
In the previous subsections, we focused on projections in settings of manifolds where the underlying geometry of the space is well understood and can be explicitly exploited. However, in many practical scenarios, such as when working with images, functions, or probability measures, the geometry of the space may not be known \textit{a priori} or may be too complex to model explicitly. We now review strategies for defining projections in such settings. 

\begin{definition}[Convolution projection for images]
    \label{def:convolution_projection_images}
    Given an image $x \in \Re^{c\times d\times d}$ ($c \geq 1, d \geq 1$), the convolution projection~\citep{nguyen2022revisiting} can be defined as follows:
    \begin{align}
        C_\theta(x) = x * \theta^{(1)} * \ldots * \theta^{(H)},
    \end{align}
    where $\theta = (\theta^{(1)},\ldots,\theta^{(H)}) \in \Theta$ ($H \geq 1$, $\Theta$ be the space of discrete convolution kernels~\citep{nguyen2022revisiting}), $*$ denotes the discrete convolution function:
    \begin{align}
       (x*y)_{i,j} = \sum_{m=1}^{c} \sum_{u=1}^{k} \sum_{v=1}^{k} x_{m, i+u-1, j+v-1} y_{m, u, v},
    \end{align}
    with $y \in \Re^{c\times k\times k}$, and $\theta^{(1)},\ldots,\theta^{(H)}$ are designed such that $C_\theta(x) \in \Re$ with $\|\theta^{(h)}\|_2^2 = 1, \ \forall h = 1,\ldots,H$. 
\end{definition}

The authors in~\citet{nguyen2022revisiting} also introduce other types of convolution with stride, dilation, and padding. Convolution can preserve local geometry and might reduce the number of projection parameters compared to vectorization of images. We can then define the corresponding SW variant as follows:

\begin{definition}[Convolution Sliced Wasserstein distance for images]
    \label{def:CSW_images} For $p \geq 1$, the convolution sliced Wasserstein-$p$ distance between $\mu \in \setP_p(\Re^{c\times d\times d})$ and $\nu \in \setP_p(\Re^{c\times d\times d})$ is defined as follows:
    \begin{align}
        CSW_p^p(\mu,\nu) = \mathbb{E}_{\theta \sim \setU(\Theta)}[W_p^p(C_\theta \sharp \mu, C_\theta \sharp \nu)],
    \end{align}
    where $\Theta$ is the space of values for convolution kernels.
\end{definition}
    
The discussed convolution projection for images is discrete convolution. For functions, we can also utilize the convolution operator to map them into one scalar.

\begin{definition}[Convolution projection for functions]
    \label{def:convolution_projection_functions}
    Let $\setS$ be a compact subset of $\Re^d$, $\theta \in \setS$, and $k$, called the kernel function, be a bounded continuous function from $\setS \times \setS \to [0, \infty)$. We define the convolution projection for a function $f:\setS \to \Re$ as follows~\citep{garrett2024validating}:
\begin{align}
    C_\theta(f) = \int_\setS f(s) k(\theta, s)\, \diff s.
\end{align}
Since $\setS$ is compact and $k(\theta, s)$ is bounded, we have $k(s, u) \in L^2(\setS)$, and thus $C_\theta(f)$ is a bounded linear map. 
\end{definition}

Similar to the case of images, we can also extend the projection into an SW variant:

\begin{definition}[Convolution Sliced Wasserstein distance for functions]
    \label{def:CSW_functions} For $p \geq 1$, the convolution sliced Wasserstein-$p$ distance between $\mu \in \setP_p(L^2(\setS))$ and $\nu \in \setP_p(L^2(\setS))$ is defined as follows:
    \begin{align}
        CSW_p^p(\mu,\nu) = \mathbb{E}_{\theta \sim \setU(\setS)}[W_p^p(C_\theta \sharp \mu, C_\theta \sharp \nu)],
    \end{align}
    where $L^2(\setS)$ is the set of square-integrable functions.
\end{definition}

\begin{proposition}[Metricity of Convolution Sliced Wasserstein distances]
    \label{proposition:metricity_CSW}
    For $p \geq 1$, convolution sliced Wasserstein distances are pseudo-metrics for the space of distributions over images or functions.
    \begin{proof}
        We refer the reader to~\citet{nguyen2022revisiting,garrett2024validating} for detailed proofs. As before, the key challenge is in proving the identity of indiscernibles of the distances, which is controlled by the injectivity of the corresponding integral transforms.
    \end{proof}
\end{proposition}

For probability measures, a natural way to project them into a scalar is by using moments:

\begin{definition}[Moment transform projection for probability measures]
    \label{def:moment_transform_projection}
    Given a probability measure $\mu \in \setP(\Re^d)$, the moment transform projection~\citep{nguyen2025lightspeed} is defined as follows:
    \begin{align}
        MTP_{\theta,\lambda}(\mu) = \int_{\Re^d} \frac{(\mathcal{FP}_\theta(x))^\lambda}{\lambda!} \, \diff \mu(x),
    \end{align}
    where $\theta \in \mathbb{S}^{d-1}$ and $\lambda \in \Lambda \subset \N$.
\end{definition}

\begin{remark}[Injectivity of moment transform projection's integral transform]
    \label{remark:injectivity_MTP} 
    When the Hamburger moment problem~\citep{reed1975ii,chihara2011introduction} is solvable, the corresponding integral transform of the moment transform projection is injective~\citep{nguyen2025lightspeed}.
\end{remark}

The moment transform projection can be used itself to define a SW metric. However, it was originally introduced to define a dataset distance~\citep{nguyen2025lightspeed}.

\subsection{Projections on Product of Metric Spaces  }
\label{subsec:projections_product_spaces}
We now discuss comparing probability measures on $\setP(\setX_1 \times \setX_2)$, where $\setX_1$ and $\setX_2$ support different metrics. We start with the case where $\setX_1$ and $\setX_2$ are manifolds. In this case, we can extend the notion of geodesic projection and Busemann projection to their generalized versions.

\begin{definition}[Generalized Geodesic projection]
    \label{def:generalized_geodesic_projection}
    Given a product manifold $\setM_1 \times \setM_2 \times \ldots \times \setM_n$ with origin $o = (o_1, o_2, \ldots, o_n)$, a generalized curve passing through the origin with velocity vector $\theta_w = (w_1 \theta_1, w_2 \theta_2, \ldots, w_n \theta_n)$ where $\langle \theta_i, \theta_i \rangle_{o_i} = 1$ for all $i=1,\ldots,n$, and weights satisfying $\sum_{i=1}^n w_i^2 = 1$, the generalized geodesic projection~\citep{nguyen2024summarizing} onto the generalized curve generated by $\theta_w$ is defined as:
    \begin{align}
        \label{eq:generalized_geodesic_projection}
        P_{\theta_w}(x) = \arg\min_{t \in \Re} c\big(x, \exp_o(t \theta_w)\big),
    \end{align}
    where $c$ is the geodesic distance on the product manifold:
    \begin{align}
        c\big((x_1, \ldots, x_n), (y_1, \ldots, y_n)\big)^2 = \sum_{i=1}^n c_i(x_i, y_i)^2,
    \end{align}
    with $c_i$ being the geodesic distance on $\setM_i$~\citep{gu2018learning}$, i=1,\ldots,n$. When $w_1 = \ldots = w_n = 1/\sqrt{n}$, we recover the conventional geodesic projection~\citep{bonet2025sliced}. Authors in~\citet{nguyen2024summarizing} show that choosing weights $w_1, \ldots, w_n$ is crucial for the injectivity of the corresponding integral transform for products of Euclidean spaces and the manifold of symmetric positive definite matrices.
\end{definition}

\begin{definition}[Generalized Busemann projection]
    \label{def:generalized_buseman_projection}
    Given a product manifold $\setM_1 \times \setM_2 \times \ldots \times \setM_n$ with origin $o = (o_1, o_2, \ldots, o_n)$, a generalized curve passing through the origin with velocity vector $\theta_w = (w_1 \theta_1, w_2 \theta_2, \ldots, w_n \theta_n)$ where $\langle \theta_i, \theta_i \rangle_{o_i} = 1$ for all $i=1,\ldots,n$, and weights satisfying $\sum_{i=1}^n w_i^2 = 1$, the generalized Busemann projection~\citep{bonet2025sliced} onto the generalized curve generated by $\theta_w$ is defined as:
    \begin{align}
        B_{\theta_w}(x) = \sum_{i=1}^n w_i B_{\theta_i}(x_i),
    \end{align}
    where $x = (x_1, \ldots, x_n)$ and $B_{\theta_i}(\cdot)$ is the Busemann projection on $\setM_i$.
\end{definition}

Using these generalized projections, we define Sliced Wasserstein variants on product manifolds as follows:

\begin{definition}[Sliced Wasserstein on product manifolds]
    \label{def:SW_product_manifolds}
    For $p \geq 1$, the product manifolds sliced Wasserstein-$p$ (PMSW) distance between $\mu \in \setP_p(\setM_1 \times \ldots \times \setM_n)$ and $\nu \in \setP_p(\setM_1 \times \ldots \times \setM_n)$ is defined as:
    \begin{align}
        PMSW_p^p(\mu, \nu) = \mathbb{E}_{(\theta, w) \sim \setU(\Theta \times \Sm^{n-1})} \big[ W_p^p(P_{\theta_w} \sharp \mu, P_{\theta_w} \sharp \nu) \big],
    \end{align}
    where $\Theta$ is the joint set of unit directions on the tangent spaces of all manifolds, and $P_{\theta_w}$ is the generalized geodesic projection.
\end{definition}

 Similarly, $P_{\theta_w}$ can be replaced by the Busemann projection $B_{\theta_w}$ from Definition~\ref{def:generalized_buseman_projection}. As usual, the expectation is approximated by Monte Carlo sampling. PMSW appears in~\citet{nguyen2024summarizing} under the name mixed sliced Wasserstein for comparing mixtures of Gaussians. The metricity of PMSW depends on the specific manifolds; however, it has been shown to be a metric in the case of the product of a Euclidean manifold and the manifold of symmetric positive definite matrices~\citep{nguyen2024summarizing}.

We now discuss a more general approach to constructing projection functions for products of spaces beyond manifolds. We first start with partial Radon Transform.

\begin{definition}[Partial Radon Transform]
    \label{def:Partial_Radon_Transform}
    Given an integrable function $f : \Re^{d_1} \times \Re^{d_2} \to \Re$, the partial Radon transform (PRT)~\citep{liang1997partial} is defined as:
    \begin{align}
        \setPR f(\theta, y, t) = \int_{\Re^{d_1}} f(x, y) \, \delta(t - \langle x, \theta \rangle) \, dx,
    \end{align}
    where we denote the conditional function of $(t,y)$ given $\theta$ as:
    \begin{align}
        \setPR_\theta f(y, t) := \setPR f(\theta, y, t).
    \end{align}
\end{definition}

Since PRT is a Radon transform on the conditional function $f_y$, it is also bijective, inheriting bijectivity from the classical Radon transform (Remark~\ref{remark:bijectivity_Radon}).  It can be extended to the Partial Generalized Radon Transform (PGRT)~\citep{nguyen2024hierarchical} defined by:
    \begin{align}
        \mathcal{PGR} f(\theta, y, t) = \int_{\Re^{d_1}} f(x, y) \, \delta(t - g(x, \theta)) \, dx,
    \end{align}
    for a suitable defining function $g$ (Definition~\ref{def:defining_functions}). Using multiple PGRTs and PRT, we define the hierarchical hybrid Radon transform:

\begin{definition}[Hierarchical Hybrid Radon Transform]
    \label{def:Hierarchical_Hybrid_Radon_Transform}
    Given an integrable function $f : \Re^{d_1} \times \ldots \times \Re^{d_n} \to \Re$ with $n \geq 2$, the hierarchical hybrid Radon transform (HHRT)~\citep{nguyen2024hierarchical} is:
    \begin{align}
        & \mathcal{HRT}(t, \theta_1, \ldots, \theta_n, w; g_1, \ldots, g_n) \nonumber \\
        &= \int_{\Re^{d_1}} \ldots \int_{\Re^{d_n}} f(x_1, \ldots, x_n) \, \delta \left( t - \sum_{i=1}^n w_i g_i(x_i, \theta_i) \right) \, dx_1 \ldots dx_n,
    \end{align}
    where $w = (w_1, \ldots, w_n)$ with $\sum_{i=1}^n w_i^2 = 1$, $g_1, \ldots, g_n$ are defining functions corresponding to the subspace structures, and $\theta_1, \ldots, \theta_n$ are the associated projection parameters.
\end{definition}

\begin{remark}[HHRT as composition of partial Radon and generalized Radon transforms]
    By Fubini's theorem, we can rewrite~\citep{nguyen2024hierarchical}:
    \begin{align}
        & \mathcal{HRT}(t, \theta_1, \ldots, \theta_n, w; g_1, \ldots, g_n) \nonumber \\
        &= \int_{\Re^{d_1}} \ldots \int_{\Re^{d_n}} f(x_1, \ldots, x_n) \delta\left(t - \sum_{i=1}^n w_i g_i(x_i, \theta_i)\right) dx_1 \ldots dx_n \nonumber \\
        &= \int_{\Re} \ldots \int_{\Re} \int_{\Re^{d_1}} \ldots \int_{\Re^{d_n}} f(x_1, \ldots, x_n) \prod_{i=1}^n \delta(\psi_i - g_i(x_i, \theta_i)) \, dx_1 \ldots dx_n \nonumber \\
        &\quad \times \delta\left(t - \sum_{i=1}^n w_i \psi_i \right) d\psi_1 \ldots d\psi_n,
    \end{align}
    showing that HHRT is a partial Radon transform applied after multiple partial generalized Radon transforms. When all PGRTs are injective, HHRT is also injective.
\end{remark}

\begin{remark}[HHRT as integral transform of generalized Busemann projection]
    When $g_1, \ldots, g_n$ are Busemann projections on manifolds $\setM_1, \ldots, \setM_n$, HHRT (Definition~\ref{def:Hierarchical_Hybrid_Radon_Transform}) corresponds to the integral transform of the generalized Busemann projection (Definition~\ref{def:generalized_buseman_projection}). However, HHRT applies beyond manifolds.
\end{remark}

Based on HHRT, we define a general sliced Wasserstein variant for product spaces:

\begin{definition}[Hierarchical Hybrid Sliced Wasserstein distance]
    \label{def:H2SW}
    For $p \geq 1$ and measures $\mu, \nu \in \setP_p(\Re^{d_1} \times \ldots \times \Re^{d_n})$, the hierarchical hybrid sliced Wasserstein-$p$ (H2SW) distance is:
    \begin{align}
        &H2SW_p^p(\mu, \nu; g_1, \ldots, g_n) \nonumber\\&= \mathbb{E}_{(w, \theta_1, \ldots, \theta_n) \sim \setU(\Sm^{n-1} \times \Omega_1 \times \ldots \times \Omega_n)} \left[ W_p^p \big( P_{w, \theta_{1:n}, g_{1:n}} \sharp \mu, P_{w, \theta_{1:n}, g_{1:n}} \sharp \nu \big) \right],
    \end{align}
    where $w = (w_1, \ldots, w_n)$ and 
   \begin{align}
        P_{w, \theta_{1:n}, g_{1:n}}(x) = \sum_{i=1}^n w_i g_i(x_i, \theta_i).
     \end{align}
\end{definition}

Theoretical properties of H2SW depend on the injectivity of HHRT. In particular:

\begin{proposition}[Metricity of Hierarchical Hybrid Sliced Wasserstein distance]
    \label{proposition:metricity_H2SW}
    If HHRT is injective, then H2SW is a valid metric.
    \begin{proof}
        See the proof in~\citet{nguyen2024hierarchical}.
    \end{proof}
\end{proposition}

For some specific cases of HHRT, the sample complexity of H2SW has been derived~\citep{nguyen2024hierarchical}.


\begin{remark}[Some special cases of H2SW]
    \label{remark:special_cases}
    The hierarchical hybrid sliced Wasserstein (H2SW) distance was initially introduced for comparing 3D meshes represented as discrete measures on $\Re^3 \times \Sm^2$. Moreover, certain instances of the product manifold sliced Wasserstein (PMSW), such as the mixed sliced Wasserstein distance (Mix-SW)~\citep{nguyen2024summarizing}, can be seen as special cases of H2SW.  Variants of H2SW also appear in the form of sliced Mixture Wasserstein (SMix-W), and for comparing mixtures of Gaussians~\citep{nguyen2024summarizing,piening2025slicing}.  By further generalization and leveraging the moment transform projection (Definition~\ref{def:moment_transform_projection}), the sliced optimal transport dataset distance (s-OTDD)~\citep{nguyen2025lightspeed} can be interpreted as a form of H2SW for measures on $\Re^{c \times d \times d} \times \setP(\Re^{c \times d \times d})$, useful for comparing datasets composed of images and their labels.
\end{remark}

\section{Numerical Estimation for Uniform Slicing}
\label{sec:MC:chapter:advances}
In previous sections, SW variants appear in the form of:
\begin{align}
\mathbb{E}_{\theta \sim \mathcal{U}(\Theta)}[W_p^p(P_\theta \sharp \mu,P_\theta \sharp \nu)],
\end{align}
where $P_\theta$ is the projection function, and $\Theta$ is the space of projection parameters. The expectation is often approximated by Monte Carlo estimation:
\begin{align}
   \mathbb{E}_{\theta \sim \mathcal{U}(\Theta)}[W_p^p(P_\theta \sharp \mu,P_\theta \sharp \nu)] \approx \frac{1}{L}\sum_{l=1}^L W_p^p(P_{\theta_l} \sharp \mu,P_{\theta_l} \sharp \nu),
\end{align}
where $\theta_1,\ldots,\theta_L \simiid \, \mathcal{U}(\Theta)$. The approximation error is the standard MC estimation error, i.e., $\mathcal{O}(L^{-1/2})$. 
In this section, we will discuss methods to improve the approximation of $SW_p^p(\theta)$. For convenience, we use the following notation:
\begin{align}
I_f(\theta)= \mathbb{E}[f(\theta)]; \quad \hat{I}_f(\theta_1,\ldots,\theta_L) = \frac{1}{L}\sum_{l=1}^L f(\theta_l),
\end{align}
when $f(\theta)= W_p^p(P_\theta \sharp \mu,P_\theta \sharp \nu)$, we have the case of SW. 
\subsection{Quasi-Monte Carlo}
\label{subsec:QMC}

In addition to standard Monte Carlo estimation, Quasi-Monte Carlo (QMC) methods can be employed for numerical integration and related problems by using low-discrepancy sequences to achieve more accurate approximations. Rather than drawing $\theta_1,\ldots,\theta_L \stackrel{\text{i.i.d.}}{\sim} \mathcal{U}(\Theta)$ as in Monte Carlo, QMC uses a deterministic set of points $\theta_1,\ldots,\theta_L \in \Theta$ that are more ``uniformly distributed" over the space.

The construction of such point sets depends on the structure of the domain $\Theta$. We begin with the traditional QMC setup, where $\Theta = [0,1]^d$. Although there is no specific SW variant that directly uses this projection space, it serves as an essential building block for the subsequent discussion. To proceed, we introduce a notion of uniformity to assess the quality of these point sets.

\begin{definition}[Star discrepancy]
    \label{def:star_discrepancy}
    A point set $\theta_1,\ldots,\theta_L$ in $[0,1]^d$ is such that $\frac{1}{L}\sum_{l=1}^L f(\theta_l) \to \mathbb{E}_{\theta \sim \mathcal{U}([0,1]^d)}[f(\theta)]$ as $L \to \infty$ for any given integrable function $f$, and aims to obtain high uniformity. To measure the uniformity, the star discrepancy~\citep{owen2013monte} has been used: 
\begin{align}
    D^*(\theta_1,\ldots,\theta_L) = \sup_{\theta \in [0,1)^d} |F_L(\theta|\theta_1,\ldots,\theta_L) - F_{\mathcal{U}([0,1]^d)}(\theta)|,
\end{align}
where $F_L(\theta|\theta_1,\ldots,\theta_L)=\frac{1}{L}\sum_{l=1}^L \mathbf{1}_{\theta_l \leq \theta}$ (the empirical CDF) and \\$F_{\mathcal{U}([0,1]^d)}(\theta) = \text{Vol}([0,\theta]) = \prod_{i=1}^d \theta_i $ is the CDF of the uniform distribution over the unit hypercube. We say the point set $\theta_1,\ldots,\theta_L$ is asymptotically uniformly distributed if $D^*(\theta_1,\ldots,\theta_L) \to 0$, since the star discrepancy is the sup-norm between the empirical CDF and the CDF of the uniform distribution.
\end{definition}

The connection between the approximation error and the star discrepancy can be described with the Koksma-Hlawka inequality.

\begin{remark}[Koksma-Hlawka inequality]
    \label{remark:KH_inequality}
    The approximation error can be bounded via the Koksma-Hlawka inequality~\citep{hlawka1961funktionen}:
\begin{align}
|\hat{I}_f(\theta_1,\ldots,\theta_L)-I_f| \leq D^*(\theta_1,\ldots,\theta_L) \, Var_{HK}(f),
\end{align}
where $Var_{HK}(f)$ is the total variation of $f$ in the sense of Hardy and Krause~\citep{niederreiter1992random}. 
\end{remark}
From the inequality, we can see that a sequence with a lower star discrepancy can lead to a lower approximation error. Therefore, we can use the discrepancy as a quality measure of a point set.

\begin{definition}[Low-discrepancy sequences on hypercube]
\label{def:QMC_sets_cube}
    The point set $\theta_1,\ldots,\theta_L$ is called a \textit{low-discrepancy sequence} if $D^*(\theta_1,\ldots,\theta_L) = \mathcal{O}(L^{-1} \log(L)^d)$. Therefore, QMC integration can achieve better approximation than its MC counterpart if $L \geq 2^{d}$, since the error rate of MC is $\mathcal{O}(L^{-1/2})$. Some constructions can be named, such as the Halton sequence~\citep{halton1964radical}, the Hammersley point set~\citep{hammersley2013monte}, the Faure sequence~\citep{faure1982discrepance}, the Niederreiter sequence~\citep{niederreiter1992random}, the Sobol sequence~\citep{sobol1967distribution}, and the sliced optimal transport sequence~\citep{paulin2020sliced}.
\end{definition}

QMC relies on deterministic low-discrepancy sequences; hence, the approximation error is hard to estimate. To address this issue, randomized QMC is introduced by randomizing low-discrepancy sequences from QMC. The key idea is to incorporate a source of randomness into the sequences while preserving their uniformity.

\begin{remark}[Randomized low-discrepancy sequences on hypercube]
\label{remark:randomized_sequences}
    Using deterministic point sets might not be optimal, and their corresponding error might be hard to estimate. To solve that, the idea is to inject randomness into a given low-discrepancy sequence. For example, we do such randomization for $\theta_1,\ldots,\theta_L$ by shifting~\citep{cranley1976randomization}, i.e., $\theta'_i = (\theta_i + U) \bmod 1$ for all $i=1,\ldots,L$ and $U \sim \mathcal{U}([0,1]^d)$. The other approach is scrambling~\citep{owen1995randomly}. In greater detail, $\theta$ is rewritten into $\theta = \sum_{k=1}^\infty b^{-k} a_k$ for base $b$ digits and $a_k \in \{0,1,\ldots,b-1\}$. After that, we permute $a_1,\ldots,a_k$ randomly to obtain the scrambled version of $\theta$. It is worth noting that the randomized samples are still uniformly distributed.
\end{remark}

We now discuss the case $\Theta = \Sm^{d-1}$, which is the standard projection space for many SW variants, e.g., SW, S3W, and so on. For $\Sm^{d-1}$, we need a different discrepancy for measuring the uniformity of point sets.

\begin{definition}[Spherical Cap Discrepancy]
\label{def:spherical_cap_discrepancy}

To measure the uniformity on the unit hypersphere, spherical cap discrepancy~\citep{brauchart2012quasi} is often used:
\begin{align}
    D^*_{\mathbb{S}^{d-1}}(\theta_1,\ldots,\theta_L) = \sup_{w \in \mathbb{S}^{d-1}, t\in[-1,1]} \left|\frac{1}{L}\sum_{l=1}^L \mathbf{1}_{\theta_l \in C(w,t)} - \sigma_0(C(w,t))\right|,
\end{align}
where $C(w,t)=\{x\in \mathbb{S}^{d-1} \mid \langle w,x \rangle \leq t \}$ is a spherical cap, and $\sigma_0$ is the uniform measure on $\Sm^{d-1}$. 
\end{definition}

It has been shown that the points $\theta_1, \ldots, \theta_L$ are asymptotically uniformly distributed if $D^*_{\mathbb{S}^{d-1}}(\theta_1, \ldots, \theta_L) \to 0$~\citep{brauchart2012quasi}. Based on the definition of the star discrepancy, one can define the concept of low-discrepancy sequences on the 2D sphere. To the best of our knowledge, existing definitions of low-discrepancy sequences are currently limited to the 2D sphere.

\begin{figure}[!t]
    \centering
    \includegraphics[width=1\linewidth]{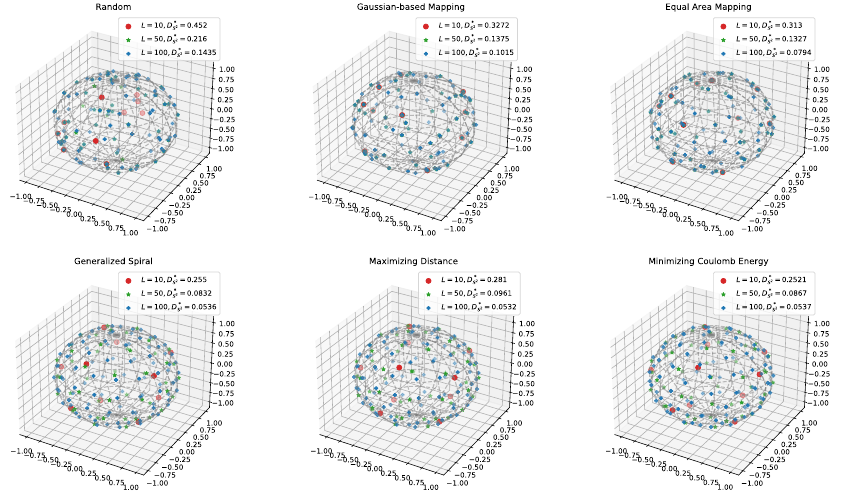}
    \caption{Potential low-discrepancy point sets on 2D sphere (Figure 6 in~\citet{nguyen2024quasimonte}).}
    \label{fig:QMC}
\end{figure}

\begin{remark}[Low-discrepancy sequences on 2D sphere]
\label{remark:low_discrepancy_sequence_sphere}

A point set $\theta_1,\ldots,\theta_L$ is called a low-discrepancy sequence on $\mathbb{S}^2$ if $D^*_{\mathbb{S}^{2}}(\theta_1,\ldots,\theta_L) \in \mathcal{O}(L^{-3/4} \sqrt{\log (L)})$. It is worth noting that there is no KH inequality in this case. However, a lower spherical cap discrepancy can lead to a better worst-case error~\citep{brauchart2012quasi,brauchart2014qmc} for some functions belonging to suitable Sobolev spaces. There are some practical ways to construct potential point sets that can be low-discrepancy sequences:

\textbf{Mapping from the hypercube.}  We transform a low-discrepancy sequence $x_1,\ldots,x_L$ on $[0,1]^d$ to a potentially low-discrepancy sequence $\theta_1,\ldots,\theta_L$ on $\mathbb{S}^{d-1}$ through the mapping. For example, we can use $\theta=f(x) =  \Phi^{-1}(x)/\|\Phi^{-1}(x)\|_2$, where $\Phi^{-1}$ is the inverse CDF of $\mathcal{N}(0,1)$ (entry-wise). This technique is mentioned in~\citep{basu2016quasi} and can be used in any dimension. Another option on $\Sm^2$ is to use equal area mapping. For example, we can use the Lambert cylindrical mapping $f(x,y)=(2\sqrt{y-y^2}\cos(2\pi x), 2\sqrt{y-y^2} \sin (2\pi x), 1-2y)$, which can generate a low-discrepancy asymptotically uniform sequence~\citep{aistleitner2012point}.  

\textbf{Direct construction.}  On $\Sm^2$, we can explicitly construct a set of $L$ points with spherical coordinates  $(\phi_1,\phi_2)$~\citep{rakhmanov1994minimal}: $z_i = 1-\frac{2i-1}{L}, \phi_{i1} = \cos^{-1}(z_i), \phi_{i2} = 1.8 \sqrt{L} \, \phi_{i1} \mod 2\pi$ for $i=1,\ldots,L$. The Euclidean coordinates can be obtained by using the mapping $(\phi_1,\phi_2) \mapsto (\sin(\phi_1)\cos(\phi_2), \sin(\phi_1)\sin(\phi_2), \cos(\phi_1))$. The point set is called a generalized spiral and is an asymptotically uniform sequence~\citep{hardin2016comparison} and can achieve optimal worst-case integration error~\citep{brauchart2014qmc} for properly defined Sobolev integrands empirically. Another way to construct a point set is by using optimization. For example, in~\citet{brauchart2014qmc,hardin2016comparison}, the authors suggest choosing a point set $\theta_1,\ldots,\theta_L$ which maximizes the distance $\sum_{i=1}^L \sum_{j=1}^L |\theta_i - \theta_j|$ or minimizes the Coulomb energy $\sum_{i=1}^L \sum_{j=1}^L \frac{1}{|\theta_i - \theta_j|}$ to create a potentially low-discrepancy sequence. They are empirically shown to achieve optimal worst-case error~\citep{brauchart2014qmc}. Moreover, minimizing the Coulomb energy is proven to create an asymptotically uniform sequence~\citep{gotz2000distribution} and can achieve the error of $\mathcal{O}(L^{-1})$ on $\Sm$~\citep{sisouk2025user}. Another optimization objective is sliced Wasserstein variants in Section~\ref{subsec:spherical_projections}~\citep{genest2024non}. Finally, there are also orthogonal constructions~\citep{feng2023determinantal,rowland2019orthogonal,lin2020demystifying}, i.e., sampling random orthonormal bases.  We visualize the discussed point sets in Figure~\ref{fig:QMC} (Figure 6 in~\citet{nguyen2024quasimonte}).
\end{remark}

\begin{figure}[!t]
    \centering
    \includegraphics[width=1\linewidth]{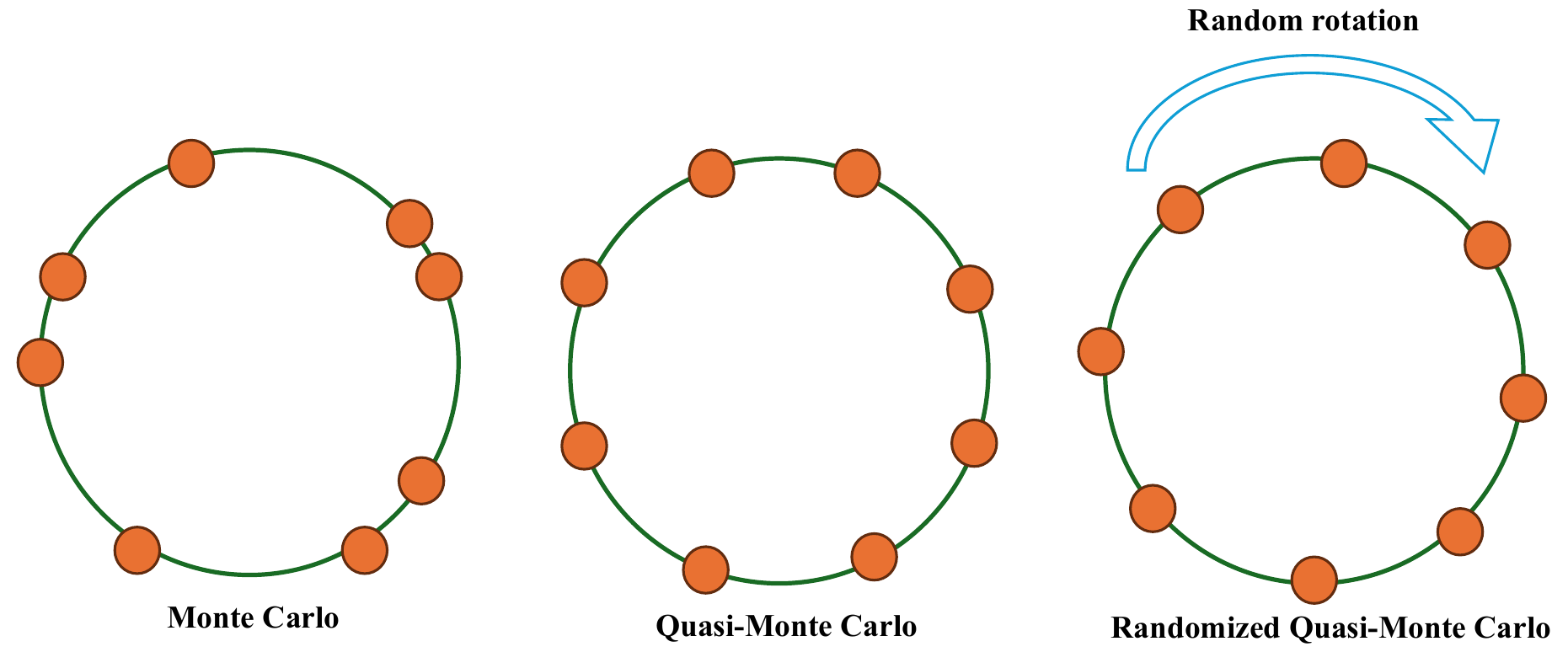}
    \caption{Monte Carlo, Quasi-Monte Carlo, and Randomized Quasi-Monte Carlo (with random rotation) on a sphere.}
    \label{fig:QMC2D}
\end{figure}

Similar to the case of the unit cube, we can also randomize low-discrepancy sequences on the hypersphere. 

\begin{remark}[Randomized Low-discrepancy sequences on hypersphere]
\label{remark:randomized_low_discrepancy_sequence_sphere}
    From~\citet{nguyen2024quasimonte}, there are two ways of randomizing:

    \textbf{Mapping from the hypercube.} Given a randomized low-discrepancy point set $x_1',\ldots,x_L'$ on the unit cube (unit grid), we can map them to $\theta_1',\ldots,\theta_L'$ as discussed in Remark~\ref{remark:low_discrepancy_sequence_sphere}.

    \textbf{Random rotation.} Given a point set $\theta_1,\ldots,\theta_L$ on the unit hypersphere $\mathbb{S}^{d-1}$, we can apply a uniform random rotation as randomization. We form the new sequence $\theta_1',\ldots,\theta_L'$ with $\theta_i' = U \theta_i$ for all $i=1,\ldots,L$ where $U \sim \mathcal{U}(\mathbb{V}_d(\mathbb{R}^d))$, and $\mathbb{V}_d(\mathbb{R}^d)$ is the Stiefel manifold. The sampling can be done by applying the Gram-Schmidt orthogonalization process to $z_1,\ldots,z_L \overset{\text{iid}}{\sim} \mathcal{N}(0,I_d)$ (Bartlett decomposition theorem~\citep{muirhead2009aspects}). We demonstrate the idea in Figure~\ref{fig:QMC2D}.
\end{remark}

For other projection spaces, there is no well-known notion of low-discrepancy sequences. However, we can still construct "low-discrepancy" point sets by transforming sequences from $[0,1]^d$ or $\Sm^{d-1}$.

\begin{remark}[Low-discrepancy sequence for general projection spaces]
\label{remark:low_discrepancy_sequence_general}
    Some SW variants might require different projection spaces. While constructing low-discrepancy sequences for general spaces is still an open area~\citep{basu2016quasi}, there are two practical approaches: mapping low-discrepancy sequences from $[0,1]^d$ or $\Sm^{d-1}$, and direct construction using optimization with uniformity objectives (repulsive objectives). 
\end{remark}

To form an estimation of SW variants with low-discrepancy sequences, we just need to use them as a replacement for conventional Monte Carlo samples. For example, QMC estimation of SW can be defined as follows:

\begin{definition}[Quasi Monte Carlo for sliced Wasserstein variants]
\label{def:QMC_SW}
Given a (randomized) low-discrepancy sequence $\theta_1,\ldots,\theta_L$, a SW variant can be approximated (estimated) as:
\begin{align}
\widehat{QSW}_p^p(\mu,\nu;\theta_1,\ldots,\theta_L) =  \frac{1}{L}\sum_{l=1}^L W_p^p(P_{\theta_l}\sharp \mu,P_{\theta_l} \sharp \nu).
\end{align}
\end{definition}
The approximation/estimation  $\widehat{QSW}_p^p(\mu,\nu;\theta_1,\ldots,\theta_L)$ often asymptotically converges to the corresponding sliced Wasserstein variant for deterministic sequences and is an unbiased estimate for randomized sequences~\citep{nguyen2024quasimonte}.

Similar to the idea of QMC, \citet{petrovic2025repulsive} propose using determinantal point processes (DPPs)~\citep{kulesza2012determinantal} and repulsive point processes to generate random point sets for estimating the SW distance. However, as they point out, QMC methods still achieve better estimation performance in relatively low dimensions.

\subsection{Control Variates}
\label{subsec:control_variates}
We remind that our problem is to estimate $I_f(\theta) = \mathbb{E}[f(\theta)]$ with $\hat{I}_f(\theta_1, \ldots, \theta_L) = \frac{1}{L} \sum_{l=1}^L f(\theta_l)$ for $f(\theta) = W_p^p(P_\theta \sharp \mu, P_\theta \sharp \nu)$ in the case of SW. From Remark~\ref{remark:MC_SW}, we know that the approximation error is bounded by the variance $\mathrm{Var}[f(\theta)]$. Therefore, if we can replace $f(\theta)$ with another integrand that has the same expectation but lower variance, we can achieve a more accurate Monte Carlo estimate. There is a principled way to construct such an integrand, which is known as control variate. In this section, we review some ways to construct control variates for SW. We first start with the definition of control variate and controlled variable, and explain why they have lower variance.

\begin{definition}[Control variate and controlled variable]
    \label{def:control_variate}
    A \textit{control variate}~\citep{owen2013monte} is a random variable $C(\theta)$ with tractable expectation $\mathbb{E}[C(\theta)] = B$. With the control variate, we can construct the controlled variable:
    \begin{align}
        f_{C,\gamma}(\theta) = f(\theta)- \gamma (C(\theta)-B),
    \end{align}
    where $\gamma \in \mathbb{R}$. 
\end{definition}

\begin{remark}[Variance reduction with control variate]
\label{remark:variance_reduction_control_variate}
As the error of MC estimation depends on $\mathrm{Var}[f(\theta)]$ (Remark~\ref{remark:MC_SW}), we obtain a lower error if we have lower $\mathrm{Var}[f(\theta)]$. We can check that:
\begin{align}
    \mathrm{Var}[f_{C,\gamma}(\theta)] = \mathrm{Var}[f(\theta)] - 2 \gamma \mathrm{Cov}[f(\theta), C(\theta)] + \gamma^2 \mathrm{Var}[C(\theta)].
\end{align}
Optimizing for $\gamma$ to minimize $\mathrm{Var}[f_{C,\gamma}(\theta)]$, we have $\gamma^\star =  \frac{\mathrm{Cov}[f(\theta), C(\theta)]}{\mathrm{Var}[C(\theta)]}$. With $\gamma^\star$, we have:
\begin{align}
    \mathrm{Var}[f_{C,\gamma^\star}(\theta)] = \mathrm{Var}[f(\theta)]\left(1 - \frac{\mathrm{Cov}[f(\theta), C(\theta)]^2}{\mathrm{Var}[f(\theta)] \mathrm{Var}[C(\theta)]}\right),
\end{align}
which is always lower than $\mathrm{Var}[f(\theta)]$ and is especially lower when $f(\theta)$ has high correlation with $C(\theta)$. In practice, $\gamma^\star$ can be estimated via Monte Carlo methods if it is intractable.
\end{remark}

We now discuss some constructions of control variates when $\theta \sim \setU(\Sm^{d-1})$, which is the case for many SW variants. The first construction is based on Gaussian approximation. We first review a special case of the Wasserstein-2 distance between two Gaussians.

\begin{remark}[Wasserstein-2 distance between Gaussians]
\label{remark:Wasserstein_Gaussians}
Given two Gaussian distributions $\mu = \mathcal{N}(\mathbf{m}_1,\Sigma_1)$ and $\nu = \mathcal{N}(\mathbf{m}_2,\Sigma_2)$, from~\citet{dowson1982frechet}, we have:
\begin{align}
    W_2^2(\mu,\nu) = \|\mathbf{m}_1 - \mathbf{m}_2\|_2^2 + \mathrm{Trace}\left(\Sigma_1 + \Sigma_2 - 2\left(\Sigma_1^{1/2} \Sigma_2 \Sigma_1^{1/2}\right)^{1/2}\right).
\end{align}
In one dimension, i.e., $\mu = \mathcal{N}(m_1,\sigma_1^2)$ and $\nu = \mathcal{N}(m_2,\sigma_2^2)$, we have a special case:
\begin{align}
     W_2^2(\mu,\nu) = (m_1 - m_2)^2 + (\sigma_1 - \sigma_2)^2.
\end{align}
\end{remark}

We now discuss how to use this closed form to construct control variates for enhancing MC estimation for SW.

\begin{remark}[Control variates with Gaussian approximation]
\label{remark:control_variates_Gaussian_approximation}
In order to make $C(\theta)$ correlated with $f(\theta) = W_p^p(P_\theta \sharp \mu, P_\theta \sharp \nu)$, authors in~\citet{nguyen2024sliced} propose to approximate $P_\theta \sharp \mu$ and $P_\theta \sharp \nu$ by univariate Gaussians (Laplace approximation, moment methods, and so on). Let $\mathcal{N}(m_{\theta,\mu}, \sigma^2_{\theta,\mu})$ and $\mathcal{N}(m_{\theta,\nu}, \sigma^2_{\theta,\nu})$ be the corresponding Gaussian approximations of $P_\theta \sharp \mu$ and $P_\theta \sharp \nu$. A control variate can be constructed as:
\begin{align}
    C(\theta) &= W_2^2(\mathcal{N}(m_{\theta,\mu}, \sigma^2_{\theta,\mu}), \mathcal{N}(m_{\theta,\nu}, \sigma^2_{\theta,\nu})) \\
    &= (m_{\theta,\mu} - m_{\theta,\nu})^2 + (\sigma_{\theta,\mu} - \sigma_{\theta,\nu})^2. \nonumber
\end{align}
When $P_\theta(x) = \langle \theta, x \rangle$, $\mu = \sum_{i=1}^n \alpha_i \delta_{x_i}$, and $\nu = \sum_{i=1}^m \beta_i \delta_{y_i}$ are discrete measures, we have:
\begin{align}
    &m_{\theta,\mu} = \sum_{i=1}^n \alpha_i \theta^\top x_i, \quad m_{\theta,\nu} = \sum_{i=1}^m \beta_i \theta^\top y_i, \\
    &\sigma^2_{\theta,\mu} = \sum_{i=1}^n \alpha_i \left(\theta^\top x_i - m_{\theta,\mu}\right)^2, \quad \sigma^2_{\theta,\nu} = \sum_{j=1}^m \beta_j \left(\theta^\top y_j - m_{\theta,\nu}\right)^2.
\end{align}
Utilizing the fact that $\mathbb{E}[\theta \theta^\top] = (1/d) I_d$ for $\theta \sim \setU(\Sm^{d-1})$~\citep{nadjahi2021fast,nguyen2024sliced}, we can obtain:
\begin{align}
    &\mathbb{E}[(m_{\theta,\mu} - m_{\theta,\nu})^2] = \frac{1}{d} \left\| \sum_{i=1}^n \alpha_i x_i - \sum_{i=1}^m \beta_i y_i \right\|^2, \\
    &\mathbb{E}[\sigma^2_{\theta,\mu}] = \frac{1}{d} \sum_{i=1}^n \alpha_i \left\| x_i - \sum_{i'=1}^n \alpha_{i'} x_{i'} \right\|^2, \\
    &\mathbb{E}[\sigma^2_{\theta,\nu}] = \frac{1}{d} \sum_{i=1}^m \beta_i \left\| y_i - \sum_{i'=1}^m \beta_{i'} y_{i'} \right\|^2.
\end{align}

Since $\mathbb{E}[\sigma_{\theta,\mu} \sigma_{\theta,\nu}]$ is intractable, authors in~\citet{nguyen2024sliced} propose two alternative control variates that are lower and upper bounds of $C(\theta)$:
\begin{align}
    &C_{low}(\theta) = (m_{\theta,\mu} - m_{\theta,\nu})^2, \\
    &C_{up}(\theta) = (m_{\theta,\mu} - m_{\theta,\nu})^2 + \sigma^2_{\theta,\mu} + \sigma^2_{\theta,\nu},
\end{align}
where 
\[
C_{low}(\theta) \leq W_2^2(\mathcal{N}(m_{\theta,\mu}, \sigma^2_{\theta,\mu}), \mathcal{N}(m_{\theta,\nu}, \sigma^2_{\theta,\nu})) \leq C_{up}(\theta).
\]
\end{remark}

We can use more than one control variate to capture more information about the integrand, i.e., multiple control variates. 

\begin{remark}[Multiple control variates and linear regression]
    \label{remark:multiple_control_variate}
    From~\citet{owen2013monte}, without loss of generality, we use more than one control variate, i.e., $C_1(\theta), \ldots, C_n(\theta)$, where we can assume $\mathbb{E}[C_i(\theta)] = 0$ for $i=1,\ldots,n$ (if $\mathbb{E}[C_i(\theta)] = B_i$, set $C'_i(\theta) = C_i(\theta) - B_i$). Let $\mathbf{C}(\theta) = (C_1(\theta), \ldots, C_n(\theta))^\top$ and $\boldsymbol{\gamma} = (\gamma_1,\ldots,\gamma_n)^\top$, then our controlled variable can be defined as:
    \begin{align}
        f_{\mathbf{C}}(\theta) = f(\theta) - \boldsymbol{\gamma}^\top \mathbf{C}(\theta).
    \end{align}
    We can compute the variance of $f_{\mathbf{C}}(\theta)$ as:
    \begin{align}
        \mathrm{Var}[f_{\mathbf{C}}(\theta)] &= \mathbb{E}\left[(f(\theta) - \boldsymbol{\gamma}^\top \mathbf{C}(\theta))^2\right] - \left(\mathbb{E}[f(\theta) - \boldsymbol{\gamma}^\top \mathbf{C}(\theta)]\right)^2 \\
        &= \mathbb{E}\left[(f(\theta) - \boldsymbol{\gamma}^\top \mathbf{C}(\theta))^2\right] - \mathbb{E}[f(\theta)]^2,
    \end{align}
    hence
    \begin{align}
        \min_{\boldsymbol{\gamma}} \mathrm{Var}[f_{\mathbf{C}}(\theta)] = \min_{\boldsymbol{\gamma}} \mathbb{E}\left[(f(\theta) - \boldsymbol{\gamma}^\top \mathbf{C}(\theta))^2\right],
    \end{align}
    which is a least squares estimation problem. With samples $\theta_1,\ldots,\theta_L \overset{\mathrm{iid}}{\sim} \setU(\Sm^{d-1})$ and vectors $\boldsymbol{f}(L) = (f(\theta_1), \ldots, f(\theta_L))^\top$ and matrix $\mathbf{C}(L) = (\mathbf{C}(\theta_1), \ldots, \mathbf{C}(\theta_L))^\top$, the optimal $\boldsymbol{\gamma}$ is:
    \begin{align}
        \boldsymbol{\gamma}^\star_L = \big(\mathbf{C}(L)^\top \mathbf{C}(L)\big)^{-1} \mathbf{C}(L)^\top \boldsymbol{f}(L),
    \end{align}
    which is the well-known least squares estimator for linear regression.
\end{remark}

Next, we review a recent approach for constructing control variates using spherical harmonics. We first discuss the definition of spherical harmonics and a related concept, i.e., polynomial spaces.

\begin{definition}[Polynomial spaces]
\label{def:polynomial_space}
From~\citet{leluc2024sliced}, let $\mathcal{P}^d_\ell$ be the space of homogeneous polynomials of degree $\ell \geq 0$ on $\mathbb{R}^d$, i.e.,
\begin{align}
\mathcal{P}^d_\ell = \mathrm{Span} \left\{ x_1^{a_1} \cdots x_d^{a_d} \mid a_k \in \mathbb{N},\ \sum_{k=1}^d a_k = \ell \right\}.
\end{align}
Let $\mathcal{H}^d_\ell$ be the space of real harmonic polynomials that are homogeneous of degree $\ell$ on $\mathbb{R}^d$, i.e.,
\[
\mathcal{H}^d_\ell = \left\{ Q \in \mathcal{P}^d_\ell \mid \Delta Q = 0 \right\},
\]
where $\Delta$ denotes the Laplace operator.
\end{definition}

\begin{definition}[Spherical harmonics]
    \label{def:spherical_harmonic}
    Spherical harmonics~\citep{atkinson2012spherical,dai2013approximation,muller2012analysis,leluc2024sliced} of degree (or level) $\ell \geq 0$ are defined as the restriction of elements in $\mathcal{H}^d_\ell$ to the unit sphere $\mathbb{S}^{d-1}$; i.e., they are the restrictions to the sphere of harmonic homogeneous polynomials in $d$ variables of degree $\ell$. Let $N^d_\ell$ denote the number of linearly independent spherical harmonics of degree $\ell$ in dimension $d$, that is,
\[
N^d_\ell = \dim \mathcal{H}^d_\ell = \dim \mathcal{P}^d_\ell - \dim \mathcal{P}^d_{\ell-2},
\]
where it is agreed that $\dim \mathcal{P}^d_{\ell-2} = 0$ for $\ell = 0, 1$. The explicit formula is:
\begin{align}
    N^d_\ell = \frac{(2\ell + d - 2)(\ell + d - 3)!}{\ell! \, (d - 2)!}.
\end{align}

Let $\{\varphi_{\ell,k} : \ell \geq 0,\, 1 \leq k \leq N^d_\ell \}$ denote the set of spherical harmonics on $\mathbb{S}^{d-1}$. Then $\{\varphi_{\ell,k}\}$ form an orthonormal eigenbasis of the Hilbert space $L^2(\mathbb{S}^{d-1}, \sigma)$, where $\sigma$ is the uniform probability measure on $\mathbb{S}^{d-1}$:
\begin{align}
f(\theta) = \sum_{\ell=0}^{\infty} \sum_{k=1}^{N^d_\ell} \widehat{f}_{\ell,k} \, \varphi_{\ell,k}(\theta), \quad \forall f \in L^2(\mathbb{S}^{d-1}),
\end{align}
where the Fourier coefficients are given by
\[
\widehat{f}_{\ell,k} = \int_{\mathbb{S}^{d-1}} f(\theta) \varphi_{\ell,k}(\theta) \, d\sigma(\theta).
\]
\end{definition}

For all $\ell \geq 1$ and $1 \leq k \leq N^d_\ell$, we have
\[
\int_{\mathbb{S}^{d-1}} \varphi_{\ell,k}(\theta) \, d\sigma(\theta) = \mathbb{E}_{\theta \sim \setU(\Sm^{d-1})}[\varphi_{\ell,k}(\theta)] = 0,
\]
which is the criterion for being a control variate.

\begin{remark}[Control variates with spherical harmonics]
\label{remark:control_variates_spherical_harmonics}
For the case of SW, the function $f(\theta) = W_p^p(P_\theta \sharp \mu, P_\theta \sharp \nu)$ is often even, i.e., $f(\theta) = f(-\theta)$, hence it is composed only of spherical harmonics with even degree. Therefore, the number of spherical harmonics of even degree up to maximal degree $2M$ is:
\begin{align}
    s_{M,d} = \sum_{\ell=1}^M N_{2\ell}^d = {2M + d - 1 \choose d - 1} - 1.
\end{align}
Authors in~\citet{leluc2024sliced} propose to use $\varphi_1(\theta), \ldots, \varphi_{s_{M,d}}(\theta)$ as the control variates for constructing the controlled variable as in Remark~\ref{remark:multiple_control_variate}. The approximation error is $\mathcal{O}(M^{-1/2} L^{-1})$ or $\mathcal{O}(\ell_n L^{-\frac{1}{2} \frac{d}{d-1}})$ where $\ell_n > 0$ diverges to infinity but can do so arbitrarily slowly. Therefore, it yields better approximation than standard MC.
\end{remark}

\begin{remark}[Control variates for general spaces]
\label{remark:control_variates_general}
The control variates discussed above are constructed under the assumption that $\theta \sim \setU(\Sm^{d-1})$. When $\theta$ lies in a different projection space, one can still apply a suitable transformation to map $\theta$ back to the unit sphere $\Sm^{d-1}$ and then utilize the same control variates defined on the sphere. However, it should be noted that the resulting variance reduction and improvement in approximation accuracy might not be guaranteed in this more general setting.
\end{remark}

\begin{remark}[Other variance reduction techniques]
\label{remark:other_variance_reduction}
In addition to control variates, other variance reduction techniques can be employed to improve Monte Carlo estimations for sliced Wasserstein variants. For instance, \citet{leluc2025speeding} propose a control neighbor estimator applicable to SW. Furthermore, Orthogonal Monte Carlo~\citep{lin2020demystifying} can also be interpreted as a variance reduction method motivated by stratification principles. \citet{petrovic2025repulsive} propose using determinantal point processes (DPPs)~\citep{kulesza2012determinantal} and repulsive point processes to reduce variance through negative dependence. Finally, \citet{boss2025reswd} use reservoir sampling to achieve variance reduction.
\end{remark}

While the discussed techniques offer practical benefits, a theoretical understanding of the approximation error has only been established for control variates with spherical harmonics (Remark~\ref{remark:control_variates_spherical_harmonics}).

    

\subsection{Fast Deterministic Approximation}
\label{subse:fast_approximation}
In this section, we review an alternative approach for approximating SW. Instead of relying on numerical methods to estimate the intractable expectation, we derive a fast deterministic approximation of SW that does not require computing projections or solving one-dimensional optimal transport problems. This approximation is based on the concentration properties of Gaussian projections. We begin by defining a variant of SW where the projection directions are sampled from a Gaussian distribution.

\begin{definition}[Sliced Wasserstein with Gaussian projections]
For $p \geq 1$, the sliced Wasserstein distance with Gaussian projections~\citep{nadjahi2021fast} between $\mu \in \setP_p(\Re^d)$ and $\nu \in \setP_p(\Re^d)$ is defined as:
\begin{align}
    SWG_{p}^p (\mu,\nu) =  \mathbb{E}_{\theta \sim \mathcal{N}(0,I_d/d)}[W_p^p(\theta \sharp \mu,\theta \sharp \nu)],
\end{align}
where $\theta \sharp \mu$ denotes the Radon conditional measure or the push-forward measure of $\mu$ through $P_\theta(x) = \langle x,\theta\rangle$.
\end{definition}

\begin{remark}[Connection of SWG to SW]
\label{remark:connection_SWG_SW}
From \citet[Proposition 1]{nadjahi2021fast}, we have:
\begin{align}
    SWG_p(\mu,\nu) =  (2/d)^{1/2} \left(\frac{\Gamma(d/2+p/2)}{\Gamma(d/2)}\right)^{1/p} SW_p(\mu,\nu),
\end{align}
where $\Gamma$ is the Gamma function. This means that SWG and SW are identical when $p=2$.
\end{remark}

In addition, there is an interesting property of the $SW_2$ distance, i.e., it can be centered:

\begin{remark}[Centered sliced Wasserstein]
    \label{remark:centered_SW}
    Given any $\mu, \nu \in \setP_2(\Re^d)$, from~\citet[Proposition 2]{nadjahi2021fast}, we have:
    \begin{align}
        SW_2^2(\mu,\nu) = SW_2^2(\bar{\mu},\bar{\nu}) + \frac{1}{d} \left\| \int_{\Re^d} x \diff \mu(x) - \int_{\Re^d} y \diff \nu(y) \right\|_2^2,
    \end{align}
    where $\bar{\mu} = f \sharp \mu$ ($f(x)= x - \int_{\Re^d} x' \diff \mu(x')$) is the centered version of $\mu$, and similarly for $\bar{\nu}$.
\end{remark}

We now review the key result for constructing the approximation, i.e., concentration of Gaussian projections.

\begin{remark}[Concentration of Gaussian projections]
    \label{remark:concentration_Gaussian_projection}
    From~\citet[Theorem 1]{reeves2017conditional}, there exists a universal constant $C>0$ such that:
    \begin{align}
        \mathbb{E}_{\theta \sim \mathcal{N}(0,I_d/d)} [W_2^2(\theta \sharp \mu, \mathcal{N}(0,d^{-1} m_{2,\mu}))] \leq C E_{\mu},
    \end{align}
    where  
    \begin{align}
        &m_{2,\mu} =  \int_{\Re^d} \|x\|_2^2 \diff \mu(x),  \\
        &E_{\mu} = d^{-1} \bigl(\alpha_\mu + (m_{2,\mu} \beta_{1,\mu})^{1/2} + m_{2,\mu}^{1/5} \beta_{2,\mu}^{4/5}\bigr),\\
        &\alpha_{\mu} =  \int_{\Re^d} (\|x\|_2^2 - m_{2,\mu}) \diff \mu(x), \\
        &\beta_{q,\mu}^q =  \int_{\Re^d} \int_{\Re^d}  (\langle x,x'\rangle)^q \diff \mu(x) \diff \mu(x').
    \end{align}
\end{remark}

From Remark~\ref{remark:concentration_Gaussian_projection}, we can approximate $\theta \sharp \mu$ by a Gaussian distribution which does not depend on $\theta$. By doing the same for $\theta \sharp \nu$, the SW distance can be approximated by the Wasserstein distance between two Gaussians, which has a closed form (Remark~\ref{remark:Wasserstein_Gaussians}). By replacing $SW_2^2(\bar{\mu},\bar{\nu})$ in Remark~\ref{remark:centered_SW} with the discussed Wasserstein distance between the corresponding Gaussians, we obtain the final fast approximation of SW with a guarantee on the approximation error.

\begin{definition}[Fast approximation of sliced Wasserstein]
\label{def:fast_SW}
For $p \geq 1$, the fast approximation of sliced Wasserstein distance between $\mu \in \setP_2(\Re^d)$ and $\nu \in \setP_2(\Re^d)$ is:
\begin{align}
    \widehat{SW}_2^2(\mu,\nu) &= \frac{1}{d}\left\| \int_{\Re^d} x \diff \mu(x) - \int_{\Re^d} y \diff \nu(y)\right\|_2^2 \nonumber \\
    & \quad + \frac{1}{d} \left(\sqrt{m_{2,\bar{\mu}}} - \sqrt{m_{2,\bar{\nu}}}\right)^2,
\end{align}
where $\bar{\mu}$ and $m_{2,\bar{\nu}}$ are defined in Remark~\ref{remark:concentration_Gaussian_projection} and Remark~\ref{remark:centered_SW}.
\end{definition}

\begin{proposition}[Approximation error for fast approximation]
\label{proposition:fast_approximation_error} 
Given $\mu, \nu \in \setP_2(\Re^d)$, we have the following bound:
\begin{align}
    \left|\widehat{SW}_2(\mu,\nu) - SW_2(\mu,\nu)\right| \leq C \sqrt{E_{\bar{\mu}} + E_{\bar{\nu}}},
\end{align}
where $E_{\bar{\mu}}$ is defined in Remark~\ref{remark:concentration_Gaussian_projection}. This follows from~\citet[Theorem 2]{nadjahi2021fast}, which is derived using the triangle inequality and Remark~\ref{remark:concentration_Gaussian_projection}. With additional assumptions~\citep[Corollary 1]{nadjahi2021fast}, the error is shown to be $\mathcal{O}(d^{-1/8})$.
\end{proposition}

\begin{remark}[Computational complexity of fast approximation for discrete measures]
\label{remark:complexity_fast_discrete}
When $\mu$ and $\nu$ are discrete measures with at most $n$ supports, the time complexity of the fast approximation is $\mathcal{O}(d n)$, which corresponds to computing the means and second moments.
\end{remark}

\begin{remark}[Fast approximation for generalized sliced Wasserstein]
\label{remark:fast_GSW}
Authors in~\citet{le2024fast} propose a fast approximation of GSW with polynomial defining functions by casting it into SW between pushforward measures.
\end{remark}

\section{One-dimensional Optimal Transport Estimation}
\label{sec:quantile:chapter:advances}
\begin{figure}[!t]
    \centering
    \begin{tabular}{c}
         \includegraphics[width=1\linewidth]{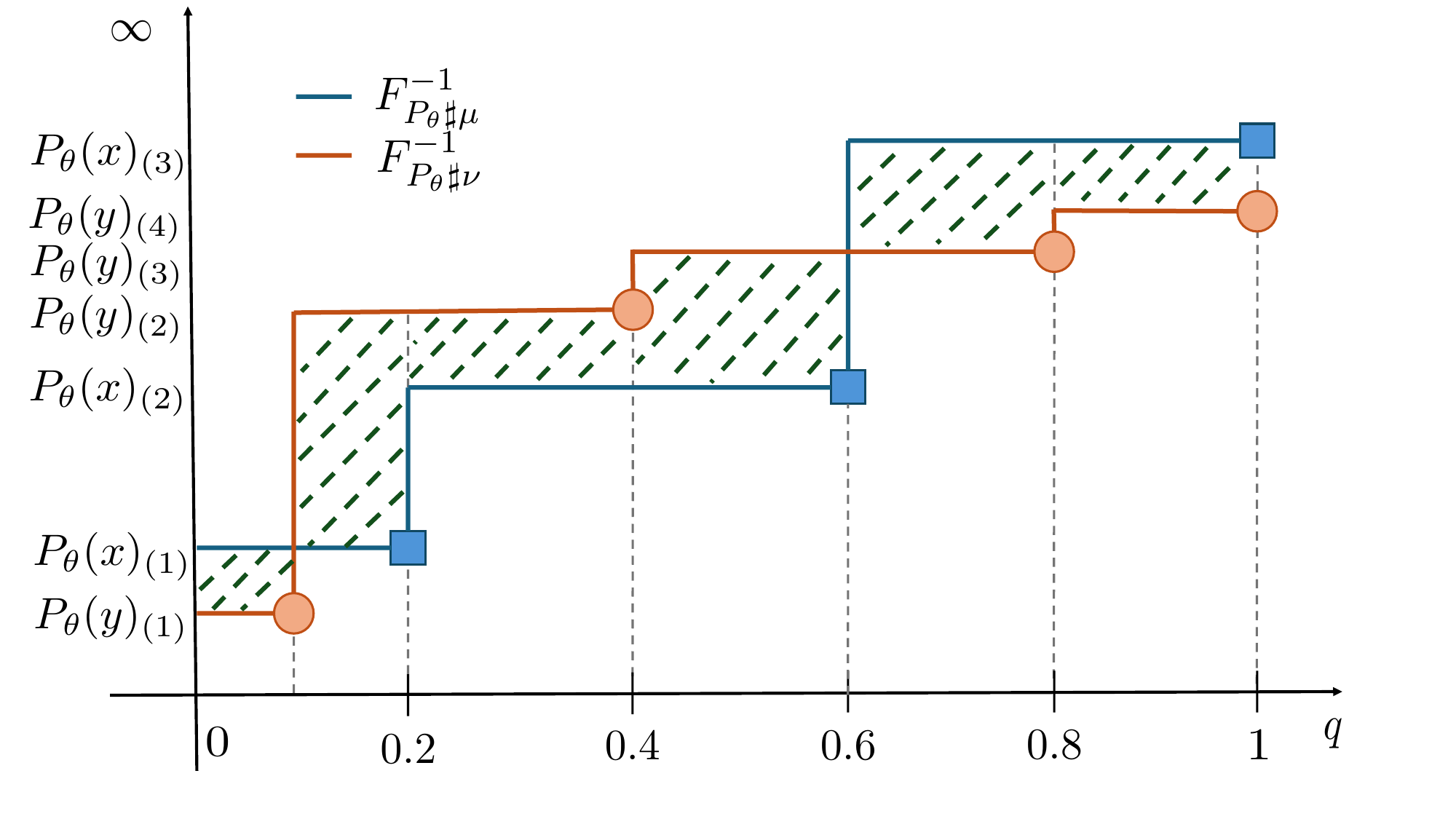}
    \end{tabular}
    \caption{Empirical quantile functions of projected distributions.} 
    \label{fig:projected_empirical_quantile}
\end{figure}

We know that computing SOT relies on computing one-dimensional optimal distances between projected measures. From Remark~\ref{remark:1DWasserstein_general}, everything boils down to computing the inverse CDFs (quantile functions) and CDFs. In the case of empirical or discrete measures, i.e.,
$P_\theta \sharp \mu = \sum_{i=1}^n \alpha_i \delta_{P_\theta(x_i)}$ and $P_\theta \sharp \nu = \sum_{j=1}^m \beta_j \delta_{P_\theta(y_j)}$, quantile functions of $\mu$ and $\nu$ can be written as follows (Remark~\ref{remark:1DWasserstein_as_quantile_approximation}):
\begin{align*}
    &F_{P_\theta \sharp \mu}^{-1}(t) =  \sum_{i=1}^n P_\theta(x)_{(i)} \, I\left(\sum_{j=1}^{i-1} \alpha_{(j)} \leq t \leq \sum_{j=1}^{i} \alpha_{(j)}\right),
    \\& F_{P_\theta \sharp \nu}^{-1}(t) =  \sum_{j=1}^m P_\theta(y)_{(j)} \, I\left(\sum_{i=1}^{j-1} \beta_{(i)} \leq t \leq \sum_{i=1}^{j} \beta_{(i)}\right),
\end{align*}
where $P_\theta(x)_{(1)} \leq \ldots \leq P_\theta(x)_{(n)}$ and $P_\theta(y)_{(1)} \leq \ldots \leq P_\theta(y)_{(m)}$ are the sorted supports (or order statistics). Similarly, the CDFs can be written as follows:
\begin{align*}
    &F_{P_\theta \sharp \mu}(z) =  \sum_{i=1}^n \alpha_i I(P_\theta(x_i) \leq z),
    \\& F_{P_\theta \sharp \nu}(z) =  \sum_{j=1}^m \beta_j I(P_\theta(y_j) \leq z),
\end{align*}
For a special case with multivariate Gaussian and linear projection, we can obtain closed-form expressions for the CDFs and quantile functions of projected distributions. 
When $\mu$ and $\nu$ are unknown (or have intractable quantile functions and CDFs) and we observe (or can sample) from them, we can use empirical CDFs and empirical quantiles (Figure~\ref{fig:projected_empirical_quantile}) as approximations~\citep{shorack2009empirical} (the estimation rate of the one-dimensional map is explored in~\citep{ponnoprat2024uniform}).

\begin{figure}[!t]
    \centering
    \begin{tabular}{c}
         \includegraphics[width=1\linewidth]{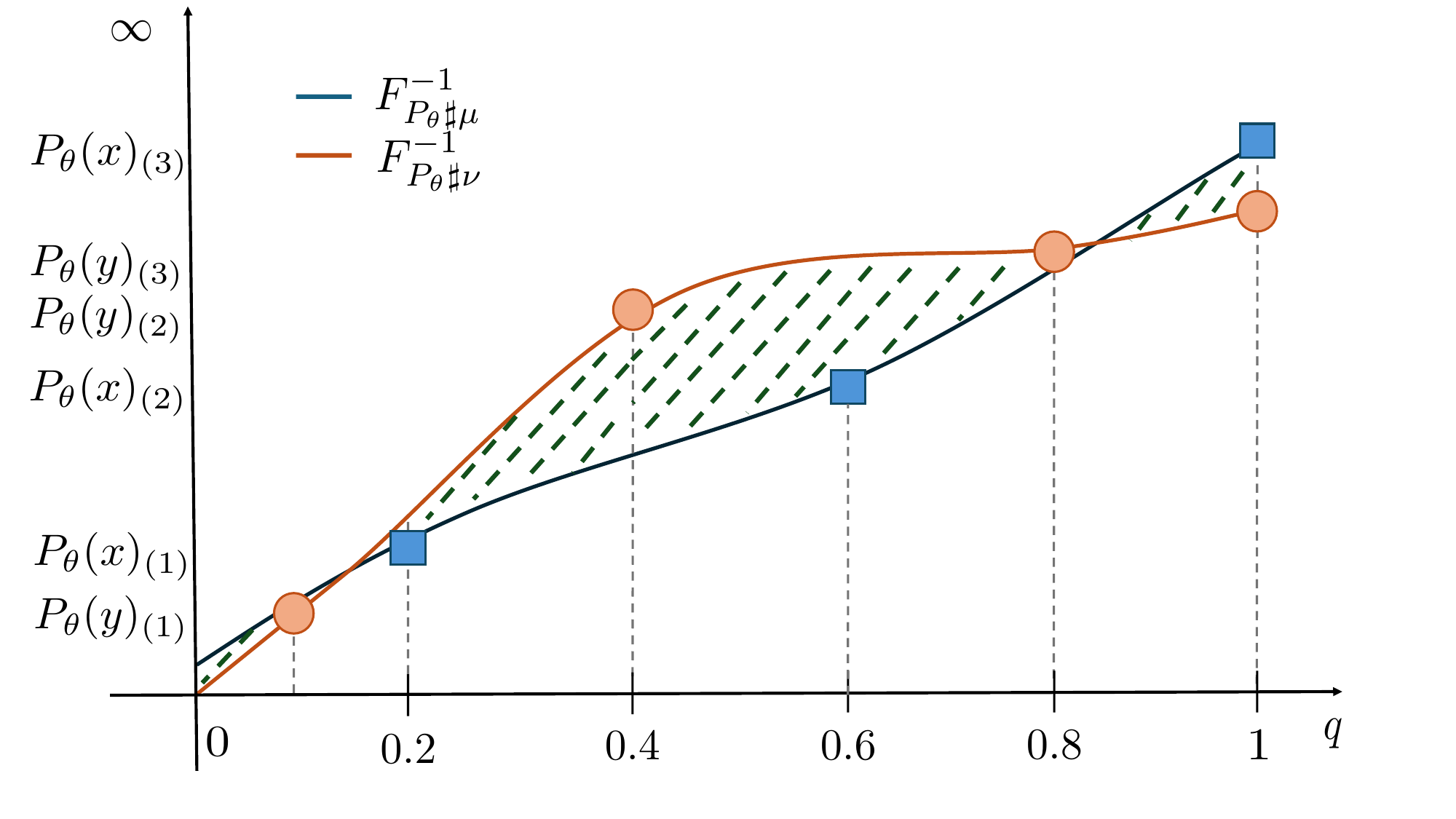}
    \end{tabular}
    \caption{Smooth estimation of quantile functions of projected distributions.} 
    \label{fig:smooth_projected_empirical_quantile}
\end{figure}

However, using empirical quantile functions might not be the best option for accuracy and robustness. In this section, we review other choices for estimating quantile functions and CDFs, including kernel estimation and monotonic rational quadratic splines (Figure~\ref{fig:smooth_projected_empirical_quantile}). 

\begin{remark}[Kernel estimation]
\label{remark:kernel_quantile}
    A natural extension is to use a kernel~\citep{jones1992estimating,sheather1990kernel} $\phi:\Re \to \Re_+$ to replace the indicator function $I(\cdot)$:
    \begin{align}
       & F_{P_\theta \sharp \mu}^{-1}(t) \approx   \sum_{i=1}^n P_\theta(x)_{(i)} \int_{\sum_{j=1}^{i-1}\alpha_{(j)}}^{\sum_{j=1}^{i}\alpha_{(j)}} \phi(t - z) \diff z, \\
        &F_{P_\theta \sharp \mu}(z) \approx  \sum_{i=1}^n \alpha_i \int_{-\infty}^{z} \phi(t - P_\theta(x_i)) \diff t.
    \end{align}
    Authors in~\citet{ponnoprat2024uniform} show that kernel smoothing enables uniform inference for the transport map.
\end{remark}

\begin{remark}[Monotonic rational quadratic spline estimation]
\label{remark:monotonic_spline_estimation}
From \\ \citep{gregory1982piecewise,durkan2019neural,kohler2021smooth}, 
Monotonic Rational Quadratic Splines approximate the function in each bin with the quotient of two quadratic polynomials. The resulting approximation is monotonic, continuously differentiable, and invertible. The splines are parametrized by the coordinates and derivatives of $M$ knots: $\{(t_i, x_i, x'_i)\}_{i=1}^M$, with $t_{i+1} > t_i$, $x_{i+1} > x_i$, and $x'_i > 0$. Given these parameters, the function $f$ in bin $i$ can be written as
\begin{align}
f_i(t) = x_i + (x_{i+1} - x_i)  \frac{s_i \xi^2 + x'_i \xi(t)(1 - \xi(t))}{s_i + \sigma_i \xi(t)(1 - \xi(t))},
\end{align}
where
\[
s_i = \frac{x_{i+1} - x_i}{t_{i+1} - t_i}, \quad \sigma_i = x'_{i+1} + x'_i - 2 s_i, \quad \xi(t) = \frac{t - t_i}{t_{i+1} - t_i}.
\]

The derivative is given by
\begin{align}
\frac{\diff x}{\diff t} = \frac{s_i^2 \left[x'_{i+1} \xi^2(t) + 2 s_i \xi(t)(1 - \xi(t)) + x'_i (1 - \xi(t))^2 \right]}{\left[s_i + \sigma_i \xi(t)(1 - \xi(t))\right]^2}.
\end{align}

The inverse of the function is:
\begin{equation}
f_i^{-1}(x) = t_i + (t_{i+1} - t_i)  \frac{2 c}{-b - \sqrt{b^2 - 4 a c}},
\end{equation}
where
\[
a = (s_i - x'_i) + \zeta(x) \sigma_i, \,\, b = x'_i - \zeta(x) \sigma_i, \,\, c = -s_i \zeta(x), \,\, \zeta(x) = \frac{x - x_i}{x_{i+1} - x_i}.
\]

The derivative $x'_i$ for $1 < i < n$ is determined by fitting a local quadratic polynomial to the neighboring knots 
$(t_{i-1}, x_{i-1})$, $(t_i, x_i)$, and $(t_{i+1}, x_{i+1})$:
\begin{align}
    x'_i = \frac{s_{i-1}(t_{i+1} - t_i) + s_i (t_i - t_{i-1})}{t_{i+1} - t_{i-1}}.
\end{align}
For $P_\theta \sharp \mu = \sum_{i=1}^n \alpha_i \delta_{P_\theta(x_i)}$, authors in~\citet{dai2021sliced} propose to estimate the quantile function with Monotonic Rational Quadratic Splines. In particular, $n$ knots are used, i.e., 
\begin{align}
\left\{ \left(\sum_{j=1}^{i-1} \alpha_{(j)}, P_\theta(x)_{(i)}, P_\theta(x)_{(i)}' \right) \right\}_{i=1}^n,
\end{align}
 where $ P_\theta(x)_{(i)}$ denotes the order statistics of the projected atoms $P_\theta(x)$. The CDF can be estimated by taking the inverse or fitting again with 
\begin{align}
\left\{ \left( P_\theta(x)_{(i)}, \sum_{j=1}^{i-1} \alpha_{(j)}, \left(\sum_{j=1}^{i-1} \alpha_{(j)} \right)' \right) \right\}_{i=1}^n.
\end{align}
\end{remark}

In practice, empirical estimation is often used because of its simplicity and computational efficiency. To the best of our knowledge, there is no explicit comparison between the discussed types of estimation. In addition, we can perform streaming estimation, where streaming samples arrive and there are memory constraints, using quantile sketches.

\begin{remark}[Streaming estimation]
A quantile sketch $S$~\citep{greenwald2001space,karnin2016optimal} is a data structure that stores samples such that for any $x \in \mathbb{R}$, we have:
\begin{align}
        |F(x; S) - F_{\mu_n}(t)| \leq \epsilon ,
\end{align}
    where $\epsilon > 0$ is the precision, $F_{\mu_n}$ is the CDF of $\mu_n$, the empirical distribution of streaming samples $x_1, \ldots, x_n$ (observed in arbitrary order). Here, $F(\cdot; S)$ is the approximate CDF given by $S$. The construction of $S$ can also be randomized with a noise source $\xi$, yielding the definition of a randomized quantile sketch $S_\xi$:
    \begin{align}
        P\left( |F(x; S) - F_{\mu_n}(t)| > \epsilon \right) \leq \delta,
    \end{align}
    where $\delta > 0$ is the failure probability. 
    For SW, authors in~\citet{nguyen2025streaming} suggest using the KKL sketch~\citep{karnin2016optimal}, which is the optimal randomized comparison-based quantile sketch. The key component of a KKL sketch is a \textit{compactor} $C$, which is a set $C = \{x_1, \ldots, x_k\}$ with an associated weight $w$. A compactor can compact $k$ elements into $k/2$ elements of weight $2w$, e.g., by sorting elements then keeping only either the even or the odd elements. For the KKL sketch, there are $H$ hierarchical compactors $C_h$ for $h=1, \ldots, H$ with associated weights $w_h = 2^{h-1}$ and capacities $k_h = \left\lceil k (2/3)^{H - h} \right\rceil + 1$ for a given initial capacity $k$. When a compactor $C_h$ performs compaction (triggered upon reaching its capacity), the selected elements are passed to the next compactor $C_{h+1}$. Given a data stream of $n > k$ samples, the second-highest compactor must have compacted at least once. This implies that the total number of compactors $H$ is less than $\log(n/k)$. The total capacity across all compactors satisfies
\[
\sum_{h=1}^H k_h \leq \sum_{h=1}^H \left(k \left(\frac{2}{3}\right)^{H-h} + 2\right) \leq 3k + 2H \leq 3k + 2 \log\left(\frac{n}{2k/3}\right),
\]
which yields a space complexity of $\mathcal{O}(k + \log(n/k))$. By choosing $k = \mathcal{O}\left(\frac{1}{\epsilon} \sqrt{\log\left(\frac{2}{\delta}\right)}\right)$ and assuming $\delta = \Omega(\epsilon)$, we obtain a failure probability of $\delta$ with space complexity $\mathcal{O}\left(\frac{1}{\epsilon} \sqrt{\log\left(\frac{1}{\epsilon}\right)} + \log(\epsilon n)\right)$. To eliminate the dependency on $n$, one can either manually cap the number of compactors $H$ or apply random sampling to compactors of capacity 2, as proposed in~\cite{karnin2016optimal}. In this case, we can set $k = \mathcal{O}\left(\frac{1}{\epsilon} \log \log\left(\frac{1}{\delta}\right)\right)$, and with $\delta = \Omega(\epsilon)$, this results in space complexity $\mathcal{O}\left(\frac{1}{\epsilon} \log^2 \log\left(\frac{1}{\delta \epsilon}\right)\right)$.

    Once the KKL sketch is built, the authors in~\citet{nguyen2025streaming} propose to approximate the quantile function and CDF using the discrete distribution from the sketch:
\begin{align}
\sum_{h=1}^H \sum_{x \in C_h} \frac{w_h}{\sum_{h'=1}^H |C_{h'}| w_{h'}} \delta_x,
\end{align}
where $|C_h|$ denotes the size of compactor $C_h$. It is worth noting that we can apply discussed techniques such as kernels and splines. However, the standard empirical quantile estimation is used in~\citet{nguyen2025streaming}.
\end{remark}

After obtaining estimates of the quantile functions and CDFs, we can estimate the transport map and the one-dimensional Wasserstein distance:
\begin{align}
    &T_{\mu, \nu} = F_{\nu}^{-1} \circ F_\mu, \\
    &W_p^p(P_\theta \sharp \mu, P_\theta \sharp \nu) = \int_0^1 c\bigl(F_{P_\theta \sharp \mu}^{-1}(t), F_{P_\theta \sharp \nu}^{-1}(t)\bigr) \diff t.
\end{align}
We now discuss how to evaluate the integral for the one-dimensional Wasserstein distance, including analytical evaluation, stochastic evaluation, equally-spaced evaluation, and trimming evaluation.

\begin{remark}[Analytical evaluation]
\label{remark:analytical_evaluation_1DW}
When the empirical quantile function is used, analytical evaluation leads to the North-West corner algorithm (Remark~\ref{remark:1DWasserstein_as_quantile_approximation}). For other estimations, analytical evaluation might not be tractable.
\end{remark}

\begin{remark}[Stochastic evaluation]
\label{remark:stochastic_evaluation_1DW}
We can use Monte Carlo estimation to evaluate:
\begin{align}
    \int_0^1 c(F_{P_\theta \sharp \mu}^{-1}(t), F_{P_\theta \sharp \nu}^{-1}(t)) \diff t \approx \frac{1}{K} \sum_{k=1}^K c(F_{P_\theta \sharp \mu}^{-1}(t_k), F_{P_\theta \sharp \nu}^{-1}(t_k)),
\end{align}
where $t_1, \ldots, t_K \overset{i.i.d.}{\sim} \mathcal{U}([0,1])$.
\end{remark}

\begin{remark}[Equally-spaced evaluation]
\label{remark:equally_spaced_evaluation_1DW}
Authors in~\citet{perez2024gaussian} propose to evaluate:
\begin{align}
    \int_0^1 c(F_{P_\theta \sharp \mu}^{-1}(t), F_{P_\theta \sharp \nu}^{-1}(t)) \diff t \approx \frac{1}{K} \sum_{k=1}^K c(F_{P_\theta \sharp \mu}^{-1}(t_k), F_{P_\theta \sharp \nu}^{-1}(t_k)),
\end{align}
where $t_1, \ldots, t_K$ is a sequence of equally-spaced points, e.g., quasi-Monte Carlo.
\end{remark}

\begin{remark}[Trimming-based evaluations]
\label{remark:trimming_evaluating_1DW}
When observed samples might be contaminated, we may want to drop projected samples around the tails of projected probability measures in the hope of removing outliers. Authors in~\citet{munk1998nonparametric,manole2022minimax} propose to use trimming versions of quantile functions:
\begin{align}
    F_{P_\theta \sharp \mu, \delta}^{-1}(t) = F_{P_\theta \sharp \mu}^{-1}(t)  I(\delta < t < 1 - \delta),
\end{align}
for $\delta \in [0, 1/2]$, which is equivalent to trimming evaluation of the one-dimensional Wasserstein distance:
\begin{align}
    \int_0^1 c(F_{P_\theta \sharp \mu}^{-1}(t), F_{P_\theta \sharp \nu}^{-1}(t)) \diff t \approx \frac{1}{1 - 2\delta} \int_{\delta}^{1 - \delta} c(F_{P_\theta \sharp \mu}^{-1}(t), F_{P_\theta \sharp \nu}^{-1}(t)) \diff t,
\end{align}
which can be evaluated again using the discussed methods.
\end{remark}

Overall, analytical evaluation is preferred. When it is intractable or computationally expensive (e.g., for a large number of atoms), stochastic or equally-spaced evaluations can be employed. Trimming-based estimations are useful for achieving robust results in the presence of contamination.

\section{Weighted Radon Transform and Non-Uniform Slicing}
\label{sec:slicing_measure:chapter:advances}
From previous sections, we know that SOT uses the Radon transform to map an original measure into a measure over the product of the unit hypersphere and the real line, with a uniform marginal on the unit hypersphere under the disintegration of measure. While this transform is natural and simple, the requirement of a uniform marginal measure is quite restrictive. In particular, when measures lie on a linear subspace, some projecting directions on the unit hypersphere, specifically those belonging to the orthogonal complement, are redundant, as the Radon conditional measures for such directions reduce to Dirac measures at zero. Including these directions leads to unnecessary computation.

Moreover, the goal of the transformation is to ensure that the Wasserstein distance in the transformed space closely approximates the Wasserstein distance in the original space. Therefore, restricting the slicing measure to be uniform, as in the standard Radon transform, can limit the ability to achieve this goal. In this section, we discuss the weighted Radon transform, which results in a non-uniform slicing measure. In contrast to the generalized Radon transform, which focuses on non-linear defining functions, the weighted Radon transform emphasizes reweighting the projecting directions. While we only review the weighted Radon transform with linear projection in this section, it is possible to extend it to any discussed non-linear projection in Section~\ref{subsec:non_linear_projections}.

We start with a review of the definitions of weighted Radon transform for functions and measures, and their connection.

\begin{definition}[Weighted Radon transform]
    \label{def:weighted_Radon_Transform}
    Given an integrable function $f:\Re^d \to \Re$, the weighted Radon transform (WRT)~\citep{beylkin1984inversion,boman1987support} is an operator that maps $f$ to a function on the product of the unit hypersphere in $d$ dimensions and the real line ($\Sm^{d-1} \times \Re$), which is defined as:
    \begin{align}
        \setWR_w f(\theta,t) = \int_{\Re^d} f(x) w(x,\theta) \delta(t-\langle x,\theta \rangle) \diff x,
    \end{align}
    where $\langle x,\theta\rangle = \theta^\top x$ is the inner product, $\theta \in \Sm^{d-1}$ is referred to as the direction, $t \in \Re$, and $w:\Re^d \times \Sm^{d-1}$ is a weight function which is typically smooth and non-negative.
\end{definition}
 For example, $w(x,\theta) = \exp\left(-\int_{-\infty}^0 a(x+\tau \theta) \diff \tau\right)$ leads to the attenuated Radon transform~\citep{natterer2001inversion}. By changing the inner product to a defining function, we can extend this definition to the weighted generalized Radon transform. 
\begin{definition}[Weighted Radon Transform of measures]
    \label{def:Weighted_Radon_Transform_measures} Given a measure $\mu \in \setM(\Re^d)$, the weighted RT measure $\setWR_w \mu \in \setM(\Sm^{d-1} \times \Re)$ is defined as:
    \begin{align}
    \label{eq:WRT_measure}
        \int_{\Sm^{d-1} \times \Re} f(\theta,t) \diff \setWR_w \mu(\theta,t) =  \int_{\Re^d} \int_{\Sm^{d-1}} f(\theta, \langle x,\theta \rangle) w(x,\theta) \diff \theta \diff \mu(x),
    \end{align} 
    for any integrable function $f$ on $\Sm^{d-1} \times \Re$, and $w:\Re^d \times \Sm^{d-1}$ is a weight function which is typically smooth and non-negative.
\end{definition}

\begin{remark}[Connection between weighted Radon Transform of measures and weighted Radon Transform of functions]
    \label{remark:connection_weighted_Radon_function_measure} Given a measure $\mu \in \setM(\Re^d)$ with a density $p_\mu(x)$, we have:
    \begin{align*}
        &\int_{\Re^d} \int_{\Sm^{d-1}} f(\theta, \langle x,\theta \rangle) w(x,\theta) \diff \theta \diff \mu(x)  \\
        &=  \int_{\Re^d} \int_{\Sm^{d-1}} f(\theta, \langle x,\theta \rangle) w(x,\theta) \diff \theta \, p_\mu(x) \diff x  \\
        &= \int_{\Sm^{d-1}} \int_{\Re^d} f(\theta, \langle x,\theta \rangle) w(x,\theta) p_\mu(x) \diff x \diff \theta \\
        &= \int_{\Sm^{d-1}} \int_{\Re^d} \int_\Re f(\theta, t) p_\mu(x) w(x,\theta) \delta(t-\langle x,\theta \rangle) \diff t \diff x \diff \theta \\
        &= \int_{\Sm^{d-1}} \int_\Re f(\theta, t) \int_{\Re^d} p_\mu(x) w(x,\theta) \delta(t-\langle x,\theta \rangle) \diff x \diff t \diff \theta \\
        &= \int_{\Sm^{d-1}} \int_\Re f(\theta, t) \setWR_w p_\mu(\theta,t) \diff t \diff \theta,
    \end{align*}
    which means that 
    \begin{align}
        \int_{\Sm^{d-1} \times \Re} f(\theta,t) \diff \setWR_w \mu(\theta,t) = \int_{\Sm^{d-1}} \int_\Re f(\theta, t) \setWR_w p_\mu(\theta,t) \diff t \diff \theta,
    \end{align}
    implying that $\setWR_w p_\mu(\theta,t)$ is the density of $\setWR_w \mu(\theta,t)$.
\end{remark}

In Definition~\ref{def:weighted_Radon_Transform} and Definition~\ref{def:Weighted_Radon_Transform_measures}, the weight function depends on both $x$ and $\theta$—that is, it assigns a weight to each projected atom and each projecting direction. While this choice is expressive, it can lead to intractable marginal measures when applying the disintegration of measure. As a result, such a weight function cannot be directly used to define a variant of SOT. For simplicity and tractability, we restrict the weight function to depend only on $\theta$.

\begin{remark}[Restricted weight function]
    \label{remark:restricted_weight_function}
    In this section, we focus on the case where $w(x,\theta) = w(\theta)$, i.e., weighting only based on $\theta$. In this case, the weighted RT becomes:
    \begin{align}
      \int_{\Sm^{d-1} \times \Re} f(\theta,t) \diff \setWR_w \mu(\theta,t) =  \int_{\Re^d} \int_{\Sm^{d-1}} f(\theta, \langle x,\theta \rangle) w(\theta) \diff \theta \diff \mu(x).
    \end{align}
    By further restricting $w(\theta) \diff \theta = \diff \sigma(\theta)$ where $\sigma$ is a continuous measure on $\Sm^{d-1}$, we have the weighted RT becomes:
    \begin{align}
       \int_{\Sm^{d-1} \times \Re} f(\theta,t) \diff \setWR_w \mu(\theta,t) &=  \int_{\Re^d} \int_{\Sm^{d-1}} f(\theta, \langle x,\theta \rangle) \diff \sigma(\theta) \diff \mu(x) \\
       &= \int_{\Sm^{d-1}} \int_{\Re} f(\theta,t) \diff \setR_\theta \mu(t) \diff \sigma(\theta),
    \end{align}
    where $\setR_\theta \mu$ is the pushforward measure of $\mu$ through the function $f_\theta(x) = \langle \theta,x \rangle$, which is also denoted as $\theta \sharp \mu$.
\end{remark}

\begin{remark}[Injectivity of Weighted Radon Transform]
\label{remark:injectivity_weighted_RT}
     The Weighted Radon Transform is injective with smooth and strictly positive weight functions~\citep{boman1987support} or the exponential function in the attenuation case~\citep{beylkin1984inversion}.
\end{remark}

By comparing weighted Radon transform measures, we can obtain a new SW variant which is named distributional sliced Wasserstein. We will review its properties and discuss its specific instances.

\begin{definition}[Distributional sliced Wasserstein distance]
    \label{def:DSW} Given $p \geq 1$, a continuous positive slicing measure $\sigma \in \setM_+(\Sm^{d-1})$, the distributional sliced Wasserstein-$p$~\citep{nguyen2021distributional} distance between $\mu \in \setP_p(\Re^d)$ and $\nu \in \setP_p(\Re^d)$ is defined as follows:
    \begin{align}
        DSW_p^p(\mu,\nu;\sigma) = \mathbb{E}_{\theta \sim \sigma}[W_p^p(\theta \sharp \mu, \theta \sharp \nu)],
    \end{align}
    where $\theta \sharp \mu$ and $\theta \sharp \nu$ are Radon conditional measures of $\mu$ and $\nu$.
\end{definition}

It is worth noting that the original definition of DSW is based on selecting $\sigma$ as a probability measure. However, we adapt the definition to make it more general. We will discuss the slicing measure selection part later.

\begin{proposition}[Distributional sliced Wasserstein distance is the Wasserstein distance between weighted Radon transform measures]
    \label{proposition:DSW_is_W_Radon} Given $\mu \in \setP_p(\Re^d)$ and $\nu \in \setP_p(\Re^d)$, we have:
    \begin{align}
        W_p^p(\setWR_w \mu, \setWR_w \nu) = DSW_p^p(\mu,\nu;\sigma),
    \end{align}
    where $\setWR_w \mu$ is the weighted Radon transform of $\mu$ (Definition~\ref{def:Weighted_Radon_Transform_measures}) and $w(\theta) \diff \theta = \diff \sigma(\theta)$.
\begin{proof}
    The proof is similar to Proposition~\ref{proposition:SW_is_W_Radon}.
\end{proof}
\end{proposition}

\begin{remark}[Motivation of distributional sliced Wasserstein]
    \label{remark:motivation_of_sw}
    As in Proposition~\ref{proposition:SW_is_W_Radon}, Sliced Wasserstein is the Wasserstein distance between Radon transform measures of the two measures. The Radon transform sets the measure on the projection parameter (on $\Sm^{d-1}$) to be uniform, which might not be optimal in terms of keeping the distance between transformed measures close to the original one:
    \begin{align}
        W_p^p(\mu,\nu) \approx W_p^p(\setR \mu,\setR \nu) = SW_p^p(\mu,\nu).
    \end{align}
    Therefore, weighted Radon transform is used to generalize the Radon transform, i.e., using any continuous measure $\sigma(\theta) \in \setM(\Sm^{d-1})$. We will discuss approaches for choosing $\sigma(\theta)$ later in this section.
\end{remark}

\begin{proposition}[Metricity of Distributional Sliced Wasserstein distance]
    \label{proposition:metricity_DSW} Distributional sliced Wasserstein distance is a valid metric when the slicing measure is continuous.
    \begin{proof}
        The proof is similar to Proposition~\ref{proposition:SW_metricity} with the injectivity of the weighted Radon transform.
    \end{proof}
\end{proposition}

\begin{proposition}[Connection of distributional sliced Wasserstein to Wasserstein distance]
    \label{proposition:connection_DSW_to_Wasserstein} Given $p \geq 1$, $\sigma \in \setP(\Sm^{d-1})$, $\mu \in \setP_p(\Re^d)$, and $\nu \in \setP_p(\Re^d)$, we have:
    \begin{align}
        DSW_p(\mu,\nu;\sigma) \leq W_p(\mu,\nu).
    \end{align}
    \begin{proof}
    From the definition of DSW, we have:
        \begin{align*}
            DSW_p^p(\mu,\nu;\sigma) &= \mathbb{E}_{\theta \sim \sigma}[W_p^p(\theta \sharp \mu, \theta \sharp \nu)] \\ 
            &\leq \max_{\theta \in \Sm^{d-1}} W_p^p(\theta \sharp \mu, \theta \sharp \nu) \\
            &= \max_{\theta \in \Sm^{d-1}} \inf_{\pi \in \Pi(\mu,\nu)} \int_{\Re^d \times \Re^d} |\theta^\top x - \theta^\top y|^p \diff \pi(x,y) \\
            &\leq \max_{\theta \in \Sm^{d-1}} \inf_{\pi \in \Pi(\mu,\nu)} \int_{\Re^d \times \Re^d} \|\theta\|^p \|x - y\|^p \diff \pi(x,y) \\
            &= \inf_{\pi \in \Pi(\mu,\nu)} \int_{\Re^d \times \Re^d} \|x - y\|^p \diff \pi(x,y) \\
            &= W_p^p(\mu, \nu),
         \end{align*}
         where the second inequality is due to the Cauchy–Schwarz inequality and the last equality is due to $\|\theta\| = 1$ for $\theta \in \Sm^{d-1}$.
    \end{proof}
\end{proposition}

\begin{remark}[Sample complexity of distributional sliced Wasserstein distance]
\label{remark:sample_complexity_DSW}
    Given $x_1,\ldots,x_n \sim \mu \in \setP_p(\Re^d)$, $y_1,\ldots,y_m \sim \nu \in \setP_p(\Re^d)$, and $\mu_n$ and $\nu_m$ are corresponding empirical measures ($n > m$), assume that $M_q(\mu) = \int_{\Re^d} \|x\|^q \diff \mu(x) < \infty$ and $M_q(\nu) = \int_{\Re^d} \|y\|^q \diff \nu(y) < \infty$ for some $q > p$. Then, there exists a constant $C_{p,q}$ such that:
    \begin{align}
        &\mathbb{E}[|DSW_p(\mu_n,\nu_n;\sigma) - DSW_p(\mu,\nu;\sigma)|] \nonumber \\
        &\leq C_{p,q}^{1/p} \, \bigl(M_q(\mu)^{1/q} + M_q(\nu)^{1/q}\bigr) \times
        \begin{cases}
        n^{-1/(2p)}, & \text{if } q > 2p, \\
        n^{-1/(2p)} \log(1 + n)^{1/p}, & \text{if } q = 2p, \\
        n^{-(q - p)/(p q)}, & \text{if } q \in (p, 2p),
        \end{cases}
    \end{align}
    where $\sigma \in \setP(\Sm^{d-1})$.
    This follows from Remark~\ref{remark:sample_complexity_1DWasserstein} with the fact that the moment of the projected measure $\theta \sharp \mu$ is smaller than the moment of the original measure $\mu$ due to the Cauchy–Schwarz inequality. By restricting the support space to a compact set, we can improve the result as in~\citet{nguyen2021distributional}.
\end{remark}

\begin{remark}[Monte Carlo estimation for distributional sliced Wasserstein]
    \label{remark:MC_DSW} When $\sigma \in \setP(\Sm^{d-1})$ and we can sample from it, we can use Monte Carlo estimation to approximate DSW. In particular, we sample $\theta_1,\ldots,\theta_L \simiid \sigma$, then form the following estimator: 
    \begin{align}
        \widehat{DSW}(\mu,\nu;\sigma,\theta_1,\ldots,\theta_L) = \frac{1}{L}\sum_{l=1}^L W_p^p(\theta_l \sharp \mu, \theta_l \sharp \nu).
    \end{align}
    When $\sigma \in \setM(\Sm^{d-1})$ is a positive and continuous measure with density $w(\theta)$, we can use importance sampling to estimate DSW using a proposal distribution $\sigma_0(\theta)$ with density $w_0$ as follows:
    \begin{align}
        \widehat{DSW}(\mu,\nu;\sigma,\sigma_0,\theta_1,\ldots,\theta_L) = \frac{1}{L} \sum_{l=1}^L \frac{w(\theta_l)}{w_0(\theta_l)} W_p^p(\theta_l \sharp \mu, \theta_l \sharp \nu),
    \end{align}
    where $\theta_1,\ldots,\theta_L \simiid \sigma_0$. The importance sampling estimation is also used when sampling from $\sigma$ is nontrivial. When $\sigma_0$ is the uniform distribution, quasi Monte Carlo and control variate methods can also be applied (Section~\ref{sec:MC:chapter:advances}).
\end{remark}

DSW is defined above with a given fixed slicing measure $\sigma$. A natural extension is to make $\sigma$ dependent on $\mu$ and $\nu$. Making $\sigma$ measure-dependent can help DSW capture more meaningful or discriminative directions.

\begin{definition}[Measures-dependent distributional sliced Wasserstein distance]
    \label{def:MDDSW} Given $p \geq 1$, the measures-dependent distributional sliced Wasserstein-$p$ distance between $\mu \in \setP_p(\Re^d)$ and $\nu \in \setP_p(\Re^d)$ is defined as follows:
    \begin{align}
        MDDSW_p^p(\mu,\nu) = DSW_p^p(\mu,\nu;\sigma_{\mu,\nu}),
    \end{align}
    where $\theta \sharp \mu$ and $\theta \sharp \nu$ are Radon conditional measures of $\mu$ and $\nu$, and $\sigma_{\mu,\nu} \in \setM_+(\Sm^{d-1})$ is a continuous positive measure which depends on $\mu$ and $\nu$.
\end{definition}

\begin{remark}[Input-Dependent Weighted Radon Transform]
    \label{remark:input_depedent_WRT}
    Measures-dependent distributional sliced Wasserstein can be seen as the application of a weighted Radon transform with the weight function $w_{\mu,\nu}(\theta)$ depending on input measures $\mu$ and $\nu$. Here, input-dependency creates a nonlinear structure which leads to the loss of injectivity of the resulting integral transform. 
\end{remark}

\begin{remark}[Metricity of measures-dependent distributional sliced Wasserstein]
    \label{remark:metricity} While input-dependent weighted Radon transform is not injective, measures-dependent distributional sliced Wasserstein can still be a valid metric. 
    Non-negativity remains as long as $\sigma_{\mu,\nu}(\theta)$ is positive. Symmetry holds if $\sigma_{\mu,\nu}(\theta)$ is symmetric in terms of $\mu$ and $\nu$. The triangle inequality can also be obtained if  
    $DSW_p(\mu,\nu;\sigma_{\mu,\nu}) \geq DSW_p(\mu,\nu;\sigma_{\mu,\nu})$ for any $\gamma$. The identity of indiscernibles can still be obtained if $w_{\mu,\nu}(x,\theta)$ can be lower bounded (up to a constant) by a measures-dependent distributional sliced Wasserstein with a fixed $w(\theta)$ as the weight function for the corresponding integral transform. We will discuss this in detail later when we discuss specific forms of $\sigma_{\mu,\nu}$.
\end{remark}

We now discuss some practical ways to construct a measures-dependent slicing measure $\sigma_{\mu,\nu}$. There are two major directions: optimization-based slicing measure and energy-based slicing measure. We first review recent works on developing optimization-based slicing measures.

\begin{remark}[Optimization-based slicing measure]
    \label{remark:optimization_based_slicing}
    Given $\mu \in \setP_p(\Re^d)$ and $\nu \in \setP_p(\Re^d)$, the slicing measure can be found by solving:
    \begin{align}
        \sup_{\sigma \in \bar\setM(\Sm^{d-1})} \mathbb{E}_{\theta \sim \sigma}[W_p^p(\theta \sharp \mu, \theta \sharp \nu)],
    \end{align}
    where $\bar\setM(\Sm^{d-1})$ is a set of measures on $\Sm^{d-1}$. Authors in~\citet{nguyen2021distributional} set 
    \[
        \bar\setM(\Sm^{d-1}) = \{ f_\phi \sharp \sigma_0 \mid \sigma_0 \in \setP(\Sm^{d-1}), \phi \in \Phi, \mathbb{E}_{\theta, \theta' \sim \sigma}[f_\phi(\theta)^\top f_\phi(\theta')] \leq C \}
    \]
    for $C \geq 0$. With Lagrange duality, the optimization problem is turned into:
    \begin{align}
        \max_{\phi \in \Phi} \mathbb{E}_{\theta \sim \sigma_0}[W_p^p(f_\phi(\theta) \sharp \mu, f_\phi(\theta) \sharp \nu)] - \lambda_C \mathbb{E}_{\theta, \theta' \sim \sigma}[f_\phi(\theta)^\top f_\phi(\theta')],
    \end{align}
    which is solved using stochastic gradient ascent methods. Authors in~\citet{nguyen2021improving} use a more explicit family of probability distributions, i.e., von Mises-Fisher~\citep{jupp1979maximum} and power spherical~\citep{de2020power}. Let $vMF(\xi, \kappa)$ be the von Mises-Fisher distribution with location $\xi \in \Sm^{d-1}$ and concentration $\kappa$ ($\kappa > 0$), the optimization is turned into:
    \begin{align}
        \max_{\xi \in \Sm^{d-1}} \mathbb{E}_{\theta \sim vMF(\xi, \kappa)}[W_p^p(\theta \sharp \mu, \theta \sharp \nu)],
    \end{align}
    which is solved using projected stochastic gradient ascent~\citep{nguyen2021improving}. Without explicit regularization or implicit regularization (some parametric family), searching $\sigma \in \setP(\Sm^{d-1})$ will lead to $\delta_{\theta^\star}$ where:
    \begin{align}
        \theta^\star = \arg\max_{\theta \in \Sm^{d-1}} W_p^p(\theta \sharp \mu, \theta \sharp \nu),
    \end{align}
    which results in the max sliced Wasserstein (Max-SW) distance~\citep{deshpande2019max}, which is either solved using projected gradient ascent~\citep{kolouri2019generalized,nietert2022statistical} or Riemannian optimization~\citep{lin2020projection}. Authors in~\citet{ohana2023shedding} discuss a PAC-Bayesian theory aspect of changing the slicing measure.
\end{remark}

\begin{proposition}[Metricity with optimization-based slicing measure]
    \label{proposition:metricity_optimization}
    The non-negativity and symmetry are trivial. For the identity of indiscernibles, if $\mu = \nu$, we directly have $MDDSW_p(\mu,\nu) = 0$ for all discussed slicing measures. Now, we discuss when $MDDSW_p(\mu,\nu) = 0$. For Max-SW, we know that it is an upper bound of SW. Hence, we have $SW_p(\mu,\nu) = 0$ which leads to $\mu = \nu$. For the case of searching for the maximal slicing measure in a continuous family of measures, we know that $MDDSW_p(\mu,\nu) \geq DSW_p(\mu,\nu; \sigma_0)$ for some fixed measure $\sigma_0$ in the family, hence, $\mu = \nu$. Lastly, we discuss the triangle inequality. 
    For any $\mu, \nu, \lambda \in \mathcal{P}_p(\Re^d)$, we have:
    \begin{align*}
        &MDDSW_p(\mu,\nu) \\ &= \sup_{\sigma}\left( \mathbb{E}_{\theta \sim \sigma}[W_p^p(\theta \sharp \mu, \theta \sharp \nu)]\right)^{\frac{1}{p}} \\
        &\leq \sup_\sigma \left( \mathbb{E}_{\theta \sim \sigma}[(W_p(\theta \sharp \mu, \theta \sharp \lambda) + W_p(\theta \sharp \lambda, \theta \sharp \nu))^p] \right)^{\frac{1}{p}}  \\
        &\leq \sup_\sigma \left( \left( \mathbb{E}_{\theta \sim \sigma}[W_p^p(\theta \sharp \mu, \theta \sharp \lambda)] \right)^{\frac{1}{p}} + \left( \mathbb{E}_{\theta \sim \sigma}[W_p^p(\theta \sharp \lambda, \theta \sharp \nu)] \right)^{\frac{1}{p}} \right) \\
        &\leq \sup_\sigma \left( \mathbb{E}_{\theta \sim \sigma}[W_p^p(\theta \sharp \mu, \theta \sharp \lambda)] \right)^{\frac{1}{p}} + \sup_\sigma \left( \mathbb{E}_{\theta \sim \sigma}[W_p^p(\theta \sharp \lambda, \theta \sharp \nu)] \right)^{\frac{1}{p}} \\
        &= MDDSW_p(\mu,\lambda) + MDDSW_p(\lambda,\nu),
    \end{align*}
    where the supremum is taken with respect to a set of slicing measures. The first inequality is due to the triangle inequality of Wasserstein distance and the second inequality is due to Minkowski's inequality.  
\end{proposition}

Overall, we have the following relationship between DSW, Max-SW, and Wasserstein distance:

\begin{proposition}
\label{proposition:relationship_optimization_based}
    Given $\mu, \nu \in \setP_p(\Re^d)$ ($p \geq 1$), we have:
    \begin{align}
        DSW_p(\mu,\nu;\sigma) \leq \text{Max-}SW_p(\mu,\nu) \leq W_p(\mu,\nu),
    \end{align}
    for any $\sigma \in \setP(\Sm^{d-1})$.
    \begin{proof}
        The first inequality is due to $\sigma \in \setP(\Sm^{d-1})$, while the second inequality is due to the Cauchy-Schwarz inequality with $\theta \in \Sm^{d-1}$.
    \end{proof}
\end{proposition}

Max-SW has been shown to have a sample complexity of $\mathcal{O}(n^{-1/2})$ \\\citep{nguyen2021distributional,boedihardjo2025sharp}. As a result, DSW also has the same low sample complexity since it is a lower bound of Max-SW. We now discuss the computational aspect of finding optimization-based slicing measures. As discussed, projected gradient ascent algorithms are often used in practice. To use such algorithms, we need to obtain the gradient $\nabla_\theta W_p^p(\theta \sharp \mu, \theta \sharp \nu)$.

\begin{proposition}[Regularity of the map $\theta \mapsto W_p^p(\theta \sharp \mu, \theta \sharp \nu)$]
    \label{proposition:regularity_theta}
    From~\citet{bayraktar2021strong,nietert2022statistical}, the functions $\theta \mapsto W_p^p(\theta \sharp \mu, \theta \sharp \nu)$ and $\theta \mapsto W_p(\theta \sharp \mu, \theta \sharp \nu)$ are Lipschitz with constants bounded by
    \[
    M_p^{\mu,\nu} = 3p 2^p \max_{\theta \in \mathbb{S}^{d-1}} \left( \mathbb{E}_{x \sim \mu} |\theta^\top x|^p + \mathbb{E}_{x \sim \nu} |\theta^\top x|^p \right)
    \]
    and
    \[
    L_p^{\mu,\nu} = \max_{\theta \in \mathbb{S}^{d-1}} \left[ \left( \mathbb{E}_{x \sim \mu} |\theta^\top x|^p \right)^{1/p} + \left( \mathbb{E}_{x \sim \nu} |\theta^\top x|^p \right)^{1/p} \right].
    \]
\end{proposition}

\begin{remark}[Gradient of $\theta \mapsto W_p^p(\theta \sharp \mu, \theta \sharp \nu)$] 
Since slicing measure optimization requires the gradient $\nabla_\theta W_p^p(\theta \sharp \mu, \theta \sharp \nu)$, we discuss its calculation in two cases: continuous and discrete.

\textbf{Continuous case.} Let $\mu$ and $\nu$ be continuous measures on $\mathbb{R}^d$, and $\theta \in \mathbb{S}^{d-1}$ a unit vector. Define
\[
F(\theta) := W_p^p(\theta \sharp \mu, \theta \sharp \nu).
\]
Using the closed-form expression of Wasserstein distance in terms of quantile functions, we have
\[
F(\theta) = \int_0^1 \big| F_{\theta \sharp \mu}^{-1}(t) - F_{\theta \sharp \nu}^{-1}(t) \big|^p \, \diff t.
\]

To compute $\nabla_\theta F(\theta)$, differentiate under the integral sign:
\begin{align*}
\nabla_\theta F(\theta) &= p \int_0^1 \left| F_{\theta \sharp \mu}^{-1}(t) - F_{\theta \sharp \nu}^{-1}(t) \right|^{p-1} \operatorname{sign}\big( F_{\theta \sharp \mu}^{-1}(t) - F_{\theta \sharp \nu}^{-1}(t) \big) \\
&\quad \times \nabla_\theta \big( F_{\theta \sharp \mu}^{-1}(t) - F_{\theta \sharp \nu}^{-1}(t) \big) \, \diff t.
\end{align*}

We now compute $\nabla_\theta F_{\theta \sharp \mu}^{-1}(t)$. Since
\[
F_{\theta \sharp \mu}(F_{\theta \sharp \mu}^{-1}(t)) = t,
\]
differentiating both sides with respect to $\theta$ gives
\[
\nabla_\theta F_{\theta \sharp \mu}(F_{\theta \sharp \mu}^{-1}(t)) + p_{\theta \sharp \mu}(F_{\theta \sharp \mu}^{-1}(t)) \, \nabla_\theta F_{\theta \sharp \mu}^{-1}(t) = 0,
\]
where $p_{\theta \sharp \mu}$ is the density of $\theta \sharp \mu$. Hence,
\[
\nabla_\theta F_{\theta \sharp \mu}^{-1}(t) = - \frac{\nabla_\theta F_{\theta \sharp \mu}(F_{\theta \sharp \mu}^{-1}(t))}{p_{\theta \sharp \mu}(F_{\theta \sharp \mu}^{-1}(t))}.
\]

Next, observe that
\[
F_{\theta \sharp \mu}(x) = \int_{\mathbb{R}^d} \mathbf{1}_{\langle \theta, z \rangle \leq x} \, \diff \mu(z).
\]
Differentiating with respect to $\theta$ (formally, using the chain rule and distributional derivative), we get
\[
\nabla_\theta F_{\theta \sharp \mu}(x) = - \int_{\mathbb{R}^d} z \, \delta(\langle \theta, z \rangle - x) \, \diff \mu(z).
\]

Thus,
\[
\nabla_\theta F_{\theta \sharp \mu}^{-1}(t) = \frac{\int_{\mathbb{R}^d} z \, \delta\big(\langle \theta, z \rangle - F_{\theta \sharp \mu}^{-1}(t)\big) \, \diff \mu(z)}{p_{\theta \sharp \mu}(F_{\theta \sharp \mu}^{-1}(t))},
\]
and similarly,
\[
\nabla_\theta F_{\theta \sharp \nu}^{-1}(t) = \frac{\int_{\mathbb{R}^d} y \, \delta\big(\langle \theta, y \rangle - F_{\theta \sharp \nu}^{-1}(t)\big) \, \diff \nu(y)}{p_{\theta \sharp \nu}(F_{\theta \sharp \nu}^{-1}(t))}.
\]

Putting it all together, the gradient is
\begin{align}
& \nabla_\theta W_p^p(\theta \sharp \mu, \theta \sharp \nu)  \nonumber\\ &= p \int_0^1 \left| F_{\theta \sharp \mu}^{-1}(t) - F_{\theta \sharp \nu}^{-1}(t) \right|^{p-1} \operatorname{sign}\big( F_{\theta \sharp \mu}^{-1}(t) - F_{\theta \sharp \nu}^{-1}(t) \big) \nonumber \\
&\quad \times \left( \frac{\int z \, \delta(\langle \theta, z \rangle - F_{\theta \sharp \mu}^{-1}(t)) \, \diff \mu(z)}{p_{\theta \sharp \mu}(F_{\theta \sharp \mu}^{-1}(t))} - \frac{\int y \, \delta(\langle \theta, y \rangle - F_{\theta \sharp \nu}^{-1}(t)) \, \diff \nu(y)}{p_{\theta \sharp \nu}(F_{\theta \sharp \nu}^{-1}(t))} \right) \diff t,
\end{align}
which can be estimated via Monte Carlo integration.


  \textbf{Discrete case.} When $\mu$ and $\nu$ are discrete measures, i.e., 
\[
\mu = \sum_{i=1}^n \alpha_i \delta_{x_i} \quad \text{and} \quad \nu = \sum_{j=1}^m \beta_j \delta_{y_j},
\]
their projections satisfy
\[
\theta \sharp \mu = \sum_{i=1}^n \alpha_i \delta_{\langle \theta, x_i \rangle}, \quad \theta \sharp \nu = \sum_{j=1}^m \beta_j \delta_{\langle \theta, y_j \rangle}.
\]

The gradient can be expressed as
\[
\nabla_\theta W_p^p(\theta \sharp \mu, \theta \sharp \nu) = \nabla_\theta \left( \min_{\pi \in \Pi(\mu,\nu)} \sum_{i=1}^n \sum_{j=1}^m |\langle \theta, x_i \rangle - \langle \theta, y_j \rangle|^p \pi_{ij} \right).
\]

Since the map 
\[
\theta \mapsto \sum_{i=1}^n \sum_{j=1}^m |\langle \theta, x_i \rangle - \langle \theta, y_j \rangle|^p \pi_{ij}
\]
is continuous and the set $\Pi(\mu,\nu)$ is compact, Danskin’s theorem applies. Let 
\[
\pi^\star = \arg\min_{\pi \in \Pi(\mu,\nu)} \sum_{i=1}^n \sum_{j=1}^m |\langle \theta, x_i \rangle - \langle \theta, y_j \rangle|^p \pi_{ij}
\]
(e.g., computed using the Northwest corner algorithm, see Remark~\ref{remark:1DKantorovich_discrete}). Then,
\begin{align}
\nabla_\theta W_p^p(\theta \sharp \mu, \theta \sharp \nu) 
&= \nabla_\theta \sum_{i=1}^n \sum_{j=1}^m |\langle \theta, x_i \rangle - \langle \theta, y_j \rangle|^p \pi_{ij}^\star \nonumber \\
&= \sum_{i=1}^n \sum_{j=1}^m \pi_{ij}^\star \nabla_\theta |\langle \theta, x_i \rangle - \langle \theta, y_j \rangle|^p \nonumber \\
&= p \sum_{i=1}^n \sum_{j=1}^m \pi_{ij}^\star \left| \langle \theta, x_i - y_j \rangle \right|^{p-1} \operatorname{sign}(\langle \theta, x_i - y_j \rangle) (x_i - y_j),
\end{align}
which parallels the continuous case, replacing the empirical CDF with discrete sums.
\end{remark}

Optimization-based variants often lead to minimax optimization problems in statistical inference contexts. These problems can be computationally expensive and difficult to solve, especially in high-dimensional settings or with complex data structures. Consequently, alternative approaches that avoid such heavy optimization procedures have been developed.  Next, we discuss the second approach to defining measures-dependent slicing measures: energy-based slicing measures.

\begin{remark}[Energy-based slicing measure]
\label{remark:energy_based_slicing_measure}
For any $p \geq 1$, dimension $d \geq 1$, an energy function $f: [0,\infty) \to \Theta \subset (0,\infty)$, and two probability measures $\mu, \nu \in \mathcal{P}_p(\mathbb{R}^d)$, the energy-based slicing measure $\sigma_{\mu,\nu}(\theta)$ supported on $\mathbb{S}^{d-1}$ is defined~\citep{nguyen2023energy} by
\begin{align}
p_{\sigma_{\mu,\nu}}(\theta; f) \propto f\big( W_p^p(\theta \sharp \mu, \theta \sharp \nu) \big) := \frac{f\big( W_p^p(\theta \sharp \mu, \theta \sharp \nu) \big)}{\int_{\mathbb{S}^{d-1}} f\big( W_p^p(\theta \sharp \mu, \theta \sharp \nu) \big) \, \diff \theta},
\end{align}
where the image of $f$ lies in $(0, \infty)$ to ensure $\sigma_{\mu,\nu}$ is continuous on $\mathbb{S}^{d-1}$. The function $f$ controls the weighting of each projection parameter $\theta$. 

Similar to the optimization-based slicing measure, there is an implicit assumption that \textit{"a higher projected Wasserstein distance corresponds to a better projection direction."} Therefore, a natural choice for $f$ is a monotonically increasing function, for example, the exponential function $f_e(x) = e^x$, or a shifted polynomial function 
\[
f_q(x) = x^a + \varepsilon,
\]
with $a, \varepsilon > 0$. The shift $\varepsilon$ prevents the slicing distribution from becoming undefined when the two input measures are equal.
\end{remark}

The use of the energy-based slicing measure leads to the energy-based sliced Wasserstein (EBSW) distance~\citep{nguyen2023energy}.

\begin{proposition}[Metricity of energy-based sliced Wasserstein]
    \label{proposition:metricity_EBSW}
    From~\citet{nguyen2023energy}, for any $p \geq 1$ and any energy function, the energy-based sliced Wasserstein (EBSW) is a semi-metric on the probability space over $\mathbb{R}^d$, meaning it satisfies non-negativity, symmetry, and the identity of indiscernibles.
\end{proposition}

\begin{proposition}[Relationship between EBSW, SW, Max-SW, and Wasserstein]
    \label{proposition:relationship_EBSW}
    From  \citet{nguyen2023energy}, the following inequalities hold for any $p \geq 1$ and increasing energy functions $f$:
    \begin{enumerate}
        \item $\displaystyle \mathrm{SW}_p(\mu, \nu) \leq \mathrm{EBSW}_p(\mu, \nu; f)$.
        \item $\displaystyle \mathrm{EBSW}_p(\mu, \nu; f) \leq \mathrm{Max\text{-}SW}_p(\mu, \nu) \leq W_p(\mu, \nu)$.
    \end{enumerate}
\end{proposition}

Since EBSW is a lower bound of Max-SW, it inherits the favorable sample complexity rate of $\mathcal{O}(n^{-1/2})$, thereby avoiding the curse of dimensionality. 

\begin{remark}[Estimation of energy-based sliced Wasserstein]
    \label{remark:computation_EBSW}
    There are two main approaches to estimate the EBSW distance:

    \textbf{Importance Sampling.}  
    The EBSW distance can be expressed as
    \begin{align}
        \label{eq:ISEBSW}
        \mathrm{EBSW}_p^p(\mu, \nu) 
        &= \frac{\mathbb{E}_{\theta \sim \sigma_0} \left[ W_p^p(\theta \sharp \mu, \theta \sharp \nu) \, w_{\mu,\nu,\sigma_0}(\theta) \right]}{\mathbb{E}_{\theta \sim \sigma_0} \left[ w_{\mu,\nu,\sigma_0}(\theta) \right]},
    \end{align}
    where $\sigma_0 \in \mathcal{P}(\mathbb{S}^{d-1})$ is a proposal distribution, and the importance weight function is
    \[
    w_{\mu,\nu,\sigma_0}(\theta) := \frac{f\big( W_p^p(\theta \sharp \mu, \theta \sharp \nu) \big)}{p_{\sigma_0}(\theta)},
    \]
    with $p_{\sigma_0}$ denoting the density of $\sigma_0$. Monte Carlo estimation can then be performed by sampling $\theta$ from $\sigma_0$.

    \textbf{Sampling Importance Resampling (SIR) and Markov Chain Monte Carlo (MCMC).}  
    Another approach is to sample directly from the energy-based slicing measure $\sigma_{\mu,\nu}(\theta)$. Suppose $\theta_1, \ldots, \theta_L \sim \sigma_{\mu,\nu}(\theta)$, then
    \[
    \frac{1}{L} \sum_{l=1}^L W_p^p(\theta_l \sharp \mu, \theta_l \sharp \nu)
    \]
    serves as an unbiased estimator of the EBSW distance.

    \textit{Sampling Importance Resampling (SIR):} Starting with samples $\theta_1', \ldots, \theta_L' \sim \sigma_0$, compute normalized weights
    \[
    \hat{w}_{\mu,\nu,\sigma_0}(\theta_l') := \frac{w_{\mu,\nu,\sigma_0}(\theta_l')}{\sum_{i=1}^L w_{\mu,\nu,\sigma_0}(\theta_i')}.
    \]
    Then form the discrete approximation
    \[
    \hat{q}(\theta) = \sum_{l=1}^L \hat{w}_{\mu,\nu,\sigma_0}(\theta_l') \delta_{\theta_l'},
    \]
    from which the final samples $\theta_1, \ldots, \theta_L \sim \hat{q}$ are drawn.

    \textit{Markov Chain Monte Carlo (MCMC):}  
    Initialize $\theta_1 \sim \sigma_0$ and use a transition kernel $\sigma_t(\theta_t | \theta_{t-1})$ for $t > 1$ to propose candidates $\theta_t'$. Accept the candidate with probability
    \[
    \alpha = \min \left(1, \frac{p_{\sigma_{\mu,\nu}}(\theta_t'; f)}{p_{\sigma_{\mu,\nu}}(\theta_{t-1}; f)} \frac{p_{\sigma_t}(\theta_{t-1} | \theta_t')}{p_{\sigma_t}(\theta_t' | \theta_{t-1})} \right),
    \]
    else keep the previous sample. For good mixing, $T$ (the number of iterations) should be sufficiently large.

    Despite producing dependent samples, both SIR and MCMC yield unbiased estimators of the EBSW distance with sequential slicing measures~\citep{nguyen2023energy, nguyen2023markovian}.
\end{remark}

\begin{figure}[!t]
    \centering
    \includegraphics[width=1\linewidth]{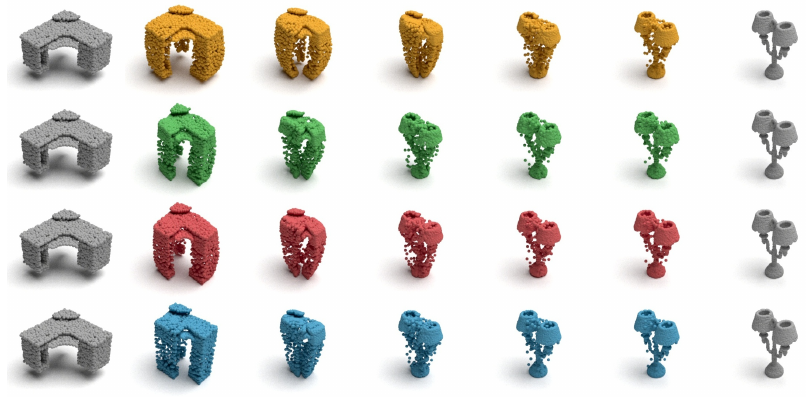}
    \caption{Gradient flows from SW, Max-SW, DSW, and EBSW (Figure 2 in~\citet{nguyen2023energy}).}
    \label{fig:EBSW_flow}
\end{figure}

Changing the slicing measure induces significantly different geometry compared to the original sliced Wasserstein distance. As shown in Figure~\ref{fig:EBSW_flow}, the gradient flows differ substantially. Parameter estimation using gradient-based algorithms with Max-SW, DSW, and EBSW can be interpreted as automatic learning rate adaptation schemes~\citep{tran2024understanding}.

\section{Sliced Optimal Transport Plans}
\label{sec:map:chapter:advances}

In this final section of this chapter, we extend the discussion on using SOT for transporting measures. In addition to iterative distribution transfer (IDT) and Knothe’s transport, which are discussed in Section~\ref{sec:IDT_Konthe:chapter:foundations}, there are some ways to define transportation on the original space with sliced optimal transport. 

\paragraph{Implicit map.} In terms of implicit transportation maps like the iterative distributional transfer algorithm and gradient flow with SW energy, it is possible to improve them. For example, we can choose good projecting directions across iterations. 

\begin{definition}[Projection pursuit Monge map]
    \label{def:PPMM} Authors in~\citet{meng2019large} propose an iterative map from an empirical measure $\mu=\frac{1}{n}\sum_{i=1}^n \delta_{x_i}$ to an empirical $\nu = \frac{1}{m} \sum_{j=1}^m \delta_{y_j}$. In particular, the map at iteration $t+1$ can be defined as:
    \begin{align}
        & x_{i}(t+1) = x_i(t) + (T_{\theta(t)}^{(t)}( \langle \theta(t) ,x_i(t)\rangle ) - \langle \theta(t) ,x_i(t)\rangle) \theta(t)^\top, \\
        & x_i(0)=x_i, \quad \forall i=1,\ldots,n, \nonumber
    \end{align}
    where $T_\theta^{(t)}( \cdot )$ is the optimal transport map from $\theta \sharp \mu(t)$ to $\theta \sharp \nu$, and $\theta(t)$ is found using projection pursuit methods e.g.,  sliced inverse regression~\citep{li1991sliced}, principal Hessian directions~\citep{li1992principal}, sliced average variance
estimator~\citep{cook1991sliced}, directional regression (DR)~\citep{li2007directional}, and so on. 
\end{definition}
This approach can be seen as the gradient flow with one-dimensional Wasserstein functional energy where the projecting direction is selected based on projection pursuit methods. Therefore, it is quite similar to the case of Max-SW (Remark~\ref{remark:optimization_based_slicing}). We refer the reader to~\citet{meng2019large} for a detailed discussion on asymptotic convergence of the map under some assumptions. 
\begin{figure}[!t]
    \centering
    \includegraphics[width=1\linewidth]{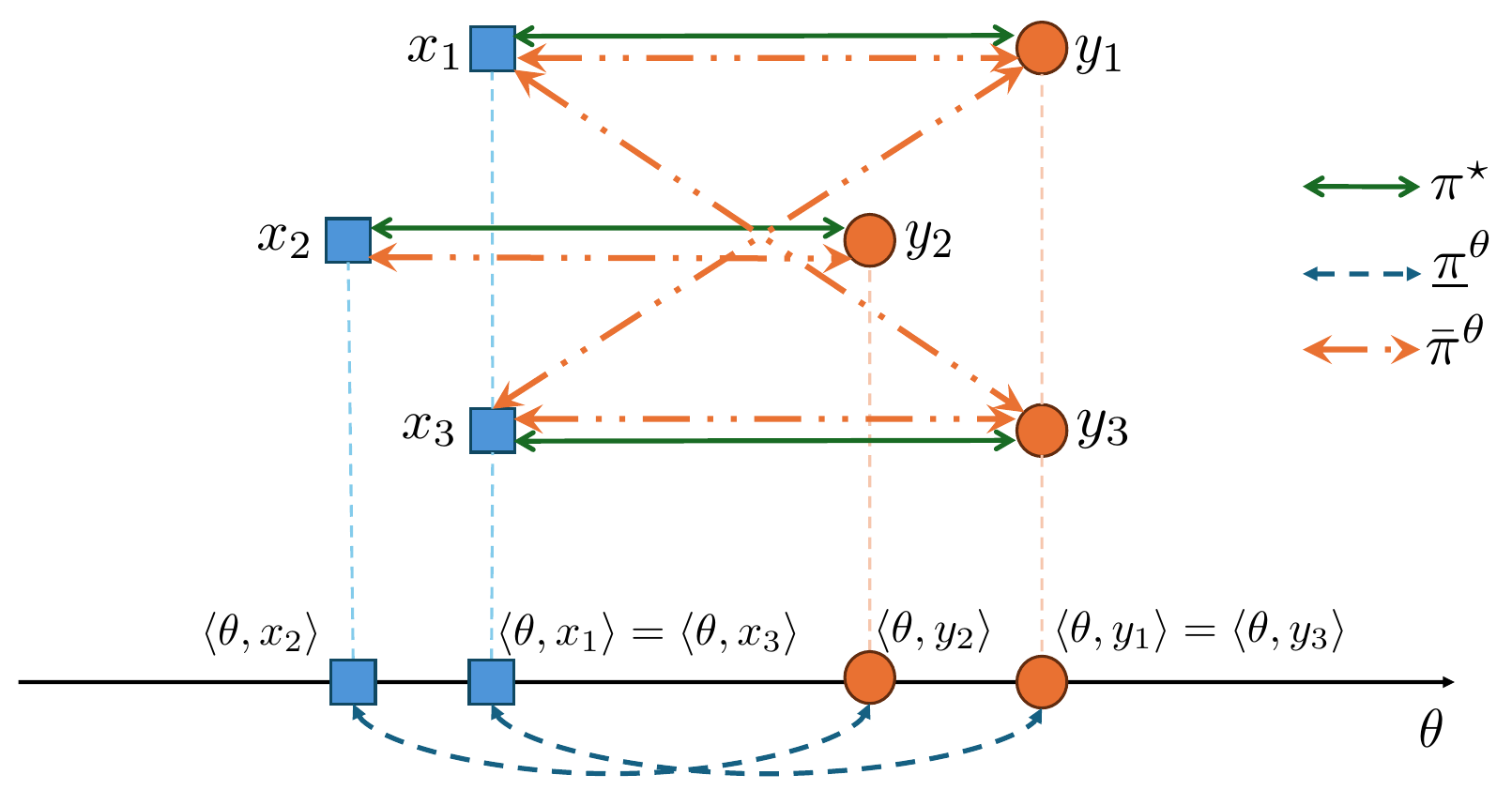}
    \caption{Optimal transport plan, one-dimensional optimal transport plan, and lifted transport plan.}
    \label{fig:lifted_plan}
\end{figure}

\paragraph{Explicit map for discrete measures.} In the discrete case, we can perform "lifting" of a one-dimensional transportation plan to the original space. In particular, the transportation plan in the discrete case can be defined via a doubly stochastic matrix. As a result, we can utilize the transportation matrices from one-dimensional cases for the original dimension.
 
\begin{definition}[Lifted Sliced Optimal Transport Plans]
    \label{def:lifed_SOT_plan}
    Given two discrete measures $\mu=\sum_{i=1}^n \alpha_i \delta_{x_i}$ and $\nu=\sum_{j=1}^m \beta_j \delta_{y_j}$, $\theta \in \Sm^{d-1}$,  and $\underline{\pi}^\theta$ is the optimal transportation plan between $\theta\sharp \mu$ and $\theta \sharp \nu$, the lifted sliced optimal transport plan is defined as~\citet{liu2025expected}:
    \begin{align}
    \bar{\pi}^\theta = \sum_{i=1}^n \sum_{j=1}^m \frac{\alpha_i \beta_j}{\alpha_i^\theta \beta_j^\theta} \underline{\pi}_{ij}^\theta(\{(\langle \theta,x_i\rangle,\langle \theta,y_j\rangle )\}) \delta_{(x_i,y_j)},
    \end{align}
    where $\alpha_i^\theta=\sum_{i'=1}^n \alpha_{i'} I(\langle \theta,x_i\rangle = \langle \theta,x_{i'}\rangle)$ and $\beta_j^\theta=\sum_{j'=1}^m \beta_{j'} I(\langle \theta,y_j\rangle = \langle \theta,y_{j'}\rangle)$. 
\end{definition}

From the lifted plan, we can also define a discrepancy between two measures. The discrepancy is coined as Sliced Wasserstein Generalized Geodesics (SWGG)~\citep{mahey2023fast} with empirical measures of $n$ atoms and then generalized in~\citet{liu2025expected} for discrete measures.
\begin{definition}[Sliced Wasserstein Generalized Geodesics]
    \label{def:SWGG}
    Given $p \geq 1$, two discrete measures $\mu=\sum_{i=1}^n \alpha_i \delta_{x_i}$ and $\nu=\sum_{j=1}^m \beta_j \delta_{y_j}$, and $\theta \in \Sm^{d-1}$, sliced Wasserstein generalized geodesics (SWGG)~\citep{mahey2023fast,liu2025expected} is defined as:
    \begin{align}
        SWGG_p^p(\mu,\nu;\theta) = \sum_{i=1}^n \sum_{j=1}^m \bar{\pi}_{ij}^{\theta}(\{(x_i,y_j)\}) \|x_i - y_j\|_p^p,
    \end{align}
    where $\bar{\pi}_{ij}^\theta$ is the lifted plan discussed in Definition~\ref{def:lifed_SOT_plan}. 
\end{definition}

We show the idea of lifting tranport plan in Figure~\ref{fig:lifted_plan}. By averaging SWGG over all possible $\theta$, we obtain the projected Wasserstein distance, which is a valid metric for discrete measures.

\begin{definition}[Projected Wasserstein distance and its plan]
    \label{def:PW} 
    Sliced Wasserstein is defined based on transportation costs between projected measures as the value. When $\mu$ and $\nu$ are discrete measures, i.e., $\mu=\sum_{i=1}^n \alpha_i \delta_{x_i}$ and $\nu=\sum_{j=1}^m \beta_j \delta_{y_j}$, we can define the Projected Wasserstein (PW) distance~\citep{rowland2019orthogonal} as follows:
    \begin{align}
        PW_p^p(\mu,\nu) = \mathbb{E}_{\theta \sim \setU(\Sm^{d-1})}\left[ SWGG_p^p(\mu,\nu;\theta) \right],
    \end{align}
    where $SWGG_p^p(\mu,\nu;\theta)$ is defined in Definition~\ref{def:SWGG}. Projected Wasserstein distance can be seen as the transport cost with the plan:
    \begin{align}
        \mathbb{E}_{\theta \sim \setU(\Sm^{d-1})}[\bar{\pi}^\theta].
    \end{align}
\end{definition}

\begin{proposition}[Projected Wasserstein distance is a valid metric for discrete measures]
    \label{proposition:PW_metricty} 
    For $p \geq 1$, the projected Wasserstein distance ($PW_{p}$) is a metric for discrete measures on $\setP_{p}(\Re^d)$. 
    \begin{proof}
        The proof for empirical measures is given in~\citep{rowland2019orthogonal}, while the proof for discrete measures is given in~\citep{liu2025expected}.
    \end{proof}
\end{proposition}

Similar to SW, PW also requires some numerical approximation for the expectation. For example, we can use Monte Carlo estimation or quasi-Monte Carlo estimation (Section~\ref{subsec:QMC}).

\begin{remark}[Computational Complexity of projected Wasserstein distance]
    \label{remark:computational_complexity_pw}
    For two discrete measures $\mu=\sum_{i=1}^n \alpha_i \delta_{x_i}$ and $\nu=\sum_{j=1}^m \beta_j \delta_{y_j}$ with at most $n$ supports, the computation of PW involves three components: sampling directions $\theta_1,\ldots,\theta_L$, conducting projections $\theta_l \sharp \mu$ and $\theta_l \sharp \nu$ for $l=1,\ldots,L$, and calculating
    \[
    \sum_{i=1}^n \sum_{j=1}^m \bar{\pi}_{ij}^{\theta_l}(\{(x_i,y_j)\}) \|x_i - y_j\|_p^p
    \]
    for each $l=1,\ldots,L$. The time complexity of the first step is $\mathcal{O}(L d)$ and space complexity is $\mathcal{O}(L d)$. The second step has time complexity $\mathcal{O}(L d n)$ and space complexity $\mathcal{O}(L n)$. The final step has time complexity $\mathcal{O}((L + d) n^2)$ and space complexity $\mathcal{O}(n^2)$. Overall, the total time complexity is $\mathcal{O}((L + d) n^2)$ and space complexity is $\mathcal{O}(n^2)$.
\end{remark}

As in the case of SW, we can also use slicing measures other than the uniform slicing measure. However, in contrast to SW, where the goal is to maximize the one-dimensional distance, here we aim to minimize the distance since it is computed in the original space.

\begin{definition}[Min Sliced Wasserstein Generalized Geodesics]
    \label{def:MinSWGG}  
    When $\mu$ and $\nu$ are discrete measures, i.e., $\mu=\sum_{i=1}^n \alpha_i \delta_{x_i}$ and $\nu=\sum_{j=1}^m \beta_j \delta_{y_j}$, we can define the Min Sliced Wasserstein Generalized Geodesics (Min-SWGG) distance~\citep{mahey2023fast} as:
    \begin{align}
        \text{Min-}SWGG_p^p(\mu,\nu) = \min_{\theta \in \Sm^{d-1}} SWGG_p^p(\mu,\nu;\theta),
    \end{align}
    where $SWGG_p^p(\mu,\nu;\theta)$ is defined in Definition~\ref{def:SWGG}. The transportation plan of Min-SWGG is $\bar{\pi}^{\theta^\star}$, where $\theta^\star$ is the minimizer of the optimization.
\end{definition}

\begin{proposition}[Metricity of Min-SWGG]
    \label{proposition:metricity_SWGG}
    Similar to PW, Min-SWGG is a valid metric for discrete measures~\citep{mahey2023fast}.
\end{proposition}

\begin{remark}[Computation of Min-SWGG]
    \label{remark:computation_minswgg}
    The optimization in Min-SWGG is challenging since the function $\theta \mapsto SWGG_p^p(\mu,\nu;\theta)$ is not differentiable. Authors in~\citet{mahey2023fast} propose to use simulated annealing or a smooth surrogate.
\end{remark}

\begin{proposition}[Connection between PW, Min-SWGG and Wasserstein distance]
    \label{proposition:PW_connect_Wasserstein} 
    For $p \geq 1$, and $\mu$, $\nu$ two discrete measures on $\setP_p(\Re^d)$, we have:
    \begin{align}
        W_p(\mu,\nu) \leq \text{Min-}SWGG_p^p(\mu,\nu) \leq PW_p(\mu,\nu).
    \end{align}
    The inequality is straightforward since
    \[
    W_p^p(\mu,\nu) \leq \sum_{i=1}^n \sum_{j=1}^m \bar{\pi}_{ij}^{\theta}(\{(x_i,y_j)\}) \|x_i - y_j\|_p^p,
    \]
    for any $\pi^\theta$ (any $\theta$) due to the optimality of the transportation plan in the Wasserstein distance.
\end{proposition}

\begin{definition}[Expected Sliced Transport]
    \label{def:EST}  
    When $\mu$ and $\nu$ are discrete measures, i.e., $\mu=\sum_{i=1}^n \alpha_i \delta_{x_i}$ and $\nu=\sum_{j=1}^m \beta_j \delta_{y_j}$, we define the Expected Sliced Transport (EST) distance~\citep{liu2025expected} as:
    \begin{align}
        EST_p^p(\mu,\nu) = \mathbb{E}_{\theta \sim \sigma}[SWGG_p^p(\mu,\nu;\theta)],
    \end{align}
    where $SWGG_p^p(\mu,\nu;\theta)$ is defined in Definition~\ref{def:SWGG} and $\sigma \in \setP(\Sm^{d-1})$. The authors in~\citep{liu2025expected} construct $\sigma$ as an energy-based slicing measure. In contrast to EBSW (Remark~\ref{remark:energy_based_slicing_measure}), the energy here is based on SWGG instead of the one-dimensional Wasserstein distance:
    \begin{align}
        p_\sigma(\theta) \propto \exp(-\tau SWGG_p^p(\mu,\nu;\theta)),
    \end{align}
    where $\tau \geq 0$ is the temperature parameter. The corresponding transportation plan is:
    \begin{align}
        \mathbb{E}_{\theta \sim \sigma}[\bar{\pi}^\theta].
    \end{align}
\end{definition}

\begin{proposition}[Connection between EST and Wasserstein]
    \label{proposition:connection_EST_Wasserstein}
    For $p \geq 1$, and $\mu, \nu$ two discrete measures in $\setP_p(\Re^d)$, we have:
    \begin{align}
        W_p(\mu,\nu) \leq EST_p(\mu,\nu).
    \end{align}
\end{proposition}

\begin{proposition}[Metricity of EST]
    \label{proposition:metricity_EST}
    EST is a valid metric for discrete measures~\citep{liu2025expected}.
\end{proposition}


Lastly, we can also go beyond linear projections by applying the projection methods discussed in Section~\ref{sec:generalized_slicing:chapter:advances}, which are designed to better adapt to the geometry of the space.

\begin{definition}[Generalized Sliced Wasserstein Plan]
    \label{def:generalized_SW_plan} 
    When $\mu$ and $\nu$ are discrete measures, i.e., $\mu=\sum_{i=1}^n \alpha_i \delta_{x_i}$ and $\nu=\sum_{j=1}^m \beta_j \delta_{y_j}$, we define the Generalized Sliced Wasserstein Plan (GSWP) distance~\citep{chapel2025differentiable} as:
    \begin{align}
        GSWP_p^p(\mu,\nu;\theta) = \sum_{i=1}^n \sum_{j=1}^m \bar{\pi}_{ij}^\theta(\{(x_i,y_j)\}) \| P_\theta(x_i) - P_\theta(y_j) \|_p^p,
    \end{align}
    where $\bar{\pi}_{ij}^\theta$ is the optimal discrete transportation matrix between $P_\theta \sharp \mu$ and $P_\theta \sharp \nu$, and $P_\theta$ ($\theta \in \Theta$) is a general projection function, which can be nonlinear (e.g., a neural network) as discussed in Section~\ref{sec:generalized_slicing:chapter:advances}.
\end{definition}

\begin{definition}[Min Generalized Sliced Wasserstein Plan]
    \label{def:min_generalized_SW_plan} 
    When $\mu$ and $\nu$ are discrete measures, i.e., $\mu=\sum_{i=1}^n \alpha_i \delta_{x_i}$ and $\nu=\sum_{j=1}^m \beta_j \delta_{y_j}$, we define the Min Generalized Sliced Wasserstein Plan (Min-GSWP) distance~\citep{chapel2025differentiable} as:
    \begin{align}
        \text{Min-}GSWP_p^p(\mu,\nu) = \min_{\theta \in \Theta} GSWP_p^p(\mu,\nu;\theta).
    \end{align}
    The transportation plan of Min-GSWP is $\bar{\pi}^{\theta^\star}$, where $\theta^\star$ is the minimizer of the optimization.
\end{definition}

\begin{remark}[Computation of Min-GSWP]
    \label{remark:computation_mingswp}
    Authors in~\citet{chapel2025differentiable} cast the problem as a bilevel optimization and propose a differentiable approximation scheme to efficiently identify the optimal projection parameter.
\end{remark}

\paragraph{Explicit plans for general measures.}  
To obtain explicit transportation plans with sliced optimal transport (SOT), authors in~\citet{muzellec2019subspace} propose ways to extrapolate
from an optimal transport map that is defined on a subspace to the entire space. We now adapt their proposal to the one-dimensional case as follows:

\begin{definition}[Monge-Independent Plan]
    \label{def:MI_plan}
    Given two measures $\mu$ and $\nu$, and $\theta \in \Sm^{d-1}$, let $T_\theta$ denote the Monge map from $\theta \sharp \mu$ to $\theta \sharp \nu$. The Monge-Independent Plan is defined as:
    \begin{align}
        \pi_{MI} = \mu_{t} \otimes \nu_{T_\theta(t)} \otimes (Id, T_\theta) \sharp (\theta \sharp \mu),
    \end{align}
    where $\mu$ is disintegrated as $\{\mu_t\}_{t}$ with respect to $(\theta \sharp \mu)(t)$, and $\nu$ is disintegrated as $\{\nu_{T_\theta(t)}\}_{t}$ with respect to $(\theta \sharp \nu)(t)$ (see Definition~\ref{def:disintegration_measures} for the disintegration of measures).
\end{definition}

\begin{definition}[Monge-Knothe Plan]
    \label{def:MK_plan}
    Given two measures $\mu$ and $\nu$, and $\theta \in \Sm^{d-1}$, let $T_\theta$ be the Monge map from $\theta \sharp \mu$ to $\theta \sharp \nu$. The Monge-Knothe Plan is defined as:
    \begin{align}
        \pi_{MK} = \pi_{OT}(\mu_t, \nu_{T_\theta(t)}) \otimes (Id, T_\theta) \sharp (\theta \sharp \mu),
    \end{align}
    where $\pi_{OT}(\mu_t, \nu_{T_\theta(t)})$ is the optimal transport plan between the conditional measures $\mu_t$ and $\nu_{T_\theta(t)}$. As before, $\mu$ and $\nu$ are disintegrated with respect to $\theta \sharp \mu$ and $\theta \sharp \nu$ (see Definition~\ref{def:disintegration_measures}).
\end{definition}

Authors in~\citet{muzellec2019subspace} derive explicit forms for the above plans when working with Gaussian distributions in the Bures metric space. It turns out that the lifted discrete plan in Definition~\ref{def:lifed_SOT_plan} is a special case of the Monge-Independent Plan, where $(Id,T_\theta) \sharp (\theta \sharp \mu)$ is replaced by $\pi_{OT}(\theta \sharp \mu, \theta \sharp \nu)$ to avoid issues when the Monge map does not exist. The second construction is recently proposed by~\citet{tanguy2025sliced}. These authors generalize EST to general measures and show that it satisfies non-negativity, symmetry, the triangle inequality, and identity of indiscernibles. Additionally, they propose the pivot sliced discrepancy and its corresponding transportation plan, which are based on the $\nu$-based Wasserstein distance.

\begin{definition}[$\nu$-based Wasserstein distance]
    \label{def:nubased_Wasserstein}
    Given a measure $\nu \in \setP_2(\Re^d)$, the $\nu$-based Wasserstein distance between $\mu, \mu' \in \setP_2(\Re^d)$ is defined as:
    \begin{align}
        W_\nu^2(\mu, \mu') = \min_{\pi \in \Pi(\nu, \mu, \mu')} \int_{\Re^d \times \Re^d \times \Re^d} \|x_1 - x_2\|_2^2 \, \diff \pi(y, x_1, x_2),
    \end{align}
    where $\Pi(\nu, \mu, \mu')$ denotes the set of multi-marginal transportation plans with marginals $\nu$, $\mu$, and $\mu'$. 
\end{definition}

The ``3-plan" set $\Pi(\nu, \mu, \mu')$ is tight (pre-compact), which implies that any sequence of measures in this set has a weakly convergent subsequence~\citep[Lemma 1]{tanguy2025sliced}. Using the $\nu$-based Wasserstein distance, SWGG can be generalized as follows:

\begin{definition}[Pivot sliced discrepancy]
    \label{def:SWGG_general}
    Given $\mu, \mu' \in \setP_2(\Re^d)$, the pivot sliced (PS) discrepancy between $\mu$ and $\mu'$ is defined as:
    \begin{align}
        PS(\mu, \mu'; \theta) = W_{\nu^\theta}(\mu, \mu'),
    \end{align}
    where $\nu^\theta$ is the Wasserstein barycenter of $Q_\theta \sharp \mu$ and $Q_\theta \sharp \mu'$, i.e.,
    \begin{align}
        \nu^\theta = \arg\min_{\nu \in \setP_2(\Re^d)} W_2^2(Q_\theta \sharp \mu, \nu) + W_2^2(Q_\theta \sharp \mu', \nu).
    \end{align}
    Here, $Q_\theta(x) = \langle \theta, x \rangle \theta$ is the projection of $x$ onto the direction $\theta$ in the original space (rather than the coordinate on the projected line). It is possible to generalize this to nonlinear projections. The pivot measure $\nu^\theta$ is unique and given by
     \begin{align}
        \nu^\theta = \left(\frac{1}{2} F_{\mu}^{-1} + \frac{1}{2} F_{\mu'}^{-1} \right) \theta \sharp \mathcal{U}([0,1]),
     \end{align}
    where $F_{\mu}^{-1}$ and $F_{\mu'}^{-1}$ are quantile functions.
\end{definition}

In~\citet{tanguy2025sliced}, the authors prove that PS is a semi-metric and generalizes SWGG for empirical measures with $n$ atoms. They also define Min-PS, which finds the direction $\theta$ minimizing $PS(\mu, \mu'; \theta)$. The transportation plan for PS is given by the optimal 3-plan $\pi \in \Pi(\nu^\theta, \mu, \mu')$.

\chapter{Variational Sliced Wasserstein Problems}
\label{chapter:varitational_SW}

In data analysis and machine learning, traditional divergences such as the Euclidean distance, Kullback–Leibler divergence, and total variation are commonly used to compare probability measures and estimate their parameters. However, these divergences often fail to capture the geometric structure or spatial relationships inherent in the data. The Wasserstein distance addresses this limitation by incorporating a ground cost that reflects the geometry of the underlying space. As discussed in earlier chapters, sliced Wasserstein (SW) distances can serve as efficient alternatives to the full Wasserstein distance, especially in settings with computational constraints. This chapter explores how SW distances provide practical and versatile tools for problems involving probability measures. Specifically, we review statistical estimators based on SW, analyze the differentiability of SW distances, solve barycenter problems using SW, study flows in the SW space, construct kernels between probability measures using SW, and investigate embedding techniques that preserve the SW geometry.

\section{Minimum Sliced Wasserstein Estimators}
\label{sec:MSWE:chapter:varitational_SW}
In parametric statistical inference, we want to estimate some parameters of interest from the data. Minimum distance estimation (MDE)~\citep{wolfowitz1957minimum} is a generalization of maximum log-likelihood inference. Using SW as the distance, we can obtain the minimum sliced Wasserstein estimator (MSWE).

\begin{definition}[Minimum sliced Wasserstein estimator]
    \label{def:MSWE}
    Given $p \geq 1$, data $X_1,\ldots,X_n \simiid \mu$ ($\mu \in \setP_p(\setX)$) and a parametric family $\mu_\phi$ with $\phi \in \Phi$, the minimum sliced Wasserstein estimator (MSWE)~\citep{nadjahi2019asymptotic} is formally defined as follows:
    \begin{align}
        \hat{\phi}_n = \operatorname{argmin}_{\phi \in \Phi} SW_p(\mu_n,\mu_\phi),
    \end{align}
    where $\mu_n = \frac{1}{n}\sum_{i=1}^n\delta_{X_i}$ is the empirical distribution over the data.
\end{definition}

\begin{figure}[!t]
    \centering
    \includegraphics[width=1\linewidth]{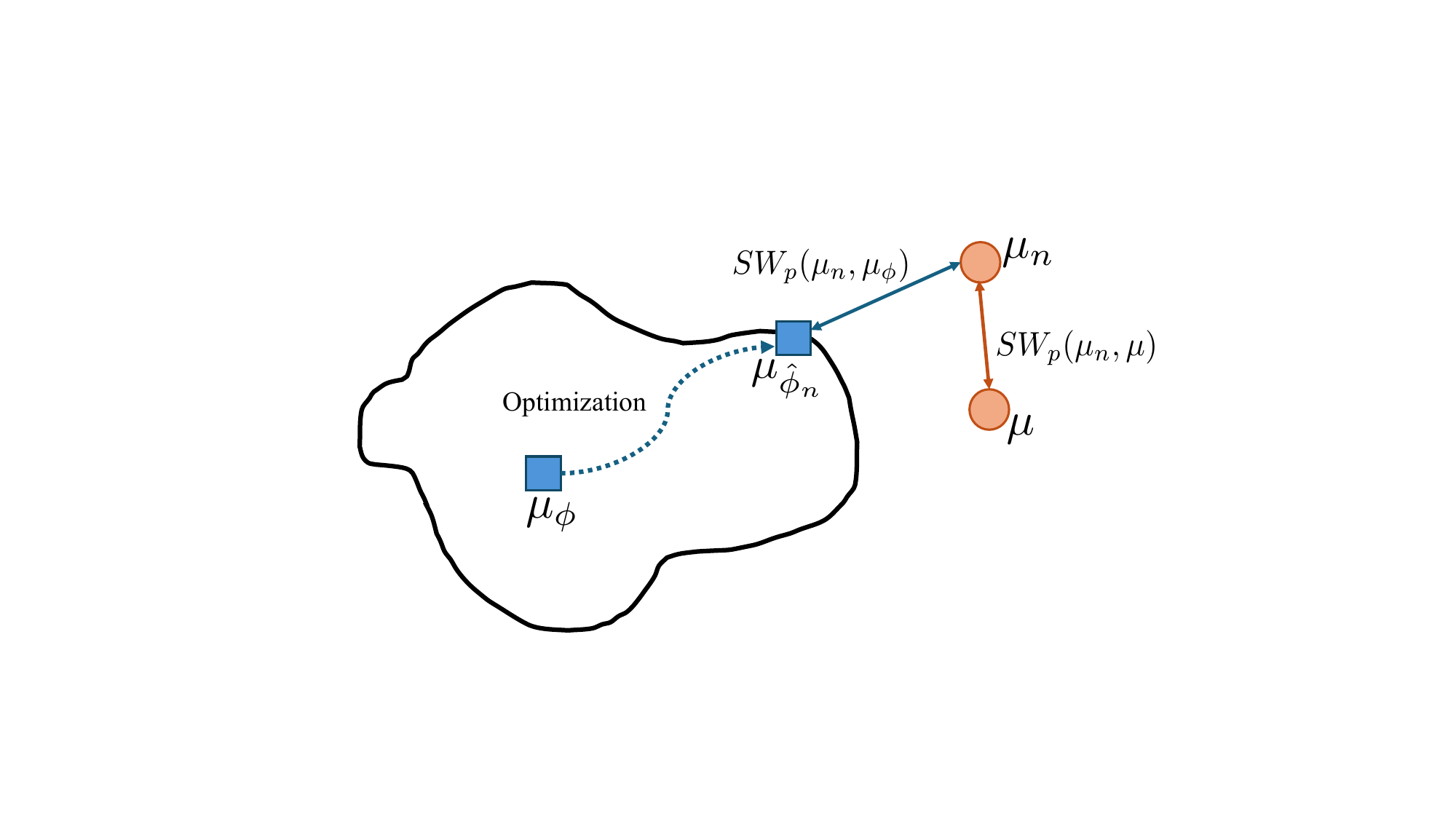}
    \caption{Minimum sliced Wasserstein estimators.}
    \label{fig:MEDE}
\end{figure}

We show the idea of MSWE in Figure~\ref{fig:MEDE}. When the form of $\mu_\phi$ is not available, we can still do inference if we can generate samples from the distribution $\mu_\phi$. The resulting estimator is known as the minimum expected distance estimator (MEDE)~\citep{bernton2019parameter}.

\begin{definition}[Minimum expected sliced Wasserstein estimator]
    \label{def:MESWE}
    Given $p \geq 1$, data $X_1,\ldots,X_n \simiid \mu$ ($\mu \in \setP_p(\setX)$) and a parametric family $\mu_\phi$ with $\phi \in \Phi$, the minimum expected sliced Wasserstein estimator (MESWE)~\citep{nadjahi2019asymptotic} is formally defined as follows:
    \begin{align}
        \hat{\phi}_{n,m} = \operatorname{argmin}_{\phi \in \Phi} \mathbb{E}[ SW_p(\mu_n,\mu_{\phi,m}) \mid X_1,\ldots,X_n],
    \end{align}
    where $\mu_n = \frac{1}{n}\sum_{i=1}^n\delta_{X_i}$ is the empirical distribution over the data, and $\mu_{\phi,m} = \frac{1}{m}\sum_{j=1}^m \delta_{Y_{\phi,j}}$ for $Y_{\phi,1},\ldots,Y_{\phi,m} \simiid \mu_\phi$.
\end{definition}

Both MSWE and MESWE play important roles in applications of SW which will be discussed later in Chapter~\ref{chapter:applications}. Next, we review existence and consistency of MSWE and MESWE, and their relationship. First, we state some assumptions:

\textbf{A1.} Let $c_\Phi$ is a metric on $\Phi$ and $c_\setP$ is the Lévy–Prokhorov metric. The map $\phi \mapsto \mu_\phi$ is continuous from $(\Phi,c_\Phi)$ to $(\setP_p(\setX),c_\setP)$; i.e., if $\lim_{n \to \infty} c_\Phi(\phi_n,\phi) = 0$, then $(\mu_{\phi_n})_{n \in \mathbb{N}} \overset{d}{\to} \mu_\phi$ (weak convergence or convergence in distribution). 

\textbf{A2.} The data-generating process is such that $\lim_{n \to \infty} SW_p(\mu_n,\mu) = 0$ almost surely.

\textbf{A3.} There exists $\epsilon > 0$ such that $\epsilon_\star = \inf_{\phi \in \Phi} SW_p(\mu,\mu_\phi)$, and the set $\Phi^\star_\epsilon = \{ \phi \in \Phi \mid SW_p(\mu,\mu_\phi) \leq \epsilon + \epsilon_\star \}$ is bounded.

\begin{proposition}[Existence and consistency of MSWE]
    \label{proposition:existence_consistency_MSWE}
    Assume A1, A2, and A3. There exists $E \in \setF$ with $P(E) = 1$ ($\setF$ is the $\sigma$-algebra on $\setX$) such that for all $w \in E$,
    \begin{align}
        &\lim_{n \to \infty} \inf_{\phi \in \Phi} SW_p(\mu_n(w), \mu_\phi) = \inf_{\phi \in \Phi} SW_p(\mu, \mu_\phi), \\
        &\limsup_{n \to \infty} \operatorname{argmin}_{\phi \in \Phi} SW_p(\mu_n(w), \mu_\phi) \subset \operatorname{argmin}_{\phi \in \Phi} SW_p(\mu, \mu_\phi).
    \end{align}
    Furthermore, for all $w \in E$, there exists $n(w)$ such that for all $n \geq n(w)$, the set $\operatorname{argmin}_{\phi \in \Phi} SW_p(\mu_n(w), \mu_\phi)$ is non-empty.
    The proof is given in~\citet[Theorem 2]{nadjahi2019asymptotic}.
\end{proposition}

\textbf{A4.} If $\lim_{n \to \infty} c_\Phi(\phi_n,\phi) = 0$, then $\lim_{n \to \infty} \mathbb{E}[SW_p(\mu_{\phi_n}, \mu_{\phi_n,n}) \mid X_1,\ldots,X_n] = 0$.

\begin{proposition}[Existence and consistency of MESWE]
    \label{proposition:existence_consistency_MESWE}
    Assume A1, A2, A3, and A4. Let $(m(n))_{n \in \mathbb{N}\setminus \{0\}}$ be an increasing sequence satisfying $\lim_{n \to \infty} m(n) = \infty$. There exists a set $E \in \setX$ with
    $P(E) = 1$ such that, for all $w \in E$,
    \begin{align}
        &\lim_{n \to \infty} \inf_{\phi \in \Phi} \mathbb{E}[SW_p(\mu_n, \mu_{\phi,m(n)}) \mid X_1,\ldots,X_n] = \inf_{\phi \in \Phi} SW_p(\mu, \mu_\phi), \\
        &\limsup_{n \to \infty} \operatorname{argmin}_{\phi \in \Phi} \mathbb{E}[SW_p(\mu_n, \mu_{\phi,m(n)}) \mid X_1,\ldots,X_n] \nonumber \\ 
        &\quad \quad \quad \subset \operatorname{argmin}_{\phi \in \Phi} SW_p(\mu, \mu_\phi),
    \end{align}
    Furthermore, for all $w \in E$, there exists $n(w)$ such that for all $n \geq n(w)$, the set
    \[
    \operatorname{argmin}_{\phi \in \Phi} \mathbb{E}[SW_p(\mu_n, \mu_{\phi,m(n)}) \mid X_1,\ldots,X_n]
    \]
    is non-empty.
    The proof is given in~\citet[Theorem 3]{nadjahi2019asymptotic}.
\end{proposition}

\textbf{A5.} For some $\epsilon > 0$, and $\epsilon_n = \inf_{\phi \in \Phi} SW_p(\mu_n, \mu_\phi)$, the set $\Phi_{\epsilon,n} = \{ \phi \in \Phi \mid SW_p(\mu_n, \mu_\phi) \leq \epsilon_n + \epsilon \}$ is bounded almost surely.

\begin{proposition}[Convergence of MESWE to MSWE]
    \label{proposition:convergence_MESW_MSWE}
    Assume A1, A4, and A5. Then,
    \begin{align}
        &\lim_{m \to \infty} \inf_{\phi \in \Phi} \mathbb{E}[SW_p(\mu_n, \mu_{\phi,m}) \mid X_1,\ldots,X_n] = \inf_{\phi \in \Phi} SW_p(\mu_n,\mu_\phi), \\
        &\limsup_{m \to \infty} \operatorname{argmin}_{\phi \in \Phi} \mathbb{E}[SW_p(\mu_n, \mu_{\phi,m}) \mid X_1,\ldots,X_n] \nonumber \\
        & \quad \quad \quad \subset \operatorname{argmin}_{\phi \in \Phi} SW_p(\mu_n, \mu_\phi),
    \end{align}
    Furthermore, there exists $m^\star$ such that, for any $m \geq m^\star$, the set
    \[
    \operatorname{argmin}_{\phi \in \Phi} \mathbb{E}[SW_p(\mu_n, \mu_{\phi,m}) \mid X_1,\ldots,X_n]
    \]
    is non-empty.
    The proof is given in~\citet[Theorem 4]{nadjahi2019asymptotic}.
\end{proposition}

We refer the reader to~\citet[Theorems 5 and 6]{nadjahi2019asymptotic} for results on the asymptotic distribution of MSWE and the rate of convergence.

\section{Differentiating Sliced Wasserstein Losses}
\label{sec:differentiatingSW:chapter:varitational_SW}

In many applications, we are interested in solving optimization problems of the form:
\begin{align}
\label{eq:min_SW}
    \min_{\phi \in \Phi} SW_p^p(\mu_\phi,\nu),
\end{align}
where $\mu_\phi$ is a parameterized probability measure (e.g., generated by a model with parameters $\phi$), and $\nu$ is a target distribution. The objective is to minimize the $p$-th power of the SW distance between $\mu_\phi$ and $\nu$, thereby aligning the two distributions as closely as possible. This formulation appears in various contexts, such as in the minimum SW estimator discussed in Section~\ref{sec:MSWE:chapter:varitational_SW}. To perform optimization procedures such as gradient descent, we are interested in obtaining the gradient $\nabla_\phi SW_p^p(\mu_\phi,\nu)$. The gradient depends on the mapping $\phi \mapsto \mu_\phi$, which is problem-specific. When the parameter $\phi$ is itself a histogram (a probability vector),~\eqref{eq:min_SW} is a convex problem.

\begin{proposition}[Convexity of $SW_p^p$]
    \label{proposition:convexity_SW}
    Let $\mu_1, \mu_2, \nu \in \mathcal{P}_p(\mathbb{R}^d)$ for $p \geq 1$, and let $\lambda \in [0,1]$. Define the convex combination $\mu_\lambda = \lambda \mu_1 + (1 - \lambda)\mu_2$. Then the sliced Wasserstein distance is convex in its first argument:
    \begin{align}
    \mathrm{SW}_p^p(\mu_\lambda, \nu) \leq \lambda \, \mathrm{SW}_p^p(\mu_1, \nu) + (1 - \lambda)\, \mathrm{SW}_p^p(\mu_2, \nu).
    \end{align}
    \begin{proof}
      For each direction $\theta \in S^{d-1}$, we have:
      \[
      \theta \sharp \mu_\lambda = \lambda \theta \sharp \mu_1 + (1 - \lambda) \theta \sharp \mu_2.
      \]
      Since $W_p^p(\cdot, \cdot)$ is convex in one of its arguments, it follows that:
      \[
      W_p^p(\theta \sharp \mu_\lambda, \theta \sharp \nu) \leq \lambda W_p^p(\theta \sharp \mu_1, \theta \sharp \nu) + (1 - \lambda) W_p^p(\theta \sharp \mu_2, \theta \sharp \nu).
      \]

      Now, from the definition of SW:
      \begin{align*}
      &SW_p^p(\mu_\lambda, \nu) \\
      &= \mathbb{E}_{\theta \sim \setU(\Sm^{d-1})}[ W_p^p(\theta \sharp \mu_\lambda, \theta \sharp \nu)] \\
      &\leq \mathbb{E}_{\theta \sim \setU(\Sm^{d-1})}\left[ \lambda W_p^p(\theta \sharp \mu_1, \theta \sharp \nu) + (1 - \lambda) W_p^p(\theta \sharp \mu_2, \theta \sharp \nu) \right] \\
      &= \lambda \mathbb{E}_{\theta \sim \setU(\Sm^{d-1})}[ W_p^p(\theta \sharp \mu_1, \theta \sharp \nu)] + (1 - \lambda) \mathbb{E}_{\theta \sim \setU(\Sm^{d-1})}[ W_p^p(\theta \sharp \mu_2, \theta \sharp \nu)] \\
      &= \lambda\, \mathrm{SW}_p^p(\mu_1, \nu) + (1 - \lambda)\, \mathrm{SW}_p^p(\mu_2, \nu).
      \end{align*}
    \end{proof}
\end{proposition}

For a general mapping $\phi \mapsto \mu_\phi$,~\eqref{eq:min_SW} is in general not convex. We now discuss some other parameterizations of $\mu_\phi$ including discrete measures with learnable weights, discrete measures with learnable atoms, and pushforward measures.

\begin{remark}[Derivative with respect to weights in discrete cases]
    \label{remark:derivative_weight}
    Let $\mu_\phi = \sum_{i=1}^n \phi_i \delta_{x_i}$ and $\nu = \sum_{j=1}^m \beta_j \delta_{y_j}$. We have:
    \begin{align*}
        \nabla_\phi SW_p^p(\mu_\phi,\nu) &= \nabla_\phi \mathbb{E}_{\theta \sim \setU(\Sm^{d-1})}[W_p^p(\theta \sharp \mu_\phi,\theta \sharp \nu)] \\
        &= \mathbb{E}_{\theta \sim \setU(\Sm^{d-1})}[\nabla_\phi W_p^p(\theta \sharp \mu_\phi,\theta \sharp \nu)].
    \end{align*}
    We have:
    \begin{align*}
        \nabla_\phi W_p^p(\theta \sharp \mu_\phi,\theta \sharp \nu) = \nabla_\phi
        \left(\max_{(\fb,\gb) \in \Rb(\Cb_\theta)} \sum_{i=1}^n f_i \phi_i + \sum_{j=1}^m g_j \beta_j \right),
    \end{align*}
    where we use Kantorovich duality (Definition~\ref{def:Kantorovich_discrete_dual}) with $\fb = (f_1,\ldots,f_n)$, $\gb = (g_1,\ldots,g_m)$, $\Cb_\theta = [[C_{ij}]]$ with $C_{\theta,ij} = |\langle \theta, x_i \rangle - \langle \theta, y_j \rangle|$, and
    \[
    \Rb(\Cb_\theta) = \{ (\fb,\gb) \in \Re^n \times \Re^m \mid f_i + g_j \leq C_{\theta,ij} \quad \forall i \in \intset{n}, j \in \intset{m} \}.
    \]
    We can solve for $\fb_{\theta,p}^\star$ and $\gb_{\theta,p}^\star$ efficiently (Remark~\ref{remark:1DKantorovich_discrete}). Using Danskin's envelope theorem (assuming compactness), we have:
    \begin{align*}
        \nabla_\phi W_p^p(\theta \sharp \mu,\theta \sharp \nu) = \fb_{\theta,p}^\star.
    \end{align*}
    Therefore, we obtain:
    \begin{align}
         \nabla_\phi SW_p^p(\mu_\phi,\nu) =  \mathbb{E}_{\theta \sim \setU(\Sm^{d-1})}[\nabla_\phi \fb_{\theta,p}^\star],
    \end{align}
    which can be estimated using Monte Carlo methods.
\end{remark}

\begin{remark}[Derivative with respect to atoms in discrete cases]
    \label{remark:derivative_atoms}
    Let $\mu_\phi = \sum_{i=1}^n \alpha_i \delta_{\phi_i}$ and $\nu = \sum_{j=1}^m \beta_j \delta_{y_j}$. We have:
    \begin{align*}
        \nabla_\phi SW_p^p(\mu_\phi,\nu) &= \nabla_\phi \mathbb{E}_{\theta \sim \setU(\Sm^{d-1})}[W_p^p(\theta \sharp \mu_\phi,\theta \sharp \nu)] \\
        &= \mathbb{E}_{\theta \sim \setU(\Sm^{d-1})}[\nabla_\phi W_p^p(\theta \sharp \mu_\phi,\theta \sharp \nu)].
    \end{align*}
    We have:
    \begin{align*}
        \nabla_{\phi_i} W_p^p(\theta \sharp \mu_\phi,\theta \sharp \nu) = \nabla_{\phi_i}
        \left( \min_{\pi \in \Pi(\mu_\phi,\nu)} \sum_{i=1}^n \sum_{j=1}^m |\langle \theta,\phi_i\rangle - \langle \theta,y_j\rangle |^p \pi_{ij} \right).
    \end{align*}
    We can solve for $\pi_{\theta}^\star$ efficiently (Remark~\ref{remark:1DKantorovich_discrete}). Using Danskin's envelope theorem~\citep{danskin2012theory}, we have:
    \begin{align*}
        \nabla_{\phi_i} W_p^p(\theta \sharp \mu_\phi,\theta \sharp \nu) &= \nabla_{\phi_i}
        \left( \sum_{i=1}^n \sum_{j=1}^m |\langle \theta, \phi_i \rangle - \langle \theta, y_j \rangle|^p \pi_{\theta,ij}^\star \right) \\
        &= p \sum_{j=1}^m \left| \langle \theta, \phi_i - y_j \rangle \right|^{p-1} \operatorname{sign}\left( \langle \theta, \phi_i - y_j \rangle \right) \theta \, \pi_{\theta,ij}^\star.
    \end{align*}
    Therefore, we obtain:
    \begin{align}
         &\nabla_{\phi_i} SW_p^p(\mu_\phi,\nu) \nonumber \\ &=  \mathbb{E}_{\theta \sim \setU(\Sm^{d-1})} \left[
         p \sum_{j=1}^m \left| \langle \theta, \phi_i - y_j \rangle \right|^{p-1} \operatorname{sign}\left( \langle \theta, \phi_i - y_j \rangle \right) \theta \, \pi_{\theta,ij}^\star
         \right],
    \end{align}
    which can be estimated using Monte Carlo methods. 
\end{remark}
For more details on the regularity and optimization properties of the function $\phi \mapsto SW_p^p(\mu_\phi,\nu)$ in this case, we refer the reader to the recent work~\citet{tanguy2025properties}.

\begin{remark}[Derivative with respect to pushforward function]
    Let $\mu_\phi = f_\phi \sharp \mu$ for a function $f_\phi$ with $\mu$ and $\nu$ being continuous measures on $\mathbb{R}^d$. We have
    \[
    \nabla_\phi W_p^p(\theta \sharp f_\phi \sharp \mu, \theta \sharp \nu) =  \nabla_\phi \int_0^1 \left| F_{\theta \sharp f_\phi \sharp \mu}^{-1}(t) - F_{\theta \sharp \nu}^{-1}(t) \right|^p \diff t,
    \]
    where we differentiate under the integral:
    \begin{align*}
        \nabla_\phi W_p^p(\theta \sharp f_\phi \sharp \mu, \theta \sharp \nu) &= p \int_0^1 \left| F_{\theta \sharp f_\phi \sharp \mu}^{-1}(t) - F_{\theta \sharp \nu}^{-1}(t) \right|^{p-1} \\
        &\quad \times \operatorname{sign}\left( F_{\theta \sharp f_\phi \sharp \mu}^{-1}(t) - F_{\theta \sharp \nu}^{-1}(t) \right) \nabla_\phi F_{\theta \sharp f_\phi \sharp \mu}^{-1}(t) \diff t.
    \end{align*}
    Next, we calculate $\nabla_\phi F_{\theta \sharp f_\phi \sharp \mu}^{-1}(t)$. Since 
    \[
    F_{\theta \sharp f_\phi \sharp \mu}(F_{\theta \sharp f_\phi \sharp \mu}^{-1}(t)) = t,
    \]
    differentiating both sides with respect to $\phi$ gives:
    \[
    \nabla_\phi F_{\theta \sharp f_\phi \sharp \mu}(F_{\theta \sharp f_\phi \sharp \mu}^{-1}(t)) + p_{\theta \sharp f_\phi \sharp \mu}(F_{\theta \sharp f_\phi \sharp \mu}^{-1}(t)) \, \nabla_\phi F_{\theta \sharp f_\phi \sharp \mu}^{-1}(t) = 0,
    \]
    where $p_{\theta \sharp f_\phi \sharp \mu}$ is the density of $\theta \sharp f_\phi \sharp \mu$. This leads to:
    \[
    \nabla_\phi F_{\theta \sharp f_\phi \sharp \mu}^{-1}(t) = - \frac{\nabla_\phi F_{\theta \sharp f_\phi \sharp \mu}(F_{\theta \sharp f_\phi \sharp \mu}^{-1}(t))}{p_{\theta \sharp f_\phi \sharp \mu}(F_{\theta \sharp f_\phi \sharp \mu}^{-1}(t))}.
    \]
    Next, observe that
    \[
    F_{\theta \sharp f_\phi \sharp \mu}(x) = \int_{\mathbb{R}^d} \mathbf{1}\big(\langle \theta, f_\phi(z) \rangle \leq x \big) \diff \mu(z).
    \]
    Differentiating with respect to $\phi$ (formally, using the chain rule and distributional derivative), we get:
    \[
    \nabla_\phi F_{\theta \sharp f_\phi \sharp \mu}(x) = - \int_{\mathbb{R}^d} \delta\big(\langle \theta, f_\phi(z) \rangle - x \big) J_{f_\phi}(z)^\top \theta \, \diff \mu(z),
    \]
    where $J_{f_\phi}$ is the Jacobian of $f_\phi$.
    
    Thus,
    \[
    \nabla_\phi F_{\theta \sharp f_\phi \sharp \mu}^{-1}(t) = - \frac{- \int_{\mathbb{R}^d} \delta\big(\langle \theta, f_\phi(z) \rangle - F_{\theta \sharp f_\phi \sharp \mu}^{-1}(t) \big) J_{f_\phi}(z)^\top \theta \, \diff \mu(z)}{p_{\theta \sharp f_\phi \sharp \mu}(F_{\theta \sharp f_\phi \sharp \mu}^{-1}(t))}.
    \]
    Putting it all together, we obtain the gradient:
    \begin{align}
        &\nabla_\phi W_p^p(\theta \sharp f_\phi \sharp \mu, \theta \sharp \nu) \nonumber\\ &= p \int_0^1 \left| F_{\theta \sharp f_\phi \sharp \mu}^{-1}(t) - F_{\theta \sharp \nu}^{-1}(t) \right|^{p-1} \nonumber \\
        &\quad \times \operatorname{sign}\left( F_{\theta \sharp f_\phi \sharp \mu}^{-1}(t) - F_{\theta \sharp \nu}^{-1}(t) \right) \nonumber \\
        &\quad \times \left(- \frac{- \int_{\mathbb{R}^d} \delta\big(\langle \theta, f_\phi(z) \rangle - F_{\theta \sharp f_\phi \sharp \mu}^{-1}(t) \big) J_{f_\phi}(z)^\top \theta \, \diff \mu(z)}{p_{\theta \sharp f_\phi \sharp \mu}(F_{\theta \sharp f_\phi \sharp \mu}^{-1}(t))} \right) \diff t.
    \end{align}
    Taking the expectation with respect to $\theta \sim \setU(\Sm^{d-1})$, we obtain the final gradient, which can be estimated with Monte Carlo methods.
\end{remark}

\section{Sliced Wasserstein Barycenter}
\label{sec:SWB:chapter:varitational_SW}
The Wasserstein barycenter~\citep{agueh2011barycenters} extends the classical notion of averaging to the space of probability measures. It is defined as the probability distribution that minimizes a weighted sum of Wasserstein distances to a given set of input measures. This construction can be understood as the Fréchet mean~\citep{grove1973conjugate} with respect to the Wasserstein metric. Although Wasserstein barycenters are geometrically meaningful, computing them in high-dimensional spaces is often computationally intensive. To address this, Sliced Wasserstein barycenters offer a more tractable alternative by replacing the full Wasserstein distance with its sliced counterpart. The result is a more efficient yet expressive method for barycenter computation. We first review the definition of the Wasserstein barycenter.

\begin{definition}[Wasserstein Barycenter]
    \label{def:WB}  Given $p\geq1$, $K\geq 1$ marginals $\mu_1,\ldots,\mu_{K} \in \mathcal{P}_p(\mathbb{R}^d)$, weights $w_1,\ldots,w_K \in \Re_+$  ($\sum_{k=1}^K w_k=1$), the Wasserstein barycenter~\citep{agueh2011barycenters} is defined as:
    \begin{align}
        WB_p(\mu_{1:K},w_{1:K})  = \text{argmin}_{\mu \in \setP_p(\Re^d)} \sum_{k=1}^K w_k W_p^p(\mu_k,\mu).
    \end{align}
\end{definition}
It is possible to generalize the Wasserstein distance to other ground metrics. The Wasserstein barycenter problem is convex but does not necessarily have a unique solution.  In one dimension, the Wasserstein barycenter has a closed form.

\begin{proposition}[One-dimensional Wasserstein barycenter]
    \label{proposition:1DWB} When $p=2$ and $d=1$,  the Wasserstein barycenter admits a closed form:
\begin{align}
    \mu=WB_p(\mu_{1:K},w_{1:K}) = \left(\sum_{k=1}^K w_k T_k\right)\sharp \mu_0,
\end{align}
where $\mu_0$ is a continuous measure and $T_k = F^{-1}_{\mu_k} \circ F_{\mu}$ is the optimal transport map from $\mu_0$ to $\mu_k$. If $\mu_0 = \mathcal{U}([0,1])$, then $T_k = F_{\mu_k}^{-1}$, and we recover the classical formula for the quantile function of the barycenter:
\begin{align}
F_{\mu}^{-1}(t) = \sum_{k=1}^K w_k F_{\mu_k}^{-1}(t).
\end{align}
\begin{proof}
 
We seek the Wasserstein-2 barycenter $\mu=WB_2(\mu_{1:K},w_{1:K})$ defined by
$$
\mu = \text{argmin}_{\mu\in \setP_2(\Re)} \sum_{k=1}^K w_k W_2^2(\mu, \mu_k).
$$
Assume $\mu = T \sharp \mu_0$ for some measurable map $T$. Then
$$
W_2^2(\mu, \mu_k) = \int_{\Re} |T(t) - T_k(t)|^2 \, d\mu_0(t),
$$
so the barycenter problem becomes:
$$
\min_T \sum_{k=1}^K w_k \int_\Re |T(t) - T_k(t)|^2 \, d\mu_0(t).
$$
This functional is minimized pointwise in $t$, so for each $t \in \Re$, we solve:
$$
T(t) = \arg\min_{x \in \mathbb{R}} \sum_{k=1}^K w_k |x - T_k(t)|^2.
$$

This is a standard quadratic minimization in $x$, whose unique minimizer is the weighted average (taking the derivative and setting it to 0):
$$
T(t) = \sum_{k=1}^K w_k T_k(t).
$$
Therefore, the Wasserstein-2 barycenter is given by
\begin{align}
\mu = T \sharp \mu_0 = \left( \sum_{k=1}^K w_k T_k \right) \sharp \mu_0,
\end{align}
which completes the proof.
\end{proof}
\end{proposition}

To utilize the closed-form benefit in one dimension, the Radon transform is again used. There are two ways of finding barycenters given the transformed Radon space. The first way is to find a Wasserstein barycenter in the transformed space, then use the inverse Radon transform to map the barycenter back to the original space. The second way is to find the barycenter in the original space using the SW distance. We first start with the first approach, which is known as the Radon Wasserstein barycenter.

\begin{definition}[Radon Wasserstein Barycenter]
    \label{def:Radon_Wasserstein_barycenter} Given $p\geq1$, $K\geq 1$ marginals $\mu_1,\ldots,\mu_{K} \in \mathcal{P}_p(\mathbb{R}^d)$, weights $w_1,\ldots,w_K \in \Re_+$  ($\sum_{k=1}^K w_k=1$), the Radon Wasserstein barycenter~\citep{bonneel2015sliced} is defined as:
    \begin{align}
        &RWB_p(\mu_{1:K},w_{1:K})  = \setR^{-1} WB_p(\setR \mu_{1:K},w_{1:K}), \\
       & WB_p(\setR \mu_{1:K},w_{1:K})  =  \text{argmin}_{\mu \in \setP_p(\Sm^{d-1} \times \Re)} \sum_{k=1}^K w_k W_p^p(\setR \mu_k, \mu).
    \end{align}
    where $\setR$ denotes the Radon transform operator (Definition~\ref{def:Radon_Transform_measures}) and $\setR^{-1}$ denotes its inverse (Definition~\ref{def:Inverse_Radon_Transform_measures}).
\end{definition}

Radon Wasserstein barycenter is invariant to scaling, translation, and rotation, as is the classical Wasserstein barycenter. Since $\setR \mu_1, \ldots, \setR \mu_K$ have the same marginal on $\Sm^{d-1}$ (uniform measure), the barycenter $\mu$ has the same marginal on $\Sm^{d-1}$. For the marginal on $\Re$, the closed-form solution in Proposition~\ref{proposition:1DWB} is applied. The Radon Wasserstein barycenter can be used to define a geodesic between two measures.

\begin{definition}[Radon Wasserstein Geodesic]
    \label{def:Radon_Wasserstein_Geodesic} When $K=2$ and $t \in [0,1]$, the curve
    \begin{align}
        \mu_t \to  RWB_p(\mu_1,\mu_2,t,1-t),
    \end{align}
    can be seen as the geodesic curve between $\mu_1 \in \mathcal{P}_p(\mathbb{R}^d)$ and $\mu_2 \in \mathcal{P}_p(\mathbb{R}^d)$ in the Radon Wasserstein space~\citep{kolouri2017optimal}. 
\end{definition}

Next, we review the second approach, which is known as the sliced Wasserstein barycenter. Similar to the Radon Wasserstein barycenter, we can also use this type of barycenter to define a geodesic between two measures using a different metric space.

\begin{definition}[Sliced Wasserstein Barycenter]
    \label{def:SWB}
 Given $p\geq1$, $K\geq 1$ marginals $\mu_1,\ldots,\mu_{K} \in \mathcal{P}_p(\mathbb{R}^d)$, weights $w_1,\ldots,w_K \in \Re_+$  ($\sum_{k=1}^K w_k=1$), the Wasserstein barycenter~\citep{bonneel2015sliced} (SWB) is defined as:
    \begin{align}
        SWB_p(\mu_{1:K},w_{1:K})  = \text{argmin}_{\mu \in \setP_p(\Re^d)} \sum_{k=1}^K w_k SW_p^p(\mu_k,\mu).
    \end{align}
\end{definition}
The sliced Wasserstein barycenter belongs to the set of Radon Wasserstein barycenters~\citep[Proposition 10]{bonneel2015sliced}.

\begin{figure}[!t]
    \centering
    \includegraphics[width=1\linewidth]{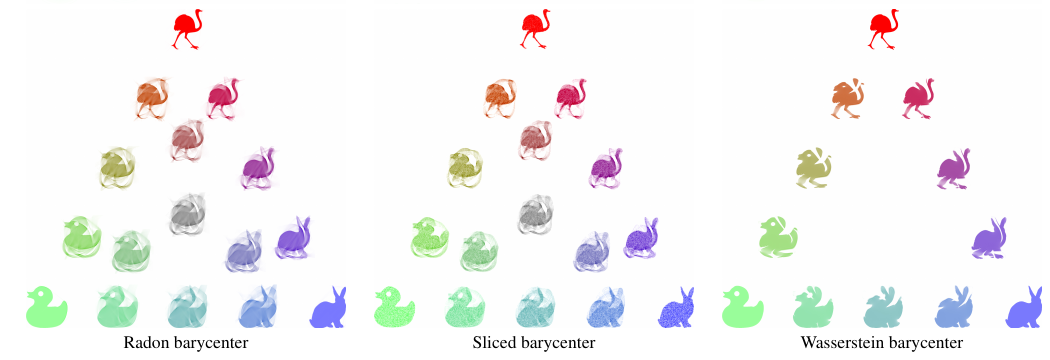}
    \caption{Radon barycenter, sliced Wasserstein barycenter, and Wasserstein barycenter (Figure 6 in~\citet{bonneel2015sliced}).}
    \label{fig:SWbarycneter}
\end{figure}

\begin{definition}[Sliced Wasserstein Geodesic]
    \label{def:sliced_Wasserstein_Geodesic} When $K=2$ and $t \in [0,1]$, the curve
    \begin{align}
        \mu_t \to  SWB_p(\mu_1,\mu_2,t,1-t),
    \end{align}
    can be seen as the geodesic curve between $\mu_1 \in \mathcal{P}_p(\mathbb{R}^d)$ and $\mu_2 \in \mathcal{P}_p(\mathbb{R}^d)$ in the sliced Wasserstein space. 
\end{definition}

In contrast to the Radon Wasserstein barycenter, SWB does not provide a closed-form solution for the barycenter (it still has the computational benefits of SW). Therefore, we need to solve the optimization problem. In practice, one way to solve it is to parameterize the barycenter. In particular, let $\mu_\phi$ be parameterized by $\phi \in \Phi$, and we solve the following problem:
\begin{align}
        \text{argmin}_{\phi \in \Phi} \sum_{k=1}^K w_k SW_p^p(\mu_k,\mu_\phi).
\end{align}
We can solve the optimization problem using stochastic gradient descent methods. For example, a simple Euler method is:
\begin{align}
    \phi^{(t+1)} = \phi^{(t)} - \eta \nabla_{\phi^{(t)}} \sum_{k=1}^K w_k \, \text{SW}_p^p(\mu_{\phi^{(t)}},\mu_k),
\end{align}
for $t=1,\ldots,T$ with $T\geq 1$ the number of steps and $\eta \geq 0$ the step size. Other methods like the Newton method can also be used~\citep{bonneel2015sliced}. We re-show the Radon barycenter, sliced Wasserstein barycenter, and Wasserstein barycenter of 3 two-dimensional shapes with various choices of weights in Figure~\ref{fig:SWbarycneter} (Figure 6 in~\citet{bonneel2015sliced}).

\begin{remark}[Parameterizations of Barycenter]\label{def:parameterization_barycenter}
We are interested in calculating the quantity given by the gradient
$$
\nabla_\phi \sum_{k=1}^K w_k \text{SW}_p^p(\mu_\phi,\mu_k) = \sum_{k=1}^K w_k \nabla_\phi \text{SW}_p^p(\mu_\phi,\mu_k).
$$ 
The gradient
$$
\nabla_\phi \text{SW}_p^p(\mu_\phi,\mu_k) = \mathbb{E}_{\theta \sim \mathcal{U}(\mathbb{S}^{d-1})} \left[ \nabla_\phi \text{W}_p^p (\theta \sharp \mu_\phi, \theta \sharp \mu_k) \right]
$$
for any $k=1,\ldots,K$ can be estimated by Monte Carlo methods. For example, with $\theta_1,\ldots,\theta_L$ i.i.d. samples from $\mathcal{U}(\mathbb{S}^{d-1})$, the stochastic gradient estimator is formed as follows:
\begin{align}
\label{eq:grad_SW}
    \nabla_\phi \text{SW}_p^p(\mu_\phi,\mu_k) \approx \frac{1}{L} \sum_{l=1}^L \nabla_\phi \text{W}_p^p (\theta_l \sharp \mu_\phi, \theta_l \sharp \mu_k).
\end{align}

As discussed in Section~\ref{sec:differentiatingSW:chapter:varitational_SW}, we consider some parameterizations of the barycenter. \textit{Free supports barycenter:} we have $\mu_{\phi}= \frac{1}{n} \sum_{i=1}^n \alpha_i \delta_{\phi_i}$. In this setting, the gradient can be obtained as in Remark~\ref{remark:derivative_atoms}. \textit{Fixed supports barycenter:} In this setting, we set $\mu_{\phi} = \sum_{i=1}^n \phi_i \delta_{x_i}$, and the corresponding gradient can be obtained as in Remark~\ref{remark:derivative_weight}.
\end{remark}

There is a variant of SWB that restricts the distances from the barycenter to all marginals to be approximately the same, which is named the marginal fairness sliced Wasserstein barycenter.

\begin{definition}[Marginal fairness sliced Wasserstein barycenter]
    \label{def:MFSWB}  Given $K\geq 2$ marginals $\mu_1,\ldots,\mu_K \in \mathcal{P}_p(\mathbb{R}^d)$, admissible $\epsilon\geq 0$, for $i=1,\ldots,K$ and $j=i+1,\ldots,K$, the marginal fairness sliced Wasserstein barycenter (MFSWB) is defined as~\citep{nguyen2025towards}:
    \begin{align}
       & MFSWB_p(\mu_1,\ldots,\mu_K) = \text{argmin}_{\mu \in \setP_p(\Re^d)} \frac{1}{K}\sum_{k=1}^K \text{SW}_p^p (\mu,\mu_k) \\
       & \quad \text{s.t. } \frac{2}{(K-1)K} \sum_{i=1}^{K-1} \sum_{j=i+1}^K \left| \text{SW}_p^p (\mu,\mu_i) - \text{SW}_p^p (\mu,\mu_j) \right| \leq \epsilon. \nonumber
    \end{align}
\end{definition}

In Definition~\ref{def:MFSWB}, we want $\epsilon \approx 0$, i.e., $\mu_1,\ldots,\mu_K$ lie on the $SW_p$-sphere with center $\mu$. However, there might not exist a solution $\mu$ for very small $\epsilon$. For admissible $\epsilon>0$, we can use Lagrange multipliers to solve the problem. However, the duality is only weak duality and might lead to hyperparameter tuning in practice (for the Lagrange multiplier). Authors in~\citet{nguyen2025towards} provide some surrogate problems that are hyperparameter-free.

\begin{definition}[Surrogate marginal Fairness Sliced Wasserstein Barycenter]
    \label{def:sMFSWB}
    Given $K\geq 2$ marginals $\mu_1,\ldots,\mu_K \in \mathcal{P}_p(\mathbb{R}^d)$, the surrogate marginal fairness sliced Wasserstein barycenter (s-MFSWB) problem is defined as~\citep{nguyen2025towards}:
    \begin{align}
    &\text{s-}MFSWB_p(\mu_1,\ldots,\mu_K) \nonumber\\
    &= \text{argmin}_{\mu \in \setP_p(\Re^d)} \max_{k \in \{1,\ldots,K\}} SW_p^p(\mu,\mu_k).
    \end{align}
\end{definition}

\begin{definition}[Unbiased surrogate marginal Fairness Sliced Wasserstein Barycenter]
    \label{def:usMFSWB}
    Given $K\geq 2$ marginals $\mu_1,\ldots,\mu_K \in \mathcal{P}_p(\mathbb{R}^d)$, the unbiased surrogate marginal fairness sliced Wasserstein barycenter (us-MFSWB) problem is defined as~\citep{nguyen2025towards}:
    \begin{align}
    &\text{us-}MFSWB_p(\mu_1,\ldots,\mu_K) \nonumber\\
    &= \text{argmin}_{\mu \in \setP_p(\Re^d)} \mathbb{E}_{\theta \sim \mathcal{U}(\mathbb{S}^{d-1})} \left[ \max_{k \in \{1,\ldots,K\}} W_p^p(\theta \sharp \mu, \theta \sharp \mu_k) \right].
    \end{align}
\end{definition}

By Jensen's inequality, we have
\begin{align*}
\max_{k \in \{1,\ldots,K\}} SW_p^p(\mu,\mu_k) \leq \mathbb{E}_{\theta \sim \mathcal{U}(\mathbb{S}^{d-1})} \left[ \max_{k \in \{1,\ldots,K\}} W_p^p(\theta \sharp \mu, \theta \sharp \mu_k) \right].
\end{align*}
The nice feature of the second surrogate is that the gradient with respect to $\mu$ can be estimated unbiasedly since the expectation is outside the maximization. Authors in~\citet{nguyen2025towards} also propose to use the energy-based slicing measure (similar to Definition~\ref{remark:energy_based_slicing_measure}) with the density
$$
f_\sigma(\theta;\mu,\mu_1,\ldots,\mu_K) \propto \exp \left( \max_{k \in \{1,\ldots,K\}} W_p^p(\theta \sharp \mu, \theta \sharp \mu_k) \right),
$$
to obtain a third variant. All discussed problems can be solved using gradient-based optimization with parametric barycenters.

\section{Sliced Wasserstein Gradient Flows}
\label{sec:SWgradientflow:chapter:varitational_SW}
Minimizing functionals over probability measures is a central problem in modern machine learning, particularly in generative modeling and variational inference. While Wasserstein gradient flows offer a principled framework for such optimization, their numerical implementation is often limited by the high computational cost of evaluating and differentiating the Wasserstein distance. Sliced Wasserstein (SW) gradient flows offer a scalable alternative by replacing the full Wasserstein distance with its sliced counterpart, an approximation based on averaging one-dimensional projections. Thanks to its closed-form expression and differentiability, the SW distance allows for efficient implementation of gradient flow dynamics in high dimensions. In this section, we study the behavior of flows induced by the SW distance. We first start by reviewing gradient flows in Euclidean spaces and Wasserstein gradient flows in probability spaces.

\begin{definition}[Gradient flows in Euclidean spaces]
    \label{definition:gradient_flow_eculidean} Given a functional $F: \Re^d\to \Re$ ($d\geq 1$), a gradient flow of $F$ is a curve that decreases $F$ as much as possible along it. For differentiable $F$, a gradient flow $x:\Re_+ \to \Re^d$ solves the following Cauchy problem~\citep{santambrogio2017euclidean}:
    \begin{align}
        \begin{cases}
            &\frac{\diff x(t)}{\diff t} = -\nabla F(x(t)), \\
            &x(0)=x_0.
        \end{cases}
    \end{align}
    The gradient flow can be unique if $F$ is convex and $\nabla F$ is Lipschitz continuous. Numerical schemes such as the Euler method are often used to approximate the solution. For the explicit Euler method, it leads to the standard gradient descent algorithm. For the implicit Euler method, it leads to the proximal point algorithm~\citep{parikh2014proximal}:
    \begin{align}
        x^{\tau}_{k+1} \in \text{argmin}_{x} \frac{\|x-x^\tau_{k}\|_2^2}{2\tau} +F(x) = \text{Prox}_{\tau,F}(x_k^\tau), 
    \end{align}
    where $\tau>0$ and $k$ denotes the discretization step.
\end{definition}

\begin{definition}[Wasserstein gradient flows in probability spaces]
    \label{definition:gradient_flow_eculidean_probablity}
    Given a functional $F: \setP_2(\Re^d)\to \Re$ ($d\geq 1$), the Wasserstein gradient flow $\{\mu_t\}_{t \in \Re_+}$ solves the following continuity equation~\citep{ambrosio2005gradient}:
    \begin{align}
            \frac{\partial \mu_t}{\partial t} =\text{div}(\mu_t \nabla_x F'(\mu_t)),
    \end{align}
    with known $\mu_0$. Here, $\text{div}$ is the divergence operator and $F'$ is the first variation of $F$~\citep{ambrosio2005gradient}. There is a connection between Wasserstein gradient flows and stochastic differential equations (SDEs). Consider the following Fokker–Planck equation with $\Phi:\Re^d \to \Re$ and a fixed diffusion coefficient:
    \begin{align}
        \frac{\partial \mu_t}{\partial t} =\text{div}(\nabla \Phi(x) \mu_t) +\beta^{-1}\Delta\mu_t,
    \end{align}
with known $\mu_0$ and $\beta>0$, which is equivalent to the following Itô SDE:
\begin{align}
    \diff X_t  = -\nabla \Phi(X_t) \diff t  + \sqrt{2\beta^{-1}} \diff W_t, \quad X_t \sim \mu_t,
\end{align}
where $\Phi:\Re^d \to \Re$ is the potential function, $W_t$ is the standard Wiener process, and $\beta > 0$ is the magnitude. The above two equations are equivalent to the Wasserstein flow with the Fokker–Planck free energy functional~\citep{jordan1998variational}:
\begin{align}
    F (\mu) = U(\mu) -\beta^{-1} E(\mu),
\end{align}
where $U(\mu) = \int_{\Re^d} \Phi(x)\diff \mu(x)$ is the potential energy, and $E(\mu)  = - \int_{\Re^d} \log \frac{\diff \mu}{\diff x }(x)\diff \mu(x)$ is the
entropy. Starting from any $\mu_0$, the advection–diffusion process of the Fokker–Planck equation converges to the
unique stationary solution with density proportional to $\exp(-\beta \Phi(x))$ \\\citep{risken1989fokker}.
\end{definition}

Wasserstein gradient flow is carried out in practice using the Jordan, Kinderlehrer, and Otto (JKO) scheme, which is a discretization scheme.

\begin{definition}[Jordan, Kinderlehrer, and Otto (JKO) Scheme]
    \label{definition:JKO_scheme} Jordan, Kinderlehrer, and
Otto propose the JKO integration to approximate the dynamics of the Wasserstein gradient flow~\citep{jordan1998variational}:
\begin{align}
    \mu_{k+1}^\tau  \in \text{argmin}_{\mu \in \setP_2(\Re^d)} \frac{W_2^2(\mu,\mu_{k}^\tau)}{2\tau } +F(\mu),
\end{align}
with $\tau>0$, and $k$ denoting the discretization step.
\end{definition}

We now discuss sliced Wasserstein gradient flows, which define steepest descent paths in the space of probability measures endowed with the Sliced Wasserstein (SW) distance. In contrast to classical Wasserstein gradient flows, which can often be characterized by partial differential equations (PDEs), SW gradient flows are defined implicitly via the Jordan–Kinderlehrer–Otto (JKO) scheme.

\begin{figure}[!t]
    \centering

    \begin{tabular}{c}
         \includegraphics[width=1\linewidth]{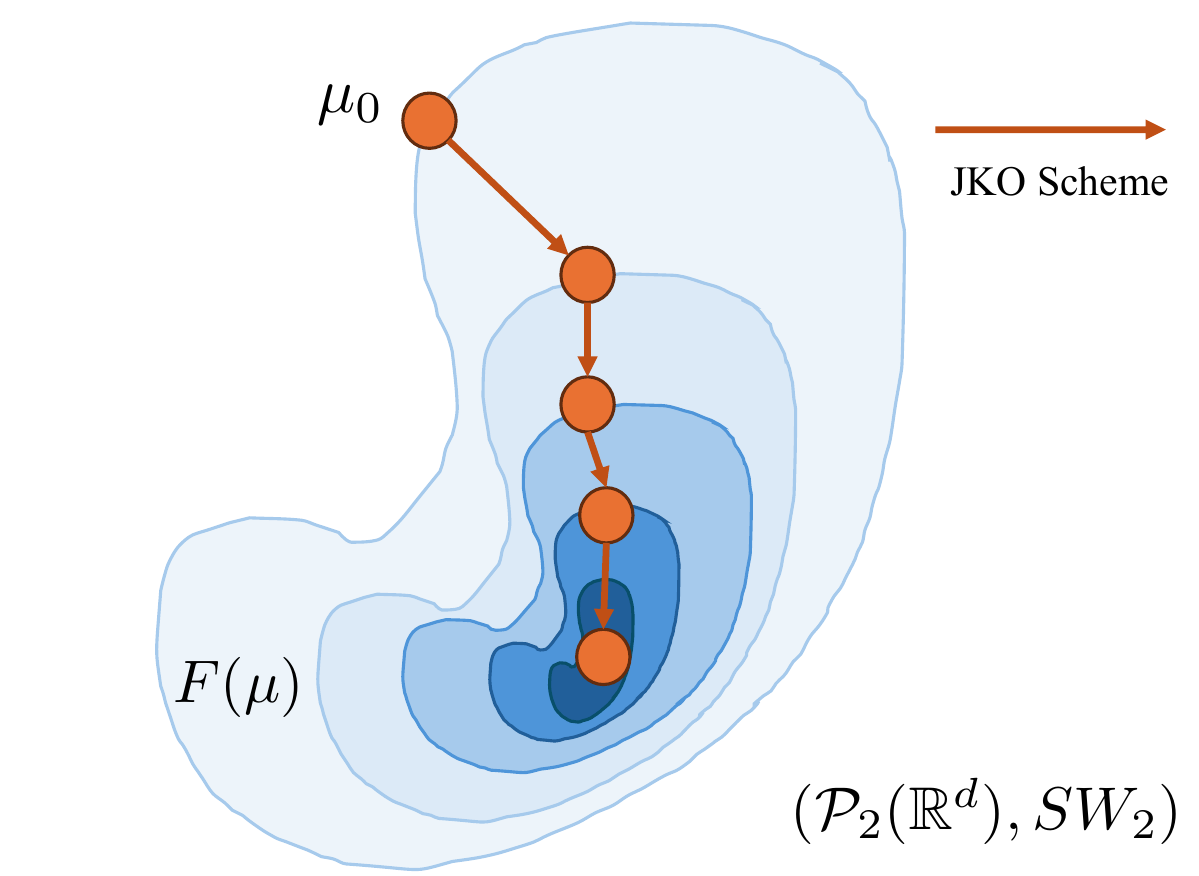}
    \end{tabular}
    \caption{Sliced Wasserstein gradient flows.} 
    \label{fig:SWGF}
\end{figure}

\begin{definition}[Sliced Wasserstein Gradient Flows]
    \label{definition:SW_gradient_flow} Sliced Wasserstein gradient flow is defined implicitly through its JKO scheme as follows~\citep{bonet2022efficient}:
    \begin{align}
    \mu_{k+1}^\tau  \in \text{argmin}_{\mu \in \setP_2(\Re^d)} \frac{SW_2^2(\mu,\mu_{k}^\tau)}{2\tau } +F(\mu),
\end{align}
with $\tau>0$, $k$ denoting the discretization step, and known $\mu_0$. From \\\citet[Propositions 1 and 2]{bonet2022efficient}, if $F$ is a lower semi-continuous functional, then the scheme admits a
minimizer. The minimizer is unique if $\mu_k^\tau$ is absolutely continuous and $F$ is convex, or if $F$ is strictly convex. In addition, the functional $F$ is non-increasing along the sequence of minimizers $(\mu_k^\tau)_k$. We visualize the sliced Wasserstein gradient flow in Figure~\ref{fig:SWGF}.
\end{definition}

\begin{remark}[Solving Sliced Wasserstein Gradient Flows]
    \label{remark:solving_SWGF}
    There are three practical ways to approximate the SW gradient flows~\citep{bonet2022efficient}:
    \begin{enumerate}
        \item \textbf{Discretized Grid.} The first approach is to model the distribution on a regular fixed grid, i.e., $\mu_k^\tau= \sum_{i=1}^n \phi_i^{(k)}\delta_{x_i}$ ($\sum_{i=1}^n \phi_i^{(k)}=1\, \forall k$) and $\mu = \sum_{i=1}^n \phi_i \delta_{x_i}$~\citep{peyre2015entropic}. The potential energy and the entropy for the Fokker–Planck equation can be expressed as follows:
        \begin{align}
           &U(\mu) = \int_{\Re^d} \Phi(x)\diff \mu(x) \approx \sum_{i=1}^n \Phi(x_i)\phi_i, \\
           &E(\mu)  = - \int_{\Re^d} \log \frac{\diff \mu}{\diff x}(x)\diff \mu(x) \approx \sum_{i=1}^n \log\left( \frac{\phi_i}{l}\right) \phi_i,
        \end{align}
        where the Lebesgue measure is approximated as $l \sum_{i=1}^n \delta_{x_i}$, where $l$ represents a volume. Hence, the Lebesgue density can be approximated by $\frac{\phi_i}{l}$ for $i=1,\ldots,n$~\citep{carlier2017convergence,peyre2015entropic}. Here, discrete SW can be computed easily.

        \item \textbf{With Particles.} The second approach is to model the distribution as a set of particles with uniform weights, i.e., $\mu_k^\tau= \frac{1}{n}\sum_{i=1}^n \delta_{\phi_i^{(k)}}$ and $\mu = \frac{1}{n}\sum_{i=1}^n  \delta_{x_i}$. In this case, we need to use nonparametric estimators~\citep{beirlant1997nonparametric} for approximating the entropy. Here, discrete SW can be computed easily.

        \item \textbf{Generative Models with Tractable Density.} The third approach is to model the distribution as a generative model, i.e., $\mu_k^\tau = f_{\gamma^{(k)}}\sharp \nu_0$ and $\mu = f_\gamma \sharp \nu_0$ with $\nu_0$ being a fixed distribution (e.g., Gaussian) and $f_\gamma: \Re^d \to \Re^d$ an invertible function (e.g., 
normalizing flows~\citep{papamakarios2021normalizing}). The potential energy and the entropy for the Fokker–Planck equation can be expressed as follows:
        \begin{align}
           &U(\mu) = \int_{\Re^d} \Phi(x)\diff \mu(x) \approx \frac{1}{n}\sum_{i=1}^n \Phi(x_i), \\
           &E(\mu)  = - \int_{\Re^d} \log \frac{\diff \mu}{\diff x}(x)\diff \mu(x)  \nonumber\\& \quad \quad \approx \frac{1}{n} \sum_{i=1}^n \left(\log \frac{\diff \nu_0}{\diff z} (z_i) -\log | \det (J_{f_\gamma}(z_i))|\right),
        \end{align}
        where $z_1,\ldots,z_n \stackrel{\mathrm{i.i.d.}}{\sim} \nu_0$. Here, the change-of-variables formula is used for the entropy approximation. In addition, we also need to approximate SW between two continuous distributions using the plug-in estimator.
    \end{enumerate}
\end{remark}
We refer the reader to~\citep{park2025geometry} for a more detailed discussion of the geometry of the Sliced Wasserstein space. In practice, the SW distance can also be used as an energy functional in Wasserstein gradient flows for applications involving the comparison of probability measures.

\begin{remark}[Wasserstein Gradient Flows with Sliced Wasserstein Functional]
    \label{remark:WGF_SWfunctional}
    As discussed in Section~\ref{sec:IDT_Konthe:chapter:foundations}, Euclidean gradient flow with SW functional, i.e., $F(\mu) = SW_2^2(\mu,\nu)$ for a given $\nu$, can be used to derive a transportation method between two measures. Authors in~\citet{liutkus2019sliced} generalize it into the Wasserstein gradient flow. However, the flow does not give an optimal map and might not converge to the target measure even for the case of $n=m$~\citep{cozzi2025long}. We refer the reader to~\citet{vauthiertowards,cozzi2025long} for a more detailed discussion.
\end{remark}

\section{Sliced Wasserstein Kernels}
\label{sec:SWKernel:chapter:varitational_SW}

While the Wasserstein distance provides a powerful geometric measure between probability distributions, it does not readily induce positive definite kernels, which are fundamental for kernel-based learning methods such as support vector machines and Gaussian processes. This limitation arises because the Wasserstein distance is \emph{not} Hilbertian, making it difficult to construct valid kernels directly from it. In contrast, SW kernels overcome this obstacle by leveraging the one-dimensional projections underlying the SW distance. By averaging kernels computed on these projected distributions, SW kernels inherit positive definiteness and computational efficiency. This makes SW kernels a highly attractive tool for embedding and comparing probability measures within kernel-based frameworks, enabling scalable applications in classification, regression, and clustering that fully exploit the geometric structure of distributions. In this section, we review SW kernels and related concepts. We first start by reviewing positive definite kernels, conditional negative definite kernels, and their connection.

\begin{definition}[Positive definite kernel]
    \label{def:positive_Definite_kernel}
    A positive definite kernel on a set $M$ is a symmetric function $k:M \times M \to \Re$ such that:
    \begin{align}
        \sum_{i=1}^n \sum_{j=1}^n c_ic_jk(x_i,x_j) \geq 0,
    \end{align}
    where $x_1,\ldots,x_n \in M$ and $c_1,\ldots,c_n \in \Re$.
\end{definition}

\begin{definition}[Conditional negative definite kernel]
    \label{def:conditional_negative_Definite_kernel}
    A conditional negative definite kernel on a set $M$ is a symmetric function $k:M \times M \to \Re$ such that:
    \begin{align}
        \sum_{i=1}^n \sum_{j=1}^n c_ic_jk(x_i,x_j) \leq 0,
    \end{align}
    where $x_1,\ldots,x_n \in M$, $c_1,\ldots,c_n \in \Re$, and $\sum_{i=1}^n c_i = 0$.
\end{definition}

\begin{proposition}[Connection between positive definite kernels and conditional negative definite kernels]
    \label{proposition:connection_PD_CND_kernel}
    From~\citet[Theorem 2.2]{cowling1983harmonic}, a kernel $k(x,y) = \exp(-\gamma g(x,y))$ is positive definite for all $\gamma > 0$ if and only if $g(\mu,\nu)$ is conditionally negative definite.
\end{proposition}

There is an interesting connection between positive definite kernels and distance functions. A distance can be used to define a positive definite kernel if it satisfies a certain property, which can be described as follows:

\begin{proposition}[Positive definite kernel and distance function]
    \label{proposition:PD_kernel_and_distance}
    From~\citet[Theorem 6.1]{jayasumana2015kernel}, if $d$ is a metric on $M$, a kernel $k(x,y) = \exp(-\gamma d^2(x,y))$ is positive definite for all $\gamma > 0$ if and only if there exists an inner product space $V$ and a function $f:M \to V$ such that $d(x,y) = \|f(x)-f(y)\|_V$.
\end{proposition}

We now review an important concept: ``Hilbertian". A space is called Hilbertian if it behaves like a Hilbert space, meaning it has an inner product that defines its geometry.

\begin{definition}[Hilbertian]
    Let $D$ be a distance between probability measures. $D$ is called Hilbertian if there exists a Hilbert space $\mathcal{H}$ and a map $f : \setP(\setX) \to \mathcal{H}$ such that $D(\mu,\nu) = \|f(\mu) - f(\nu)\|_\mathcal{H}$.
\end{definition}
While the Wasserstein distance is not Hilbertian in multivariate settings, it is Hilbertian in one-dimensional cases.

\begin{proposition}[One-dimensional Wasserstein is Hilbertian]
    \label{proposition:1DW_Hilbertian}
     From Remark~\ref{remark:1DWasserstein_general}, with $\mu_0$ as a reference probability measure, we have:
\begin{align}
    W_2(\mu,\nu) & = \left(\int_0^1 |F^{-1}_\mu(t)-F_{\nu}^{-1}(t)|^2 \diff t \right)^{\frac{1}{2}} \nonumber \\
    &= \|f(\mu)-f(\nu)\|_{L^2(\Re)},
\end{align} 
where an example of the function is $f(\mu)(t)= F^{-1}_\mu(t) \in L^2([0,1])$, with $ L^2([0,1])$ being the Hilbert space of all real-valued functions whose squared absolute value is integrable over $[0,1]$. 
\end{proposition}

Being Hilbertian, from Proposition~\ref{proposition:PD_kernel_and_distance}, we can derive a kernel for one-dimensional probability measures using the Wasserstein distance.

\begin{definition}[One-dimensional Wasserstein kernel for probability measures]
    \label{def:Wasserstein_kernel_measures}
    Let $M \in \setP_2(\Re)$ be the set of one-dimensional probability measures, $\gamma > 0$, the Wasserstein kernel between $\mu \in M$ and $\nu \in M$ is defined as follows:
    \begin{align}
        k_W(\mu,\nu) = \exp\left(-\gamma W_2^2(\mu,\nu)\right).
    \end{align}
    $k_W(\mu,\nu)$ is a positive definite kernel~\citep[Theorem 4]{kolouri2016sliced}. In particular, given a set $\mu_1,\ldots,\mu_n \in M$, we have:
    \begin{align}
        \sum_{i=1}^n \sum_{j=1}^n c_ic_jk_W(\mu_i,\mu_j) \leq 0,
    \end{align}
    where $\sum_{i=1}^n c_i = 0$. The proof follows from the fact that $W_2^2$ is Hilbertian, which satisfies~\citet[Theorem 6.1]{jayasumana2015kernel} (Proposition~\ref{proposition:PD_kernel_and_distance}). In~\citet{kolouri2016sliced}, the authors restrict the kernel to measures with densities; however, the SW kernel is still valid for discrete measures~\citep{carriere2017sliced}.
\end{definition}

Now, we discuss how to obtain a kernel for probability measures in higher dimensions using the SW distance. We first review the definition of the SW kernel, then show that it is a positive definite kernel.

\begin{definition}[Sliced Wasserstein kernel for probability measures]
    \label{remark:SW_kernel_measures}
    Let $M \in \setP_2(\Re^d)$ and $\gamma > 0$, the sliced Wasserstein kernel~\citep{kolouri2016sliced,carriere2017sliced} between $\mu \in M$ and $\nu \in M$ is defined as follows:
    \begin{align}
        k_{SW}(\mu,\nu) = \exp\left(-\gamma SW_2^2(\mu,\nu)\right).
    \end{align}
\end{definition}

\begin{proposition}[Sliced Wasserstein kernel is positive definite]
    \label{proposition:SWK_positive_definite} The sliced Wasserstein kernel is a positive definite kernel.
    \begin{proof}
        We follow the proof in~\citet[Theorem 5]{kolouri2016sliced}. We show that $SW_2^2$ is conditionally negative definite. Given $\mu_1,\ldots,\mu_n\in \setP_2(\Re^d)$, we have:
        \begin{align*}
    \sum_{i=1}^n\sum_{j=1}^n c_ic_jSW_2^2(\mu_i,\mu_j) &=\sum_{i=1}^n\sum_{j=1}^n c_ic_j \mathbb{E}_{\theta \sim \setU(\Sm^{d-1})}[W_2^2(\theta \sharp \mu_i,\theta \sharp \mu_j)] \\
    &=\mathbb{E}_{\theta \sim \setU(\Sm^{d-1})}\left[\sum_{i=1}^n\sum_{j=1}^n c_ic_j W_2^2(\theta \sharp \mu_i,\theta \sharp \mu_j)\right],
        \end{align*}
        due to Fubini's theorem. Moreover, we know that $$\sum_{i=1}^n\sum_{j=1}^n c_ic_j W_2^2(\theta \sharp \mu_i,\theta \sharp \mu_j) \leq 0$$ due to the positive definiteness of the Wasserstein kernel. Therefore, we obtain the desired inequality:
        \begin{align}
            \sum_{i=1}^n\sum_{j=1}^n c_ic_jSW_2^2(\mu_i,\mu_j) \leq 0,
        \end{align}
        which completes the proof.
    \end{proof}
\end{proposition}

Since the SW kernel is positive definite, we can conjecture that the SW distance is Hilbertian. 

\begin{proposition}[Sliced Wasserstein is Hilbertian]
    \label{proposition:SW_Hilbertian} For $\mu,\nu \in \setP_2(\Re^d)$, we have:
\begin{align}
    SW_2(\mu,\nu) &= \left(\mathbb{E}_{\theta \sim \setU(\Sm^{d-1})} \left[\int_0^1 |F^{-1}_{\theta \sharp \mu}(t)-F_{\theta \sharp \nu}^{-1}(t)|^2 \diff t\right] \right)^{\frac{1}{2}} \nonumber\\
    &= \|f ( \mu)-f (\nu)\|_{L^2([0,1]\times \Sm^{d-1})},
\end{align} 
where $f(\mu)(t,\theta) =F_{\theta\sharp \mu}^{-1}(t ) \in L^2([0,1] \times \Sm^{d-1})$. Other constructions of the function $f$ exist, such as using a reference measure, as in~\citet{kolouri2016sliced}.
\end{proposition}
It is worth noting that we can use $\sqrt{SW_1}$ as a replacement for $SW_2^2$ to define the SW kernel since $\sqrt{SW_1}$ is also Hilbertian~\citep{carriere2017sliced,meunier2022distribution}. The SW kernel is often approximated by Monte Carlo samples. However, this leads to a biased estimate since the kernel is the exponential function of an expectation. In particular, we have:
\begin{align}
    \hat{k}_{SW}(\mu,\nu;\theta_1,\ldots,\theta_L) = \exp\left(-\gamma \widehat{SW}_2^2(\mu,\nu;\theta_1,\ldots,\theta_L)\right), 
\end{align}
where $\theta_1,\ldots,\theta_L \simiid \setU(\Sm^{d-1})$ and 
$$\widehat{SW}_2^2(\mu,\nu;\theta_1,\ldots,\theta_L) = \frac{1}{L}\sum_{l=1}^L W_2^2(\theta_l \sharp \mu,\theta_l \sharp \nu).$$ 
We can check that
$\mathbb{E}[\hat{k}_{SW}(\mu,\nu)] \neq k_{SW}(\mu,\nu).$ The reason for the bias is that the expectation (SW) is inside the exponential function.
By moving the expectation outside of the exponential function, we can obtain a new type of kernel that can be estimated without bias.

\begin{definition}[Unbiased sliced Wasserstein kernel for probability measures]
    \label{remark:USW_kernel_measures}
    Let $M \in \setP_2(\Re^d)$ be the set of absolutely continuous probability measures on $\Re^d$, $\gamma > 0$, the unbiased sliced Wasserstein kernel~\citep{luong2025unbiased} between $\mu \in M$ and $\nu \in M$ is defined as follows:
    \begin{align}
        k_{USW}(\mu,\nu) = \mathbb{E}_{\theta \sim \setU(\Sm^{d-1})} \left[\exp\left(-\gamma W_2^2(\theta \sharp \mu,\theta \sharp \nu)\right) \right],
    \end{align}
\end{definition}

By Jensen's inequality, we have that $k_{USW}(\mu,\nu) \geq k_{SW}(\mu,\nu)$ for all $\mu,\nu \in M \subset \setP_2(\Re^d)$.

\begin{proposition}[Unbiased sliced Wasserstein kernel is positive definite]
    \label{proposition:USWK_positive_definite} The unbiased sliced Wasserstein kernel is a positive definite kernel.
    \begin{proof}
        We follow the proof in~\citet{luong2025unbiased}.  We have:
        \begin{align*}
    \sum_{i=1}^n\sum_{j=1}^n c_ic_j k_{USW}(\mu_i,\mu_j) &=\sum_{i=1}^n\sum_{j=1}^n c_ic_j \mathbb{E}_{\theta \sim \setU(\Sm^{d-1})}[\exp\left(-\gamma W_2^2(\theta \sharp \mu,\theta \sharp \nu)\right)] \\
    &=\mathbb{E}_{\theta \sim \setU(\Sm^{d-1})}\left[\sum_{i=1}^n\sum_{j=1}^n c_ic_j\exp\left(-\gamma W_2^2(\theta \sharp \mu,\theta \sharp \nu)\right)\right],
        \end{align*}
        due to Fubini's theorem. Moreover, we know that $$\sum_{i=1}^n\sum_{j=1}^n c_ic_j \exp\left(-\gamma W_2^2(\theta \sharp \mu,\theta \sharp \nu)\right) \geq 0$$ due to the positive definiteness of the Wasserstein kernel. Therefore, we obtain the desired inequality:
        \begin{align}
           \sum_{i=1}^n\sum_{j=1}^n c_ic_j k_{USW}(\mu_i,\mu_j) \geq 0,
        \end{align}
        which completes the proof.
    \end{proof}
\end{proposition}

For Monte Carlo approximation, we have
\begin{align}
    \hat{k}_{USW}(\mu,\nu;\theta_1,\ldots,\theta_L) = \frac{1}{L}\sum_{l=1}^L\exp\left(-\gamma W_2^2(\theta_l\sharp \mu,\theta_l\sharp \nu)\right), 
\end{align}
where $\theta_1,\ldots,\theta_L \simiid \setU(\Sm^{d-1})$. We can check that: 
\begin{align*}
    \mathbb{E}[\hat{k}_{USW}(\mu,\nu)] = k_{USW}(\mu,\nu),
\end{align*}
which is an unbiased estimate.


It is worth noting that other SW variants using non-linear projections (Section~\ref{sec:generalized_slicing:chapter:advances}) can also be used to define kernels. For example, authors in~\citet{bonet2023sliced} use SW between measures over a symmetric positive definite matrix to define a kernel. In addition, the weighted Radon transform (Section~\ref{sec:slicing_measure:chapter:advances}) can also be adapted to obtain variants of the SW kernel.

\section{Sliced Wasserstein Embeddings}
\label{sec:SWEmbedding:chapter:varitational_SW}

Beyond the SW kernel, the Hilbertian property of the SW distance (Proposition~\ref{proposition:SW_Hilbertian}) can be used to derive a representation extraction framework for probability measures, which is named Sliced Wasserstein Embeddings (SWE). The idea of Sliced Wasserstein Embeddings (SWE) is rooted in Linear Optimal Transport~\citep{wang2013linear,moosmuller2023linear}. SWE was first introduced for pattern recognition from
2D probability density functions (e.g., images)~\citep{kolouri2015radon} in the form of the Radon cumulative distribution transform~\citep{park2018cumulative}. Recently, authors in~\citet{naderializadeh2021pooling} defined it for high-dimensional measures. We first review the definition of SWE.

\begin{figure}[!t]
    \centering
    \includegraphics[width=1\linewidth]{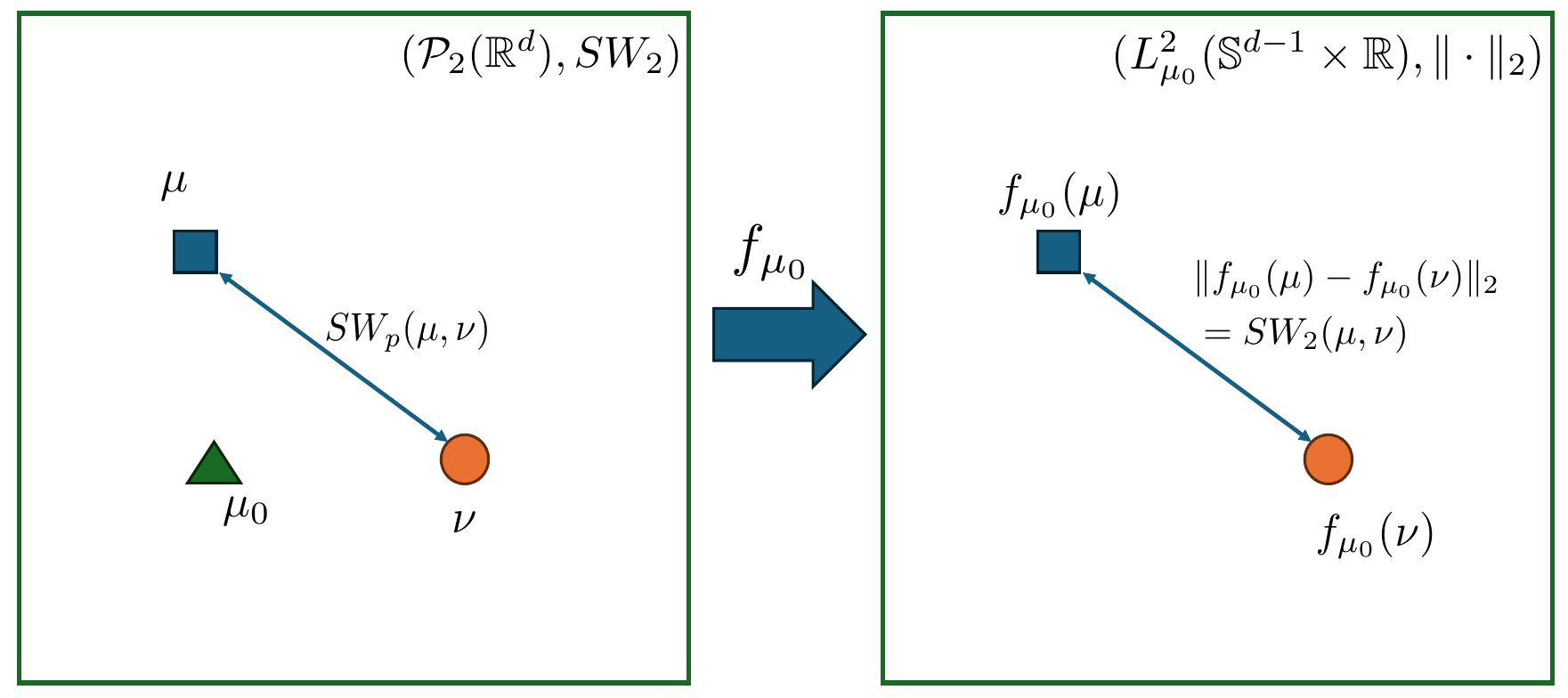}
    \caption{Sliced Wasserstein embedding of a measure is a function.}
    \label{fig:SWE}
\end{figure}

\begin{definition}[Sliced Wasserstein Embedding]
    \label{def:SWE} Let $P_\theta:\Re^d \to \Re$ be a projection function and let $\mu \in \setP_2(\Re^d)$ be a probability measure. A slice of $\mu$ with $P_\theta$ ($\theta \in \Theta$) is defined as $P_\theta\sharp \mu$. Given a reference measure $\mu_0 \in \setP_2(\Re^d)$, the optimal transport map between $P_\theta \sharp \mu$ and $P_\theta \sharp \mu_0$ is $T_{\mu,\mu_0}^\theta= F_{P_\theta \sharp \mu}^{-1}\circ F_{P_\theta \sharp \mu_0}$. The sliced Wasserstein embedding of $\mu$~\citep{naderializadeh2021pooling} is defined as:
    \begin{align}
        f_{\mu_0}(\mu)(x,\theta) = T_{\mu,\mu_0 }^\theta(x) - x.
    \end{align}
    We can check that:
    \begin{align}
       & \|f_{\mu_0}(\mu)-f_{\mu_0}(\nu)\|_{L^2_{\mu_0}(\Re\times \Theta)}^2 \nonumber\\
       &= \int_{\Theta} \int_\Re |f_{\mu_0}(\mu)(x,\theta)-f_{\mu_0}(\nu)(x,\theta)|^2 \diff \mu_0(x)\diff\theta  \nonumber\\
        &= \int_{\Theta} \int_\Re |T_{\mu,\mu_0}^\theta(x)-T_{\nu,\mu_0}^\theta(x)|^2 \diff \mu_0(x)\diff\theta  \nonumber\\
        &= \int_{\Theta} \int_\Re | F_{P_\theta \sharp \mu}^{-1}\circ F_{P_\theta \sharp \mu_0} (x)- F_{P_\theta \sharp \nu}^{-1}\circ F_{P_\theta \sharp \mu_0} (x)|^2 \diff \mu_0(x)\diff\theta  \nonumber \\
        &= \int_{\Theta} \int_0^1 | F_{P_\theta \sharp \mu}^{-1} (t)- F_{P_\theta \sharp \nu}^{-1} (t)|^2 \diff t\diff\theta  \nonumber \\
        &= \mathbb{E}_{\theta \sim \setU(\Theta)}[W_2^2(P_\theta \sharp \mu,P_\theta \sharp \nu)],
    \end{align}
    where we use the normalized integral for $\Theta$. When $P_\theta(x)=\langle \theta,x\rangle$ and $\theta \in \Sm^{d-1}$, we have:
    \begin{align}
        \|f_{\mu_0}(\mu)-f_{\mu_0}(\nu)\|_{L^2_{\mu_0}(\Re\times \Sm^{d-1})}=SW_2(\mu,\nu).
    \end{align}
    The mapping $f_{\mu_0}$ embeds a probability measure into a function. 
\end{definition}

We illustrate the idea of SWE in Figure~\ref{fig:SWE}. In practice, we might not want to use an ``infinite" representation of the embedding function. Therefore, the discretization of such an embedding should be obtained for applications. As in SW, we can use Monte Carlo methods for that purpose.

\begin{remark}[Numerical Approximation of Sliced Wasserstein Embedding]
    \label{remark:numerical_approximation_SWE}
    In practice, it is desirable to have a vector-based embedding. We can then define the following embedding:
    \begin{align}
        &\hat{f}_{\mu_0}(\mu)(x;\theta_1,\ldots,\theta_L) = (f_{\mu_0}(\mu)(x,\theta_1),\ldots,f_{\mu_0}(\mu)(x,\theta_L)), \\
        &f_{\mu_0}(\mu)(x,\theta_l) \nonumber \\
        &= ( T_{\mu,\mu_0 }^{\theta_l}(P_{\theta_l}(y_1)) - P_{\theta_l}(y_1),\ldots,T_{\mu,\mu_0 }^{\theta_l}(P_{\theta_l}(y_n)) - P_{\theta_l}(y_n))\, \forall l,
    \end{align}
    where  $\hat{f}_{\mu_0}(\mu)(x;\theta_1,\ldots,\theta_L)  \in \Re^{Ln}$, $\theta_1,\ldots,\theta_L \sim \setU(\Theta)$, and $y_1,\ldots,y_n \sim \mu_0$ (can be non-i.i.d.\ sampling). For $T_{\mu,\mu_0 }^{\theta_l}$, quantile function and CDF approximations are used (see Section~\ref{sec:quantile:chapter:advances}). It is worth noting that other approximation methods like Quasi-Monte Carlo (Section~\ref{sec:MC:chapter:advances}) can also be used for $\Theta$. In practice, the reference measure $\mu_0$ can be estimated using some task-specific objective~\citep{naderializadeh2021pooling}.
\end{remark}

Definition~\ref{def:SWE} defines SWE using a reference measure. As discussed in Proposition~\ref{proposition:SW_Hilbertian}, another possible choice of embedding function for a probability measure $\mu$ is $F^{-1}_{P_\theta \sharp \mu}$. However, this embedding function may produce discrete-valued embeddings for discrete distributions. Therefore, additional smoothing is required, e.g., via the cosine transform as follows:

\begin{definition}[Fourier Sliced Wasserstein Embedding]
    \label{def:FSWE} 
    Authors in~\citet{amir2025fourier} propose an alternative embedding:
    \begin{align}
        f(\mu)(\xi,\theta) = 2(1+\epsilon) \int_0^1 F_{P_\theta \sharp \mu}^{-1}(t)\cos (2\pi \xi t) \diff t,
    \end{align}
    where $\int_0^1F_{P_\theta \sharp \mu}^{-1}(t)\cos (2\pi \xi t) \diff t$ is the cosine transform (a variant of the Fourier
transform)~\citep{jones2001lebesgue} of the quantile function. 
\end{definition}

From~\citet[Theorem 3.2]{amir2025fourier}, $f(\mu)(\xi,\theta)$ yields an SW embedding, i.e., $\|f(\mu)-f(\nu)\|=SW_2(\mu,\nu)$. Again, we need to discretize the embedding in practice:

\begin{remark}[Numerical Approximation of Fourier Sliced Wasserstein Embedding]
    \label{remark:numerical_approximation_FSWE}
    To obtain a vector-based embedding for Fourier SWE, we can define the following embedding:
    \begin{align}
        &\hat{f}(\mu)(\xi;\theta_1,\ldots,\theta_L) = (f(\mu)(\xi,\theta_1),\ldots,f(\mu)(\xi,\theta_L)), \\
        &f(\mu)(\xi,\theta_l) = \left( 2(1+\epsilon) \int_0^1 F_{P_{\theta_l} \sharp \mu}^{-1}(t)\cos (2\pi \xi_1 t) \diff t, \right. \nonumber \\ 
        &\quad\quad\quad\quad\quad \left. \ldots,\, 2(1+\epsilon) \int_0^1 F_{P_{\theta_l} \sharp \mu}^{-1}(t)\cos (2\pi \xi_n t) \diff t \right)\, \forall l,
    \end{align}
    where  $\hat{f}(\mu)(\xi;\theta_1,\ldots,\theta_L) \in \Re^{Ln}$, $\theta_1,\ldots,\theta_L \sim \setU(\Theta)$, and $\xi_1,\ldots,\xi_n \sim p(\xi) \propto (1+\xi)^{-2}$ for $\xi>0$. For $\int_0^1 F_{P_{\theta_l} \sharp \mu}^{-1}(t)\cos (2\pi \xi_1 t) \diff t$, a quantile function approximation is used (see Section~\ref{sec:quantile:chapter:advances}).
\end{remark}

We have discussed SWE using the projection function that maps to the real line. For the spherical case, as discussed in Section~\ref{subsec:spherical_projections}, there is an alternative option: mapping to a circle. For this option, we can also form an embedding function.

\begin{remark}[Spherical Sliced Wasserstein Embedding]
\label{remark:spherical_SWE}
    Authors in~\citet{liu2025linear} utilize such a mapping $P_\theta: \Sm^{d-1} \to \Sm$, e.g., $P_\theta (x) = \frac{\theta x}{\|\theta x\|_2}$ with $\theta\in \mathbb{V}_2(\Re^d)$ (see Definition~\ref{def:SSW}), to obtain a new embedding for spherical probability measures:
    \begin{align}
        f_{\mu_0}(\mu)(x,\theta) = T_{\mu,\mu_0 }^\theta(x) - x,
    \end{align}
    where $\mu_0 \in \setP(\Sm^{d-1})$ is a reference measure and 
    \begin{align}
        T_{\mu,\mu_0 }^\theta (x) = F_{P_\theta \sharp \mu}^{-1}  \circ \left(F_{\mu_0} -\alpha_{P_\theta \sharp \mu,P_\theta \sharp \mu_0}\right)(x) - x, \quad  \forall x\in [0,1],
    \end{align}
    where $\alpha_{P_\theta \sharp \mu,P_\theta \sharp \mu_0}=\text{argmin}_{\alpha \in \Re} \int_0^1 h\left(|F_{P_\theta \sharp \mu}^{-1}(t) - (F_{P_\theta \sharp \mu_0}-\alpha)^{-1}(t)|\right)\diff t$, and $h$ is an increasing convex function.
\end{remark}

\chapter{Extensions of Sliced Optimal Transport}
\label{chapter:extension}

This chapter reviews several extensions of Sliced Optimal Transport (SOT). We begin with multi-marginal problems, where transport involves more than two probability measures. Next, we discuss unbalanced settings, in which the source and target measures may have different total masses. We then examine sliced partial optimal transport, which restricts transportation to only a fraction of the total mass. After that, we explore sliced variants of the Gromov--Wasserstein distance, which are designed to compare measures defined on different metric spaces. Finally, we present smooth SOT, which mitigates local irregularities in the measures by applying Gaussian convolution.

\section{Sliced Multi-marginal Optimal Transport}
\label{sec:SMOT:chapter:extension}
In this section, we focus on the multi-marginal optimal transport (OT) problem, which involves more than two probability measures. We now directly present the resulting metric, known as the multi-marginal Wasserstein distance.

\begin{definition}[Multi-marginal Wasserstein distance]
    \label{def:MW}
     Given $K\geq 1$ marginals $\mu_1,\ldots,\mu_{K} \in \mathcal{P}_2(\mathbb{R}^d)$, the multi-marginal Wasserstein distance~\citep{gangbo1998optimal} (MW) is defined as:
     \begin{align}
        &MW_{c}^2(\mu_1,\ldots,\mu_K) \nonumber\\
        &= \inf_{\pi \in \Pi(\mu_1,\ldots,\mu_K)}\int_{\Re^d \times \ldots\times \Re^d} c(x_1,\ldots,x_K )^2 \, d\pi(x_1,\ldots,x_K),
     \end{align}
     where $c:\Re^d\times \ldots\times \Re^d \to \Re_+$  is the multi-marginal ground metric and $\Pi(\mu_1,\ldots,\mu_K)$ is the set of multi-marginal transportation plans:
     \begin{align}
         &\Pi(\mu_1,\ldots,\mu_K)=\{\pi \in \mathcal{P}_2(\Re^d\times\ldots\times \Re^d)\mid \nonumber \\ &\quad \quad\pi(\Re^d,\ldots,A_k,\ldots,\Re^d) =\mu_k(A_k) \,\forall k\in \{1,\ldots,K\}\}.
     \end{align}
\end{definition}

There are many options for the ground metric, e.g., Coulomb interaction cost 
$$c(x_1,\ldots,x_K) = \sum_{i=1}^K \sum_{j=1}^K \frac{I(i\neq j)}{\|x_i-x_j\|_2},$$ 
which arises from quantum chemistry problems in density functional theory~\citep{cotar2013density,buttazzo2012optimal,gori2009density}, and barycentric cost~\citep{agueh2011barycenters,gangbo1998optimal}: 
\begin{align}
c(x_1,\ldots,x_K)^2=\sum_{k=1}^K \beta_k \left\| x_k-\sum_{k'=1}^K \beta_{k'} x_{k'}\right\|^2_2,
\end{align}
for $\beta_k>0$ for all $k=1,\ldots,K$ and $\sum_{k=1}^K\beta_k =1$. We refer the reader to~\citet{pass2015multi} for a more detailed review of multi-marginal optimal transport and the multi-marginal Wasserstein distance. With the barycentric cost, one-dimensional multi-marginal Wasserstein distance also admits a closed form.

\begin{proposition}[One-dimensional multi-marginal Wasserstein distance]
    \label{proposition:1DMWD}
    For $\mu_1,\ldots,\mu_K \in \mathcal{P}_2(\Re)$ and the barycentric cost 
    \begin{align}
    c(x_1,\ldots,x_K)^2=\sum_{k=1}^K \beta_k \left( x_k-\sum_{k'=1}^K \beta_{k'} x_{k'}\right)^2,
    \end{align}
    we have~\citep{cohen2021sliced}:
    \begin{align}
        MW_c^2(\mu_1,\ldots,\mu_K) = \int_0^1 \sum_{k=1}^K \beta_k\left|F_{\mu_k}^{-1}(t) - \sum_{k'=1}^K \beta_{k'}F_{\mu_{k'}}^{-1}(t)\right|^2 \diff t, 
    \end{align}
    where $F_{\mu_k}^{-1}$ is the inverse CDF or the quantile function of $\mu_k$.
    \begin{proof}
        We restate the proof in~\citet[Proposition 1]{cohen2021sliced}. We first rewrite:
        \begin{align*}
             c(x_1,\ldots,x_K )^2 &= \sum_{k=1}^K \beta_k \left( x_k-\sum_{k'=1}^K \beta_{k'} x_{k'}\right)^2  \\
             &= \sum_{k=1}^K \beta_k  x_k^2 - \sum_{k=1}^K \sum_{k'=1}^K \beta_k\beta_{k'}x_k x_{k'},
        \end{align*}
        hence 
        \begin{align*}
            MW_c^2(\mu_1,\ldots,\mu_K) &= \sum_{k=1}^K \beta_k \int_{\Re \times \ldots\times \Re} x_k^2 \diff \pi(x_1,\ldots,x_K) \\
            &\quad - \sum_{k=1}^K \sum_{k'=1}^K \beta_k\beta_{k'} \int_{\Re \times \ldots\times \Re} x_k x_{k'} \diff \pi(x_1,\ldots,x_K) \\
            &= \sum_{k=1}^K \beta_k \int_{\Re \times \ldots\times \Re} x_k^2 \diff \pi_k(x_k) \\
            &\quad - \sum_{k=1}^K \sum_{k'=1}^K \beta_k\beta_{k'} \int_{\Re \times \ldots\times \Re} x_k x_{k'} \diff \pi_{k,k'}(x_k,x_{k'}),
        \end{align*} 
        where $\pi_k$ corresponds to marginalizing $\pi$ onto component $k$, and $\pi_{k,k'}$ corresponds to marginalizing $\pi$ onto components $k$ and $k'$. This can be formalized by
        defining the map $P_{k}(x_1,\ldots,x_K)=x_k$ and $P_{k,k'}(x_1,\ldots,x_K)=(x_k,x_{k'})$, hence, $\pi_k = P_k\sharp \pi$ and $\pi_{k,k'} = P_{k,k'}\sharp \pi$.

        Now, we define $\pi^\star = (F_{\mu_1}^{-1},\ldots,F_{\mu_K}^{-1}) \sharp \sigma_0$ where $\sigma_0 := \mathcal{U}([0,1])$. Next, we show that $\pi^\star$ is optimal. We have $P_{k,k'}\sharp \pi^\star =(F_{\mu_k}^{-1},F_{\mu_{k'}}^{-1}) \sharp \sigma_0$ is the optimal transportation plan between $\mu_k$ and $\mu_{k'}$ (Remark~\ref{remark:1DWasserstein_general}). Therefore, we have:
        \begin{align}
            &\sum_{k=1}^K \sum_{k'=1}^K \beta_k\beta_{k'} \int_{\Re \times \ldots\times \Re} x_k x_{k'} \diff \pi_{k,k'}(x_k,x_{k'}) \nonumber \\
            &\leq \sum_{k=1}^K \sum_{k'=1}^K \beta_k\beta_{k'} \int_{\Re \times \ldots\times \Re} x_k x_{k'} \diff \pi^\star_{k,k'}(x_k,x_{k'}),
        \end{align}
        which implies that for any $\pi$:
        \begin{align}
            &\int_{\Re \times \ldots\times \Re} c(x_1,\ldots,x_K )^2 \diff \pi(x_1,\ldots,x_K) \nonumber \\
            &= \int_{\Re \times \ldots\times \Re} \left( x_k-\sum_{k'=1}^K \beta_{k'} x_{k'}\right)^2 \diff \pi(x_1,\ldots,x_K)\nonumber \\
            &= \sum_{k=1}^K \beta_k \int_{\Re \times \ldots\times \Re} x_k^2 \diff \pi(x_1,\ldots,x_K) \nonumber\\
            &\quad - \sum_{k=1}^K \sum_{k'=1}^K \beta_k\beta_{k'} \int_{\Re \times \ldots\times \Re} x_k x_{k'} \diff \pi(x_1,\ldots,x_K) \nonumber\\
            &\geq \sum_{k=1}^K \beta_k \int_{\Re \times \ldots\times \Re} x_k^2 \diff \pi^\star(x_1,\ldots,x_K) \nonumber\\
            &\quad - \sum_{k=1}^K \sum_{k'=1}^K \beta_k\beta_{k'} \int_{\Re \times \ldots\times \Re} x_k x_{k'} \diff \pi^\star(x_1,\ldots,x_K) \nonumber\\
            &= \int_{\Re \times \ldots\times \Re} \left( x_k-\sum_{k'=1}^K \beta_{k'} x_{k'}\right)^2 \diff \pi^\star(x_1,\ldots,x_K)\nonumber \\
            &= \int_{\Re \times \ldots\times \Re} c(x_1,\ldots,x_K )^2 \diff \pi^\star(x_1,\ldots,x_K),
        \end{align}
        which means that $\pi^\star$ is optimal. Plugging in $\pi^\star$, we have:
        \begin{align}
            &\int_{\Re \times \ldots\times \Re} c(x_1,\ldots,x_K )^2 \diff \pi^\star(x_1,\ldots,x_K) \nonumber \\
            &= \int_{\Re \times \ldots\times \Re} \left( x_k-\sum_{k'=1}^K \beta_{k'} x_{k'}\right)^2 \diff (F_{\mu_1}^{-1},\ldots,F_{\mu_K}^{-1}) \sharp \sigma_0 \nonumber \\
            &= \int_0^1 \sum_{k=1}^K \beta_k \left|F_{\mu_k}^{-1}(t) - \sum_{k'=1}^K \beta_{k'} F_{\mu_{k'}}^{-1}(t)\right|^2 \diff t,
        \end{align}
        which completes the proof.
    \end{proof}
\end{proposition}

From the one-dimensional case, we are again able to define the sliced version of the multi-marginal Wasserstein distance.

\begin{definition}[Sliced Multi-marginal Wasserstein distance]
    \label{def:SMW}
    Given $K\geq 1$ marginals $\mu_1,\ldots,\mu_{K} \in \mathcal{P}_2(\mathbb{R}^d)$ and the barycentric cost 
    $$c(x_1,\ldots,x_K)^2=\sum_{k=1}^K \beta_k \left( x_k-\sum_{k'=1}^K \beta_{k'} x_{k'}\right)^2,$$ 
    the sliced multi-marginal Wasserstein distance~\citep{cohen2021sliced} (SMW) is defined as:
    \begin{align}
    \label{eq:SMW}
        &SMW_c^2(\mu_1,\ldots,\mu_K) = \mathbb{E}_{\theta \sim \mathcal{U}(\mathbb{S}^{d-1})}\left[MW_c^2(\theta\sharp \mu_1,\ldots,\theta\sharp \mu_K)\right],
    \end{align}
    where  $\mathcal{U}(\mathbb{S}^{d-1})$ is the uniform distribution over the unit hypersphere in $d$ dimensions. 
\end{definition}

With the barycentric cost, the sliced multi-marginal Wasserstein distance is equivalent to a weighted sum of sliced Wasserstein distances. Therefore, minimizing the distance with respect to a marginal leads to the sliced Wasserstein barycenter problem.

\begin{proposition}[Connection to sliced Wasserstein distance]
    \label{proposition:connection_SMW_SW} 
    Given marginals $\mu_1,\ldots,\mu_K \in \mathcal{P}_2(\Re^d)$ and the barycentric cost  
    $$c(x_1,\ldots,x_K)^2=\sum_{k=1}^K \beta_k \left( x_k-\sum_{k'=1}^K \beta_{k'} x_{k'}\right)^2,$$ 
    we have:
    \begin{align}
        SMW_c^2(\mu_1,\ldots,\mu_K) = \frac{1}{2}\sum_{k=1}^K \sum_{k'=1}^K \beta_k \beta_{k'} SW_2^2(\mu_k,\mu_{k'}).
    \end{align}
    
    \begin{proof} 
    We extend the proof in ~\citet{cohen2021sliced}. We first show that:
    
\begin{align}
\sum_{k=1}^K \beta_k \left( x_k - \sum_{k'=1}^K \beta_{k'} x_{k'} \right)^2 = \frac{1}{2} \sum_{k=1}^K \sum_{k'=1}^K \beta_k \beta_{k'} (x_k - x_{k'})^2.
\end{align}
Let $\bar{x} = \sum_{k=1}^K \beta_k x_k$. Then the left-hand side becomes:
\begin{align}
\sum_{k=1}^K \beta_k (x_k - \bar{x})^2 = \sum_{k=1}^K \beta_k (x_k^2 - 2x_k \bar{x} + \bar{x}^2).
\end{align}
Expanding and simplifying:
\begin{align}
\sum_{k=1}^K \beta_k x_k^2 - 2\bar{x} \sum_{k=1}^K \beta_k x_k + \bar{x}^2 \sum_{k=1}^K \beta_k = \sum_{k=1}^K \beta_k x_k^2 - \bar{x}^2.
\end{align}
Now expand the right-hand side:
\begin{align*}
&\frac{1}{2} \sum_{k=1}^K \sum_{k'=1}^K \beta_k \beta_{k'} (x_k - x_{k'})^2 \\
&= \frac{1}{2} \left( \sum_{k,k'} \beta_k \beta_{k'} x_k^2 - 2 \sum_{k,k'} \beta_k \beta_{k'} x_k x_{k'} + \sum_{k,k'} \beta_k \beta_{k'} x_{k'}^2 \right).
\end{align*}
Since
\begin{align*}
&\sum_{k,k'} \beta_k \beta_{k'} x_k^2 = \sum_k \beta_k x_k^2 \sum_{k'} \beta_{k'} = \sum_k \beta_k x_k^2, \\ 
&\sum_{k,k'} \beta_k \beta_{k'} x_{k'}^2 = \sum_{k'} \beta_{k'} x_{k'}^2 \sum_k \beta_k = \sum_{k'} \beta_{k'} x_{k'}^2, \\
&\sum_{k,k'} \beta_k \beta_{k'} x_k x_{k'} = \left( \sum_k \beta_k x_k \right)^2 = \bar{x}^2,
\end{align*}
we obtain
\begin{align}
\frac{1}{2} \left( 2 \sum_k \beta_k x_k^2 - 2 \bar{x}^2 \right) = \sum_k \beta_k x_k^2 - \bar{x}^2.
\end{align}
Hence, both sides are equal, and the identity is proved. Next, for $\mu_1,\ldots,\mu_K \in \mathcal{P}_2(\Re)$ we have:
\begin{align}
     &MW_c^2(\mu_1,\ldots,\mu_K) \nonumber\\ &= \int_{\Re \times \ldots\times \Re} \left( x_k-\sum_{k'=1}^K \beta_{k'} x_{k'}\right)^2 \diff \pi^\star(x_1,\ldots,x_K) \nonumber \\
    &= \int_{\Re \times \ldots\times \Re} \frac{1}{2} \sum_{k=1}^K \sum_{k'=1}^K \beta_k \beta_{k'} (x_k - x_{k'})^2 \diff \pi^\star(x_1,\ldots,x_K)  \nonumber\\
    &= \frac{1}{2} \sum_{k=1}^K \sum_{k'=1}^K \beta_k \beta_{k'} \int_{\Re \times \ldots\times \Re} (x_k - x_{k'})^2 \diff \pi^\star(x_1,\ldots,x_K)   \nonumber\\
    &= \frac{1}{2} \sum_{k=1}^K \sum_{k'=1}^K \beta_k \beta_{k'} W_2^2(\mu_k,\mu_{k'}).
\end{align}
Back to higher dimensions, for $\mu_1,\ldots,\mu_K \in \mathcal{P}_2(\Re^d)$ we have: 
\begin{align}
    SMW_c^2(\mu_1,\ldots,\mu_K) &= \mathbb{E}_{\theta \sim \mathcal{U}(\mathbb{S}^{d-1})}[MW_c^2(\theta \sharp \mu_1,\ldots,\theta \sharp \mu_K)] \nonumber \\
    &= \mathbb{E}_{\theta \sim \mathcal{U}(\mathbb{S}^{d-1})}\left[ \frac{1}{2} \sum_{k=1}^K \sum_{k'=1}^K \beta_k \beta_{k'} W_2^2(\theta \sharp \mu_k,\theta \sharp \mu_{k'})\right] \nonumber\\
    &= \frac{1}{2} \sum_{k=1}^K \sum_{k'=1}^K \beta_k \beta_{k'} \mathbb{E}_{\theta \sim \mathcal{U}(\mathbb{S}^{d-1})}\left[W_2^2(\theta \sharp \mu_k,\theta \sharp \mu_{k'})\right]  \nonumber\\
    &= \frac{1}{2} \sum_{k=1}^K \sum_{k'=1}^K \beta_k \beta_{k'} SW_2^2( \mu_k, \mu_{k'}),
\end{align}
which completes the proof.
    \end{proof}
\end{proposition}

Proposition~\ref{proposition:connection_SMW_SW} allows us to derive statistical and topological properties of SMW using properties of SW (see Section~\ref{sec:SOTdistances:chapter:foundations}).  In addition, computation of SMW reduces to computation of $K^2$ SW distances (see Section~\ref{sec:MC:chapter:advances}).

\begin{proposition}[Connection to sliced Wasserstein barycenter]
    \label{proposition:connection_SMW_barycenter}
    For $\mu_1,\ldots,\mu_K \in \mathcal{P}_2(\Re^d)$ and weights $\beta_1,\ldots,\beta_K > 0$ with $\sum_{k=1}^K \beta_k = 1$, we define $\hat{\beta}_k = m \beta_k$ for some $m \in [0,1]$ and $k=1,\ldots,K$, and $\hat{\beta}_{K+1} = 1 - \sum_{k=1}^K \hat{\beta}_k = 1 - m$. Then we have:
    \begin{align}
        &\text{argmin}_{\mu \in \mathcal{P}_2(\Re^d)} SMW_c^2(\mu_1,\ldots,\mu_K,\mu) \nonumber \\
        &= \text{argmin}_{\mu \in \mathcal{P}_2(\Re^d)} \sum_{k=1}^K \beta_k SW_2^2(\mu_k,\mu),
    \end{align}
    where $c$ is the barycentric ground metric. This follows from the connection between SMW and SW in Proposition~\ref{proposition:connection_SMW_SW}.
\end{proposition}

Similar to the multi-marginal Wasserstein distance, the sliced multi-marginal Wasserstein distance is a generalized metric on the space of probability measures.

\begin{proposition}[Generalized metricity of sliced multi-marginal Wasserstein distance]
    \label{proposition:generalized_metricity_SMW}
    The sliced multi-marginal Wasserstein distance is non-negative, is zero if and only if all measures are identical, is permutation-equivariant, and satisfies a generalized triangle inequality. We refer the reader to \citet[Proposition 4]{cohen2021sliced} for more details.
\end{proposition}

\section{Sliced Unbalanced Optimal Transport}
\label{sec:SUOT:chapter:extension}
Unbalanced optimal transport (UOT) extends classical optimal transport to settings where the source and target measures do not have equal total mass, a common situation in real-world applications. In many domains, such as image processing, biology, and machine learning, data distributions may vary in size, intensity, or completeness, making mass conservation an unrealistic assumption. UOT provides a principled framework to compare and interpolate between such measures by introducing mass creation or destruction penalties, typically via divergence terms. This flexibility allows UOT to model more general scenarios while preserving the geometric structure of transport, making it a powerful tool for analyzing and comparing irregular or noisy data. In this section, we will discuss its sliced version, i.e., sliced unbalanced optimal transport. We first start with the definition of $\varphi$-divergence and UOT.

\begin{definition}[$\varphi$-divergence]
    \label{def:varphi_divergence} 
    Given $\mu,\nu \in \setM_+(\Re^d)$, let $\varphi:\Re_+ \to \Re_+$ be an entropy function, i.e., $\varphi$ is convex, lower semicontinuous, and $\varphi(1)=0$. The $\varphi$-divergence~\citep{csiszar1963informationstheoretische} between $\mu$ and $\nu$ is defined as:
    \begin{align}
        D_\varphi(\mu,\nu) = \int_{\Re^d} \varphi\left(\frac{\diff \mu}{\diff \nu} (x)\right) \diff \nu(x) + \varphi'_\infty \int_{\Re^d}\diff \mu^\perp (x),
    \end{align}
    where $\varphi'_\infty = \lim_{x\to \infty} \frac{\varphi(x)}{x}$ and $\mu^\perp$ is defined by the decomposition $\mu = \frac{\diff \mu}{\diff \nu} \nu +\mu^\perp$.
\end{definition} 

Here, $D_\varphi(\mu,\nu)$ satisfies the identity of indiscernibles property. When $\varphi(x) = \lambda |x-1|$, we obtain the Total Variation (TV) distance~\citep{takezawa2005introduction} (scaled with $\lambda$). When $\varphi(x)=\lambda(x \log(x) - x +1)$, we obtain the KL divergence~\citep{csiszar1975divergence}(scaled with $\lambda$).

\begin{definition}[Unbalanced Optimal Transport]
    \label{def:UOT}
    Given $\varphi_1,\varphi_2$ be two entropy functions and a ground metric $c:\Re^d\times \Re^d \to \Re$, the unbalanced optimal transport (UOT) problem between $\mu \in \setM_+(\Re^d)$ and $\nu \in \setM_+(\Re^d)$ ($d\geq 1$) is defined as~\citep{liero2018optimal}:
    \begin{align}
        &UOT_{c,\varphi_1,\varphi_2}(\mu,\nu)  \nonumber \\&= \inf_{\pi \in \setM_+(\Re^d\times \Re^d)} \int_{\Re^d\times \Re^d} c(x,y) \diff \pi(x,y) + D_{\varphi_1}(\pi_1,\mu) + D_{\varphi_2}(\pi_2,\nu),
    \end{align}
    where $\pi_1$ and $\pi_2$ are the first and second marginal measures of $\pi$, respectively.
\end{definition}

UOT with TV distance is introduced in~\citet{chizat2018interpolating}, while UOT with KL divergence is introduced in~\citet{liero2018optimal}. We refer the reader to~\citet{sejourne2023unbalanced} for a detailed discussion of UOT. Like OT, UOT admits a dual form.

\begin{remark}[Duality of Unbalanced Optimal Transport]
    \label{remark:duality_UOT}
    The UOT problem can be rewritten as~\citep[Corollary 4.12]{liero2018optimal}:
    \begin{align}
       & UOT_{c,\varphi_1,\varphi_2}(\mu,\nu) = \sup_{(f,g) \in \setR(c)}  \mathcal{D}(f, g; \mu, \nu),  \nonumber \\
       & \mathcal{D}(f, g; \mu, \nu)=\int_{\Re^d} \varphi_1^o(f(x)) \diff \mu(x) + \int_{\Re^d} \varphi_2^o(g(x)) \diff \nu(x),
    \end{align}
    where $\varphi^0(x) = -\varphi^*(-x) = \sup_{y\in\Re_+} \langle x,y \rangle - \varphi(y)$ ($\varphi^*$ is the Legendre transform), and $\setR(c)=\{(f,g) \in \setC(\Re^d) \mid f(x) + g(y) \leq c(x,y), \forall (x,y) \in \Re^d \times \Re^d\}$.
\end{remark}

The duality of UOT is vital for theoretical aspects of UOT. It is also important to compute UOT, e.g., deriving iterative solvers like Sinkhorn. In addition to the Sinkhorn solver for UOT~\citep{benamou2015iterative}, authors in~\citet{sejourne2022faster} propose using the Frank-Wolfe~\citep{frank1956algorithm} algorithm to optimize the dual problem. The Frank-Wolfe algorithm is a popular iterative first-order optimization algorithm for solving the following optimization problem:
\[
\max_{x \in \mathcal{E}} f(x),
\]
where $\mathcal{E}$ is a compact convex set and $f : \mathcal{E} \to \Re$ is a concave, differentiable function.
At each iteration, the algorithm maximizes a linear approximation of the objective function $f$. Given the current iterate $x_t$, the Frank-Wolfe (FW) algorithm solves the linear subproblem:
\[
r_{t+1} \in \arg\max_{r \in \mathcal{E}} \langle \nabla f(x_t), r \rangle,
\]
and updates the iterate via a convex combination:
\[
x_{t+1} = (1 - \gamma_{t+1}) x_t + \gamma_{t+1} r_{t+1},
\]
where the step size $\gamma_{t+1}$ is typically set to
\[
\gamma_{t+1} = \frac{2}{t + 3}.
\]
Authors in~\citet{sejourne2022faster} show that the Frank-Wolfe algorithm enjoys speedup for UOT in one dimension with a closed-form solution for the linear  approximation. Moreover, the authors point out that we can replace the dual $\mathcal{D}(f,g;\mu,\nu)$ with $\sup_{\lambda \in \Re}\mathcal{D}(f+\lambda,g-\lambda;\mu,\nu)$ and still obtain UOT. When $\varphi_1 = \rho_1 KL$ and $\varphi_2= \rho_2 KL$, we have 
\[
\lambda^\star(f,g) = \frac{\rho_1 \rho_2}{\rho_1+\rho_2} \log \left(\frac{\langle \mu,e^{-f/\rho_1}\rangle}{\langle \nu,e^{-g/\rho_2}\rangle}\right)
\]
~\citep[Proposition 2]{sejourne2022faster}.

\begin{remark}[Linear approximation for unbalaced optimal transport]
    \label{remark:linear_oracles_OT}
    The linear minimization problem for the Frank-Wolfe algorithm  for UOT is:
    \begin{align}
        (r_t,s_t) \in \arg\min_{(r,s) \in \setR(c)} \left\langle (r,s), \nabla \sup_{\lambda \in \Re}\mathcal{D}(\bar{f}_t+\lambda,\bar{g}_t-\lambda;\mu,\nu) \right\rangle,
    \end{align}
    which is equivalent to~\citep[Proposition 1]{sejourne2022faster}:
    \begin{align}
        (r_t,s_t) \in \arg\min_{(r,s) \in \setR(c)} \left\langle (r,s), (\tilde{\mu}_t,\tilde{\nu}_t) \right\rangle,
    \end{align}
    where $\tilde{\mu}_t = \nabla \varphi_1^* (-\bar{f}_t - \lambda^\star (\bar{f}_t,\bar{g}_t)) \mu$ and $\tilde{\nu}_t = \nabla \varphi_2^* (-\bar{g}_t - \lambda^\star (\bar{f}_t,\bar{g}_t)) \nu$. It turns out to be an optimal transport problem between $\tilde{\mu}_t$ and $\tilde{\nu}_t$.
\end{remark}

We now discuss the one-dimensional case of UOT, review sliced variants of UOT and their properties and relationships. While we restrict to linear projection, it is possible to extend to non-linear projection as in Section~\ref{sec:generalized_slicing:chapter:advances}.

\begin{remark}[One-dimensional Unbalanced Optimal Transport]
    \label{remark:1DUOT}
    From Remark~\ref{remark:linear_oracles_OT}, we can solve for $(r_t,s_t)$ with the closed-form solution of OT in 1D (which can be estimated efficiently in the discrete case as in Remark~\ref{remark:1DKantorovich_discrete}). Then the following iterative update is conducted:
    \begin{align}
        (\bar{f}_{t+1},\bar{g}_{t+1}) = (1-\gamma_t)(\bar{f}_{t},\bar{g}_{t}) + \gamma_t (r_t,s_t),
    \end{align}
    where $\gamma_t$ (e.g., $\gamma_{t+1} = \frac{2}{t + 3}$) is the step size.
\end{remark}

\begin{definition}[Sliced Unbalanced Optimal Transport]
    \label{def:SUOT}
    Given $\varphi_1,\varphi_2$ be two entropy functions and a ground metric $c:\Re \times \Re \to \Re$, the sliced unbalanced optimal transport (SUOT) problem between $\mu \in \setM_+(\Re^d)$ and $\nu \in \setM_+(\Re^d)$ ($d \geq 1$) is defined as~\citep{bonet2024slicing}:
    \begin{align}
        SUOT_{c,\varphi_1,\varphi_2}(\mu,\nu) = \mathbb{E}_{\theta \sim \sigma(\theta)} \big[ UOT_{c,\varphi_1,\varphi_2}(\theta \sharp \mu, \theta \sharp \nu) \big],
    \end{align}
    where $\sigma(\theta) \in \setP(\Sm^{d-1})$, and $\theta \sharp \mu$ and $\theta \sharp \nu$ are push-forward measures of $\mu$ and $\nu$ through the function $P_\theta(x) = \langle \theta,x \rangle$ (see Section~\ref{sec:Radontransform:chapter:foundations}).
\end{definition}

\begin{proposition}[Metricity of Sliced Unbalanced Optimal Transport]
    \label{proposition:metricity_SUOT}
    If UOT is non-negative, symmetric, and/or definite, then SUOT is respectively non-negative, symmetric, and/or definite. If UOT satisfies the triangle inequality, SUOT also satisfies the triangle inequality. The proof is given in~\citet[Proposition 3.3]{bonet2024slicing}. This property is similar to other sliced divergences~\citep{nadjahi2020statistical}.
\end{proposition}

\begin{remark}[Sample complexity of Sliced Unbalanced Optimal Transport]
    \label{remark:sample_complexity_SUOT}
    From~\citet[Theorem 3.4]{bonet2024slicing}, we know that SUOT also does not suffer from the curse of dimensionality with the rate $\mathcal{O}(n^{-1/2})$, where $n$ is the sample size.
\end{remark}

SUOT is a natural way to define a sliced variant of UOT, i.e., averaging one-dimensional UOT. However, there is an alternative way, which is using SW to replace the transportation cost in UOT.

\begin{definition}[Unbalanced Sliced Optimal Transport]
    \label{def:USOT}
    Given $\varphi_1,\varphi_2$ be two entropy functions and a ground metric $c:\Re \times \Re \to \Re_+$, the unbalanced sliced optimal transport (USOT) problem between $\mu \in \setM_+(\Re^d)$ and $\nu \in \setM_+(\Re^d)$ ($d \geq 1$) is defined as~\citep{bonet2024slicing}:
    \begin{align}
        &USOT_{c,\varphi_1,\varphi_2}(\mu,\nu) \nonumber \\ &= \inf_{\pi \in \setM_+(\Re^d \times \Re^d)} SW_{c,1}(\pi_1,\pi_2) + D_{\varphi_1}(\pi_1,\mu) + D_{\varphi_2}(\pi_2,\nu),
    \end{align}
    where $SW_{c,1}(\pi_1,\pi_2)$ is the sliced Wasserstein distance with ground metric $c$ of order 1 (Definition~\ref{def:SW}).
\end{definition}

\begin{proposition}[Metricity of Unbalanced Sliced Optimal Transport]
    \label{proposition:metricity_USOT}
    USOT is non-negative. If $\varphi_1 = \varphi_2$, USOT is symmetric. If $D_{\varphi_1}$ and $D_{\varphi_2}$ are definite, then USOT is definite. If $c(x,y) = |x - y|$ and $D_{\varphi_1} = D_{\varphi_2} = \rho TV$, then USOT satisfies the triangle inequality. The proof is given in~\citet[Proposition 3.8]{bonet2024slicing}.
\end{proposition}

We now discuss the relationship between SUOT, USOT, and UOT.

\begin{proposition}[Relationship between SUOT and USOT]
    \label{proposition:relationship_SUOT_USOT}
    For any $\mu,\nu \in \setM_+(\Re^d)$, two entropy functions $\varphi_1, \varphi_2$, and ground metric $c:\Re \times \Re \to \Re_+$, we have~\citep[Theorem 3.10]{bonet2024slicing}:
    \begin{align}
        SUOT_{c,\varphi_1,\varphi_2}(\mu,\nu) \leq USOT_{c,\varphi_1,\varphi_2}(\mu,\nu).
    \end{align}
\end{proposition}

\begin{proposition}[Relationship between SUOT (USOT) and UOT]
    \label{proposition:relationship_SUOT_USOT_UOT}
    For any $\mu,\nu \in \setM_+(\Re^d)$, two entropy functions $\varphi_1,\varphi_2$, ground metric $c:\Re^d \times \Re^d \to \Re_+$ and one-dimensional ground metric $c_1:\Re \times \Re \to \Re_+$ such that
    \[
        c_1(\langle \theta,x \rangle, \langle \theta,y \rangle) \leq c(x,y), \quad \forall \theta \in \Sm^{d-1}, \ \forall (x,y) \in \Re^d \times \Re^d,
    \]
    we have~\citep[Theorem 3.10]{bonet2024slicing}:
    \begin{align}
        SUOT_{c_1,\varphi_1,\varphi_2}(\mu,\nu) \leq USOT_{c_1,\varphi_1,\varphi_2}(\mu,\nu) \leq UOT_{c,\varphi_1,\varphi_2}(\mu,\nu).
    \end{align}
\end{proposition}
 From~\citet[Theorem 3.11]{bonet2024slicing}, we can further bound UOT with SUOT by restricting the support set of the two measures.
We now discuss how to compute SUOT and USOT in practice. Again, we need to use approximation techniques like Monte Carlo estimation.

\begin{definition}[Monte Carlo estimation of SUOT and USOT]
    \label{definition:MC_SUOT_SUOT}
    As in SOT, SUOT and USOT need to be approximated. Let $\theta_1, \ldots, \theta_L \simiid \sigma(\theta)$, the Monte Carlo estimations of SUOT and USOT are defined as follows:
    \begin{align}
        &\widehat{SUOT}_{c,\varphi_1,\varphi_2}(\mu,\nu; \theta_1, \ldots, \theta_L) = \frac{1}{L} \sum_{l=1}^L UOT_{c,\varphi_1,\varphi_2}(\theta_l \sharp \mu, \theta_l \sharp \nu), \\
        &\widehat{USOT}_{c,\varphi_1,\varphi_2}(\mu,\nu; \theta_1, \ldots, \theta_L) \nonumber \\ &= \inf_{\pi \in \setM_+(\Re^d \times \Re^d)} \widehat{SW}_{c,1}(\pi_1, \pi_2; \theta_1, \ldots, \theta_L) + D_{\varphi_1}(\pi_1, \mu) + D_{\varphi_2}(\pi_2, \nu),
    \end{align}
    where $\widehat{SW}_{c,1}(\pi_1, \pi_2; \theta_1, \ldots, \theta_L)$ is the Monte Carlo estimation of SW (Remark~\ref{remark:MC_SW}).
\end{definition}

\begin{proposition}[Duality of Monte Carlo estimation of SUOT and USOT]
    \label{proposition:duality_MC_SUOT_USOT} 
    From~\citet[Theorems 3.5 and 3.9]{bonet2024slicing}, under some mild conditions, we have:
    \begin{align}
        &\widehat{SUOT}_{c,\varphi_1,\varphi_2}(\mu,\nu; \theta_1, \ldots, \theta_L) = \sup_{(f,g) \in \setR(c)} \frac{1}{L} \sum_{l=1}^L \mathcal{D}(f,g; \theta_l \sharp \mu, \theta_l \sharp \nu), \\
        &\widehat{USOT}_{c,\varphi_1,\varphi_2}(\mu,\nu; \theta_1, \ldots, \theta_L) \nonumber \\ &= \sup_{(f,g) \in \setR(c)} \mathcal{D}\left(\frac{1}{L} \sum_{l=1}^L f \circ P_{\theta_l}, \frac{1}{L} \sum_{l=1}^L g \circ P_{\theta_l}; \mu, \nu \right),
    \end{align}
    where $\mathcal{D}$ is defined in Remark~\ref{remark:duality_UOT} and $P_\theta(x) = \langle \theta, x \rangle$.
\end{proposition}

With this duality, the authors in~\citet{bonet2024slicing} apply the Frank–Wolfe algorithm to compute SUOT and USOT. For SUOT, the problem can also be decomposed into $L$ independent one-dimensional unbalanced OT problems. Overall, both SUOT and USOT require iterative application of the Frank–Wolfe algorithm, where each iteration involves solving a sliced OT problem.

\section{Sliced Partial Optimal Transport}
\label{sec:SPOT:chapter:extension}
Partial optimal transport (partial OT) addresses situations where only a portion of the total mass from the source and target measures should be transported. This is particularly useful in applications where the measure may contain outliers, noise, or irrelevant regions, such as in image alignment, shape matching, or domain adaptation, where forcing full mass matching would lead to poor or misleading correspondences. By allowing a prescribed amount of mass to remain untransported, partial OT introduces robustness to mismatches and makes the transport problem more flexible and adaptive. This partial matching framework retains the geometric interpretability of optimal transport while being better suited to real-world, imperfect data. In this section, we will review sliced variants of partial OT. We first review partial OT and its variants.

\begin{definition}[Partial optimal transport]
    \label{def:POT}
    Given measures $\mu \in \setM_a(\Re^d)$ ($a>0$) and $\nu \in \setM_b(\Re^d)$ ($b>0$), the partial optimal transport (POT)~\citep{caffarelli2010free,figalli2010optimal} problem with the ground metric $c:\Re^d \times \Re^d \to \Re_+$ and limited mass $0 < s \leq \min(a,b)$ is defined as:
    \begin{align}
        POT_{s,c}(\mu,\nu) = \min_{\pi \in \Pi_s(\mu,\nu)} \int_{\Re^d \times \Re^d} c(x,y) \diff \pi(x,y),
    \end{align}
    where 
\begin{align*}
    &\Pi_s(\mu,\nu) = \left\{ \pi \in \setM_s(\Re^d \times \Re^d) \mid \pi(A \times \Re^d) \leq \mu(A) \right. \\ &\left. \quad \quad \quad \quad \quad  \forall A \subset \Re^d, \ \pi(\Re^d \times B) \leq \nu(B) \ \forall B \subset \Re^d \right\}.
\end{align*}
    
\end{definition}

\begin{definition}[One-sided partial optimal transport]
    \label{definition:one_sided_POT}
    Authors in~\citet{bonneel2019spot} propose an adjusted version of POT. In this version, one marginal transports all its mass while the other transports only a fraction of its mass. Given measures $\mu \in \setM_a(\Re^d)$ ($a>0$) and $\nu \in \setM_b(\Re^d)$ ($b>0$), the one-sided partial optimal transport (OPOT) problem with the ground metric $c:\Re^d \times \Re^d \to \Re_+$ is defined as:
    \begin{align}
        OPOT_c(\mu,\nu) = \min_{\pi \in \bar{\Pi}(\mu,\nu)} \int_{\Re^d \times \Re^d} c(x,y) \diff \pi(x,y),
    \end{align}
    where
    \begin{align*}
   & \bar{\Pi}(\mu,\nu) = \left\{ \pi \in \setM_b(\Re^d \times \Re^d) \mid \pi(A \times \Re^d) = \mu(A) \right. \\ &  \quad \quad \quad \quad \left.\forall A \subset \Re^d, \ \pi(\Re^d \times B) \leq \nu(B) \ \forall B \subset \Re^d \right\},
     \end{align*}
    with $a \leq b$.
\end{definition}

\begin{definition}[Optimal partial transport]
    \label{def:OPT}
    Authors in~\citet{bai2023sliced} relax POT's constraints into a Lagrange duality problem called optimal partial transport (OPT). OPT can be seen as a special case of UOT (Section~\ref{sec:SUOT:chapter:extension}) with total variation (TV) distance penalties, i.e., $\varphi_1 = \rho_1 TV$ and $\varphi_2 = \rho_2 TV$.
\end{definition}

\begin{figure}[!t]
    \centering
    \includegraphics[width=1\linewidth]{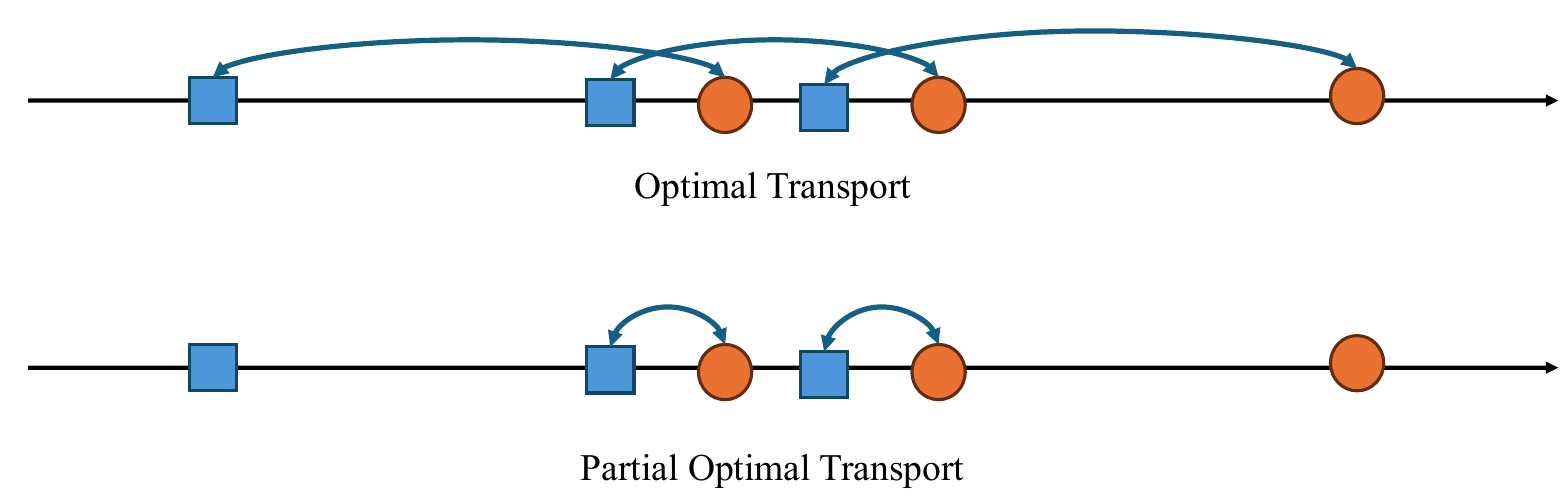}
    \caption{Optimal transport assignment and partial optimal transport assignment.}
    \label{fig:POT1D}
\end{figure}

The additional constraints in POT are also linear; hence, we can still use linear programming to compute POT in discrete cases or Sinkhorn solver with entropic regularization~\citep{benamou2015iterative}. Authors in~\citet{chapel2020partial} propose solving discrete POT by adding dummy points, which yields an OT problem. Overall, POT has at least the same computational cost as OT depending on the solvers. Practical usage of POT is restricted to its assignment problem, i.e., $\mu = \sum_{i=1}^n w \delta_{x_i}$ and $\nu = \sum_{j=1}^m w \delta_{y_j}$ for $w > 0$. We now discuss one-dimensional cases of POT variants (see Figure~\ref{fig:POT1D} for the idea of POT).

    

\begin{remark}[One-dimensional partial optimal transport assignment]
\label{remark:1DPOT}
    The support sets $\{x_1,\ldots,x_n\}$ and $\{y_1,\ldots,y_m\}$ are assumed to be disjoint. We denote $\{z_1,\ldots,z_{n+m}\} = \{x_1,\ldots,x_n,y_1,\ldots,y_m\}$. From \\\citet[Proposition 1]{chapel2025one}, $POT_s(\mu,\nu)$ can be computed using the transportation plan 
    \[
        \pi^s = \pi^k + (\pi^{k'} - \pi^k) (s \bmod w),
    \]
    where $k = \lfloor s/w \rfloor$ (floor), $k' = \lceil s/w \rceil$ (ceiling), and $\pi^k$, $\pi^{k'}$ are the corresponding transportation plans with limited masses $k$ and $k'$. We denote the active set associated to a solution $\pi^k$ as 
    \[
        \mathcal{A}_k = \left\{ x_i : \sum_j \pi^k_{ij} > 0 \right\} \cup \left\{ y_j : \sum_i \pi^k_{ij} > 0 \right\}.
    \]
    From~\citet[Theorem 1]{caffarelli2010free}, there exist $\pi^k$ and $\pi^{k'}$ such that $\mathcal{A}_k \subseteq \mathcal{A}_{k'}$ for all $k < k'$.

    \begin{enumerate}
        \item \textbf{Naive induction.} Starting from $\pi^0$ (with $\pi_{ij} = 0$ for all $i=1,\ldots,n$ and $j=1,\ldots,m$) and $\mathcal{A}_0 = \emptyset$, and assuming that we already have $\pi^k$ and $\mathcal{A}_k$, we iterate over all candidate pairs and pick the one that least increases the cost to obtain $\pi^{k+1}$ and $\mathcal{A}_{k+1}$. This approach has a poor complexity of $\mathcal{O}(n^4)$.

        \item \textbf{Restricting candidate set.} From~\citet[Proposition 2]{chapel2025one}, a sample $z_l \in \bar{\mathcal{A}}_k$ (the complement of $\mathcal{A}_k$) can only be added to the active set together with one of its neighbors in $\bar{\mathcal{A}}_k$ coming from the other measure. This property reduces the search space from $\mathcal{O}((n-k)^2)$ to $\mathcal{O}(n-k)$, resulting in an overall complexity of $\mathcal{O}(n^3)$.

        \item \textbf{Efficient computation of OT for each candidate set.} For each search step, the OT cost must be computed as the selection criterion. By limiting the ground metric to $c(x,y) = |x - y|$, authors in~\citet{chapel2025one} show that (i) all candidates form specific patterns on the real line called \emph{chains} (ordered sets of contiguous samples balanced between the two measures), which can be extracted efficiently; (ii) those chains can be decomposed into minimal sets for which the cost can be computed in constant time; and (iii) it is possible to select only the best candidate at each iteration. Overall, this reduces the complexity of the induction step $k$ from $\mathcal{O}(k(n-k))$ to $\mathcal{O}(n-k)$, leading to an overall complexity of $\mathcal{O}(n^2)$.

        \item \textbf{Choosing the best candidate.} At each step $k$, to avoid iterating over all candidates, authors in~\citet{chapel2025one} suggest maintaining an ordered list of candidates, sorted in increasing order of their marginal cost. Thus, the best candidate is always the first element of the list, reducing the overall complexity to $\mathcal{O}(n \log n)$.
    \end{enumerate}

    Recovering the transportation plan for a given $k$ requires an additional $\mathcal{O}(k \log k)$ time.
\end{remark}

\begin{remark}[One-dimensional one-sided partial optimal transport assignment]
    \label{remark:1DOPOT} 
    The following algorithm is proposed in~\citet{bonneel2019spot}. We restrict the ground metric to $c(x,y) = (x - y)^2$. Without loss of generality, assume $n < m$. We aim to find an injective assignment $\sigma: \{1,\ldots,n\} \to \{1,\ldots,m\}$ minimizing
    \[
        \sum_{i=1}^n (x_i - y_{\sigma(i)})^2.
    \]
    \begin{enumerate}
        \item \textbf{Quadratic time algorithm.} After sorting both sets of supports, compute a nearest neighbor assignment 
        \[
        t: \{1,\ldots,n\} \to \{1,\ldots,m\}
        \]
        between $X = \{x_1,\ldots,x_n\}$ and $Y = \{y_1,\ldots,y_m\}$. When $t$ is injective, it is the solution. Otherwise, resolve conflicts where $t(i) = t(j)$ for $i \neq j$. Scan $X$ from left to right and consider sub-matching problems $X' = \{x_1,\ldots,x_{n'}\}$ for $n' = 1,\ldots,n$. Assume we have solved the assignment with solution $\sigma: \{1,\ldots,n'\} \to \{1,\ldots,m\}$. Let $z$ be the index of the rightmost point in the range $\{y_{\sigma(1)}, \ldots, y_{\sigma(n')}\}$ that is unassigned to any point in $X'$. If no such $z$ exists, set $z = \sigma(1) - 1$. Let $r$ be the index in $X'$ such that $\sigma(r) = z + 1$. By construction, $\sigma([r,n'])$ is a range of consecutive integers $[z+1, \sigma(n')]$.

        Now consider the point $x_{n'+1}$. If $t(n'+1) > \sigma(n')$, set $\sigma(n'+1) = t(n'+1)$. If $t(n'+1) \leq \sigma(n')$, analyze two cases:
        \begin{itemize}
            \item \textit{Case 1:} Offset the last subsequence of consecutive values in $\sigma$ to the left to leave room for a new assignment, i.e., $\sigma([r,n']) = [z, \sigma(n')+1)$ and $\sigma(n'+1) = \sigma(n')+1$.
            \item \textit{Case 2:} Directly push the new assignment to the right: $\sigma(n'+1) = \sigma(n') + 1$.
        \end{itemize}
        Choose the case with smaller cost. This procedure finds the optimal assignment in $\mathcal{O}(\max(n+m, n^2))$.

        \item \textbf{Special cases.} Three special cases can be solved in linear time:
        \begin{itemize}
            \item \textit{Case 1:} There exists $k$ such that $x_i \leq y_1$ for all $i < k$. Then assign $\sigma(i) = i$ for all $i < k$.
            \item \textit{Case 2:} Reduce the range of $Y$ based on the number of non-injective values in $t$. It suffices to solve the sub-matching problem for 
            \[
            Y' = \{ y_j \}_{j = \max(1, t(1) - p), \ldots, \min(t(n) + p, m)},
            \]
            where 
            \[
            p = \mathrm{card}(\{ i : t(i) = t(i+1), \, i \leq n \}).
            \]
            \item \textit{Case 3:} When $n = m - 1$, there exists $k$ such that 
            \[
            \sigma([1,k-1]) = [1,k-1], \quad \sigma([k,n]) = [k+1,m].
            \]
            We can find $k$ by minimizing
            \[
                \min_k \sum_{i=1}^k (x_i - y_i)^2 + C - \sum_{i=1}^k (x_i - y_{i+1})^2,
            \]
            where $C = \sum_{i=1}^n (x_i - y_{i+1})^2$ is constant and need not be computed.
        \end{itemize}

        \item \textbf{Quasilinear time problem decomposition.} The final algorithm in~\citet{bonneel2019spot} decomposes the initial problem into many small independent subproblems that can be solved in parallel and utilize the special cases above. Each subproblem is characterized by point sets $X_k \subset X$ and $Y_k \subset Y$, and indices $z_k$ and $l_k$ indicating the first free spots to the left and right in $Y$ respectively. We keep track of flags $f_j$ for $j=1,\ldots,m$, where $f_j = \text{False}$ indicates that $y_j$ has been considered by Cases 1 or 2 (in the quadratic time algorithm) at some point, and $f_j = \text{True}$ otherwise.

        Considering $x_{n'+1}$, if $f_{t(n'+1)} = \text{True}$, then $t(n'+1)$ is masked as occupied by a candidate and we set $f_{t(n')} = \text{False}$. If $t(n'+1) = t(n')$, then we set $f_z = \text{False}$ and $f_{l+1} = \text{False}$. If $t(n'+1) \neq t(n')$ but $f_{t(n'+1)} = \text{False}$, then we set $f_{l+1} = \text{False}$.

        \textbf{Efficient implementation.} To avoid re-evaluating costs each time $t(n'+1) \leq \sigma(n')$ in the quadratic time algorithm, costs are progressively updated by storing intermediate results from previous iterations.
    \end{enumerate}
\end{remark}

\begin{remark}[One-dimensional optimal partial transport assignment]
    \label{remark:1DOPT}
    From~\citet[Proposition 4.3]{bai2023sliced}, OPT with 
    \[
        \mu = \sum_{i=1}^n \delta_{x_i} \quad \text{and} \quad \nu = \sum_{j=1}^m \delta_{y_j}, \quad \text{and} \quad \rho_1 = \rho_2 = \lambda,
    \]
    admits the following dual form:
    \begin{align}
        OPT_c(\mu,\nu) = \sup_{\substack{\fb \in \Re^n, \gb \in \Re^m \\ f_i + g_j \leq c(x_i,y_j) \, \forall i,j}} \sum_{i=1}^n \min(f_i, \lambda) + \sum_{j=1}^m \min(g_j, \lambda).
    \end{align}
    Moreover, the necessary and sufficient conditions for the optimal plan $\pi$ are as follows: for all $(x_i, y_j) \in \mathrm{supp}(\pi)$,
    \begin{align*}
        & f_i + g_j = c(x_i, y_j), \\
        & f_i < \lambda \implies \pi_{1,i} = 1, \\
        & f_i = \lambda \implies \pi_{1,i} \in [0,1], \\
        & f_i > \lambda \implies \pi_{1,i} = 0, \\
        & g_j < \lambda \implies \pi_{2,j} = 1, \\
        & g_j = \lambda \implies \pi_{2,j} \in [0,1], \\
        & g_j > \lambda \implies \pi_{2,j} = 0.
    \end{align*}

    The following algorithm~\citep{bai2023sliced} finds solutions $(\pi, \fb, \gb)$ satisfying these primal-dual optimality conditions. Assume we have solved $OPT(\mu^{k}, \nu)$, where $\mu^{k} = \sum_{i=1}^{k-1} \delta_{x_i}$, i.e., having an assignment 
    \[
        \sigma : \{1, \ldots, k-1\} \to \{-1\} \cup \{1, \ldots, m\},
    \]
    where $-1$ denotes no association. The supports are sorted and distinct. For a new point $x_k$, define
    \[
        j^* = \arg\min_{j \in \{1, \ldots, m\}} \bigl( c(x_k, y_j) - g_j \bigr),
    \]
    and set
    \[
        f_k = \min \bigl( c(x_k, y_{j^*}) - g_{j^*}, \lambda \bigr).
    \]

    \begin{itemize}
        \item \textbf{Case 1.} If $f_k = \lambda$, set $\sigma(k) = -1$ (no assignment).
        
        \item \textbf{Case 2.} If $f_k < \lambda$ and $y_{j^*}$ is unassigned, set $\sigma(k) = j^*$.
        
        \item \textbf{Case 3.} If $f_k < \lambda$ and $y_{j^*}$ is already assigned (say to $x_{k-1}$), perform the following update:
        \begin{itemize}
            \item Increase $f_i$ for $i \in [k-1, k]$ and simultaneously decrease $g_{j^*}$ at the same rate until one of the following occurs:
            \begin{itemize}
                \item \textit{Case 3.1.} Either $f_{k-1}$ or $f_k$ reaches $\lambda$. The corresponding $x$ becomes unassigned, and the other remains or becomes assigned to $y_{j^*}$.
                
                \item \textit{Case 3.2.} The condition $f_k + g_{j^*+1} = c(x_k, y_{j^*+1})$ is met, then assign $x_k$ to $y_{j^*+1}$.
                
                \item \textit{Case 3.3a.} The condition $f_{k-1} + g_{j^*-1} = c(x_{k-1}, y_{j^*-1})$ is met, and if $y_{j^*-1}$ is unassigned, assign $x_{k-1}$ to $y_{j^*-1}$ and $x_k$ to $y_{j^*}$.
                
                \item \textit{Case 3.3b.} If $y_{j^*-1}$ is assigned (say to $x_{k-2}$), generalize \textit{Case 3}:
                \begin{itemize}
                    \item Consider the consecutive subset $\{x_i\}_{i = i_{\min}}^{k-1}$ monotonically matched to contiguous $\{y_j\}_{j = j_{\min}}^{j^*}$, where $i_{\min}$ starts at $k-1$ and $j_{\min}$ is the index of $y_j$ assigned to $x_{i_{\min}}$.
                    \item Increase $f_i$ for $i \in [i_{\min}, k]$ and decrease $g_j$ for $j \in [j_{\min}, j^*]$ simultaneously until one of the following:

                         $f_{k'} = \lambda$ for some $k' \in [i_{\min}, k+1]$ (reduces to \textit{Case 3.1}),
                         
                        $f_k + g_{j^*+1} = c(x_k, y_{j^*+1})$ (reduces to \textit{Case 3.2}),
                        
                         $f_{i_{\min}} + g_{j_{\min}-1} = c(x_{i_{\min}}, y_{j_{\min}-1})$ (reduces to \textit{Case 3.3}).
                \end{itemize}
            \end{itemize}
        \end{itemize}
    \end{itemize}

    We refer the reader to~\citet[Theorem 4.4]{bai2023sliced} for the proof of optimality of this algorithm.
\end{remark}

With 1D POT, there are two ways to define a distance: 
\begin{itemize}
    \item Rely on the 1D transportation cost, similarly to sliced Wasserstein.
    \item Use the transportation cost with the ``lifted" discrete plans (see Section~\ref{sec:map:chapter:advances}).
\end{itemize}

\begin{definition}[Sliced Partial Optimal Transport]
\label{def:SPOT}
Given measures $\mu \in \setM_a(\Re^d)$ ($a > 0$) and $\nu \in \setM_b(\Re^d)$ ($b > 0$), a ground metric $c: \Re \times \Re \to \Re_+$, and a one-dimensional POT distance $\mathcal{D}_c: \setM_a(\Re) \times \setM_b(\Re) \to \Re_+$ (one of the three discussed POT variants), the \emph{sliced partial optimal transport} (SPOT) distance is defined as:
\begin{align}
    SPOT_c(\mu, \nu) = \mathbb{E}_{\theta \sim \sigma(\theta)} \bigl[ \mathcal{D}_c(P_\theta \sharp \mu, P_\theta \sharp \nu) \bigr],
\end{align}
where $\sigma(\theta) \in \setP(\Sm^{d-1})$ (e.g., the uniform distribution $\setU(\Sm^{d-1})$), and $P_\theta : \Re^d \to \Re$ is the projection operator, e.g., $P_\theta(x) = \langle \theta, x \rangle$.
\end{definition}
This definition is used in~\citet{bonneel2019spot,bai2023sliced}.

\begin{proposition}[Metricity of SPOT]
    \label{proposition:metricity_SPOT}
    SPOT satisfies symmetry and non-negativity. The triangle inequality and identity of indiscernibles hold when using the OPT formulation~\citep{bai2023sliced}.
\end{proposition}

\begin{definition}[Sliced Partial Wasserstein]
\label{def:SPW}
Given discrete measures $\mu = \sum_{i=1}^n w \delta_{x_i}$ and $\nu = \sum_{j=1}^m w \delta_{y_j}$ with $w > 0$, ground metrics $c: \Re^d \times \Re^d \to \Re_+$ and $c_1: \Re \times \Re \to \Re_+$, and a transportation plan 
\[
    \pi_{c_1}: \setM_{wn}(\Re) \times \setM_{wm}(\Re) \to \Re_+^{n \times m}
\]
corresponding to one of the three discussed one-dimensional POT problems, the \emph{sliced partial Wasserstein} (SPW) distance is defined as:
\begin{align}
    SPW_c(\mu, \nu) = \min_{\theta \in \Sm^{d-1}} \sum_{i=1}^n \sum_{j=1}^m c(x_i, y_j) \cdot (\pi_{c_1}(P_\theta \sharp \mu, P_\theta \sharp \nu))_{ij},
\end{align}
where $P_\theta : \Re^d \to \Re$ is the projection, e.g., $P_\theta(x) = \langle \theta, x \rangle$.
\end{definition}
This definition is used in~\citet{chapel2025one}.  It is also possible to replace the minimization over $\theta$ by an expectation with respect to $\theta \sim \sigma(\theta)$, where $\sigma(\theta) \in \setP(\Sm^{d-1})$, as discussed in Section~\ref{sec:map:chapter:advances}. Finally, we discuss the metricity of SPW.

\begin{proposition}[Metricity of SPW]
    \label{remark:metricity_SPW}
    SPW satisfies symmetry, non-negativity, and identity of indiscernibles~\citep{chapel2025one}.
\end{proposition}

\section{Sliced Gromov-Wasserstein}
\label{sec:SGW:chapter:extension}
Gromov-Wasserstein (GW) is a powerful extension of classical optimal transport that allows for the comparison of probability measures supported on different metric spaces, where no common ground space exists. Instead of relying on pointwise correspondences, GW focuses on aligning the intrinsic relational structures of each distribution by comparing their pairwise distance matrices. This makes it particularly well-suited for tasks like graph matching, shape comparison, and cross-domain analysis, where structural similarity is more meaningful than spatial proximity. Building on this, Fused Gromov-Wasserstein (FGW) further enhances the framework by incorporating feature-level information in addition to structural alignment. By blending the geometric structure captured by GW with a fidelity term that compares features directly, FGW enables a richer and more flexible comparison of datasets, especially in applications where both relational and descriptive information are important, such as attributed graph matching or multi-modal data integration. In this section, we review the sliced variants of Gromov-Wasserstein and Fused Gromov-Wasserstein, i.e., sliced Gromov-Wasserstein, rotation invariant sliced Gromov-Wasserstein, and sliced fused Gromov-Wasserstein. We first start by reviewing the Gromov-Wasserstein distance.

\begin{definition}[Gromov-Wasserstein distance]
    Given $\mu \in \setP_{c_1}(\Re^{d_1})$ and $\nu \in \setP_{c_2}(\Re^{d_2})$ with ground metrics $c_1:\Re^{d_1}\times \Re^{d_1}\to \Re_+$ and $c_2:\Re^{d_2}\times \Re^{d_2}\to \Re_+$, the Gromov-Wasserstein (GW) distance of order $p\geq1$ between $\mu$ and $\nu$ is defined as~\citep{memoli2011gromov}:
    \begin{align}
        &GW_{c_1,c_2,p}^p(\mu,\nu) = \inf_{\pi \in \Pi(\mu,\nu)} \int_{\Re^{d_1} \times \Re^{d_2}}\int_{\Re^{d_1} \times \Re^{d_2}}  \nonumber\\& \quad \quad |c_1(x_i,x_{i'}) - c_2(y_j,y_{j'}) |^p \diff \pi(x_i,y_j) \diff \pi(x_{i'},y_{j'}),
    \end{align}
    where $\Pi(\mu,\nu)$ is the set of transportation plans between $\mu$ and $\nu$.
\end{definition}

When $c_1$ and $c_2$ are distances on $\Re^{d_1}$ and $\Re^{d_2}$, $(\Re^{d_1}, c_1, \mu)$ and $(\Re^{d_2}, c_2 , \nu)$
are measurable metric spaces, $GW_{c_1,c_2,p}$ is a valid metric between $(\Re^{d_1}, c_1, \mu)$ and $(\Re^{d_2}, c_2 , \nu)$.  In particular, GW is symmetric, satisfies the triangle inequality
when considering three measurable metric spaces, and vanishes if and only if the measurable metric spaces are isomorphic, i.e., when
there exists a surjective function $f : \Re^{d_1} \to \Re^{d_2}$ such that $f\sharp \mu = \nu$ and $c_1(x,x') = c_2(f(x),f(x'))$ for all $x,x' \in \Re^{d_1}$.

\begin{remark}[Discrete Gromov-Wasserstein distance]
    \label{remark:discrete_GW}
Given two discrete measures $\mu=\sum_{i=1}^n \alpha_i\delta_{x_i}$ and $\nu=\sum_{j=1}^m \beta_j\delta_{y_j}$, GW can be rewritten as:
   \begin{align}
        &GW_{c_1,c_2,p}^p(\mu,\nu) \nonumber \\&= \inf_{\pi \in \Pi(\mu,\nu)} \sum_{i=1}^n\sum_{j=1}^m \sum_{i'=1}^n \sum_{j'=1}^m |c_1(x_i,x_{i'}) - c_2(y_j,y_{j'}) |^p  \pi_{ij} \pi_{i'j'},
    \end{align}
    which is a non-convex quadratic program when $p=2$. A conditional gradient approach is introduced in~\citet{titouan2019optimal}, where each iteration requires the optimization of a classical OT problem with complexity $\mathcal{O}(n^3)$. By adding entropic regularization, a projected gradient solver is proposed in~\citet{peyre2016gromov} with overall complexity $\mathcal{O}(n^3)$.
\end{remark}

Now, we discuss two variants of sliced Gromov-Wasserstein.
\begin{definition}[Sliced Gromov-Wasserstein]
\label{def:SGW}
Given $\mu \in \setP(\Re^{d_1})$ and $\nu \in \setP(\Re^{d_2})$ with ground metrics $c_1:\Re\times \Re\to \Re_+$ and $c_2:\Re\times \Re\to \Re_+$, a lifting function $\Delta:\Re^{d_1} \to \Re^{d_2}$ (assumed that $d_1\leq d_2$), the sliced Gromov-Wasserstein (SGW) discrepancy of order $p\geq1$ between $\mu$ and $\nu$ is defined as~\citep{titouan2019sliced}:
    \begin{align}
        &SGW_{\Delta,c_1,c_2,p}^p(\mu,\nu)= \mathbb{E}_{\theta \sim \setU(\Sm^{d_2-1})}[GW_{c_1,c_2,p}^p(\theta \sharp \Delta \sharp \mu, \theta \sharp \nu)],
    \end{align}
    where $\setU(\Sm^{d_2-1})$ is the uniform distribution over the unit hypersphere of dimension $d_2-1$.
\end{definition}

In the original paper~\citep{titouan2019sliced}, $\Delta$ is chosen as a padding function, i.e., $\Delta(x)= (x,0,\ldots,0)$. Monte Carlo estimation is used in practice to approximate the expectation in SGW like SW (Section~\ref{sec:MC:chapter:advances}).

\begin{definition}[Rotation Invariant Sliced Gromov-Wasserstein]
    \label{def:RISGW}
    Authors in~\citet{titouan2019sliced} propose a rotation invariant version of SGW:
    \begin{align}
        RISGW_{c_1,c_2,p}^p(\mu,\nu) = \min_{U \in \mathbb{V}_{d_1}(\Re^{d_2})}SGW_{\Delta_U,c_1,c_2,p}^p(\mu,\nu),
    \end{align}
    where $\Delta_U(x) = U^\top x$ and $\mathbb{V}_{d_1}(\Re^{d_2})$ is the Stiefel manifold.
\end{definition}

\begin{proposition}[Properties of SGW and RISGW]
\label{proposition:properties_SGWW_RISGW}
From~\citet[Theorem 3.3]{titouan2019sliced}, SGW and RISGW are translation invariant. RISGW is also rotation invariant when $d_1=d_2$. When $d_1=d_2$, SGW and RISGW are pseudo-distances, i.e., they are symmetric, satisfy the
triangle inequality, and $RISGW_{c_1,c_2,p}^p(\mu,\mu) = SGW_{\Delta,c_1,c_2,p}^p(\mu,\mu)=0$. When $d_1=d_2$, and $c_1(x,y)=c_2(x,y)=\|x-y\|_2$, $SGW_{\Delta,c_1,c_2,p}^p(\mu,\mu)=0$ implies $GW_{c_1,c_2,2}^2(\mu,\nu)=0$ (abusing notation of $c_1$ and $c_2$).
\end{proposition}

\begin{remark}[One-dimensional Gromov-Wasserstein distance]
    \label{remark:1DGW}
    In contrast to the one-dimensional Wasserstein distance, which admits a closed-form solution allowing fast computation, one-dimensional GW does not have a closed-form solution to date. Practitioners still utilize one-dimensional optimal transport mapping (using quantile functions, CDFs, and sorting) since such mapping is ``often'' the correct mapping. However, such a solution is not true in general, given counterexamples in~\citet{beinert2023assignment}. We refer the reader to~\citet{dumont2022existence} for further discussion. 
\end{remark}

Next, we review the definition of fused Gromov-Wasserstein.
\begin{definition}[Fused Gromov-Wasserstein distance]
\label{def:FGSW}
     Given $\mu \in \setP_{c_1}(\Re^d\times \Re^{d_1})$ and $\nu \in \setP_{c_2}(\Re^d\times \Re^{d_2})$ with ground metrics $c:\Re^d\times \Re^d\to \Re$, $c_1:\Re^{d_1}\times \Re^{d_1}\to \Re_+$, and $c_2:\Re^{d_2}\times \Re^{d_2}\to \Re_+$, the Fused Gromov-Wasserstein (FGW) distance of order $p\geq1$ between $\mu$ and $\nu$ is defined as~\citep{vayer2020fused}:
    \begin{align}
        &FGW_{c,c_1,c_2,\alpha,p}^p(\mu,\nu) = \inf_{\pi \in \Pi(\mu,\nu)} \int_{\Re^d\times \Re^{d_1} \times \Re^d \times \Re^{d_2}}\int_{\Re^d\times \Re^{d_1} \times \Re^d \times \Re^{d_2}} \nonumber \\& \quad  (\alpha c(a_i,b_j)+ (1-\alpha)|c_1(x_i,x_{i'}) - c_2(y_j,y_{j'}) |)^p  \nonumber \\&\quad \diff \pi(a_i,x_i,b_j,y_j) \diff \pi(a_{i'},x_{i'},b_{j'},y_{j'}),
    \end{align}
    where $\Pi(\mu,\nu)$ is the set of transportation plans between $\mu$ and $\nu$, and $\alpha \in [0,1]$.
\end{definition}

Like GW, discrete FGW can be solved using a conditional gradient approach in~\citet{titouan2019optimal}, where each iteration requires the optimization of a classical OT problem with complexity $\mathcal{O}(n^3)$. Similar to GW, sliced fused Gromov-Wasserstein is introduced in~\citet{xu2020learning}. We adapt the definition as follows:

\begin{definition}[Sliced Fused Gromov-Wasserstein]
    \label{def:SFGW}
Given $\mu \in \setP(\Re^d\times \Re^{d_1})$ and $\nu \in \setP(\Re^d\times \Re^{d_2})$ with ground metrics $c:\Re\times \Re\to \Re$, $c_1:\Re\times \Re\to \Re_+$, and $c_2:\Re\times \Re\to \Re_+$,  a lifting function $\Delta:\Re^{d_1} \to \Re^{d_2}$ (assumed that $d_1\leq d_2$), the sliced fused Gromov-Wasserstein (SFGW) discrepancy of order $p\geq1$ between $\mu$ and $\nu$ is defined as~\citep{xu2020learning}:
    \begin{align}
        SFGW_{\Delta,c,c_1,c_2,\alpha,p}^p(\mu,\nu) &= \mathbb{E}_{(\theta,\theta') \sim \setU(\Sm^{d-1})\otimes\setU(\Sm^{d_2-1})}\big[ \alpha W_c^p(\theta\sharp \mu_1,\theta \sharp \nu_1) \nonumber \\ &\quad +(1-\alpha)GW_{c_1,c_2,p}^p(\theta' \sharp \Delta \sharp \mu_2, \theta' \sharp \nu_2) \big],
    \end{align}
    where $\alpha \in [0,1]$, $\mu_1 \in \setP(\Re^d)$ and $\mu_2 \in \setP(\Re^{d_1})$ are marginals of $\mu$, and $\nu_1 \in \setP(\Re^d)$ and $\nu_2 \in \setP(\Re^{d_2})$ are marginals of $\nu$.
\end{definition}

Monte Carlo estimation is used in practice to approximate the expectation in SFGW, similar to SW (Section~\ref{sec:MC:chapter:advances}).

\begin{remark}[One-dimensional fused Gromov-Wasserstein distance]
    \label{remark:1DGGW}
    Similar to GW, the one-dimensional FGW distance does not admit a closed-form solution. In practice, one-dimensional optimal transport mapping (using quantile functions, CDFs, and sorting) is often used. 
\end{remark}

A recent attempt to obtain a fast sliced version of Gromov-Wasserstein and fused Gromov-Wasserstein appears in~\citet{piening2025novel}. In particular, the author replaces Gromov-Wasserstein by lower bounds in the form of Wasserstein distances between transformed measures.

\section{Smooth Sliced Optimal Transport}
\label{sec:SmoothSOT:chapter:extension}
Smooth optimal transport offers a principled relaxation of classical optimal transport by convolving input distributions with a Gaussian kernel before computing the transport cost. This yields the smooth Wasserstein distance, which retains the core geometric and metric properties of the original OT framework while dramatically improving statistical and computational behavior in high dimensions. In this section, we review smooth sliced optimal transport, which has a nice connection with differential privacy. We begin with the definition of smooth Wasserstein and its properties.

\begin{definition}[Smooth Wasserstein Distance]
    \label{def:smoothWD}
    Given $\sigma\in \Re_+$, $\mu \in \setP_p(\Re^d)$ and $\nu \in \setP_p(\Re^d)$ ($p\geq 1$), the $\sigma$-smooth Wasserstein-$p$ distance between $\mu$ and $\nu$ is defined as follows~\citep{nietert2021smooth}:
    \begin{align}
        W_p^{(\sigma)}(\mu,\nu)=W_p(\mu*\mathcal{N}_\sigma, \nu*\mathcal{N}_\sigma),
    \end{align}
    where $\mathcal{N}_\sigma:=\mathcal{N}(0,\sigma^2 I)$ and $*$ denotes the convolution operator.
\end{definition}

$W_p^{(\sigma)}(\mu,\nu)$ is continuous and monotonically non-increasing in $\sigma \in \Re_+$. When $\sigma \to 0$, we have $W_p^{(\sigma)}(\mu,\nu) \to W_p(\mu,\nu)$.

\begin{proposition}[Metricity of Smooth Wasserstein Distance]
    \label{proposition:metricity_smoothW}
$W_p^{(\sigma)}$ is a metric on $\setP_p(\Re^d)$ for all $\sigma \in \Re_+$, i.e., it satisfies symmetry, non-negativity, the triangle inequality, and identity of indiscernibles. $W_p^{(\sigma)}$ induces the same topology as $W_p$~\citep[Proposition 1]{nietert2021smooth}.
\end{proposition}

We refer the reader to~\citet{nietert2021smooth} for more theoretical properties of smooth Wasserstein distance. One key property is that it enjoys a parametric
empirical convergence rate $\mathcal{O}(n^{-1/2})$. We now review the smooth sliced Wasserstein distance, its metricity, its computation, and its differential privacy aspect.

\begin{definition}[Smooth Sliced Wasserstein Distance]
    \label{def:smooth_SW} 
     Given $\sigma\in \Re_+$, $\mu \in \setP_p(\Re^d)$ and $\nu \in \setP_p(\Re^d)$ ($p\geq 1$), the $\sigma$-smooth sliced Wasserstein-$p$ (Smooth-SW) distance between $\mu$ and $\nu$ is defined as follows~\citep{rakotomamonjy2021differentially,alaya2024gaussiansmoothed}:
    \begin{align}
        SW_p^{(\sigma),p}(\mu,\nu)=\mathbb{E}_{\theta \sim \setU(\Sm^{d-1})}[W_p^{(\sigma)}(\theta \sharp \mu,\theta \sharp \nu)],
    \end{align}
    where $\mathcal{N}_\sigma:=\mathcal{N}(0,\sigma^2)$. 
\end{definition}

\begin{proposition}[Metricity of Smooth Sliced Wasserstein Distance]
    \label{proposition:metricity_smoothSW}
$SW_p^{(\sigma)}$ is a metric on $\setP_p(\Re^d)$ for all $\sigma \in \Re_+$, i.e., it satisfies symmetry, non-negativity, the triangle inequality, and identity of indiscernibles~\citep{rakotomamonjy2021differentially}. The proof is similar to the case of SW using the metricity of smooth Wasserstein distance.
\end{proposition}

$SW_p^{(\sigma)}$ also induces weak convergence of probability measures on $\Re^d$~\citep{alaya2024gaussiansmoothed}.

\begin{definition}[Monte Carlo estimation of Smooth-SW]
    \label{definition:MC_smooth_SW}
    Given Monte Carlo samples $\theta_1,\ldots,\theta_L \simiid \setU(\Sm^{d-1})$ ($L\geq 1$), a Monte Carlo estimate of Smooth-SW can be formed as follows~\citep{rakotomamonjy2021differentially,alaya2024gaussiansmoothed}:
    \begin{align}
        \widehat{SW}_p^{(\sigma),p}(\mu,\nu)=\frac{1}{L} \sum_{l=1}^L W_p^{(\sigma)}(\theta_l \sharp \mu,\theta_l \sharp \nu).
    \end{align}
    Other approximation techniques (Section~\ref{sec:MC:chapter:advances}) can also be used.
\end{definition}
\begin{remark}[Empirical approximation of Smooth-SW]
    \label{remark:empirical_approximation_smoothSW}
    The computation of $W_p^{(\sigma)}(\theta \sharp \mu,\theta \sharp \nu)$ requires empirical approximation. When $\mu$ and $\nu$ are discrete measures, i.e., $\mu = \sum_{i=1}^n \alpha_i \delta_{x_i}$ and $\nu=\sum_{j=1}^m \beta_j \delta_{y_j}$, we have the smoothed projected measures $\theta \sharp \mu * \mathcal{N}_\sigma = \sum_{i=1}^n \alpha_i \mathcal{N}(\langle \theta,x_i\rangle,\sigma^2)$ and $\theta \sharp \nu * \mathcal{N}_\sigma = \sum_{j=1}^m \beta_j \mathcal{N}(\langle \theta,y_j\rangle,\sigma^2)$. We then sample $t_1,\ldots,t_n \simiid \theta \sharp \mu * \mathcal{N}_\sigma$ and $z_1,\ldots,z_m \simiid \theta \sharp \nu * \mathcal{N}_\sigma$. After that, we can calculate 
    \[
    W_p^p \left(\frac{1}{n} \sum_{i=1}^n \delta_{t_i}, \frac{1}{m} \sum_{j=1}^m \delta_{z_j}\right)
    \]
    using the closed-form solution in one dimension. When $\mu$ and $\nu$ are continuous measures, we can still sample i.i.d. from $\theta \sharp \mu * \mathcal{N}_\sigma$ and $\theta \sharp \nu * \mathcal{N}_\sigma$. In particular, we sample $x_1,\ldots,x_n \simiid \mu$, $y_1,\ldots,y_m \simiid \nu$, $\epsilon_1,\ldots,\epsilon_n \simiid \mathcal{N}(0,\sigma^2)$, and $\varepsilon_1,\ldots,\varepsilon_m \simiid \mathcal{N}(0,\sigma^2)$. Finally, we set $t_i = \langle \theta,x_i\rangle + \epsilon_i$ for $i=1,\ldots,n$ and $z_j = \langle \theta,y_j\rangle + \varepsilon_j$ for $j=1,\ldots,m$, which are i.i.d. samples from $\theta \sharp \mu * \mathcal{N}_\sigma$ and $\theta \sharp \nu * \mathcal{N}_\sigma$. The approximation rate of the empirical approximation of Smooth-SW is also $\mathcal{O}(n^{-1/2})$ as for SW~\citep{alaya2024gaussiansmoothed}.
\end{remark}

A nice property of smooth-SW is that it is a differentially private algorithm, meaning that small changes in the input data lead to only small and controlled changes in the output distribution. We first restate the definition of differential privacy (DP)~\citep{dwork2008differential}.
\begin{definition}
    Let $\varepsilon, \delta > 0$. Let $\mathcal{A} : \mathcal{D} \rightarrow \mathrm{Im}(\mathcal{A})$ be a randomized algorithm, where $\mathrm{Im}(\mathcal{A})$ is the image of $\mathcal{D}$ through $\mathcal{A}$. The algorithm $\mathcal{A}$ is \emph{$(\varepsilon, \delta)$-differentially private}, or $(\varepsilon, \delta)$-DP, if for all neighboring datasets $D, D' \in \mathcal{D}$ and for all sets of outputs $O \subseteq \mathrm{Im}(\mathcal{A})$, the following inequality holds:
\[
\mathbb{P}[\mathcal{A}(D) \in O] \leq e^{\varepsilon} \mathbb{P}[\mathcal{A}(D') \in O] + \delta,
\]
where the probability is taken over the randomness of $\mathcal{A}$.
\end{definition}

\begin{remark}[Differential privacy aspect of Smooth-SW]
\label{remark:differential_aspect_smoothsw} 
From~\citet[Proposition 2]{rakotomamonjy2021differentially}, let $\alpha > 1$ and $\delta \in [0, 1/2]$. Given a random direction projection matrix $\Theta=(\theta_1,\ldots,\theta_L) \in \mathbb{R}^{d \times L}$ ($\theta_1,\ldots,\theta_L \simiid \setU(\Sm^{d-1})$), then the Gaussian mechanism:
\[
\mathcal{M}(X) = X\Theta + E,
\]
where $E$ is a Gaussian matrix in $\mathbb{R}^{n \times k}$ with entries drawn from $\mathcal{N}(0, \sigma^2)$, is
\begin{align}
\left( \frac{\alpha w(k, \delta/2)}{2\sigma^2} + \frac{\log(2/\delta)}{\alpha - 1}, \delta \right)\text{-DP}.
\end{align}
This Gaussian mechanism on the random direction projections $\mathcal{M}(X)$ can be related to the definition of the empirical Smooth-SW.
\end{remark}

Due to this property, Smooth-SW has been utilized as a functional within the Wasserstein gradient flow framework in~\citet{sebag2025differentially}, enabling the design of generative models that preserve differential privacy through controlled sensitivity to data perturbations.

\chapter{Applications of Sliced Optimal Transport}
\label{chapter:applications}

Sliced Optimal Transport (SOT) was initially introduced and widely adopted in computer graphics and vision, where it addressed challenges such as texture mixing, color transfer, and shape interpolation, owing to its scalability and closed-form solutions in one dimension. In recent years, however, the versatility of SOT has attracted increasing attention in the broader machine learning and statistics communities due to its computational and statistical scalability. In particular, SOT metrics, such as the Sliced Wasserstein (SW) distance, have been used as risk functions for parameter estimation or as decision losses. This chapter reviews the diverse applications of SOT in machine learning (Section~\ref{sec:ML:chapter:applications}), statistics (Section~\ref{sec:stats:chapter:applications}), and computer graphics and vision (Section~\ref{sec:computer_graphic_and_vision:chapter:applications}). To the best of our knowledge, there is currently no comprehensive review focused on the applications of SOT in machine learning and statistics. For computer graphics and vision, the recent survey by~\citet{bonneel2023survey} provides a thorough overview of applications involving both OT and SOT. In Section~\ref{sec:computer_graphic_and_vision:chapter:applications}, we further contribute by adding new applications, such as neural rendering and blue noise sampling, and by updating references.

We note that our classification into machine learning, statistics, and computer graphics and vision is primarily based on publication venues, along with some subjective judgment. In many cases, the boundaries between these domains are fluid and overlapping.

\section{Machine Learning}
\label{sec:ML:chapter:applications}

In this section, we review the use of Sliced Optimal Transport (SOT) in various machine learning applications. Specifically, we discuss its role in clustering, classification, and regression with distributional predictors, as well as in generative models including autoencoders, generative adversarial networks (GANs), diffusion models, and flow matching. We also cover applications in domain adaptation, representation learning, attention mechanisms and transformers, backdoor attacks, federated learning, reinforcement learning, preference alignment for large language models, model evaluation, and model averaging.

\begin{longtable}[!t]{p{0.3\linewidth} | p{0.65\linewidth}}

\caption{Applications of sliced optimal transport in machine learning.} \label{tab:SOT_ML} \\
\toprule
\textbf{Applications} & \textbf{References} \\
\midrule
\endfirsthead

\toprule
\textbf{Applications} & \textbf{References} \\
\midrule
\endhead

\bottomrule
\endfoot

Clustering & Sliced Wasserstein means and sliced Wasserstein Gaussian mixture models~\citep{kolouri2018slicedgmm}, 
K-means with sliced Wasserstein embedding~\citep{kolouri2016sliced}. \\
\midrule
Classification & K-nearest neighbors~\citep{li2024hilbert,lu2024slosh}, kernel support vector machine~\citep{carriere2017sliced,kolouri2016sliced}. \\
\midrule
Regression & Kernel density regression~\citep{meunier2022distribution,bonet2023sliced}, kernel Gaussian process density regression~\citep{perez2024gaussian}. \\
\midrule
Generative modes & Sliced Wasserstein autoencoder~\citep{kolouri2018sliced,bonet2023spherical}, sliced Wasserstein generative adversarial network~\citep{deshpande2018generative,deshpande2019max,nguyen2021distributional,nguyen2022amortized,wu2019sliced}, sliced Wasserstein normalizing flow~\citep{coeurdoux2022sliced,bonet2022efficient}, diffusion model~\citep{nguyen2024random}, flow matching~\citep{chapel2025differentiable}, non-parametric generative flow~\citep{liutkus2019sliced,du2023nonparametric,dai2021sliced}. \\
\midrule
Domain adaptation & Unsupervised domain adaptation~\citep{lee2019sliced}. \\
\midrule
Representation learning & Distribution autoencoder~\citep{nguyen2021point,nguyen2023self,nguyen2024hierarchical}, model-free representation with sliced Wasserstein embedding~\citep{naderializadeh2021pooling,lu2024slosh,naderializadeh2025aggregating}. \\
\midrule
Attention mechanisms and Transformer & Sliced optimal transport plans for attention~\citep{shahbazi2025espformer}. \\
\midrule
Backdoor attack & Backdoor attack with latent modification~\citep{doan2021backdoor}. \\
\midrule
Federated learning & Data condensation~\citep{wang2024aggregation}. \\
\midrule
Reinforcement learning & Sliced Wasserstein trust-region policy Optimization~\citep{rowland2019orthogonal}, multi-dynamics reinforcement learning~\citep{cohen2021sliced}. \\
\midrule
Preference alignment for large language models & Preference alignment with stochastic dominance~\citep{melnyk2024distributional}. \\
\midrule
Model evaluation and model averaging & Evaluating Generative Models~\citep{karras2018progressive,piening2025slicing}, validating climate models~\citep{garrett2024validating}, and merging climate models~\citep{bonet2024slicing}.

\end{longtable}

\subsection{Clustering}
\label{subsec:clustering:sec:ML:chapter:applications}

Clustering is an unsupervised learning task in machine learning that involves grouping a set of data points into clusters such that points in the same cluster are more similar to each other than to those in other clusters~\citep{bishop2006pattern}. We now discuss some applications of SOT in clustering.

\begin{definition}[Sliced Wasserstein Means]
\label{def:SW_means}
Given data points $x_1,\ldots,x_n \in \Re^d$ ($d\geq 1$), the sliced Wasserstein mean problem is defined as~\citep{kolouri2018slicedgmm}:
\begin{align}
    \min_{z_1,\ldots,z_K \in \Re^d}  SW_p^p\left(\frac{1}{n}\sum_{i=1}^n \delta_{x_i}, \frac{1}{K}\sum_{k=1}^K \delta_{z_k}\right),
\end{align}
where $z_1,\ldots,z_K$ are centroids.
\end{definition}

It is worth noting that the name ``SW means" is somewhat abused since it is also used for the SW barycenter problem in the literature. Here, the SW mean problem can be seen as an MSWE problem (Section~\ref{sec:MSWE:chapter:varitational_SW}) and can be solved using a stochastic gradient descent algorithm (Section~\ref{sec:differentiatingSW:chapter:varitational_SW}). The cluster labels (mapping from $\{1,\ldots,n\}$ to $\{1,\ldots,K\}$) can be determined by using any assignment algorithms, e.g., nearest neighbors, optimal transport, sliced optimal transport, and so on.

\begin{definition}[Sliced Wasserstein Gaussian Mixture Models]
    \label{definition:SGMM}
    Given data points $x_1,\ldots,x_n \in \Re^d$ ($d\geq 1$), the sliced Wasserstein Gaussian mixture models problem is defined as~\citep{kolouri2018slicedgmm}:
\begin{align}
    \min_{\begin{array}{c}
         (w_1,\ldots,w_K ) \in \Delta_K, \\
          \mu_1,\ldots,\mu_K \in \Re^d, \\
          \Sigma_1,\ldots,\Sigma_K \in S_d^{++}(\Re)
    \end{array}}  SW_p^p\left(\frac{1}{n}\sum_{i=1}^n \delta_{x_i}, \sum_{k=1}^K w_k \mathcal{N}(\mu_k,\Sigma_k) \right),
\end{align}
where $\mathcal{N}(\mu,\Sigma)$ denotes a multivariate Gaussian with mean $\mu$ and covariance $\Sigma$, $\Delta_K$ denotes the $K$-simplex, and $S_d^{++}(\Re)$ denotes the set of symmetric positive definite matrices.
\end{definition}

The above problem can also be seen as an MSWE problem (Section~\ref{sec:MSWE:chapter:varitational_SW}) and can be solved using a stochastic gradient descent algorithm. Details of the optimization algorithm can be found in~\citep{kolouri2018slicedgmm}. After obtaining the estimated mixture of Gaussians, the cluster label of a data point can be obtained by selecting the component with the highest likelihood or by solving an assignment problem with the likelihood~\citep{ho2019probabilistic}. In addition to clustering, fitting a Gaussian mixture model can also be seen as a density estimation task. 

\begin{remark}[Sliced Wasserstein embedding K-means]
\label{remark:SW_kernel_Kmeans}
Given a set of measures $\mu_1,\ldots,\mu_n$, sliced Wasserstein embedding (Section~\ref{sec:SWEmbedding:chapter:varitational_SW}) can be used to embed them into a Hilbert space, and then into a finite Euclidean space via discretization. After that, we can use K-means~\citep{mcqueen1967some} algorithms to perform clustering on those measures. This approach is used for animal faces in~\citet{kolouri2016sliced}. With the Euclidean representation, we can further use any other clustering algorithms.
\end{remark}

\subsection{Classification}
\label{subsec:classification:sec:ML:chapter:applications}
Classification is a supervised learning task in machine learning where the goal is to predict a categorical label for a given input based on learned patterns from labeled training data. Each input sample is assigned to one of several predefined classes~\citep{bishop2006pattern}. We now discuss some applications of SOT in classification.

\begin{remark}[$K$-nearest neighbor]
    \label{remark:KNN_SW_classification}
     Since SW is a distance, it can be used to implement the $K$-nearest neighbors (KNN) algorithm~\citep{fix1985discriminatory}. In particular, given training data of $N$ measures with associated labels $\{(\mu_1,c_1),\ldots,(\mu_n,c_n)\}$ for $(\mu_i,c_i) \in \setP_p(\Re^d)\times C, \forall i$  ($C$ is the set of labels), the label of a new measure $\mu^*$ can be determined using the top $K$ measures in the training set that have the smallest SW distance to $\mu^*$ ($SW_p^p(\mu^\star,\cdot)$) using a voting algorithm (majority vote, weighted vote, and so on). This classification method is used for point-cloud classification and document classification~\citep{li2024hilbert,lu2024slosh}.
\end{remark}
 
\begin{remark}[Kernel Support Vector Machine]
\label{remark:SVM_SW}
    Sliced Wasserstein kernels (Section~\ref{sec:SWKernel:chapter:varitational_SW}) can be used for support vector machines (SVM) \citep{cortes1995support,hofmann2008kernel}. This approach is used to classify persistence diagrams~\citep{carriere2017sliced} and animal faces~\citep{kolouri2016sliced}.
\end{remark}

\subsection{Regression}
\label{subsec:regression:sec:ML:chapter:applications}

In machine learning, regression refers to modeling the relationship between a set of predictors (features) and a response. The goal is to learn a function that can accurately predict the response variable given new values of the predictors. We now discuss how SOT can be used for regression.

\begin{remark}[Kernel Density Regression]
    \label{remark:density_regression}
    Given training data of $N$ measures with associated responses $\{(\mu_1,y_1),\ldots,(\mu_n,y_n)\}$ for $(\mu_i,y_i) \in \setP_p(\Re^d)\times \Re, \forall i$, authors in~\citet{meunier2022distribution} propose a kernel distribution (density) regression framework:
    \begin{align}
        f_\lambda(\mu) = (y_1,\ldots,y_n)(\mathbf{K} + \lambda n I)^{-1} k_\mu,
    \end{align}
    where $\lambda>0$ is the ridge regularization coefficient, $\mathbf{K} \in \Re_+^{n\times n}$ with $K_{i,j} = k(\mu_i,\mu_j)$ is the SW kernel between $\mu_i$ and $\mu_j$, e.g., $k(\mu_i,\mu_j)=\exp(-\gamma SW_2^2(\mu_i,\mu_j))$, and $k_\mu=(k(\mu,\mu_1),\ldots,k(\mu,\mu_n))^\top$.
\end{remark}

Authors in~\citet{bonet2023sliced} utilize SW between measures over symmetric positive definite matrices to perform regression for M/EEG signals.

\begin{remark}[Kernel Gaussian Process Density Regression]
    \label{remark:kernel_GPregression}
    With the same setup for density regression, authors in~\citet{perez2024gaussian} adapt Gaussian process (GP) regression~\citep{williams1995gaussian} by using the SW kernel to construct the covariance function in GP. In particular, we assume that our data follows the model $y_i = f(\mu_i) + \epsilon_i$, where $\epsilon_i \sim \mathcal{N}(0,\lambda^2)$. Let $\mathbf{y}=(y_1,\ldots,y_n)$ and $\mathbf{f}_*= (f(\mu_1^*),\ldots,f(\mu_m^*))$ with $\mu_1^*,\ldots,\mu_m^*$ being the set of test data; we have the following prior:
    \begin{align}
        \begin{bmatrix}
\mathbf{y} \\
\mathbf{f}_*
\end{bmatrix}
\sim \mathcal{N} \left(
\mathbf{0},
\begin{bmatrix}
\mathbf{K} + \eta^2 \mathbf{I} & \mathbf{K}_*^\top \\
\mathbf{K}_* & \mathbf{K}_{**}
\end{bmatrix}
\right),
    \end{align}
    where $\mathbf{K}$, $\mathbf{K}_{**}$, and $\mathbf{K}_*$ are the train, test, and test/train kernel matrices, respectively. The posterior can be written as:
    \begin{align}
        \mathbf{f}_* \mid \mu_1,\ldots,\mu_n,\mu_1^*,\ldots,\mu_m^* \sim \mathcal{N}(\mathbf{m}, \Sigma),
    \end{align}
    where $\mathbf{m}= \mathbf{K}_*(\mathbf{K}+\lambda^2 I)^{-1} \mathbf{y}$ and $\Sigma= \mathbf{K}_{**} - \mathbf{K}_* (\mathbf{K}+\lambda^2 I)^{-1} \mathbf{K}_*^\top$.
\end{remark}

In the original paper, authors in~\citet{perez2024gaussian} combine the sliced Wasserstein kernel with the Weisfeiler-Lehman graph kernel~\citep{shervashidze2011weisfeiler} to obtain the sliced Weisfeiler-Lehman graph kernel. With the new kernel, the authors can perform graph regression with the discussed framework.

\subsection{Generative Models}
\label{subsec:generative_model:sec:ML:chapter:applications}

Given a set of samples $x_1,\ldots,x_n \sim \mu \in \setP(\Re^d)$, the goal of generative modeling is to obtain a distribution $\nu$ such that we can sample from it and generate samples that are similar to $x_1,\ldots,x_n$ (in some sense). We now discuss how SOT is adapted to generative modeling.

\begin{remark}[Sliced Wasserstein Autoencoder]
    \label{remark:SWAE}
    Sliced Wasserstein autoencoder~\citep{kolouri2018sliced} (SWAE) is a variant of Wasserstein autoencoder~\citep{tolstikhin2018wasserstein} (WAE) which uses SW as the latent regularization. Let $f_\phi:\Re^h \to \Re^d$ ($\phi \in \Phi$, $h>0$ and often smaller than $d$), $g_\psi:\Re^d \to \Re^h$ ($\psi \in \Psi$), and $\nu_0 \in \setP_p(\Re^d)$; SWAE solves the following optimization problem:
    \begin{align}
        \min_{\phi \in \Phi, \psi \in \Psi} \mathbb{E}_{x \sim \mu}[c(x,f_\phi(g_\psi(x)))] + \lambda SW_p^p(\nu_0,g_\psi \sharp \mu),
    \end{align}
    where $\lambda >0$ and $c:\Re^d\times \Re^d \to \Re_+$ is a ground metric. Since we only observe $\mu$ through its samples, a plug-in estimator is used to approximate the expectation and SW. After estimating $\phi$, we can use $\nu = g_\phi \sharp \nu_0$ as the generative model. While we do not have a tractable form of $\nu$, we can still sample from it since it is a push-forward probability measure. 
\end{remark}

SWAE can also be seen as a dimension reduction technique since we can use $g_\psi$ to reduce the dimension of data. Spherical SWAE~\citep{bonet2023spherical} (replacing SW by spherical SW for spherical latent space) has also been adapted to do topic modeling in~\citet{adhya2025s2wtm}, where the decoder is represented by the topic-word matrix.

\begin{remark}[Sliced Wasserstein Generative Adversarial Network]
\label{remark:SWGAN}
    Like SWAE, sliced Wasserstein Generative Adversarial Network~\citep{deshpande2018generative} (SWGAN) also aims to estimate an implicit generative model $\nu = f_\phi \sharp \nu_0$. SWGAN performs minimum expected SW estimation (Section~\ref{sec:MSWE:chapter:varitational_SW}):
    \begin{align}
        \min_{\phi \in \Phi} \mathbb{E}_{(z_1,\ldots,z_m) \sim \nu_0^{\otimes m} } \left[SW_p^p\left(\frac{1}{n}\sum_{i=1}^n\delta_{x_i}, \frac{1}{m}\sum_{i=1}^m \delta_{f_\phi(z_i)}\right)\right].
    \end{align}
    When the number of data points ($n$) is too large, random sampling can be used:
    \begin{align}
        \min_{\phi \in \Phi} \mathbb{E}_{(z_1,\ldots,z_m) \sim \nu_0^{\otimes m}, (x_1,\ldots,x_m)\sim \mu^{\otimes m} } \left[SW_p^p\left(\frac{1}{m}\sum_{i=1}^m\delta_{x_i}, \frac{1}{m}\sum_{i=1}^m \delta_{f_\phi(z_i)}\right)\right].
    \end{align}
    This approach can be seen as an application of mini-batch optimal transport~\citep{fatras2020learning}. In practice, when having high-dimensional data that lives in unknown non-Euclidean space, a transformation function is needed to map the data to a space with known structure (Euclidean space), e.g., $g_\psi:\Re^d \to \Re^h$:
    \begin{align}
        &\min_{\phi \in \Phi} \mathbb{E}_{(z_1,\ldots,z_m) \sim \nu_0^{\otimes m}, (x_1,\ldots,x_m)\sim \mu^{\otimes m} }   \nonumber \\&\left[SW_p^p\left(\frac{1}{m}\sum_{i=1}^m\delta_{g_\psi(x_i)}, \frac{1}{m}\sum_{i=1}^m \delta_{g_\psi(f_\phi(z_i))}\right)\right],
    \end{align}
    where $g_\psi$ is estimated independently using other objectives such as adversarial training~\citep{deshpande2018generative}.
\end{remark}

Other variants of SW can also be used~\citep{deshpande2019max,nguyen2021distributional,nguyen2022amortized}. Authors in~\citet{wu2019sliced} propose to use the dual form of sliced Wasserstein instead of the primal form with stochastic approximation. Authors in~\citet{lezama2021run} propose an efficient implementation to increase the mini-batch size $m$. Normalizing flow~\citep{papamakarios2021normalizing} models (invertible $f_\phi$) can also be trained using similar estimators~\citep{coeurdoux2022sliced,bonet2022efficient}.

\begin{remark}[Implicit Diffusion Model]
    \label{remark:implicit_diffusion_model}
    Similar to SWGAN, SW is used for implicit distribution matching in diffusion models in~\citet{nguyen2024random}. Denoising diffusion models~\citep{ho2020denoising,sohl2015deep} define a generative process $x_0\sim q(x_0)$ with $T>0$ steps:
\begin{align*}
    q(x_{1:T}|x_0) = \prod_{t=1}^T q(x_t|x_{t-1}),
\end{align*}
where $q(x_t|x_{t-1}) = \mathcal{N}(x_t;\sqrt{1-\beta_t}x_{t-1},\beta_t I)$ with a known variance schedule $\beta_t$. We aim to estimate $\phi$ of a reverse denoising process defined by:
\begin{align*}
    p_\phi(x_{0:T}) = p(x_T) \prod_{t=1}^T p_\phi(x_{t-1}|x_t),
\end{align*}
where $p_\phi(x_{t-1}|x_t) = \mathcal{N}(x_{t-1};f_\phi(x_t,t),\sigma_t^2 I)$. Authors in~\citet{nguyen2024sliced} carry out the following optimization:
\begin{align}
    \min_{\phi} \sum_{t=1}^T \mathbb{E}_{q(x_t)} [D(q(x_{t-1}|x_t) , p_\phi(x_{t-1}|x_t))],
\end{align}
with 
\begin{align}
    D(\mu,\nu) = 2 \mathbb{E}[SW_p(\mu_{X},\nu_{Y}) ] - \mathbb{E}[SW_p(\mu_{X},\mu_{X'})] - \mathbb{E}[SW_p(\nu_{Y},\nu_{Y'})],
\end{align}
where $X,X' \stackrel{i.i.d.}{\sim} \mu^{\otimes m}$ and $Y,Y' \stackrel{i.i.d.}{\sim} \nu^{\otimes m}$, and $\mu_X$ denotes the empirical measure over $X$. $D$ is the mini-batch energy distance with SW as the energy function~\citep{salimans2018improving}. The authors also propose to use an augmented transformation to deal with high-dimensional data (images). After training, we can generate samples from $p_\phi(x_{0:T})$ using ancestral sampling. 
\end{remark}

\begin{remark}[Flow Matching]
    \label{remark:conditional_flow_matching}
    Conditional flow matching~\citep{lipman2023flow,pooladian2023multisample,tong2024improving} utilizes a coupling to train a vector field that captures the true vector field of (smoothed) data:
    \begin{align}
    \min_\phi \mathbb{E}_{t\sim \setU([0,1]), (x_0,x_1)\sim \pi (x_0 ,x_1)} \|v_t(x_t; \phi) - u_t(x_t|x_1)\|_2^2,
    \end{align}
    where $\pi(x_0,x_1)$ is a coupling between $q_0(x_0)$ (a chosen distribution) and $q_{1}(x_1)$ (data distribution), $v_t(\cdot;\phi)$ is a function with parameter $\phi$ and indexed by $t$, and $u_t(x_t|x_1)$ is a known vector field from data indexed by $t$. Authors in~\citet{chapel2025differentiable} propose to use sliced optimal transport plans between $q_0(x_0)$ and $q_1(x_1)$ (Section~\ref{sec:map:chapter:advances}) as $\pi(x_0,x_1)$. After training, we can generate samples by integrating through a stochastic differential equation that is created by $v_t(x_t; \phi)$.
\end{remark}

\begin{remark}[Non-parametric Generative Modeling using Gradient Flows]
\label{remark:generative_modeling_gradient_flow}
Authors in~\citet{liutkus2019sliced} propose to use Wasserstein gradient flow with SW as the functional (Section~\ref{sec:SWgradientflow:chapter:varitational_SW}) to drive a particle system towards the data distribution. Authors in~\citet{du2023nonparametric} extend the framework by using convolution slicing for images. Authors in~\citet{dai2021sliced} propose a similar approach to~\citet{liutkus2019sliced} using one-dimensional optimal transport maps.
\end{remark}

\subsection{Domain Adaptation}
\label{subsec:domain_adaptation:sec:ML:chapter:applications}

Domain adaptation aims to transfer knowledge from a source domain to a target domain, where the data distributions differ across domains. The goal is to learn domain-invariant representations that perform well on the target task despite the absence of target labels. We now review how SOT can be used for domain adaptation.

\begin{remark}[Unsupervised Domain Adaptation]
    \label{remark:UDA_SW}
    In unsupervised domain adaptation, we are given a labeled source domain $X_s=(x_1,\ldots,x_n)$, (features) and $Y_s=(y_1,\ldots,y_n)$ (labels) ($n>0$), and a target domain $X_t = (x_1',\ldots,x_m')$. We are interested in training a classifier for $X_t$ based on the information of $X_s$ and $Y_s$. Authors in~\citet{lee2019sliced} propose a three-step framework.

    \begin{enumerate}
        \item \textbf{Training on source domain.} This step involves training a feature extractor $G_\phi$ and two classifiers $C_{1,\psi}$ and $C_{2,\varphi}$ using classification loss $\mathcal{L}_s$, such as cross-entropy, on $X_s$ and $Y_s$:
        \begin{align}
            \min_{\phi,\psi,\varphi} \mathbb{E}_{(x,y) \sim \mu_s(x,y)}[L_s ( C_{1,\psi}(G_\phi(x)), y) + L_s ( C_{2,\varphi}(G_\phi(x)), y)],
        \end{align}
        where $\mu_s(x,y)$ is the empirical joint distribution on features $X_s$ and labels $Y_s$.

        \item \textbf{Maximizing the discrepancy between two classifiers.} In this step, we fix $G_\phi$ and maximize the SW distance between
the outputs of the two classifiers on the target set $X_t$:
\begin{align}
            \min_{\psi,\varphi} &\mathbb{E}_{(x,y) \sim \mu_s(x,y)}[L_s ( C_{1,\psi}(G_\phi(x)), y) + L_s ( C_{2,\varphi}(G_\phi(x)), y)]  \nonumber \\
            &- \mathbb{E}_{x \sim \mu_t(x)}[SW_1(C_{1,\psi}(G_\phi(x)),C_{2,\varphi}(G_\phi(x)))],
        \end{align}
        where $\mu_t(x)$ is the empirical distribution over $X_t$.

        \item \textbf{Updating the feature extractor.} We fix $C_{1,\psi}$ and $C_{2,\varphi}$, then update $G_\phi$ to minimize the SW distance between the
outputs of the two classifiers on the target set:
\begin{align}
    \min_\phi \mathbb{E}_{x \sim \mu_t(x)} [SW_1(C_{1,\psi}(G_\phi(x)),C_{2,\varphi}(G_\phi(x)))].
\end{align}
    \end{enumerate}
\end{remark}

\subsection{Representation Learning}
\label{subsec:representation_learning:sec:ML:chapter:applications}

Representation learning aims to automatically discover useful features or representations from raw data that capture the underlying structure or semantics. These learned representations can improve performance on downstream tasks such as classification, clustering, or regression. We now discuss how SOT can be used for extracting representations.

\begin{remark}[Distribution Autoencoder]
    Given a dataset $\mu_1,\ldots,\mu_n \in \setP_p(\Re^d)$ ($p>0$), we want to estimate two functions $f_\phi:\setP_p(\Re^d) \to \Re^h$ and $g_\psi:\Re^h \to \setP_p(\Re^d)$ where $h$ is often chosen to be relatively small. The function $f_\phi$ is called the encoder and $g_\psi$ the decoder. These two functions help us to ``compress'' $\mu_1,\ldots,\mu_n$ to a smaller representation in $\Re^h$ such that the reconstruction (distortion) cost is as small as possible. SW distance can be used as the distortion loss:
    \begin{align}
        \min_{\phi,\psi} \mathbb{E}_{\mu \sim p(\mu)} [SW_p^p(\mu,g_\psi(f_\phi(\mu)))],
    \end{align}
    where $p(\mu)$ is the meta distribution of $\mu_1,\ldots,\mu_n$. This is used for 3D point-cloud autoencoders~\citep{nguyen2021point,nguyen2023self} and 3D mesh autoencoders~\citep{nguyen2024hierarchical}. With an entropy model on the latent space $\Re^h$, a data compression model can be obtained~\citep{yang2023introduction}.
\end{remark}

\begin{remark}[Model-free Representation with Sliced Wasserstein Embedding]
\label{remark:SWE_application}
    Sliced Wasserstein embedding (Section~\ref{sec:SWEmbedding:chapter:varitational_SW}) can also serve as a representation without using any model. It has been used to extract embeddings for point clouds, graphs, and patches of images to train other deep learning models on such data in~\citet{naderializadeh2021pooling}, create hash codes for sets in~\citet{lu2024slosh}, and aggregate residue-level protein language model embeddings in~\citet{naderializadeh2025aggregating}.
\end{remark}

\subsection{Attention Mechanisms and Transformer}
\label{subsec:attention:sec:ML:chapter:applications}

Attention is a mechanism that allows models to focus on the most relevant parts of the input when making predictions, dynamically weighting different elements based on their importance. The Transformer architecture leverages multi-head self-attention and feedforward layers to process sequences in parallel, achieving state-of-the-art performance in tasks like language modeling and machine translation. SOT plans are recently adapted into developing attention mechanisms.

\begin{remark}[Sliced Optimal Transport Plans for Attention]
    \label{remark:SOT_attention}
    Self-attention plays a central role in the recent success of deep learning models such as Transformer~\citep{vaswani2017attention}. In one attention layer, let $ W_Q, W_K \in \mathbb{R}^{m \times d} $, $ W_V \in \mathbb{R}^{d \times d} $ denote the query, key, and value matrices, respectively. Then, for a sequence $ X=(x_1, x_2, \dots, x_N) $, where $ x_i \in \mathbb{R}^d$ for all $i$, the attention output with $Q=W_QX$, $K=W_KX$, and $V=W_VX$ can be written as:
\begin{align}
    \text{Attention}(Q, K, V) = \mathrm{softmax}\left( \frac{Q K^\top}{\sqrt{d}} \right) V,
\end{align}
where $\mathrm{softmax}$ is applied row-wise. Authors in~\citet{shahbazi2025espformer} propose a new type of attention:
\begin{align}
    \text{ESP-Attention}(Q, K, V) &= V G_{Q,K}, 
\end{align}
where 
\begin{align*}
    [C]_{ij} &= \| Q_{i} - K_{j} \|^2, \\
    A_l &= \text{SoftSort}_t(Q_{l}), \quad B_l = \text{SoftSort}_t(K_{l}), \\
    U_l &= \frac{1}{N} A^{\top}_l B_l, \quad 
    D_l = \sum_{i,j} C_{ij} U_{lij}, \\
    \sigma &= \text{softmax}\left( \frac{D}{\tau} \right), \quad G_{Q,K} = \sum_{l=1}^m \sigma_l U_l,
\end{align*}
for $l=1,\ldots,m$ and temperature $\tau>0$. Here, $U_l$ can be seen as the lifted SOT plan with basis vectors as  projecting directions (Section~\ref{sec:map:chapter:advances}) between the empirical distribution over $Q$ and the empirical distribution over keys $K$.
\end{remark}

\subsection{Backdoor Attack}
\label{subsec:backdoor_attack:sec:ML:chapter:applications}

A backdoor attack involves embedding hidden malicious behavior into a machine learning model during training, often by poisoning the training data with specific triggers. At inference time, the model behaves normally on clean data but produces attacker-controlled outputs when the trigger is present. Designing such attacks can lead to better defense mechanisms. We now review how SOT can contribute to this application.

\begin{remark}[Backdoor Attack with Latent Modification]
    \label{remark:backdoor_latent_SW}
    A threat model is proposed in~\citet{doan2021backdoor} using distributional SW (Remark~\ref{remark:optimization_based_slicing}) as a regularization for the latent trigger function. The trigger
    function acts as a conditional noise generator $T_\phi(x) = x + g_\phi(x)$ mapping $\Re^d \to \Re^d$. Let $f_\psi:\Re^d \to \mathcal{Z}$ be the target function, e.g., classification or regression, with corresponding training loss $\mathcal{L}$. The trigger function is trained via the optimization:
    \begin{align}
        &\min_{\psi} \mathbb{E}_{(x,y)\sim p(x,y)}\big[\alpha \mathcal{L}(f_\psi(x), y) + \beta \mathcal{L}(f_\psi(T_{\phi^*}(x)), \eta(y)) \big] \\
        &\text{s.t.} \quad \phi^* = \arg\min_{\phi} \mathbb{E}_{(x,y)\sim p(x,y)}\big[ \mathcal{L}(f_\psi(T_{\phi}(x)), \eta(y))  \nonumber\\ &\quad \quad \quad + DSW_2^2(f_\psi' \sharp p(x), f_\psi' \sharp T_{\phi} \sharp p(x)) \big], \nonumber
    \end{align}
    where $\alpha, \beta > 0$, $p(x,y)$ is the joint data distribution on features and labels, $p(x)$ is the marginal distribution on features, and $f'_\psi$ denotes the output of an intermediate hidden layer of $f_\psi$ (e.g., a neural network).
\end{remark}

\subsection{Federated Learning}
\label{subsec:federated_learning:sec:ML:chapter:applications}

Federated learning is a decentralized machine learning framework where $K$ clients jointly learn a global model $f_\phi$ without uploading local raw data. The local datasets are $\mathcal{D} = \{\mathcal{D}_1, \ldots, \mathcal{D}_K\}$, where $\mathcal{D}_k = \{(x_{k,i}, y_{k,i})\}_{i=1}^{N_k}$ is the data owned by client $k$.

\begin{remark}[Data Condensation]
    \label{remark:SW_federated_learning_data_condensation}
    Authors in~\citet{wang2024aggregation} propose an aggregation-free method for federated learning using the SW distance as an objective for data condensation. Each client $k$ learns condensed data $\mathcal{S}_k$ by minimizing the local objective:
    \begin{align}
        \mathcal{L}_{\mathrm{loc}}(\mathcal{S}_k, \mathcal{D}_k) = \mathcal{L}_{\mathrm{DM}}(\mathcal{S}_k, \mathcal{D}_k) + \lambda_{\mathrm{loc}} \sum_{c=0}^{C-1} SW_p(P_{\mathbf{u}_{k,c}}, P_{\mathbf{v}_c}),
    \end{align}
    where 
    \[
    \mathbf{v}_{k,c} = \frac{1}{N_{k,c}} \sum_{j=1}^{N_{k,c}} f_\phi(x_{k,c}^{(j)}), \quad \mathbf{u}_{k,c} = \frac{1}{M_{k,c}} \sum_{j=1}^{M_{k,c}} f_\phi(\tilde{x}_{k,c}^{(j)}),
    \]
    with $x_{k,c}^{(j)}$ the $j$-th real data sample of class $c$ on client $k$, and $\tilde{x}_{k,c}^{(j)}$ the $j$-th condensed data sample of class $c$ on client $k$. Here, $\lambda_{\mathrm{loc}} > 0$, and $\mathcal{L}_{\mathrm{DM}}$ is the matching loss:
    \[
    \mathcal{L}_{\mathrm{DM}}(\mathcal{S}_k, \mathcal{D}_k) = \sum_{c=0}^{C-1} \left\| \mathbf{u}_{k,c}^{\mathrm{real}} - \mathbf{u}_{k,c}^{\mathrm{syn}} \right\|_2^2,
    \]
    where $C$ is the number of classes, and $\mathbf{u}_{k,c}^{\mathrm{real}}$ and $\mathbf{u}_{k,c}^{\mathrm{syn}}$ represent class-wise feature means of real and condensed data using the model. The second term employs the SW distance to match the empirical distribution over all local logits $\mathbf{u}_{k,c}$, denoted $P_{\mathbf{u}_{k,c}}$, with the empirical distribution over global average logits $\mathbf{v}_c$, denoted $P_{\mathbf{v}_c}$, across clients.

    Clients send $\mathcal{S}_k$ and soft labels $\mathcal{R}_k$ to the server, which trains the global model $\phi$ by minimizing:
    \[
    \mathcal{L}_{\mathrm{glob}}(\phi, \mathcal{S}) = \mathcal{L}_{\mathrm{CE}}(\phi, \mathcal{S}) + \lambda_{\mathrm{glob}} \mathcal{L}_{\mathrm{LGKM}}(\phi, \mathcal{S}),
    \]
    where $\mathcal{S}$ is the combination of all condensed data and the local-global knowledge matching (LGKM) loss is defined as:
    \[
    \mathcal{L}_{\mathrm{LGKM}} = \frac{1}{2} \big( D_{\mathrm{KL}}(\mathcal{R}, \mathcal{T}) + D_{\mathrm{KL}}(\mathcal{T}, \mathcal{R}) \big),
    \]
    with $\mathcal{R}$ and $\mathcal{T}$ denoting soft labels derived from client and server data, respectively.
\end{remark}

\subsection{Reinforcement Learning}
\label{subsec:RL:sec:ML:chapter:applications}

SOT distances have recently found application in reinforcement learning (RL) as stable and efficient trust-region discrepancy measures between successive policies. In standard RL, trust-region methods constrain the policy update to prevent drastic changes, traditionally using the Kullback--Leibler (KL) divergence as the measure of policy difference. However, Wasserstein-based metrics offer a notion of distance that directly reflects the geometry of the policy distributions, and SOT distances provides a computationally tractable alternative to the full Wasserstein distance, particularly in high-dimensional action spaces

\begin{remark}[Sliced Wasserstein trust-region policy optimization]
    In reinforcement learning, an agent interacts with an environment in discrete time: at step \(t\), it observes a state \(s_t\), takes an action \(a_t\), receives a reward \(r_t\), and transitions to the next state \(s_{t+1}\). The objective is to learn a parameterized policy \(\pi_\theta : s_t \mapsto a_t\) that maximizes the expected discounted return
    \begin{align}
        J(\pi_\theta) = \mathbb{E}_{\pi_\theta} \left[ \sum_{t=0}^\infty \gamma^t r_t \right],
    \end{align}
    where \(\gamma \in (0,1)\) is the discount factor. Trust-region policy optimization methods improve stability by constraining the change between successive policies using a discrepancy measure \(D(\pi_{\theta_{\mathrm{old}}}, \pi_{\theta_{\mathrm{new}}}) \leq \varepsilon\).
    In a penalty-based formulation with SW~\citep{rowland2019orthogonal} (PW, Definition~\ref{def:PW}, can also be used), the update rule becomes
    \begin{align}
        \theta_{\mathrm{new}} \leftarrow \theta_{\mathrm{old}} + \alpha \nabla_{\theta_{\mathrm{old}}} \left( J(\pi_{\theta_{\mathrm{old}}}) - \lambda \, SW_1(\pi_{\theta_{\mathrm{old}}}, \pi_{\theta_{\mathrm{new}}}) \right),
    \end{align}
    where \(\alpha > 0\) is the learning rate and \(\lambda > 0\) is a penalty coefficient. This formulation maintains the stability benefits of trust regions.
\end{remark}

Multi-dynamics reinforcement learning (RL) addresses environments with multiple, possibly shifting, underlying dynamics or transition models. The goal is to train agents that can adapt to or generalize across different dynamics, enabling robust decision-making in varied or uncertain settings. Sliced multi-marginal optimal transport can be used to improve this application.

\begin{remark}[Multi-dynamics Reinforcement Learning]
    In multi-dynamics reinforcement learning, the sliced multi-marginal Wasserstein (SMW, Section~\ref{sec:SMOT:chapter:extension}) distance is used to share structure across \(P\) agents~\citep{cohen2021sliced}, each solving the same task but with different dynamics. Each agent \(p\) operates in a Markov Decision Process \((\mathcal{S}, \mathcal{A}, T_p, r^{\mathrm{env}}_p)\) and collects a trajectory 
    \[
    \mu_p = \frac{1}{T} \sum_{t=1}^T \delta_{x^{(p)}_t},
    \]
    interpreted as an empirical measure over states. To allow agents to learn even when their reward \(r^{\mathrm{env}}_p\) is missing, the reward is augmented with a shared structure term:
    \begin{align}
        R_p(x^{(p)}_t) = r^{\mathrm{env}}_p(x^{(p)}_t) + \gamma\, r^{\mathrm{mul}}(x^{(p)}_t, X),
    \end{align}
    where \(r^{\mathrm{mul}}\) is derived from the SMW distance over all agent trajectories \(X = \{x^{(p)}_t\}_{p,t}\), and \(\gamma > 0\) is a regularization parameter. The training objective is to maximize
    \begin{align}
        \mathbb{E}_{\pi_1, \dots, \pi_P} \left[ \sum_{p=1}^P \sum_{t=1}^T R_p(x^{(p)}_t) \right],
    \end{align}
    which implicitly includes a regularization term \(\mathrm{SMW}^2(\mu_1, \dots, \mu_P)\) aligning agents’ state distributions.
\end{remark}

\subsection{Preference Alignment for Large Language Models}
\label{subsec:preference_alignment:sec:ML:chapter:applications}

Preference alignment aims to ensure that machine learning models, especially large language models (LLMs), act in accordance with human values and intentions. This is often achieved through techniques like reinforcement learning with human feedback (RLHF), where models are fine-tuned based on human preferences. The goal is to learn a policy \(\pi_\phi\), parameterized by \(\phi\) in a bounded space, given a reference LLM policy \(\pi_{\mathrm{ref}}\).

\begin{remark}[Preference Alignment with Stochastic Dominance]
    \label{remark:preference_alignment}
    Authors in~\citet{melnyk2024distributional} propose performing alignment via stochastic dominance. Given two random variables \(Z_1\) and \(Z_2\), we say \(Z_1\) dominates \(Z_2\) (\(Z_1 \succcurlyeq Z_2\)) if
    \[
    F_{Z_1}^{-1}(q) \geq F_{Z_2}^{-1}(q) \quad \forall q \in [0,1].
    \]
    There are two scenarios for preference alignment: unpaired preference and paired preference.

    \textbf{Unpaired preference.} Access is available to samples of positive prompt-response pairs \((X_+, Y_+)\) and negative prompt-response pairs \((X_-, Y_-)\). The goal is to obtain:
    \begin{align}
        \log \left(\frac{\pi_\phi(Y_+ \mid X_+)}{\pi_{\mathrm{ref}}(Y_+ \mid X_+)}\right) \succcurlyeq \log \left(\frac{\pi_\phi(Y_- \mid X_-)}{\pi_{\mathrm{ref}}(Y_- \mid X_-)}\right).
    \end{align}

    \textbf{Paired preference.} Access is available to triplets of prompt, positive response, and negative response \((X, Y_+, Y_-)\). The goal is to obtain:
    \begin{align}
        \log \left(\frac{\pi_\phi(Y_+ \mid X)}{\pi_\phi(Y_- \mid X)}\right) \succcurlyeq \log \left(\frac{\pi_{\mathrm{ref}}(Y_+ \mid X)}{\pi_{\mathrm{ref}}(Y_- \mid X)}\right).
    \end{align}

    \textbf{One-dimensional OT as a relaxation of stochastic dominance.} Both settings reduce to a dominance problem \(U_\phi \succcurlyeq V_\phi\). The authors relax this to a one-dimensional OT problem:
    \begin{align}
        \min_{\phi \in \Phi} \int_{\mathbb{R}} h\big(F_{U_\phi}^{-1}(q) - F_{V_\phi}^{-1}(q)\big) \, \mathrm{d}q,
    \end{align}
    where \(h\) can be 0/1 loss, \(\beta\)-squared hinge loss, \(\beta\)-logistic loss, or \(\beta\)-least squares loss. Minimizing over \(\phi\) can be seen as finding the best implicit projection of data distributions minimizing the transportation cost.
\end{remark}
\subsection{Model Evaluation and Model Averaging}
\label{subsec:model_evaluation:sec:ML:chapter:applications}

In model evaluation, it is often necessary to quantify how closely a model-generated distribution matches the target distribution. The sliced Wasserstein (SW) distance is well-suited for this task due to its computational efficiency, enabling fast evaluation cycles.

\begin{remark}[Evaluating Generative Models]
    \label{remark:evaluating_generative_models}
    SW is used as an automated evaluation metric to compare generated images with real images across multiple scales in~\citet{karras2018progressive}. The authors construct a Laplacian pyramid for both generated and real images, extracting local patches at each resolution level, from coarse (16×16) to full resolution, to capture spatial structure at different frequencies. After normalizing these patches by color statistics, they compute the sliced Wasserstein distance (SWD) between patch distributions at each pyramid level. 
    
    Additionally, authors in~\citet{piening2025slicing} propose a hierarchical SW variant between Gaussian mixtures to evaluate the performance of image generative models. In particular, they use the distance between fitted Gaussian mixtures on features extracted from real and generated images (using the Inception network) as the evaluation score.
\end{remark}

\begin{remark}[Validating Climate Models]
    \label{remark:validating_climate_model}
    Authors in~\citet{garrett2024validating} utilize convolution sliced Wasserstein (Definition~\ref{def:convolution_projection_functions}) to compare the spatiotemporal distributions of climate variables between model outputs and reference datasets. This enables assessing how well climate models capture local patterns while accounting for spatial structure and variability.
\end{remark}

\begin{remark}[Merging Climate Models]
    \label{remark:merging_climate_models}
    Authors in~\citet{bonet2024slicing} apply unbalanced sliced optimal transport (USOT) to compute barycenters of climate model data. They employ a mirror-descent algorithm to efficiently compute USOT barycenters on large spatial grids. This approach handles differences in total mass across models without normalization, yielding a physically meaningful single-mode barycenter.
\end{remark}

\section{Statistics}
\label{sec:stats:chapter:applications}
In this section, we review applications of SOT in statistics. In particular, we discuss two-sample testing, feature screening, density regression, posterior sampling, variational inference, empirical Bayes, approximate Bayesian computation, and summarizing random partitions.

\begin{table}[!t]
    \centering
    \begin{tabular}{p{0.3\linewidth} | p{0.65\linewidth}}
        \toprule
        \textbf{Applications} & \textbf{References} \\
        \midrule
        Two-sample Testing & Permutation two-sample test~\citep{wang2022two}, bootstrap two-sample test~\citep{hu2025two}. \\
        \midrule
        Feature screening & Model-free feature screening~\citep{li2023scalable}. \\
        \midrule
        Density regression & Sliced Wasserstein regression~\citep{chen2023sliced}, Bayesian density–density regression~\citep{nguyen2025bayesian}. \\
        \midrule
        Variational inference & Sliced Wasserstein variational inference~\citep{yi2023sliced}, sliced Wasserstein joint contrastive inference~\citep{nguyen2021distributional}. \\
        \midrule
        Posterior sampling & Posterior sampling with gradient flow~\citep{bonet2022efficient}. \\
        \midrule
        Empirical Bayes & Empirical Bayes for Bayesian autoencoders~\citep{tran2021model}. \\
        \midrule
        Approximate Bayesian computation & Sliced Wasserstein approximate Bayesian computation~\citep{nadjahi2020approximate}. \\
        \midrule
        Summarizing Random Partition & Summarizing random partition via mixing measures~\citep{nguyen2024summarizing}. \\
        \bottomrule
    \end{tabular}
    \caption{Applications of sliced optimal transport in statistics.}
    \label{tab:SOT_stats}
\end{table}

\subsection{Two-sample Testing}
\label{subsec:two_sample_test:sec:stats:chapter:applications}

The two-sample hypothesis test is designed to answer the question of whether two samples come from the same distribution. Given  samples $X_1,\ldots,X_n \simiid \mu$ and $Y_1,\ldots,Y_m \simiid \nu$, we test the following hypothesis:
\begin{align}
    H_0: \mu=\nu \quad \text{ versus } \quad H_1:\mu \neq \nu,
\end{align}
where $H_0$ is the null hypothesis which suggests that the two samples are from the same distribution, while $H_1$ is the alternative hypothesis which suggests that they are from different distributions.

\begin{remark}[Permutation Two-sample Test]
\label{remark:permutation_test}
Authors in~\citet{wang2022two} propose a two-sample test with projected Wasserstein distance~\citep{paty2019subspace}, which admits max sliced Wasserstein (Max-SW) (Remark~\ref{remark:optimization_based_slicing}) as a special case. The test statistic with Max-SW can be written as:
\begin{align}
    T = \max_{\theta \in \Sm^{d-1}} W_2^2(\theta \sharp \hat{\mu}_n,\theta \sharp \hat{\nu}_m),
\end{align}
where $\hat{\mu}_n=\frac{1}{n} \sum_{i=1}^n \delta_{X_i}$ and $\hat{\nu}_m = \frac{1}{m} \sum_{j=1}^m \delta_{Y_j}$ are the corresponding empirical distributions. The test is conducted as follows: 
\begin{enumerate}
    \item We split $X_1,\ldots,X_n$ into two sets $X^{Tr} \cup X^{Te}$ and $Y_1,\ldots,Y_m$ into two sets $Y^{Tr} \cup Y^{Te}$.
    \item We find $\theta^\star = \arg\max_{\theta \in \Sm^{d-1}} W_2^2(\hat{\mu}_{Tr}, \hat{\nu}_{Tr})$, where $\hat{\mu}_{Tr}$ and $\hat{\nu}_{Tr}$ are empirical distributions over $X^{Tr}$ and $Y^{Tr}$. We compute $T^\star = W_2^2(\theta^\star \sharp \hat{\mu}_{Te}, \theta^\star \sharp \hat{\nu}_{Te})$.
    \item We shuffle $X^{Te} \cup Y^{Te}$, then compute the statistic \\ $T = SW_2^2(\theta^\star \sharp \hat{\mu}_{Te}, \theta^\star \sharp \hat{\nu}_{Te})$. We repeat this $N$ times to obtain $T_1,\ldots,T_N$. We compute the p-value $\frac{1}{N} \sum_{i=1}^N I(T_i \geq T^\star)$.
    \item We reject the null hypothesis if the p-value $\leq \alpha$, where $\alpha$ is a given significance level.
\end{enumerate}
Authors in~\citet{wang2022two} also generalize the test by introducing kernel projected Wasserstein (KPW) distance which is defined as:
\begin{align}
    KPW_2^2(\mu,\nu)=\max_{f \in \setF} W_2^2(f \sharp \mu,f\sharp \nu),
\end{align}
where $\mathcal{F}=\{f \in \mathcal{H} \mid \|f\|_{\mathcal{H}} \leq 1\}$ with $\mathcal{H}$ is the reproducing kernel Hilbert space. Metric and statistical properties of KPW when $f:\Re^d \to \Re$ (max kernel sliced Wasserstein) are investigated in~\citet{wang2025statistical}.
\end{remark}

\begin{remark}[Bootstrap Two-sample Test]
\label{remark:bootstrap_two_sample_test}
Authors in~\citet{hu2025two} consider the duality of a sparse version of Max-SW of order 1 to construct a test statistic:
\begin{align}
    &T^\star = \left(\frac{nm}{n+m}\right)^{1/2} \max_{\theta \in \Sm^{d-1}, \|\theta\|_0 \leq s} W_1(\theta \sharp \hat{\mu}_n, \theta \sharp \hat{\nu}_m), \\
    &W_1(\theta \sharp \hat{\mu}_n, \theta \sharp \hat{\nu}_m) = \sup_{f \in Lip_1} \left(\frac{1}{n} \sum_{i=1}^n f(\langle \theta, X_i \rangle) - \frac{1}{m} \sum_{j=1}^m f(\langle \theta, Y_j \rangle)\right), \nonumber
\end{align}
where $Lip_1$ is the set of all Lipschitz-1 functions and $s$ is a tuning parameter. The test is then conducted as follows:
\begin{enumerate}
    \item We first define the bootstrap statistic of $T^\star$. We sample $M = (M_1,\ldots,M_n)$ from $Multinomial(n; 1/n,\ldots,1/n)$ and $M' = (M'_1,\ldots,M'_m)$ from $Multinomial(m; 1/m,\ldots,1/m)$. We compute:
    \begin{align}
        T' &= \left(\frac{nm}{n+m}\right)^{1/2} \max_{\theta \in \Sm^{d-1}, \|\theta\|_0 \leq s} \sup_{f \in Lip_1} \nonumber \\
        & \quad \left(\frac{1}{n} \sum_{i=1}^n (M_i - 1) f(\langle \theta, X_i \rangle) - \frac{1}{m} \sum_{j=1}^m (M'_j - 1) f(\langle \theta, Y_j \rangle) \right).
    \end{align}
    We repeat this procedure $N$ times to obtain $T'_1,\ldots,T'_N$. We compute the p-value $\frac{1}{N} \sum_{i=1}^N I(T'_i \geq T^\star)$.
    \item We reject the null hypothesis if the p-value $\leq \alpha$, where $\alpha$ is a given significance level.
\end{enumerate}
We refer the reader to~\citet{hu2025two} for more details about computation and construction of simultaneous confidence intervals.
\end{remark}

\subsection{Feature Screening}
\label{subsec:feature_screening:sec:stats:chapter:applications}

Feature screening is a preprocessing technique used to identify and retain the most relevant variables from high-dimensional data before model training. It aims to reduce dimensionality, improve model interpretability, and enhance computational efficiency by filtering out irrelevant or redundant features. We now discuss the application of SOT in feature screening.

\begin{definition}[Sliced Wasserstein Dependency and Sliced Wasserstein Correlation]
    \label{def:SWD_SWC}
    Let $X \in \Re^{d_1}$ and $Y \in \Re^{d_2}$ be two random variables with corresponding distributions $\mu$ and $\nu$, and the joint distribution $\gamma$. The sliced-Wasserstein dependency (SWD) is defined as~\citep{li2023scalable}:
    \begin{align}
        SWD_p(X,Y) = SW_p(\gamma, \mu \otimes \nu),
    \end{align}
    where $\mu \otimes \nu$ is the product probability measure. The sliced-Wasserstein correlation (SWC) is defined as:
    \begin{align}
        SWC_p(X,Y) = \frac{SWD_p(X,Y)}{\sqrt{SWD_p(X,X) SWD_p(Y,Y)}}.
    \end{align}
\end{definition}

From~\citet[Theorem 1]{li2023scalable}, $SWD_p(X,Y) \geq 0$, and equality holds if and only if $X$ and $Y$ are independent.

\begin{remark}[Model-Free Feature Screening]
    \label{remark:feature_screening_SW}
    Let $X \in \Re^{d_1}$ and $Y \in \Re^{d_2}$ be two random variables over features and responses respectively with the joint distribution $\gamma$. We observe a random sample of size $2n$ from $\gamma$, i.e., $\{(x_i,y_i)\}_{i=1}^{2n}$. We randomly split them into two sets $I = \{(x_i,y_i)\}_{i=1}^{n}$ and $\tilde{I} = \{(\tilde{x}_i, \tilde{y}_i)\}_{i=1}^{n}$. For $k = 1, \ldots, d_1$, we denote $X^{(k)}$ and $\tilde{X}^{(k)}$ as the $k$-th column of $X$ and $\tilde{X}$. The active set of features is defined as~\citep{li2023scalable}:
    \begin{align}
        \hat{\mathcal{A}} = \{k \in \{1,\ldots,d_1\} \mid \widehat{SWD}_1(X_k, Y) \geq c_1 n^{-c_2} \},
    \end{align}
    where $\widehat{SWD}_1(X_k, Y) = SW_1 \left(P_{\{(x_i^{(k)}, y_i)\}}, P_{\{(\tilde{x}_i^{(k)}, \tilde{y}_i)\}} \right)$ with $P_S$ denoting the empirical distribution over set $S$, and constants $c_1 > 0$ and $0 \leq c_2 < 1/2$.
\end{remark}

\subsection{Density Regression}
\label{subsec:density_regression:sec:stats:chapter:applications}
In contrast to Section~\ref{subsec:regression:sec:ML:chapter:applications}, where we consider distributional predictors and real responses, density regression addresses settings with real predictors and distributional responses, or distributional predictors and distributional responses.

\begin{remark}[Sliced Wasserstein Regression]
\label{remark:SWRegression}
Authors in~\citet{chen2023sliced} propose an approach for regression of distributional responses on Euclidean predictors. Let $(X,\mu) \sim \setP(\Re^h \times \setP_2(\Re^d))$ be a random pair. The slice-averaged Wasserstein (SAW) regression of $\mu$ given $X = x$ is defined as:
\begin{align}
    &m(x) = \arg\min_{\nu \in \setP_2(\Re^d)} M(\nu, x), \\
    &M(\cdot,x) = \mathbb{E}[SW_2^2(\mu,\cdot) \mid X=x].
\end{align}
The global slice-averaged Wasserstein (GSAW) regression given $X = x$ is:
\begin{align}
    &m_G(x) = \arg\min_{\nu \in \setP_2(\Re^d)} M_G(\nu, x), \\
    &M_G(\cdot,x) = \mathbb{E}[s_G(X,x) SW_2^2(\mu,\cdot) \mid X=x],
\end{align}
where $s_G(X,x) = 1 + (X - \mathbb{E}[X])^\top \operatorname{Var}[X]^{-1} (x - \mathbb{E}[X])$. There is also a second regression framework that relies on using the inverse Radon transform (Remark~\ref{def:Inverse_Radon_Transform_measures}). In particular, the slice-wise Wasserstein (SWW) regression is defined as:
\begin{align}
    &\tilde{m}(x) = \mathcal{R}^{-1}\left[\arg\min_{\nu \in \setP_2(\Re \times \Sm^{d-1})} \tilde{M}(\nu, x)\right], \\
    &\tilde{M}(\cdot,x) = \mathbb{E}[W_2^2(\setR \mu,\cdot) \mid X = x],
\end{align}
where $\setR$ and $\setR^{-1}$ denote the Radon transform and its inverse. The global slice-wise Wasserstein (GSWW) regression given $X = x$ is:
\begin{align}
    &\tilde{m}_G(x) = \mathcal{R}^{-1}\left[\arg\min_{\nu \in \setP_2(\Re \times \Sm^{d-1})} \tilde{M}_G(\nu, x)\right], \\
    &\tilde{M}_G(\cdot,x) = \mathbb{E}[s_G(X,x) W_2^2(\setR \mu, \cdot) \mid X = x].
\end{align}
In practice, expectations and variances are approximated by Monte Carlo samples.
\end{remark}

\begin{remark}[Bayesian Density-Density Regression]
    \label{remark:BayesianDDR}
    Authors in~\citet{nguyen2025bayesian} use SW for a density regression framework in which both responses and predictors are distributions. Let $(\mu_1, \nu_1), \ldots, (\mu_N, \nu_N) \simiid \setP(\setP_2(\Re^h) \times \setP_2(\Re^d))$ be random pairs. SW is used to construct a generalized likelihood~\citep{bissiri2016general}:
    \begin{align}
        \nu_i \mid \mu_i, f \propto \exp\left(-w \sum_{i=1}^N SW_2^2(f \sharp \mu_i, \nu_i)\right),
    \end{align}
    where $f: \Re^h \to \Re^d$ is a random regression function with prior $f \sim p(f)$. Authors in~\citet{nguyen2025bayesian} consider a parametric version $f_\phi$ with $\phi \sim p(\phi)$ and perform inference for the generalized posterior $p(f_\phi \mid \{\nu_i, \mu_i\}_{i=1}^N)$ using a Metropolis-adjusted Langevin algorithm. Then, the predictive distribution $\nu^*$ given a predictor $\mu^*$ can be defined as:
    \begin{align}
        p(\nu^*(y) \mid \mu^*, \{\nu_i, \mu_i\}_{i=1}^N) = \mathbb{E}[p(\nu^*(y) \mid \mu^*, f) \mid \{\nu_i, \mu_i\}_{i=1}^N].
    \end{align}
\end{remark}
\subsection{Variational Inference}
\label{subsec:VI:sec:stats:chapter:applications}

Variational inference~\citep{blei2017variational} (VI) is an approximate Bayesian inference technique for models with an intractable posterior. In particular, given a joint distribution $p(x,\phi)$ where $x$ denotes observed variables and $\phi$ denotes latent variables, we are interested in approximating $p(\phi \mid x)$ since it is intractable. VI finds the best variational posterior $q(\phi \mid x)$ that is the closest member to $p(\phi \mid x)$ in a family $\mathcal{Q}$. The "closeness" is often defined via Kullback–Leibler (KL) divergence:
\begin{align}
    \min_{q \in \mathcal{Q}} KL(q(\phi \mid x), p(\phi \mid x)),
\end{align}
where $KL(q,p) = \int q(x) \log\frac{q(x)}{p(x)} \diff x$. The optimization is often carried out under the duality, i.e., by maximizing the evidence lower bound~\citep{blei2017variational}.

\begin{remark}[Sliced Wasserstein Variational Inference]
    \label{remark:SWVI}
    Authors in~\citet{yi2023sliced} replace the KL divergence by SW:
    \begin{align}
        \min_{q \in \mathcal{Q}} SW_p(q(\phi \mid x), p(\phi \mid x)).
    \end{align}
    Since $SW_p(q(\phi \mid x), p(\phi \mid x))$ is intractable, we can use discrete approximations of $q(\phi \mid x)$ and $p(\phi \mid x)$. Sampling from $q(\phi \mid x)$ is often easy due to the construction of the variational families. On the other hand, sampling from $p(\phi \mid x)$ requires using Monte Carlo methods such as Markov chain Monte Carlo.
\end{remark}

\begin{remark}[Sliced Wasserstein Joint Contrastive Inference]
    \label{remark:SWJCI}
    To avoid sampling directly from the posterior, we can utilize joint contrastive inference~\citep{dumoulin2017adversarially}. In particular, distributional SW (Remark~\ref{remark:optimization_based_slicing}) is used for joint contrastive inference in~\citet{nguyen2021distributional} with the following optimization:
    \begin{align}
        \min_{q \in \mathcal{Q}} DSW_p(q(\phi, x), p(\phi, x)),
    \end{align}
    where $q(\phi, x)$ is the joint variational posterior. Sampling from both $q(\phi, x)$ and $p(\phi, x)$ is often tractable and relatively easy. Therefore, we can approximate the optimization using discrete approximations of $q(\phi, x)$ and $p(\phi, x)$ efficiently. However, one issue of this approach is lifting the variational inference problem into a higher dimension.
\end{remark}

\subsection{Posterior Sampling}
\label{subsec:posterior_sampling:sec:stats:chapter:applications}
One approximate inference technique in Bayesian statistics is to sample
from the posterior distribution. Given a joint distribution $p(x, \phi)$ where $x$ denotes observed variables and $\phi$ denotes latent variables, we are interested in sampling from $p(\phi \mid x)$. It turns out that we can design a SW gradient flow to achieve this task. 

\begin{remark}[Posterior sampling with gradient flow]
    \label{remark:posterior_sampling}
    From Proposition~\ref{definition:gradient_flow_eculidean_probablity}, we know that the Wasserstein flow with the Fokker-Planck free energy
functional:
\begin{align}
    F(\mu) = U(\mu) - \beta^{-1} E(\mu),
\end{align}
where $U(\mu) = \int_{\Re^d} \Phi(\phi) \diff \mu(\phi)$ is the potential energy and $E(\mu) = - \int_{\Re^d} \log \frac{\diff \mu}{\diff \phi}(\phi) \diff \mu(\phi)$ is the
entropy, will converge to the
unique stationary solution with density proportional to $\exp(-\beta \Phi(\phi))$. Therefore, we can set $\Phi(\phi) = \bar{p}(\phi \mid x)$ where $\bar{p}(\phi \mid x)$ is the unnormalized density of the posterior $p(\phi \mid x)$ to obtain a flow that converges to the posterior. By flowing through the Wasserstein gradient flow using an empirical approximation of $\mu$, we can obtain an empirical approximation of the posterior. Authors in~\citet{bonet2022efficient} suggest that SW gradient flow can also be used as a fast replacement for the Wasserstein gradient flow.
\end{remark}

\subsection{Empirical Bayes}
\label{subsec:empirical_bayes:sec:stats:chapter:applications}

In the Bayesian paradigm with a model $p(x,\phi)$ ($x$ denotes observed variables and $\phi$ denotes latent variables), choosing the prior $p(\phi)$ plays a very important role. Let us parameterize the prior $p_\psi(\phi)$ for $\psi \in \Psi$. A common way to estimate hyperparameters is to rely on Bayesian
Occam’s razor: choose $\psi$ such that we maximize the marginal likelihood $p_\psi(x) = \int p(x \mid \phi) p_\psi(\phi) \diff \phi$. 

\begin{remark}[Empirical Bayes for Bayesian autoencoders]
    \label{remark:empirical_Bayes}
    To avoid overfitting, authors in~\citet{tran2021model} suggest utilizing the minimization of distributional SW instead of maximizing the marginal likelihood:
    \begin{align}
        \min_{\psi \in \Psi} DSW_p(p_\psi(x), q(x)),
    \end{align}
    where $q(x)$ is the data distribution. Again, discrete approximations of $p_\psi(x)$ and $q(x)$ are used for optimization. This framework is used to select the prior for Bayesian autoencoders in~\citet{tran2021model}.
\end{remark}

\subsection{Approximate Bayesian Computation}
\label{subsec:ABC:sec:stats:chapter:applications}

Approximate Bayesian Computation (ABC)~\citep{csillery2010approximate} is a family of methods for performing Bayesian inference when the likelihood function is computationally intractable or unavailable. Instead of evaluating the likelihood explicitly, ABC relies on simulating data from the model and comparing it to observed data using a distance metric. If the simulated data is sufficiently close to the observed data (within a tolerance level), the corresponding parameter value is accepted as a sample from the approximate posterior. ABC is particularly useful in complex models such as those in population genetics, epidemiology, and systems biology.

\begin{remark}[Sliced Wasserstein Approximate Bayesian Computation]
    ABC is designed for inference under a joint model $p(x,\phi) = p(\phi) p(x \mid \phi)$ where $p(x \mid \phi)$ is computationally intractable or unavailable; however, we are able to sample from $p(x \mid \phi)$. We observe data $x_1,\ldots,x_n \sim q(x)$ where $q(x)$ is the data distribution. Sliced Wasserstein ABC~\citet{nadjahi2020approximate} repeats the following procedure:
    \begin{enumerate}
        \item Sample a parameter $\phi$ from the prior distribution $p(\phi)$, and a synthetic dataset $\tilde{x}_1,\ldots,\tilde{x}_m \sim p(x \mid \phi)$.
        \item If $SW_p\left(\frac{1}{n}\sum_{i=1}^n \delta_{x_i}, \frac{1}{m}\sum_{j=1}^m \delta_{\tilde{x}_j}\right) \leq \epsilon$ (where $\epsilon$ is a given threshold), we keep $\phi$; otherwise, we reject it.
    \end{enumerate}
    The accepted $N$ samples drawn from the above procedure follow the ABC posterior distribution:
    \begin{align}
        &p_\epsilon(\phi \mid x_1,\ldots,x_n) \nonumber \\ 
        &\propto p(\phi) \int I\left(SW_p\left(\frac{1}{n}\sum_{i=1}^n \delta_{x_i}, \frac{1}{m}\sum_{j=1}^m \delta_{\tilde{x}_j}\right) \leq \epsilon\right) \nonumber \\
        &\quad \times p(\tilde{x}_1 \mid \phi) \diff \tilde{x}_1 \ldots p(\tilde{x}_m \mid \phi) \diff \tilde{x}_m,
    \end{align}
    Similar to Wasserstein ABC~\citep{bernton2019approximate}, SW-ABC does not require summary statistics as in Euclidean ABC while enjoying fast computation of SW. When $\epsilon \to 0$, the SW-ABC posterior converges to the true posterior. In addition to rejection ABC, we refer the reader to~\citet{nadjahi2020approximate} for sequential Monte Carlo samplers of SW-ABC.
\end{remark}

\subsection{Summarizing Random Partition}
\label{subsec:summarizing_random_partition:sec:stats:chapter:applications}

Inferring a partition of data, commonly known as clustering, aims to assign observations into distinct groups based on similarity or shared structure. A principled probabilistic framework for this task is provided by Bayesian mixture models~\citep{gelman1995bayesian}, which model the data as arising from a mixture of latent components, each corresponding to a cluster. In this framework, each data point is assumed to be generated by first sampling a latent cluster assignment from a prior distribution (such as a categorical or Dirichlet Process prior), followed by drawing the observation from a component-specific distribution (e.g., a Gaussian with cluster-specific parameters). Bayesian inference allows for the incorporation of uncertainty in the number of clusters and in the cluster assignments themselves, producing posterior distributions over partitions rather than committing to a single clustering. After having the posterior, a natural question is how to obtain a point estimate for decision making, interpretation, comparison, and so on~\citep{wade2023bayesian}. 

\begin{remark}[Summarizing Random Partition via Mixing Measures]
    \label{remark:summarizing_random_partition} 
    Consider the following mixture model:  
\begin{align}
    Y_1, \ldots, Y_n \simiid F, \quad F = f * G, \quad G \sim p(G),
\end{align}
where $f$ is a kernel and $p(G)$ denotes a prior on the random mixing
measure $G$. Authors in~\citet{nguyen2024summarizing} propose to obtain a point estimate of $G$ by a decision theoretic framework:
\begin{align}
\label{eq:posterior_expected_loss}
    \hat{G}^\star = \operatorname{argmin}_{\hat{G}} \mathbb{E}[SW_2^2(G, \hat{G})
  \mid Y_1, \ldots, Y_n], 
\end{align}
SW can be replaced by other SW variants that fit the hyperparameter space of the mixing measures. For example, the authors propose SW variants for Gaussian mixtures in the work. The optimization is approximated using posterior samples generated by Monte Carlo methods such as Markov chain Monte Carlo (MCMC) posterior simulation. In addition, the distance between two measures is approximated by the distance between
their truncated versions. After obtaining a point estimate of the mixing measure $\hat{G}$, a point estimate of the density can be obtained by convolution
with the kernel as
\begin{align}
    \hat{F} = f * \hat{G}.
\end{align}
For estimating the partition, we determine the cluster membership
indicator $z_i$ by maximum a posteriori (MAP): 
\begin{align}
  \hat{z}_i = \operatorname{argmax}_{k \in \{1,\ldots,K\}} p(z_i = k \mid \hat{G}, y_i).
\end{align}
\end{remark}

\section{Computer Graphics and Vision}
\label{sec:computer_graphic_and_vision:chapter:applications}
In this section, we extend the recent survey by~\citet{bonneel2023survey} by adding new applications of SOT, including neural radiance fields (NeRF), lidar upsampling, and perceptual color difference measures. We also add recent new works in applications mentioned by the survey, such as color transfer, style transfer, texture synthesis, image segmentation, shape comparison and retrieval, shape registration, and image segmentation, denoising, and super-resolution.

\begin{table}[!t]
    \centering
    \begin{tabular}{p{0.3\linewidth} | p{0.65\linewidth}}
        \toprule
        \textbf{Applications} & \textbf{References} \\
        \midrule
       Blue noise sampling & sliced optimal transport sampling~\citep{paulin2020sliced}, non-Euclidean sliced optimal transport sampling~\citep{genest2024non}. 
       \\
       \midrule 
       Neural radiance fields (NeRFs) & Dynamic NeRFs~\citep{ramasinghe2024improving}, style transfer for NeRFs~\citep{fujiwara2024style}.
       \\
       \midrule
       Lidar Upsampling &  Pointcloud upsampling~\citep{savkin2022lidar}.
       \\
       \midrule
       Color transfer & ~\citep{pitie2005n,pitie2007automated,bonneel2015sliced,bonneel2019spot,bai2023sliced,bonet2024slicing,lobashev2025color}.\\
       \midrule
       Style Transfer&~\citep{shu2017portrait,li2022sliced}.
\\
\midrule
Texture synthesis&~\citep{heeger1995pyramid,tartavel2016wasserstein,heitz2021sliced,yin2022long, rabin2011wasserstein,bonneel2015sliced}.
\\
\midrule
Shape Comparison, retrieval, registration, interpolation, correspondence, and reconstruction & Shape comparison and retrieval~\citep{rabin2010geodesic}, shape registration~\citep{bonneel2019spot,bai2023sliced,chapel2025one}, shape interpolation~\citep{rabin2011wasserstein}, shape correspondence~\citep{le2024integrating}, shape reconstruction~\citep{le2024diffeomorphic}.
       \\
       \midrule
       Image Segmentation, denoising, and super-resolution & Image segmentation~\citep{peyre2012wasserstein,lu2025improving}, image denoising and image super-resolution~\citep{tartavel2016wasserstein}.
       \\
       \midrule 
       Perceptual color difference measure & ~\citep{he2024multiscale}.
       \\
        \bottomrule
    \end{tabular}
    \caption{Applications of sliced optimal transport in computer graphics and vision.}
    \label{tab:SOT_graphics}
\end{table}

\subsection{Blue Noise Sampling}
\label{subsec:blue_noise_sampling:computer_graphics_and_vision:chapter:applications}

As discussed in Section~\ref{sec:MC:chapter:advances}, spherical low-discrepancy sequences can be used to improve estimation of SW (Remark~\ref{remark:low_discrepancy_sequence_sphere}). It is interesting that SW can also be used to construct low-discrepancy sequences or sample blue noise. This construction of low-discrepancy sequences also benefits applications like rendering. From~\citet{mullen2011hot}, we know that Monte Carlo estimation of a Lipschitz function can be upper-bounded by Wasserstein-1 distance. With the connection between SW and Wasserstein (Remark~\ref{remark:SW_connect_Wasserstein}), we know that minimizing SW between a distribution and its discretized version can reduce the approximation error.

\begin{remark}[Sliced Optimal Transport Sampling]
    \label{remark:SOTsampling}
    Let $\mu$ be the uniform distribution over the $d$-dimensional unit ball $B(\Re^d) = \{x \in \Re^d \mid \|x\|_2 \leq 1\}$. We want to construct a point set $x_1,\ldots,x_n$ such that~\citep{paulin2020sliced}:
    \begin{align}
        \min_{x_1,\ldots,x_n} SW_1\left(\setU(B_r(\Re^d)), \frac{1}{n}\sum_{i=1}^n \delta_{x_i}\right).
    \end{align}
    The above optimization involves a semi-discrete version of SW. For $B_r(\Re^d)$, we know that its density is $p(x) = 1/V_{d}(1)$ where $V_d(r) = \frac{\pi^{d/2} r^d}{\Gamma(d/2+1)}$. 
By symmetry, the Radon transform of $\setU(B_r(\Re^d))$ does not depend on
the chosen direction. From~\citet{paulin2020sliced}, its density is $p(s) = \frac{V_{d-1}(\sqrt{1 - s^2})}{V_d(1)}$ and its CDF is:
\begin{align}
    CDF = \begin{cases}
\displaystyle
\sum_{k=0}^{(d-1)/2} (-1)^k 
\binom{\frac{d-1}{2}}{k} 
\frac{x^{2k+1}}{2k+1}, & \text{if } d \text{ is odd}, \\[10pt]
\displaystyle
\frac{\sqrt{\pi} \, \Gamma\left(\frac{1 + d}{2}\right)}{2 \, \Gamma\left(1 + \frac{d}{2}\right)} 
+  {}_2F_1\left(\frac{1}{2}, \frac{1 - d}{2}, \frac{3}{2}, x^2 \right)x, & \text{if } d \text{ is even},
\end{cases}
\end{align}
where $\Gamma$ is the Gamma function, and ${}_2F_1$ is the so-called hypergeometric function, involving polynomials and trigonometric functions. The following iterative algorithm is used to update $x_1,\ldots,x_n$:
\begin{enumerate}
    \item We initialize the point set $x_1,\ldots,x_n$.
    \item We sample $\theta_1,\ldots,\theta_L \simiid \mathcal{U}(\Sm^{d-1})$ (or other Monte Carlo methods).
    \item For $i=1,\ldots,n$, we relocate $x_i = x_i + \frac{1}{L}\sum_{l=1}^L CDF^{-1}((\sigma_{\theta_l}(j) - 0.5)/n) \theta_l$, where $\sigma_{\theta_l}$ is the sorted permutation
of the indices of the projected point set with $\theta_l$, and $CDF^{-1}(x) = \inf \{ t \in \Re \mid CDF(t) \geq x \}$.
    \item Repeat steps 2 and 3 until convergence or reaching a maximum number of iterations.
\end{enumerate}

For sampling on the $d$-dimensional unit cube $[0,1]^d$, authors in~\citet{paulin2020sliced} utilize volume-preserving invertible mappings from $B(\Re^d)$ to $[0,1]^d$~\citep{griepentrog2008bi}. This method is also applicable to multi-class blue noise sampling (including class information). A sliced multi-class sampling technique was introduced by~\citet{salaun2022scalable}.
\end{remark}

\begin{remark}[Non-Euclidean Sliced Optimal Transport Sampling]
    \label{remark:non_Euclidean_SOTsampling}
    Authors in~\citet{genest2024non} extend SOT sampling to non-Euclidean domains including spherical, hyperbolic, and projective
spaces, and with general distributions (including non-uniform). The authors use spherical SW (Definition~\ref{def:SSW}) and hyperbolic SW (Definition~\ref{def:CHSW}) for spherical and hyperbolic domains, respectively. For non-uniform distributions, the authors use empirical distributions with a very large number of atoms to approximate them. In addition, they perform subsampling after projections to obtain $n$ points for each slice to speed up the computation. The optimization is conducted using Riemannian stochastic gradient descent. For intrinsic discrete manifolds (3D meshes), the authors also rely on using an invertible mapping to convert point sets from spherical or hyperbolic domains to intrinsic discrete manifolds.
\end{remark}

\subsection{Neural Radiance Fields}
\label{subsec:Nerf:computer_graphics_and_vision:chapter:applications}
Neural Radiance Fields~\citep{mildenhall2021nerf} (NeRF) is a method for synthesizing novel views of complex 3D scenes by learning a continuous volumetric scene function using deep neural networks. It takes a set of 2D images and their camera poses as input and optimizes a model to render photorealistic images from new viewpoints. We review some uses of SOT in this application.

\begin{remark}[Dynamic NeRFs]
    \label{remark:dynamicNeRF}
    Synthesizing novel views of dynamic scenes using NeRFs is challenging due to the under-constrained nature of estimating dynamic 3D geometry from sparse 2D observations. Authors in~\citet{ramasinghe2024improving} propose to use SW as a lightweight, architecture-agnostic regularizer that leverages the statistical stability of rendered pixel intensity distributions over short time intervals. The key insight is that, for a dynamic scene with smooth temporal changes, the pixel intensity distributions from a fixed camera pose should be similar over nearby timestamps. In particular, rendered pixel values $r_{t,p}$ at time $t$ from camera pose $p$ are treated as empirical samples from a distribution. The regularization term minimizes $SW_1(P(r_{t,p}), P(r_{t+\Delta t,p}))$, encouraging temporal consistency in the pixel distribution. This regularizer integrates seamlessly into dynamic NeRF training by augmenting the standard photometric loss:
\begin{align}
\mathcal{L}_{\text{total}} = \mathcal{L}_{\text{photo}} + \beta \cdot SW_1(P(r_{t,p}), P(r_{t+\Delta t,p})),
\end{align}
where $\beta$ is a weighting parameter.
\end{remark}

\begin{remark}[Style Transfer for NeRFs]
     Style-NeRF2NeRF~\citep{fujiwara2024style} enables artistic 3D style transfer by refining a pre-trained NeRF with stylized multi-view images generated from a text-guided diffusion model. The method decouples image stylization and 3D scene optimization, allowing users to preview stylization before 3D training. To guide the NeRF refinement, the authors replace the standard RGB pixel loss with an SW loss computed on CNN feature activations.  

\begin{align}
\mathcal{L}_{\text{style}} = SW_2^2\left( P_{\mathbf{F}^\ell}, P_{\hat{\mathbf{F}}^\ell} \right),
\end{align}
where $P_{\mathbf{F}^\ell}$ and $P_{\hat{\mathbf{F}}^\ell}$ are empirical measures over feature sets from the target stylized image and the NeRF-rendered image, respectively.
\end{remark}

\subsection{Lidar Upsampling}
\label{subsec:upsampling:computer_graphics_and_vision:chapter:applications}
Lidar upsampling is the process of increasing the point density of a lidar scan by interpolating or estimating additional points between existing measurements. This enhances spatial resolution and creates a smoother, more detailed 3D representation of the environment. Authors in~\citet{savkin2022lidar} propose to use SW as an upsampling loss for 3D point clouds from lidar. In particular, a neural network is trained to map from a point cloud to a larger point cloud. The loss is the SW distance between empirical measures over the two point clouds accordingly. The neural network is trained on a dataset of point clouds.
\subsection{Color Transfer}
\label{subsec:color_grading:computer_graphics_and_vision:chapter:applications}

The aim of color transfer is to transform the color distribution of an input image to match the color distribution of a target image. The color distribution of an input image in a given color space is extracted, while discarding any spatial information. This distribution can be represented in several ways: as a density on a predefined grid, as empirical measures over individual pixel values in color space, using clustered color representations (e.g., via k-means), or even approximated by a parametric model such as a Gaussian distribution. After obtaining discrete measures, the iterative distribution transfer (Definition~\ref{def:IDT}) algorithm can be used~\citep{pitie2005n,pitie2007automated}. The second approach is to use gradient flow with sliced Wasserstein variants as functionals (Remark~\ref{remark:gf_SW_energy}); for example, SW is used in~\citet{bonneel2015sliced}, sliced POT in~\citet{bonneel2019spot}, and sliced unbalanced OT in~\citet{bai2023sliced,bonet2024slicing}. Recently, SW has been incorporated to regularize the color distribution of the generated image and the reference palette in the sampling process of diffusion models~\citep{lobashev2025color}.

\subsection{Style Transfer}
\label{subsec:style:computer_graphics_and_vision:chapter:applications}

Style transfer is a technique that recomposes an image by separating and recombining its content with the style of a target image. Authors in~\citet{shu2017portrait} perform portrait relighting by transporting histograms of normals, positions, and colors between 3D templates fitted to 2D portraits. In addition, SW has also been applied to neural style transfer. In this setting, features of images (Gram matrix) are extracted using deep neural networks (e.g., VGG-19). For each layer, the SW distance between empirical measures over the features of the source image and empirical measures over the features of the target image is minimized using stochastic gradient descent~\citep{li2022sliced}.

\subsection{Texture Synthesis}
\label{subsec:texture:computer_graphics_and_vision:chapter:applications}

Texture synthesis is the process of generating a larger or novel image that visually resembles a given texture sample. The goal is to produce an output that maintains the statistical and structural properties of the input texture while allowing for variation and expansion. Authors in~\citet{heeger1995pyramid} decompose an image into steerable pyramid coefficients. From a Gaussian white noise image, the authors transform its pyramid coefficients to those of the target image using 1D OT. Authors in~\citet{tartavel2016wasserstein} extend the approach with an LBFGS-based optimization of a sliced optimal transport function. Similar to style transfer, neural network features are used to extract texture patterns, then SW is used to match the generated texture with the target one~\citep{heitz2021sliced,yin2022long}. SW barycenters are also used to generate textures that lie between two given textures~\citep{rabin2011wasserstein,bonneel2015sliced}.

\subsection{Shape Comparison, Retrieval, Registration, Interpolation, Correspondence, and Reconstruction}
\label{subsec:shape_retrieval:computer_graphics_and_vision:chapter:applications}

Shape retrieval is the task of finding and ranking shapes from a database that are similar to a given query shape. Authors in~\citet{rabin2010geodesic} sample points from shapes, then use SW to compute the distance between empirical distributions as the retrieval metric.

Shape registration is the process of aligning two or more shapes, typically point clouds, meshes, or volumetric data, into a common coordinate system. Sliced POT is used in~\citet{bonneel2019spot} to derive the Fast Iterative Sliced Transport algorithm (FIST), which replaces the nearest neighbor matching in the iterative closest point algorithm (ICP)~\citep{chen1992object} by a sliced partial optimal transport matching. Sliced optimal partial transport matching is used as a replacement in~\citet{bai2023sliced} and by sliced partial Wasserstein plans in~\citet{chapel2025one}.

By interpreting shapes as the boundaries of indicator functions and normalizing their total mass, one can represent them as 2D or 3D probability densities. Authors in~\citet{rabin2011wasserstein} use the SW barycenter for shape interpolation.

The shape correspondence problem involves finding a meaningful mapping between points on two shapes, such that geometric or semantic features are preserved across the correspondence. Authors in~\citet{le2024integrating} use energy-based SW (Remark~\ref{remark:energy_based_slicing_measure}) to regularize training of the functional map between shapes treated as discrete probability measures.

Shape reconstruction refers to the process of recovering a complete geometric shape, often in 2D or 3D, from partial, noisy, or indirect observations, such as point clouds, silhouettes, or images. Authors in~\citet{le2024diffeomorphic} use SW between varifold representations of 3D shapes~\citep{almgren1966plateau} to train a neural ODE that reconstructs the highly folded white matter surface region from a 3D brain MRI volume.

\subsection{Image Segmentation, Denoising, and Super-resolution}
\label{subsec:segmentation:computer_graphics_and_vision:chapter:applications}

Image segmentation is the process of partitioning an image into meaningful regions or segments, typically to isolate objects or boundaries. It assigns a label to each pixel so that pixels with similar visual characteristics (like color, intensity, or texture) belong to the same region. Authors in~\citet{peyre2012wasserstein} use SW to model regional statistics for image segmentation. Recently, SW has been used in~\citet{lu2025improving} for semi-supervised semantic segmentation. In particular, SW is used to optimize feature representation uniformity and alignment between embeddings of perturbations of input images.

Image denoising and super-resolution aim to improve image quality by removing noise and enhancing resolution, recovering clean and high-detail images from noisy or low-resolution inputs. Authors in~\citet{tartavel2016wasserstein} use SW distance in a Total Variation (TV) based image restoration framework for denoising and super-resolution tasks.

\subsection{Perceptual Color Difference Measure}
\label{subsec:perceptual_measure:computer_graphics_and_vision:chapter:applications}

SW is used to design a perceptual color difference (CD) metric for photographic images that remains robust even when images are misaligned~\citep{he2024multiscale}. Instead of comparing co-located pixels or features, it constructs multiscale Gaussian pyramids of the two images, converts them into a perceptually uniform color space (CIELAB), and represents them as patch distributions. It then projects these patches along random directions, computes one-dimensional Wasserstein distances after sorting, and averages the results across multiple projections and scales. This multiscale SW enables non-local patch comparisons, aligns with how human vision perceives color and structure together, and works effectively for tasks like color transfer and misaligned image evaluation without requiring model training.

\chapter{Discussion}
\label{chapter:discussion}

Throughout this work, we have examined the methodological and computational aspects of sliced optimal transport (SOT), including its theoretical properties, algorithmic implementations, and applications. In this chapter, we extend the discussion to topics that have not been addressed in earlier sections but are essential for broadening the scope of SOT research.

\paragraph{Optimization and Sliced Optimal Transport.} 
Optimization has a strong connection with SOT. In particular, when using SW as a loss (Section~\ref{sec:differentiatingSW:chapter:varitational_SW}), e.g., in minimum SW estimators (Section~\ref{sec:MSWE:chapter:varitational_SW}) or in solving SW barycenters (Section~\ref{sec:SWB:chapter:varitational_SW}), optimization algorithms such as stochastic gradient descent and Newton methods are used in practice. \citet{tanguy2025properties} proved that the algorithm can converge towards critical points in a certain sense and derived the corresponding convergence rate for discrete SW. From a theoretical perspective, it would be interesting to conduct similar studies for other SW variants, including non-linear projection (where Riemannian optimization appears), non-uniform slicing measures, and alternative parameterizations of measures (e.g., pushforward measures). From a methodological standpoint, exploring efficient and effective algorithms beyond conventional gradient descent, e.g., by adding momentum, could enhance a lot of applications of SW.

Optimization also appears in selecting the slicing measure or projection parameters, e.g., Max-SW, DSW (Remark~\ref{remark:optimization_based_slicing}), and Min-SWGG (Definition~\ref{def:MinSWGG}), among others. \citet{nietert2022statistical} show that a projected-gradient-based method for Max-SW can lead to a local optimum with rate $\mathcal{O}(\epsilon^{-4})$. \citet{lin2020projection} achieve the same rate using Riemannian optimization, while \citet{huang2021riemannian} improve it to $\mathcal{O}(\epsilon^{-3})$ with Riemannian block coordinate descent. Potential future directions include understanding convergence for the parameters of the slicing measure (DSW) and extending to other SW variants when the space of projection parameters is not the unit hypersphere. Moreover, solving Min-SWGG (and its variants) also presents optimization challenges since it is not differentiable. Therefore, pseudo-objectives~\citep{mahey2023fast} or differentiable approximation schemes~\citep{chapel2025differentiable} could be further developed. In addition, when computing the distance multiple times for multiple pairs of measures, amortized optimization~\citep{amos2023tutorial} becomes relevant. \citet{nguyen2022amortized,nguyen2023self} propose amortized models for empirical measures, but there is still substantial room for developing amortized models for more general cases and for advancing amortized optimization algorithms (e.g., semi-amortized optimization, objective-based amortized optimization, regression-based amortized optimization, etc.).

Finally, optimization also lies at the root of OT. For one-dimensional transport problems such as POT and UOT, finding efficient ways to solve the corresponding assignment problems remains an interesting and important topic.

\paragraph{Deep Learning and Sliced Optimal Transport.}
There is a strong mutual connection between SOT and deep learning. From SOT to deep learning, SOT can serve as a loss function for training deep learning models such as deep generative models (Section~\ref{subsec:generative_model:sec:ML:chapter:applications}), deep domain adaptation (Section~\ref{subsec:domain_adaptation:sec:ML:chapter:applications}), and other tasks (see Section~\ref{sec:ML:chapter:applications}). In these settings, deep learning models (neural networks) parameterize probability measures of interest, which are compared using SW. Furthermore, SOT maps can inspire the design of deep learning architectures, such as attention mechanisms and transformers (Section~\ref{subsec:attention:sec:ML:chapter:applications}), by solving matching problems inside deep learning modules. In addition, SW embeddings (Section~\ref{sec:SWEmbedding:chapter:varitational_SW}) can be used directly as inputs to deep learning models~\citep{naderializadeh2021pooling,naderializadeh2025aggregating}.

From deep learning to SOT, neural networks can define the projection functions used in SW, as in GRT (Definition~\ref{def:GRT}). In addition, overparameterized version of Radon transform~\citep{nguyen2023hierarchical} can also expand the set of neural networks which can be used for projection.  Deep neural networks can also approximate intractable Kantorovich duals in continuous settings~\citep{wu2019sliced}, and can serve as amortized models to predict projection parameters~\citep{nguyen2022amortized,nguyen2023self}.

\paragraph{Sliced Optimal Transport with discrete structures.}  
In this work, we focus on sliced optimal transport with continuous one-dimensional structures such as lines, curves, and circles.  
Recent works have also explored extensions of sliced optimal transport to discrete structures, particularly tree structures, where optimal transport admits closed-form solutions~\citep{le2019tree,sato2020fast,tran2025treesliced,tran2025treesliced2,tran2025distancebased}.

\paragraph{Sliced probability metrics.}  
The sliced Wasserstein distance can be viewed as an instance of sliced probability metrics~\citep{nadjahi2020statistical}.  
Other examples include the sliced integral probability metric~\citep{nadjahi2020statistical}, sliced Sinkhorn distance~\citep{nadjahi2020statistical}, sliced Cramér distance~\citep{Kolouri2020Sliced}, sliced Fisher divergence~\citep{song2020sliced}, and sliced maximum mean discrepancy~\citep{hertrich2024generative}, among others.  
Slicing techniques such as GRT (Section~\ref{sec:generalized_slicing:chapter:advances}) can be further adapted to these metrics~\citep{kolouri2022generalized}.  
Potential research directions include adapting the slicing techniques discussed in Section~\ref{sec:generalized_slicing:chapter:advances} and the approximation techniques in Section~\ref{sec:MC:chapter:advances} to these metrics.

\begin{acknowledgements}
We would like to thank Professor Peter Müller and Professor Nhat Ho for their insightful comments on the contents of this work. We are also grateful to the anonymous Foundations and Trends reviewers for their detailed and constructive feedback, which greatly improved the quality of this monograph.
\end{acknowledgements}

\backmatter  

\printbibliography

\end{document}